\documentclass{egpubl}
\usepackage{eg2022}
 
\ConferencePaper        
\usepackage[T1]{fontenc}
\usepackage{dfadobe}  

\usepackage{booktabs} 
\usepackage{amsmath}
\usepackage{amssymb}
\usepackage{subfigure} 
\usepackage{contour}
\usepackage{rotating}
\usepackage[abs]{overpic}
\usepackage{xcolor,varwidth}
\usepackage{makecell}

\usepackage{lineno}

\biberVersion
\BibtexOrBiblatex
\usepackage[backend=biber,bibstyle=EG,citestyle=alphabetic,maxbibnames=1000,backref=true]{biblatex} 
\electronicVersion
\PrintedOrElectronic

\usepackage{egweblnk} 

\title[Estimation of Spectral Biophysical Skin Properties\\ from Captured RGB Albedo]%
      {Estimation of Spectral Biophysical Skin Properties\\ from Captured RGB Albedo}

\author[Aliaga et al.]
 {\parbox{\textwidth}{\centering C. Aliaga$^{1}$\qquad
         C. Hery$^{1}$\qquad
         M.Xia$^{1,2}$ \\
         Meta Reality Labs Research$^{1}$ \qquad 
         Cornell University$^{2}$
         }
        \\
}

\begin{document}

\teaser{
  \centering
    \vspace{-12mm}
  \includegraphics[width=\linewidth]{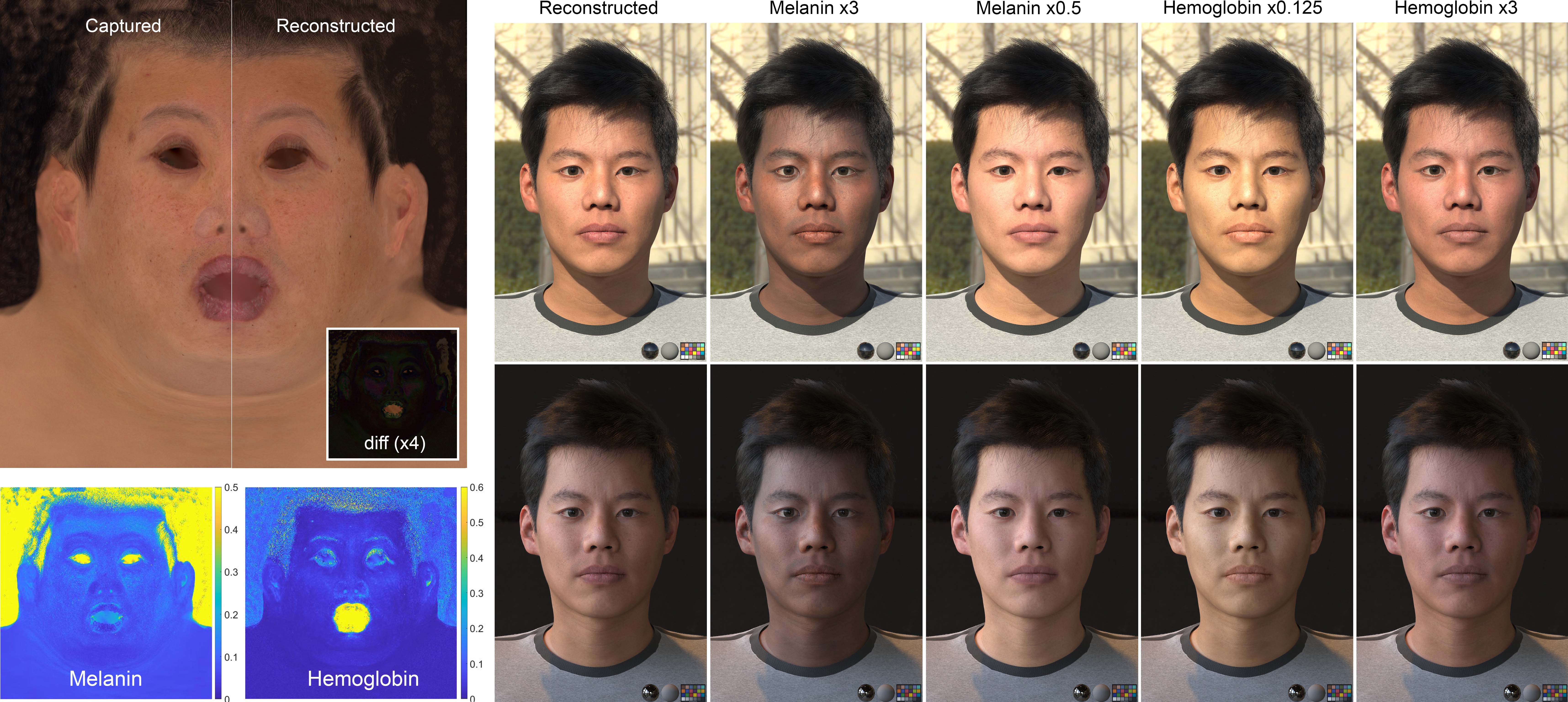}
    \vspace{-6mm}
  \centering
  \caption{Our model takes RGB albedo captures as input, and reconstructs a set of biophysical properties of human skin. On the left, an example of a captured albedo vs the reconstruction of our model (top) using the estimated skin properties, of which melanin and hemoglobin are shown in the bottom. On the right, renderings using albedos modified through edits of some of the estimated parameters, under two different lighting setups (top and bottom rows).}
  \label{fig:teaser}
}

\maketitle
\begin{abstract}
 We present a new method to reconstruct and manipulate the spectral properties of human skin from simple RGB albedo captures. To this end, we leverage Monte Carlo light simulation over an accurate biophysical human skin layering model parameterized by its most important components, thereby covering a plausible range of colors. The practical complexity of the model allows us to learn the inverse mapping from any albedo to its most probable associated skin properties. Our technique can faithfully reproduce any skin type, being expressive enough to automatically handle more challenging areas like the lips or imperfections in the face. Thanks to the smoothness of the skin parameters maps recovered, the albedo can be robustly edited through meaningful biophysical properties.

\ccsdesc[500]{Computing methodologies~Reflectance modeling}
\ccsdesc[500]{Computing methodologies~Reconstruction}

\printccsdesc   
\end{abstract}  

\definecolor{gray}{rgb}{0.5,0.5,0.5}
\definecolor{purple}{rgb}{0.7,0.3,0.7}
\definecolor{blue}{rgb}{0.3,0.5,1}
\definecolor{darkblue}{rgb}{0,0,0.6}
\definecolor{orange}{rgb}{1,0.5,0} 
\definecolor{red}{rgb}{1,0,0}
\definecolor{green}{rgb}{0,0.5,0}
\definecolor{purple}{rgb}{0.5,0,1}

\newcommand{\new}[1]{{\textcolor{darkblue}{#1}}}


\newcommand{\IGNORE}[1]{}

\definecolor{MyDarkBlue}{rgb}{0,0.08,1}
\definecolor{MyDarkGreen}{rgb}{0.02,0.6,0.02}
\definecolor{MyDarkRed}{rgb}{0.8,0.02,0.02}
\definecolor{MyDarkOrange}{rgb}{0.70,0.35,0.02}
\definecolor{MyPurple}{rgb}{0.43,0,1.}
\definecolor{MyRed}{rgb}{1.0,0.0,0.0}
\definecolor{MyGold}{rgb}{0.75,0.6,0.12}
\definecolor{MyDarkgray}{rgb}{0.66, 0.66, 0.66}

\newcommand{\chery}[1]{{\color{green}\emph{Chery: #1}}}
\newcommand{\carlos}[1]{{\color{blue}\emph{Carlos: #1}}}
\newcommand{\mandy}[1]{{\color{purple}\emph{Mandy: #1}}}


\section{Introduction}
Creating convincing photo realistic replicas of human faces has been a long-standing goal of computer graphics, usually relying on skilled artists to fine tune a large number of spatially varying shading parameters. As a crucial part of virtual humans, significant effort has been put on accurately modeling skin. Skin is not only challenging due to its complex interaction with light, but also because the human visual system has specifically evolved for strong facial perception. A number of models have targeted the effect of the biophysical composition of the skin on its final appearance, often centered in its diffuse reflectance resulting from the complex interaction of light inside the dermal tissues~\cite{donner2006spectral,krishnaswamy2004biophysically,baranoski2010light,chen2015hyperspectral,IglesiasEG15}. More recently, the interest in estimating skin properties from captures has grown, mostly focusing on inverting the biophysical properties of skin from its diffuse reflectance, leveraging biophysical constraints~\cite{alotaibi2017biophysical,gevaux2019three,zherebtsov2019hyperspectral,gitlina2020practical,gevaux2021real}.

We present a new model that, given a biophysically-inspired space of spectral skin albedos, makes use of a neural network to accurately recover, out of simple RGB albedos, the main components and structure of the skin. For this, our model builds upon a biophysical description of the main properties in human skin, which we translate into albedo by means of Monte Carlo simulations. 


Thus, our high-level contributions can be summarized as follows:
\begin{itemize}
    \item An expressive space of spectral skin albedo
    constrained by physical properties of real skin, in agreement with measurements reported in tissue optics and medical research. Such space is created through a biophysically-inspired model of human skin of practical complexity.
    \item A learned inverse mapping from skin albedo to its associated spectral biophysical skin properties, that enables the recovery of smooth, high resolution spatially varying maps of skin properties from RGB captured albedos.
    \item Altogether, a framework to faithfully reproduce captured albedos of a wide range of skin types with minimal error, also enabling editing capabilities in a robust way through the estimated biophysical properties.
\end{itemize}

The process is composed of two steps. First, we precompute an albedo space in the form of a high dimensional tensor of skin tones resulting from all possible combinations of skin parameters. The details of our biophysical skin model are described in Section \ref{sec:biophysical_model}. Then, we learn the inverse mapping from an albedo to its associated skin properties in Section~\ref{sec:parameter_estimation}. After discussion of implementation details in Section~\ref{sec:implementation}, we show how to identically reconstruct the albedo and the results of manipulating the skin parameters in Section~\ref{sec:results}, with comparisons to related work in Section~\ref{sec:comparison}.


\section{Related Work}
\label{sec:related_work}

Accurately capturing and modeling the appearance of human skin has been a very active area of research for decades. Here we focus on appearance modeling and reconstruction, and refer to a number of extensive surveys for further details in skin and face appearance modeling~\cite{igarashi2007appearance,nunes2019appearance}, face appearance capture~\cite{klehm15star,weyrich2009principles}, or a broader collection of methods acquiring the optical properties of translucent materials~\cite{frisvad2020survey}.

\emph{Modeling.} There has been a bunch or works focusing on biophysical skin components. Some of these isolated the distributions of melanin and hemoglobin in human skin~\cite{tsumura1999independent,tsumura2003image}. After an early model first introduced multi-spectral skin color~\cite{angelopoulo2001multispectral}, spectral techniques that explicitly drive the appearance through biophysical properties have been developed, spanning models of different complexity, ranging from two layers and the two most relevant components (melanin and haemoglobin)~\cite{donner2006spectral}, up to seven layers and accounting for a large number of parameters related to the chemical composition and structure of human skin~\cite{chen2015hyperspectral,IglesiasEG15,krishnaswamy2004biophysically,baranoski2010light}. Or even a model for dynamic facial color~\cite{JimenezSIGA2010}, built from in vivo measurements of melanin and hemoglobin concentrations.

\emph{Capturing.} Since early efforts in computer graphics to acquire the skin BRDF~\cite{marschner1999image}, a series of approaches focused on human face acquisition, progressively improving from the original Light Stage device~\cite{debevec2000acquiring}. Relying on such Light Stage-type scans, a number of models reconstructed the reflectance functions directly from captures~\cite{ma2007rapid, ghosh2008practical,ghosh2011multiview,graham2013measurement}, or synthesized high-resolution facial surface microgeometries~\cite{graham2013measurement}. Some other subsequent publications expanded on the facial appearance modeling and capture, including dynamic diffuse albedo encoding blood flow, specular intensity or high resolution normals~\cite{gotardo2018,gotardo2018practical}, even single-shot~\cite{riviere2020single}, with editing capabilities~\cite{weyrich2006analysis}. Another method~\cite{donner2008layered} used multi spectral photographs to recover spatially varying biophysical components over a layered skin model. A number of works opted to rely on Kubelka-Munk (KM) theory~\cite{kubelka1931article} to recover the skin parameter maps. An example~\cite{alotaibi2017biophysical} of this kind of models is tested and discussed in Supplementary Section 2.2. In skin research, 3D maps of human skin properties on the full face with shadows can be reconstructed from 3D hyper spectral imaging~\cite{gevaux2019three} via a two-flux KM model. Based on this, neural networks were trained to recover maps of oxygen rate, blood volume fraction, melanin concentration, bilirubin concentration, and epidermal thickness~\cite{gevaux2021real}. Also relying on hyperspectral imaging, a neural network and a more complex 7 layer skin model were presented~\cite{zherebtsov2019hyperspectral} to extract 2D maps of blood volume fraction, blood oxygen saturation, melanin content and epidermal thickness. It demonstrated the importance of the thickness of the superficial bloodless layer. However, both approaches ~\cite{gevaux2021real,zherebtsov2019hyperspectral} lack albedos reconstructed by the model and the resulting estimated parameters, making it difficult to assess its application in a vfx or virtual reality context. Closest to our solution, a recent paper~\cite{gitlina2020practical} reconstructs spectral skin properties through a novel measurement system, benefiting from narrow-band LEDs,  based on a simplified skin model~\cite{JimenezSIGA2010,donner2006spectral}.

Differently to previous works, our model is capable of faithfully reproducing with minimal error any skin type by a set of biophysical properties proven to be relevant. It relies on Monte Carlo simulations to accurately model light-skin interactions, avoiding assumptions (e.g. epidermal blood~\cite{gitlina2020practical}) and corrections needed for KM approaches~\cite{gevaux2021real,saunderson1942calculation}. Also, our model does not require hyperspectral captures, since it operates directly on simple RGB albedos (although it can be extended to spectral captures), and recovers smooth parameter maps that allow for intuitive edits in such parameter space.

\section{A Space of Biophysically-Based Skin Albedos}
\label{sec:biophysical_model}

In this section we describe how we create our albedo space, the details of the skin model and the data employed, together with balance of practical complexity and expressiveness.

\subsection{Skin Model: Structure and Optical Properties}



\subsubsection{Skin Structure}

\label{sec:skin_structure}
 We decided to restrict our model to two layers, epidermis and dermis, since a similar assumption has proven to be adequate in the past \cite{meglinski2002quantitative} and fits our purpose. The epidermis is composed of two parts: the living epidermis and the \emph{stratum corneum}. The latter is the outermost part, and exhibits properties (surface roughness and sebum production) that affect the specular reflectance of the skin. However, it has a minimal effect on the albedo of the skin, given its low absorption and relative small thickness (5 and 20 $\mu m$ depending on the body location~\cite{czekalla2019noninvasive}). 
 
 On the other hand, we model the dermis as a single layer of averaged properties of scattering and absorption of its two sub-layers, the reticular dermis and the papillary dermis. It is modeled as a semi infinite medium, omitting the sub-dermal fat layer. This was decided to keep the generality of the model, with a focus on faces: next to the dermis can be not only fat, but other internal tissues like cartilage or muscles, heavily depending on the anatomical location and composition of the subject. Also, we empirically found that including the dermal thickness has a minimal effect over the resulting diffuse reflectance.



\subsubsection{Absorption}

Following well known multi-layered tissue optics models~\cite{jacques2013optical}, we describe the optical properties of each layer by its spectral absorption ($\mu_a$) and scattering ($\mu_s$) coefficients. The absorption of each layer $\mu_{a_i}$ is the result of the additive contribution of each chromophore absorption $\mu_{a_c}$ present in each layer:
\begin{equation}
\label{eq:layer_absorption}
    \mu_{a_i} = \sum_{c\in C_i} \mu_{a_c} = \sum_{c\in C_i} \frac{V_c p_c \varepsilon_c }{w_c} 
\end{equation}
where $C_i$ is the set of chromophores contained in the layer $i$; $V_c$ is the volume fraction of the substance containing the chromophore $c$; $p_c$ is the concentration in g/L; $\varepsilon_c$ is the molar extinction of the chromophore; and $w_c$ is its molar weight (Table~\ref{table:chromos} in Appendix~\ref{appendix1}).

As a result, the total spectral absorption of epidermis is defined as $\mu_{a_{e}} = V_{m}\left(\varphi_{m} \mu_{a_{eu}} + \left(1 - \varphi_{m}\right) \mu_{a_{ph}}\right) + (1-V_{m})\left( \mu_{a_{\beta-c}} + \mu_{a_{base}}\right)$, and the spectral absorption of the dermis is defined as\\
$\mu_{a_{d}} = V_{b}\left(\varphi_{h} \mu_{a_{hb}} + (1-\varphi_{h})\mu_{a_{hbO2}} + \mu_{a_{bil}} + \mu_{a_{\beta-c}}\right) + (1-V_{b})\mu_{a_{base}}$.

In the same spirit of biophysical approaches~\cite{krishnaswamy2004biophysically,IglesiasEG15,chen2015hyperspectral,donner2006spectral}, we include the effect of melanin, determined by the melanosomes volume fraction $V_{m}$, containing the two types of melanin (eumelanin $\mu_{a_{eu}}$ and pheomelanin $\mu_{a_{ph}}$) in the epidermis, governed by the melanin type ratio $\varphi_{m}$, which greatly varies through skin type. We also incorporate the haemoglobin present in blood $V_{b}$ in the dermis: oxygenated haemoglobin $\mu_{a_{hbO2}}$, responsible for the saturated reddish tint, and deoxygenated haemoglobin $\mu_{a_{hb}}$, responsible for a desaturated purple color; $\varphi_{h}$ being the haemoglobin type ratio. Other chromophores included by the model are the beta-carotene $\mu_{a_{\beta-c}}$ in both epidermis and dermis and bilirubin $\mu_{a_{bil}}$ in the dermis, contained in blood. Additionally, we also include a baseline of skin absorption $\mu_{a_{base}} = 7.84x10^{8}~\lambda^{-3.255}$ \cite{saidi1992transcutaneous} already employed in previous models. 



\subsubsection{Scattering}
We treat epidermis and dermis as homogeneous mediums, the later being semi infinite. Both layers have an index of refraction of 1.4 resulting from the weighted (by thickness) sum of the corresponding sub layers: \emph{stratum corneum} (1.53), living epidermis (1.34), papillary dermis (1.39) and reticular dermis (1.395). The interfaces of the two layers, as well as other details of multiple scattering in the tissue, are discussed in Section~\ref{sec:layered_model}.

The reduced scattering coefficient $\mu_s'$ is modeled through a function of wavelength fitted by Jacques \cite{jacques2013optical}, which is generic for a wide range of human tissues :
\begin{equation}
\label{eq:scattering}
    \mu_s'(\lambda) = a\left( f_{Ray} \left(\frac{\lambda}{\lambda_r}\right)^{-4} 
    + (1 - f_{Ray})\left(\frac{\lambda}{\lambda_r}\right)^{-b_{Mie}}\right)
\end{equation}
where the wavelength $\lambda$ is normalized by a reference wavelength $\lambda_r = 500$ nm, the scaling factor $a = \mu_s'( \lambda_r)$, $f_{Ray}$ is the relative contribution of Rayleigh scattering, and $b_{Mie}$ characterizes the wavelength dependence of the Mie scattering component. We rely on the coefficients reported in the optics literature~\cite{bashkatov2011optical}, leading to $a$ = 36.4, $f_{Ray}$ = 0.48, and $b_{Mie}$ = 0.22 for human skin. In addition, we include the spectral dependency of the anisotropy factor $g$, which has been measured in the past~\cite{van1989skin}, ranging from 0.73 at 380 nm to 0.84 at 780 nm. It is expressed as $g_e \sim g_d \sim 0.62 + \lambda~0.29x10^{-3}$. 

\begin{figure}
  \centering
  \includegraphics[width=\linewidth]{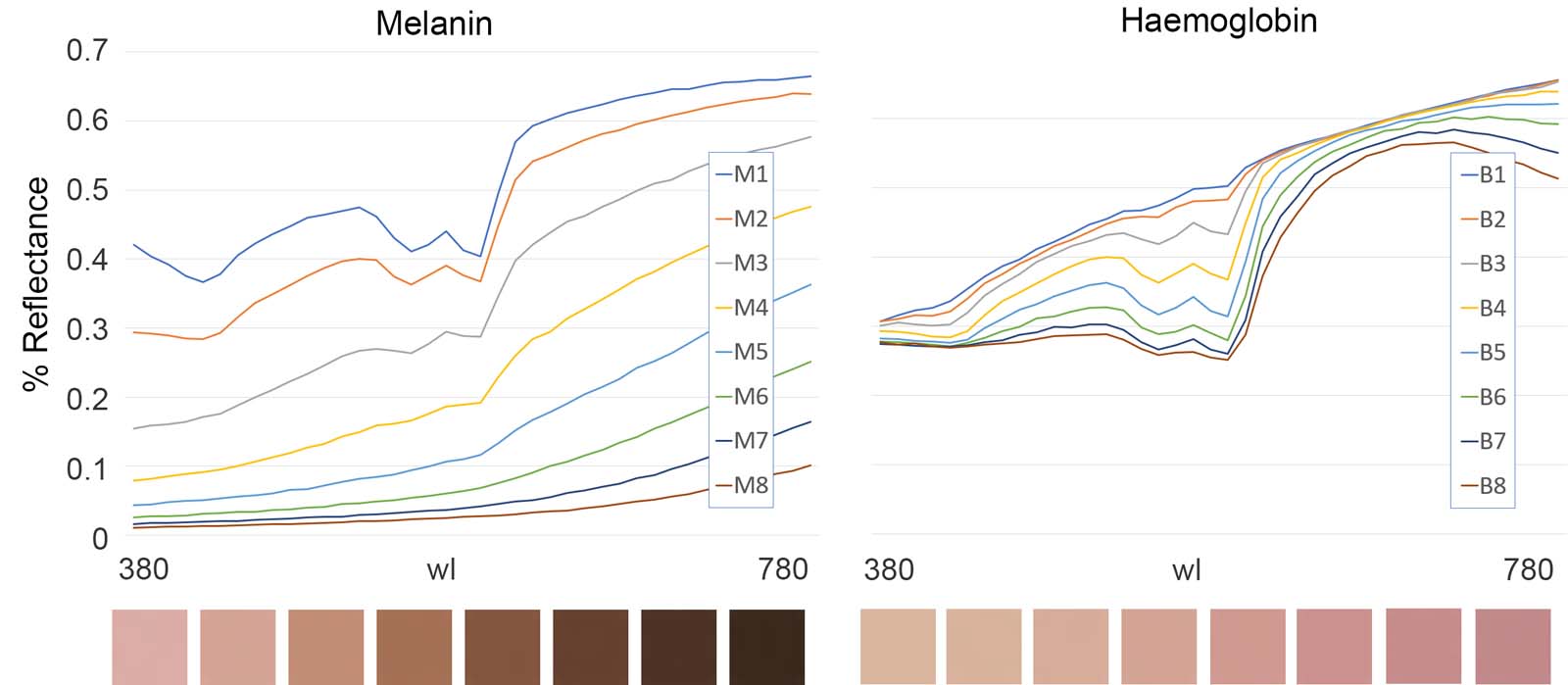}
  \caption{Melanin and haemoglobin have a highly non-linear relationship with the final albedo. In this example, we vary melanin and haemoglobin to show the effect on the spectral reflectance (top) and corresponding final albedos (bottom). \emph{Left:} Melanin is cubicly sampled with Mn = (0.2, 0.7, 2.3, 5.4, 10.6, 18.2, 29.9, 43.1)\%, while haemoglobin is fixed at 2\%. \emph{Right:} haemoglobin is quarticly sampled with Bn = (0.1, 0.2, 0.7, 2, 4.5, 9.6, 17.7, 30)\%, while melanin is fixed at 0.8\%. These spectral reflectance plots are qualitatively in agreement with previous models of computer graphics~ \cite{krishnaswamy2004biophysically}, optics~\cite{zherebtsov2019hyperspectral}, color research~\cite{VrhelSpectra}, computer vision~\cite{angelopoulo2001multispectral}, and dermatology~\cite{anderson1981optics}.
}
    \vspace{-6mm}
  \label{fig:mel_blood_reflectance}
\end{figure}

\subsection{Layered Model for Albedo Generation}
\label{sec:layered_model}

For  computing  the  diffuse  reflectance  of  a  skin  patch,  we  first attempted to use a Kubelka-Munk - based layering model~\cite{alotaibi2017biophysical}. However, it suffers from lack of expressiveness to robustly recover the parameters of a wide enough range of skin types. Details on such exploration can be found in the Supplemental, Section 2.2. Thus, we decided to shift to a brute force Monte Carlo random walk. 
At this stage, we focus on resolving the transports through the stacking of the different layers of the skin structure (\ref{sec:skin_structure}). We conduct the walk in 2D, assuming symmetry in the azimuthal planes of each layer. The interface between the epidermis and dermis only considers the change in scattering and absorption parameters, since the index of refraction is measured to be mostly the same in both layers~\cite{listeroptical}. 

For each skin tone, we run spectral simulations for the wavelengths comprised in the visible range, between 380 to 780 nm. We empirically found that steps of 10 nm is enough to produce stable, noise free albedos. For each wavelength, a million photons are launched, simulating the 2D random walk over the two-layered, semi infinite medium, to produce diffuse albedos (by opposition to the rendering step, where we make use of such uv-mapped albedos to conduct true 3D random walks, as outlined in Section~\ref{sec:results}). We treat the skin as an homogeneous medium with exponential decay, since the more recent statistical theories attempting to account for spatial correlations in some materials~\cite{wrenninge2017path,bitterli2018nonexp,deon2019nonexp} do not have corresponding measured data of particle distributions for tissue. We use the multiple scattering parameters listed previously, including the wavelength dependent anisotropy factor $g$ from the Henyey-Greenstein~\cite{henyey1941diffuse} phase function, expressed in its 2D form~\cite{deon2016ahitchhiker}. Note that the simulation starts once the photon has crossed the outermost interface and scatters diffusely into the tissue, namely choosing the initial direction from a Lambertian distribution around the inverse surface normal, in agreement with what we later use at rendering time (Section~\ref{sec:results}).


We validated the biophysical model by plotting the spectral reflectance of different skin tones, showing that our results are in agreement with existing measurements and simulations found in the literature (see Figure~\ref{fig:mel_blood_reflectance}).

\subsection{Albedo Space Parametrization and Sampling}
\label{sec:albedo_space_precomputation}

\begin{table}[t]
\footnotesize
\begin{tabular}{llll}
 Parameter & Description & Epidermis & Dermis \\
 \hline
 $V_{m}$ & Melanin Volume Fraction & [0.001, 1] &  - \\
 $V_{b}$ & Blood Volume Fraction & - &  [0.001, 1] \\
 $t$ &  Thickness [$\mu$m] & [10, 250] & 2100 \\
 $\varphi_{m}$ & Ratio of melanin types & [0.001, 1] &  - \\
 $\varphi_{h}$ & Ratio of haemoglobin types & - &  [0.001, 1] \\
\end{tabular}
\caption{Ranges for the skin properties of our 5D albedo space.}
  \vspace{-6mm}
\label{table:parameters}
\end{table}

Our biophysically-based albedo space is created by varying the skin properties in the ranges listed in Table~\ref{table:parameters}. Note that we consider epidermal thickness, which proved to be critical for parameter estimation~\cite{zherebtsov2019hyperspectral}. Accounting for this parameter over the face, along with varying values in melanin concentrations, helps us achieve local dark zones, such as moles, and generalizes over any skin type. We also allow melanin and haemoglobin to go beyond the usual values for human adults measured in the literature~\cite{meglinski2002quantitative}, in order to automatically handle outliers found in the face, such as the lips, which exhibit very thin epidermis and higher blood concentration, or other cases like underlying veins and capillary veins, or areas with abnormal melanin concentration like freckles or spots. This range expansion is also reasonable for the oxygenation level, since it can vary a lot depending on the physical state of the person, and for the melanin type ratio, where there is little agreement in the available measured data.

Thus, some parameters like bilirubin and $\beta$-carotene concentrations, remain fixed to common values of human skin measurements found in literature (see Table~\ref{table:chromos} in Appendix~\ref{appendix1}). The remaining 5D parameter space is sampled as follows: melanin and haemoglobin are respectively selected cubicly ($\sqrt[3]{V_m}$) and quarticly ($\sqrt[4]{V_b}$), to better adjust to their non linear effect on the albedo (see Figure~\ref{fig:mel_blood_reflectance}); epidermal thickness, melanin type ratio and haemoglobin type ratio all are treated uniformly.

\begin{figure}[t]
  \centering
  \includegraphics[width=\linewidth]{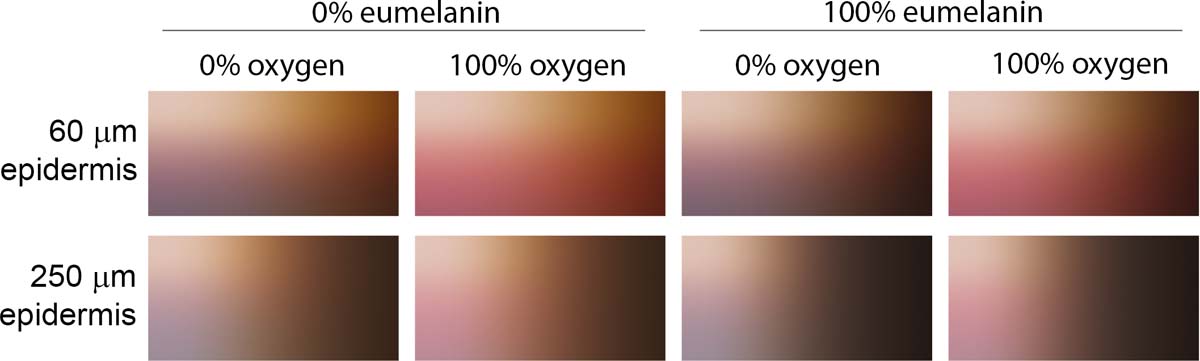}
  \caption{Matrices of resulting skin albedos, built by varying melanin and blood concentrations in subsets (here [0.1, 43] and [0.1, 30] \% for visualization purposes) of the ranges employed for our final space referenced in Table~\ref{table:parameters}. Epidermal thickness, melanin type ratio and blood oxygenation significantly affect the appearance of the skin.}
  \vspace{-6mm}
  \label{fig:palettes}
\end{figure}



\section{Recovering the Skin Properties}
\label{sec:parameter_estimation}

Our biophysical model defines the forward mapping from the skin parameters to the skin albedo. For the inverse process, we need to characterize the mapping from the skin albedo back to skin parameters. The non-bijective nature of such mapping, where many combinations of skin properties can lead to the same albedo, makes this task challenging.

\emph{Metamerism.} It is important to note that our model operates in spectral space to compute the reflectance corresponding to each skin parameter combination, but then such reflectance is integrated to RGB to learn the mapping. This is a design choice: it simplifies the capture, and provides good enough recovered skin parameter maps to both a) reconstruct the original albedo and b) perform plausible edits. However, operating in a higher dimensional space could alleviate metamerism, which could help to disentangle more accurately the skin parameters, generalize the model to any illuminant, or be robust to different levels of exposure, among others. Out of scope for this paper, further exploration can be done inspired by the literature in Spectral Reconstruction from RGB, where significant efforts have been done, specially in data-driven deep learning methods (a recent survey covers the wide variety of existing techniques~\cite{zhang2021learnable}).

\subsection{Look-up Tensor (LUT).} 
\begin{figure}[h]
  \centering
  \includegraphics[width=\linewidth]{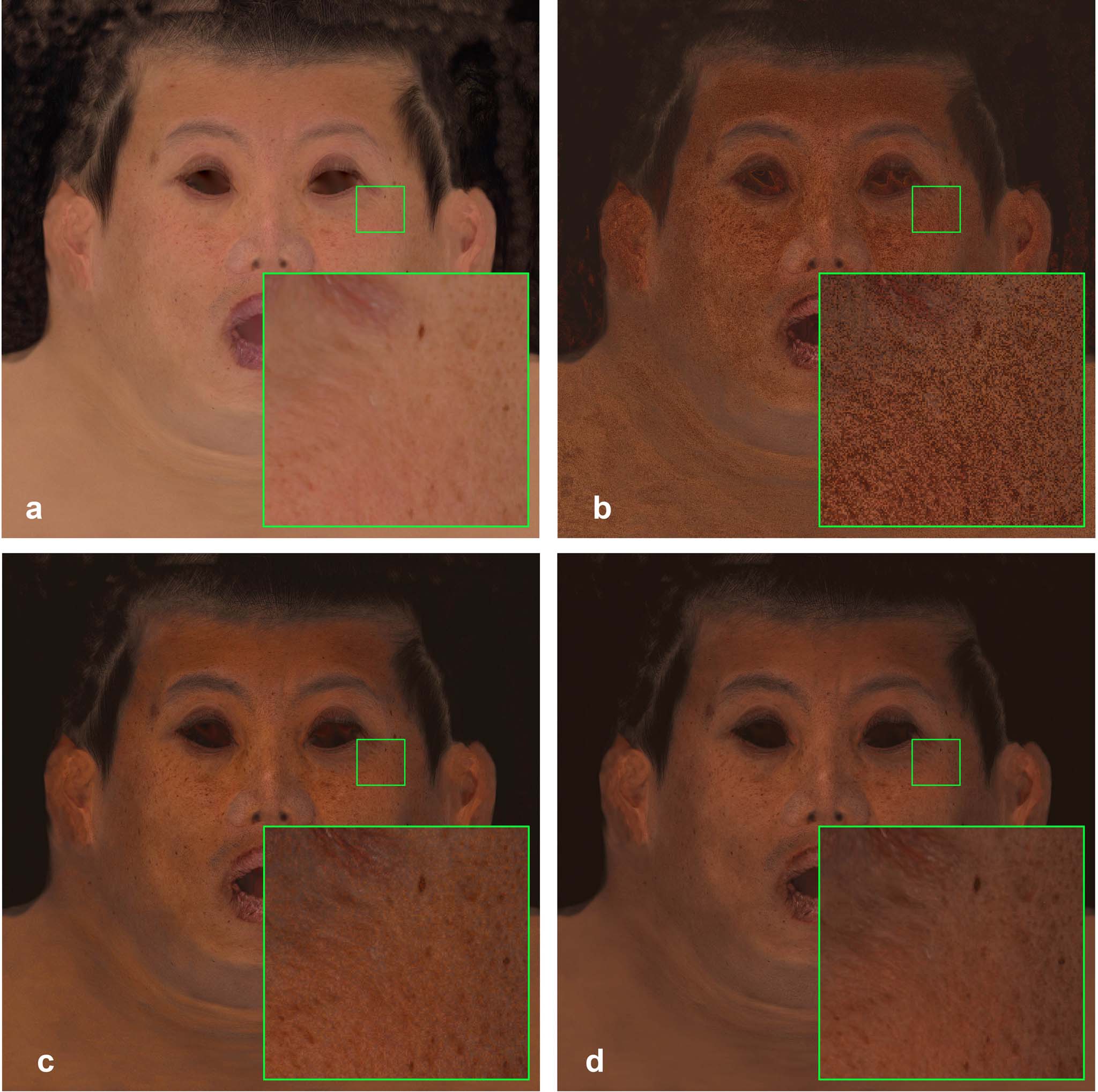}
  \caption{The LUT approach suffers from quantization in the estimated parameters maps even for large tensors. This issue does not reveal during reconstruction, but creates artifacts and even some perceptual color shifting when edits on the skin properties are performed. An example of a x3 edit over the reconstructed melanin concentration of a Type III skin is shown. From left to right: a) original image, and reconstructions (edited melanin x3) using a b) tensor of 55296 skin tones ($V_{m}, V_{b}, t, \varphi_{m},\varphi_{h}$) = (64, 32, 3, 3, 3), c) tensor of 256k skin tones (64, 32, 5, 5, 5), d) learned inverse mapping using 600k points to train the network. Whereas even for densely sampled tensors quantization appears, the neural approach provides smooth maps for the estimated parameters, which results in clean edits.}
  \label{fig:quantization}
\end{figure}

We pre-compute a wide tensor of skin tones by following the sampling strategy outlined in Section~\ref{sec:albedo_space_precomputation}. To estimate the skin parameters given an input albedo, we search the LUT on each texel of the albedo to find the best skin parameter set that minimizes a reconstruction $L_2$ error. Then, we can manipulate them and query the new corresponding albedo from the LUT. Some slices of the 5D tensor are shown in Figure~\ref{fig:palettes}. This approach is able to reconstruct the skin albedo faithfully with an error close to zero. However, the inverted parameter maps are noisy and have many discontinuities, since the mapping from RGB to skin parameters is not smooth. In turn, editing operations over neighboring pixels in such extracted components can lead to unexpected abrupt changes in the reconstructed albedos (see Figure~\ref{fig:quantization}). We leave out of scope of this paper the assessment of different representations or data structures suitable for more efficient search strategies, that could dramatically improve the performance of this approach.

\subsection{Neural Skin Parameter Estimation}
\label{sec:neural_estimation}


To overcome the limitations of the LUT approach, we opted to train an encoder-decoder network to recover smooth skin parameter maps. Both the encoder and the decoder are MLPs of four fully connected layers, with two hidden layers of 70 neurons each. The encoder maps a 3D skin albedo to a 5D skin parameter vector, while the decoder does the inverse. 

\emph{Dataset.} Using our biophysical skin model, we generate a data set of 600k pairs of 5D skin parameter vector and the corresponding 3D albedo, split 80\% and 20\% for training and validation respectively. For validation, we sample the skin parameters according to the uniform distribution. For training, we rely on Quasi-Monte Carlo for a better coverage of the skin parameters, using a low-discrepancy sequence (Halton~\cite{halton1964algorithm}). Both are subsequently non linearly remapped following the schemes detailed in Section~\ref{sec:albedo_space_precomputation}, and the corresponding albedos are computed using the biophysical model.

\emph{Training.} We used the Adam optimizer~\cite{kingma2014adam}, with an learning rate of $10^{-4}$ and a batch size of 4096 for an optimal number of 400 epochs to prevent over fitting. A grid search was performed to find the optimal hyper parameters listed, including the number of hidden neurons. The loss consists of three parts: 
\begin{equation}
   \mathcal{L} = \mathcal{L}_{param} + \mathcal{L}_{albedo} + \mathcal{L}_{cycle} 
\end{equation}



where $\mathcal{L}_{param}$ denotes the parameter loss (encoder), which computes the $L2$ difference between the predicted skin properties from the input albedo and its corresponding ground truth properties. $\mathcal{L}_{albedo}$ (decoder) computes the $L1$ differences between the predicted albedo and the ground truth albedo corresponding to the input skin properties. $\mathcal{L}_{cycle}$ evaluates the full encoder-decoder cycle, computing the $L1$ difference between the predicted albedos from the predicted skin properties, and the ground truth albedos. We chose $L1$ loss for robustness against outliers in albedo. Note that, although in theory, a given albedo can invert to different sets of skin parameters, we use the skin parameter set that generates this albedo from the Monte Carlo simulation; this input serves as weak supervision and works well in practice. 


\section{Implementation Details}
\label{sec:implementation}

\emph{Spectral Downsampling.} The albedo space is generated through spectral computations, but input albedos are in RGB space. To downsample the multi-band spectral values into RGB, we use an existing integration approach~\cite{Mallett2019} (note that we could rely on other concurrent works~\cite{Jakob2019Spectral} to this end). Most color spaces will work here (we use sRGB), considering that diffuse albedos have rather limited gamuts and dynamic ranges. We perform a change of illuminant for direct comparisons to previous work in Section~\ref{sec:comparison}, which we further detail in the Supplementary, Section 1.

 \emph{Using Albedo Maps in Rendering.} The final 3D lit geometries of our virtual faces are rendered using our own skin materials inside \emph{Blender Cycles}~\cite{blender}. Aside from the specular component of the skin, which we represent as a double lobe GGX~\cite{walterGGX}, we follow, in the spirit of a state of the art technique in production~\cite{wrenninge2017path}, a 3D random walk subsurface scattering solution that relies on a numerical albedo inversion around the mean free path and accounting for the anisotropy factor $g$ (see Supplemental Section 2.1). At this stage, we simplify the model to be single layered, dermis and epidermis combined, as a semi infinite medium. Obviously multi-layered models could be employed, for instance directly consuming the chromophores estimations, but these are out of scope of this paper. See Figure~\ref{fig:teaser}, Figure~\ref{fig:editing} and the Supplemental material for examples of path traced renders under 3 different lighting environments.

\section{Results}
\label{sec:results}
\begin{figure*}[t!]
 \contourlength{0.1em}%
 \centering
 \hspace*{-4.5mm}%
  \begin{tabular}{l@{\;}c@{\;}c@{\;}c@{\;}c@{\;}c}
 & \textsc{Subject A} & \textsc{Subject C} & \textsc{Subject D} & \textsc{Subject E} & \textsc{Subject F}
  \\
    \begin{sideways}\hspace{0.6cm}\textsc{Input albedo}\end{sideways}
  &
  \includegraphics[width=0.195\textwidth]{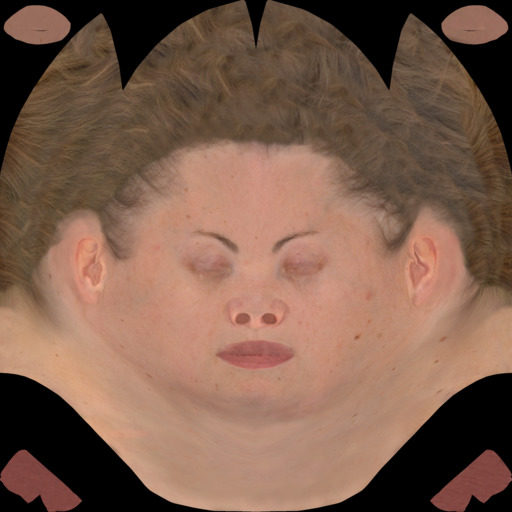}
  &
  \includegraphics[width=0.195\textwidth]{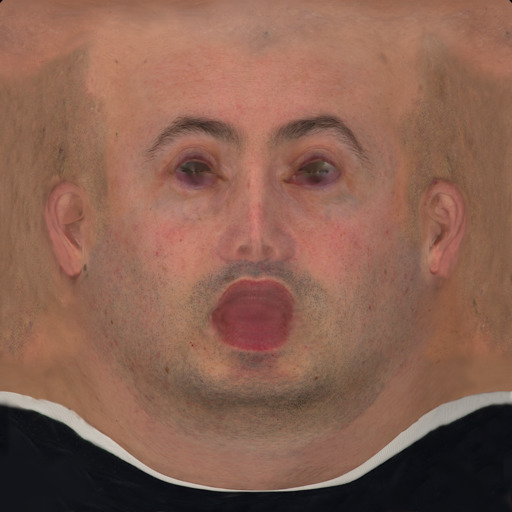}
  &
  \includegraphics[width=0.195\textwidth]{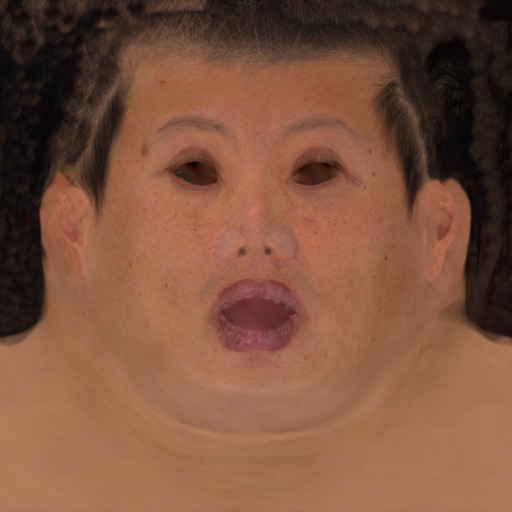}
  &
\includegraphics[width=0.195\textwidth]{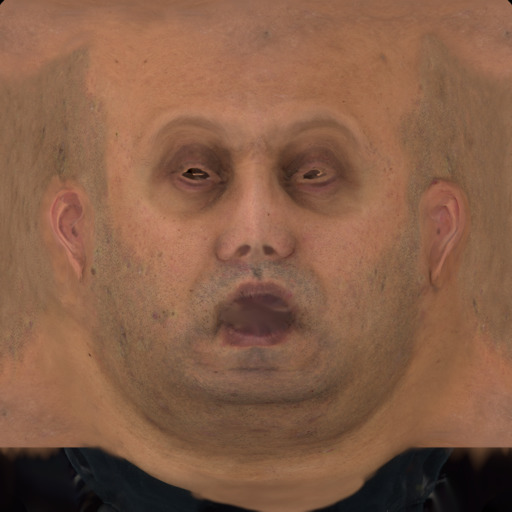}
  &
  \includegraphics[width=0.195\textwidth]{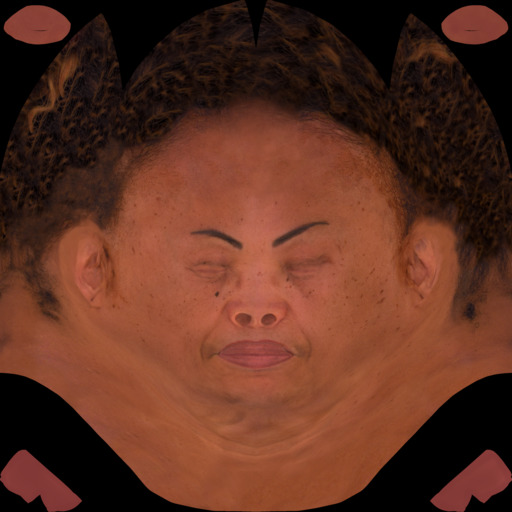}
   
  \\
  \begin{sideways}\hspace{1cm}\textsc{Neural}\end{sideways}
  &
  \includegraphics[width=0.195\textwidth]{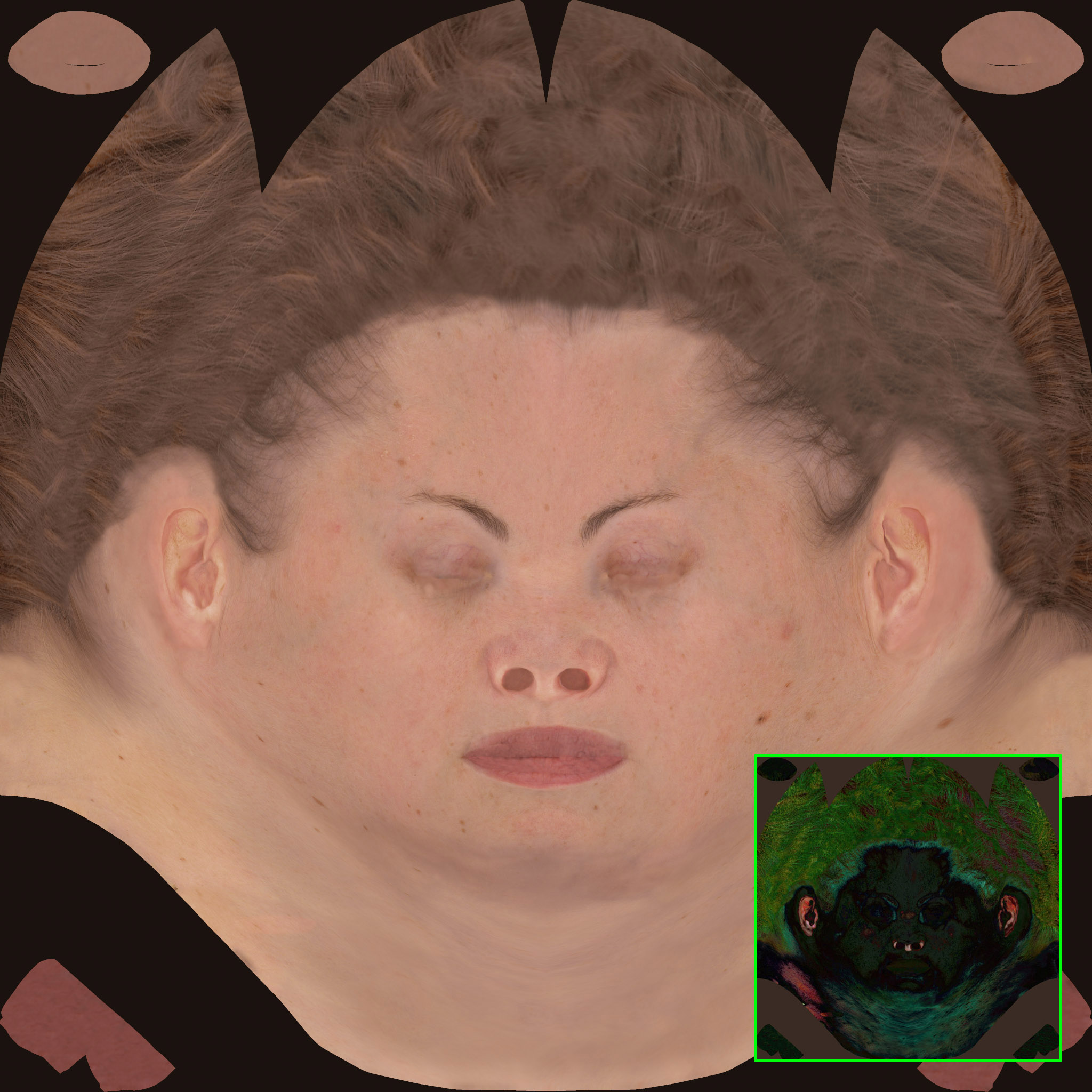}
  &
   \includegraphics[width=0.195\textwidth]{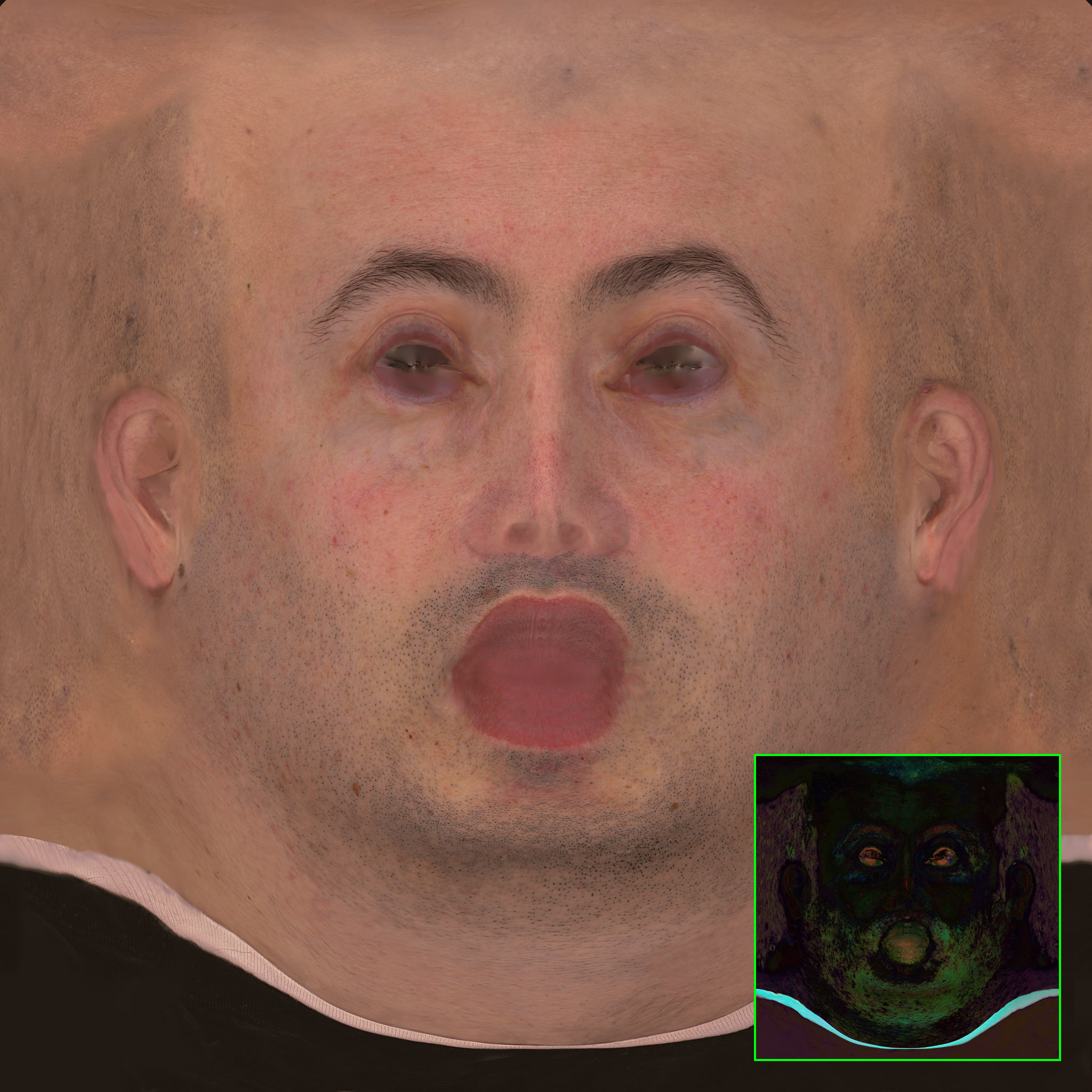}
  &
   \includegraphics[width=0.195\textwidth]{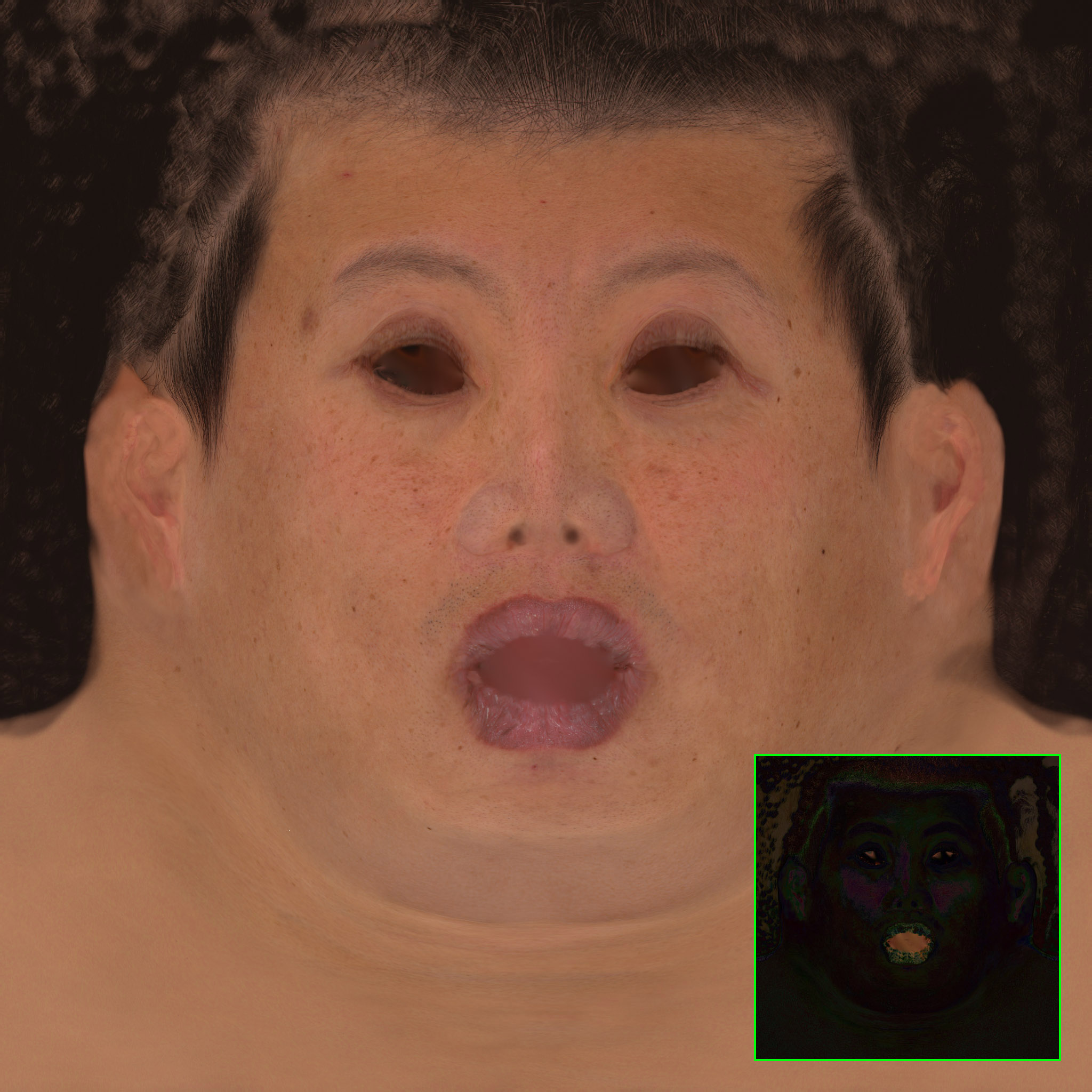}
  &
   \includegraphics[width=0.195\textwidth]{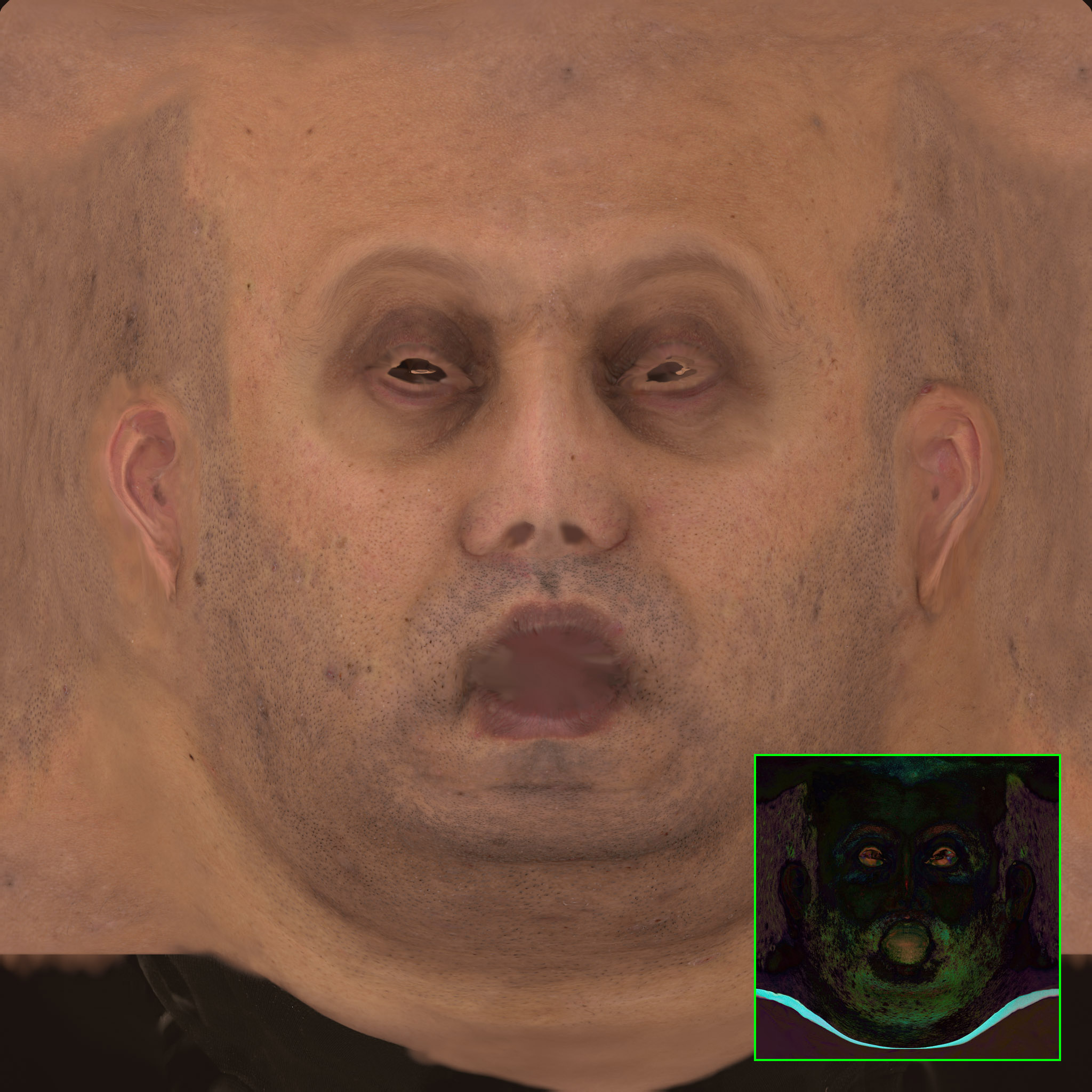}   
  &
   \includegraphics[width=0.195\textwidth]{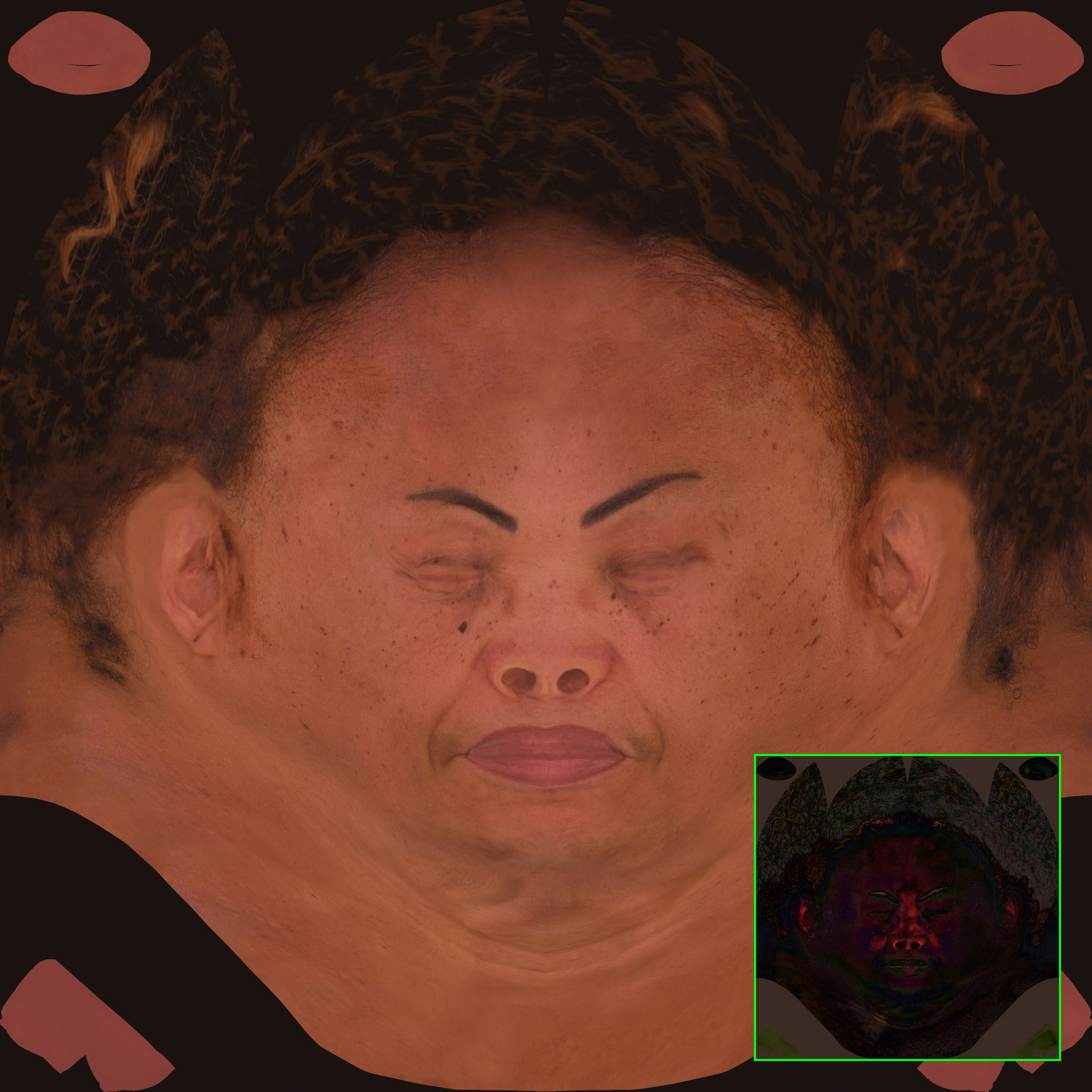}
    \\
  \begin{sideways}\hspace{1cm}\textsc{LUT}\end{sideways}
  &
  \includegraphics[width=0.195\textwidth]{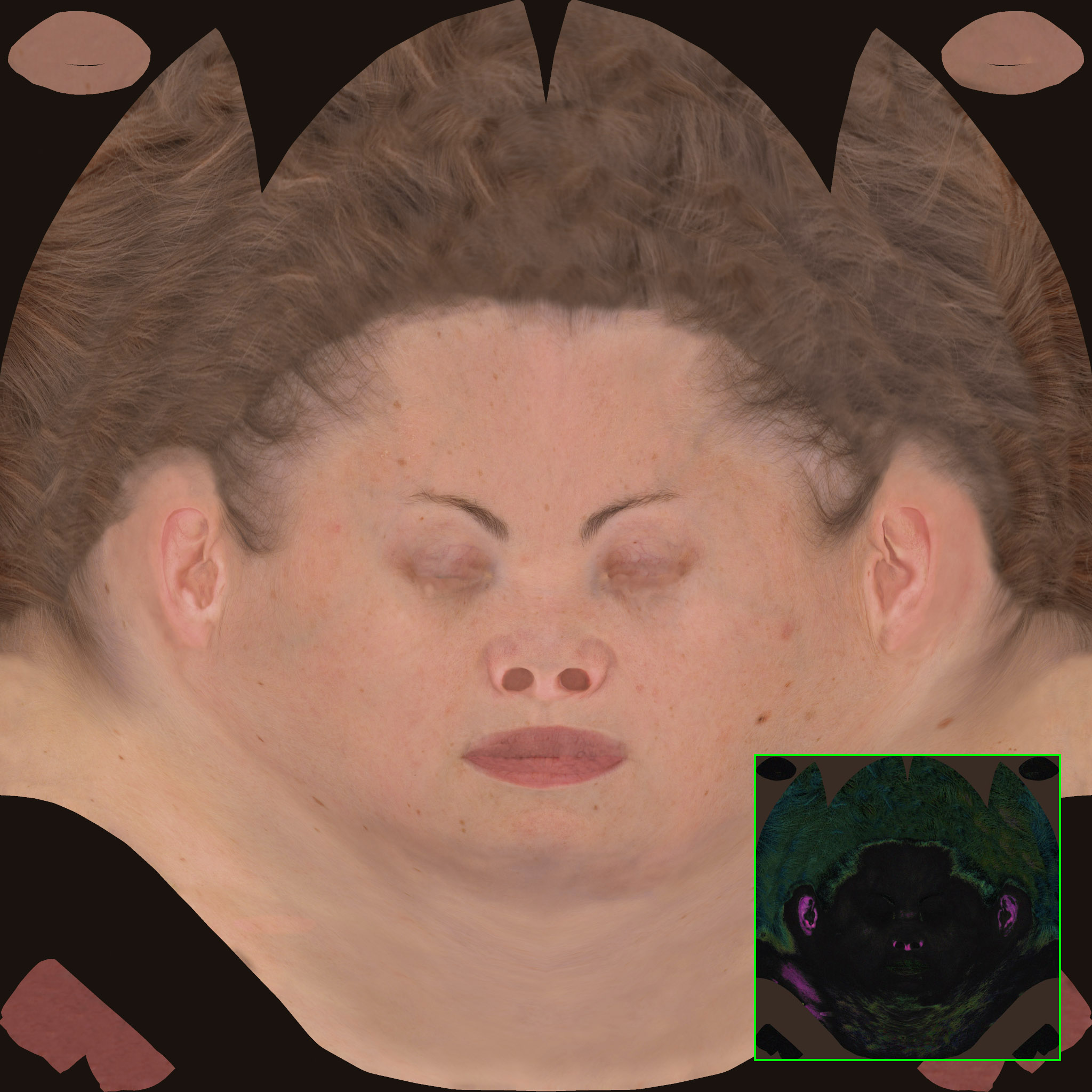}
  &
   \includegraphics[width=0.195\textwidth]{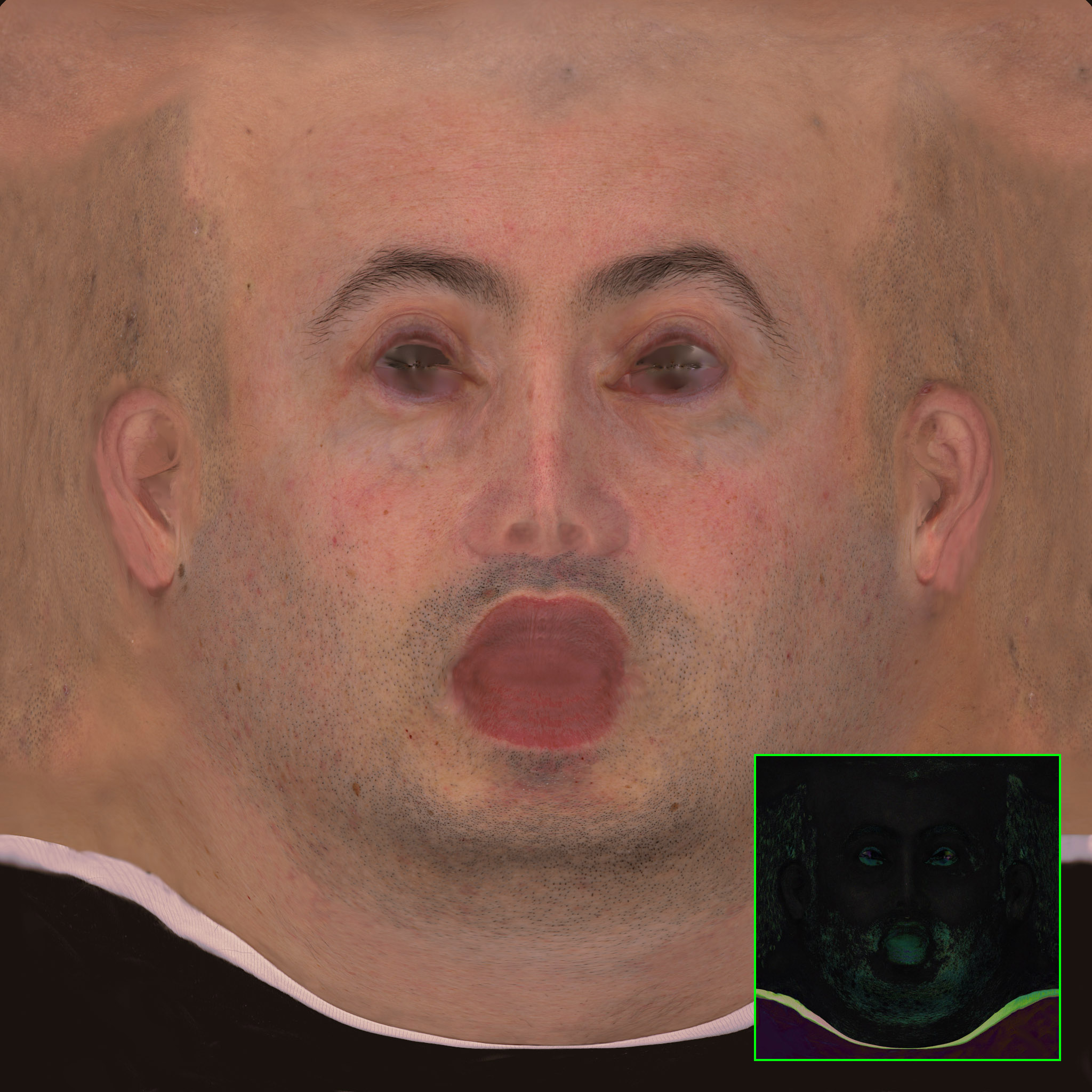}
  &
   \includegraphics[width=0.195\textwidth]{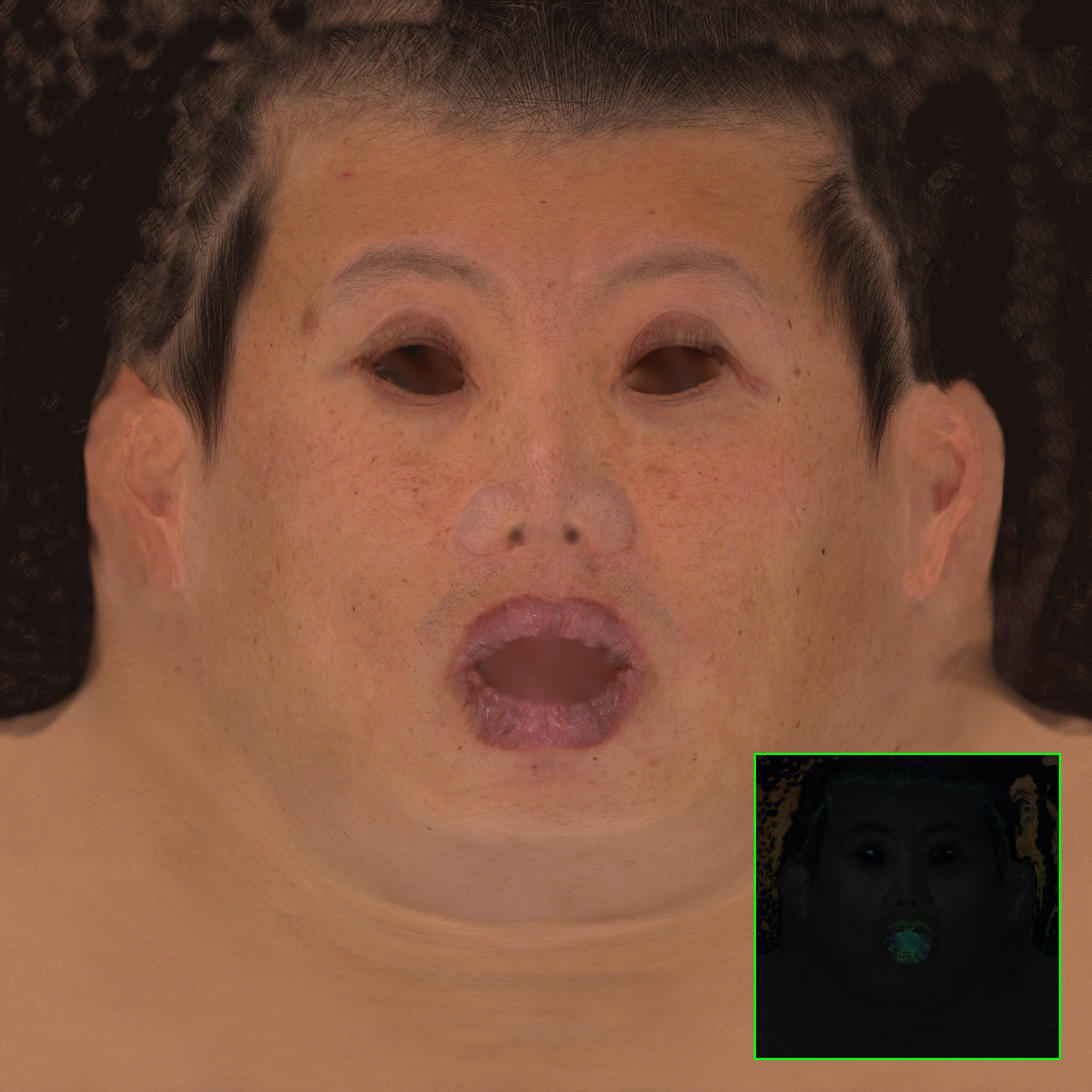}
  &
   \includegraphics[width=0.195\textwidth]{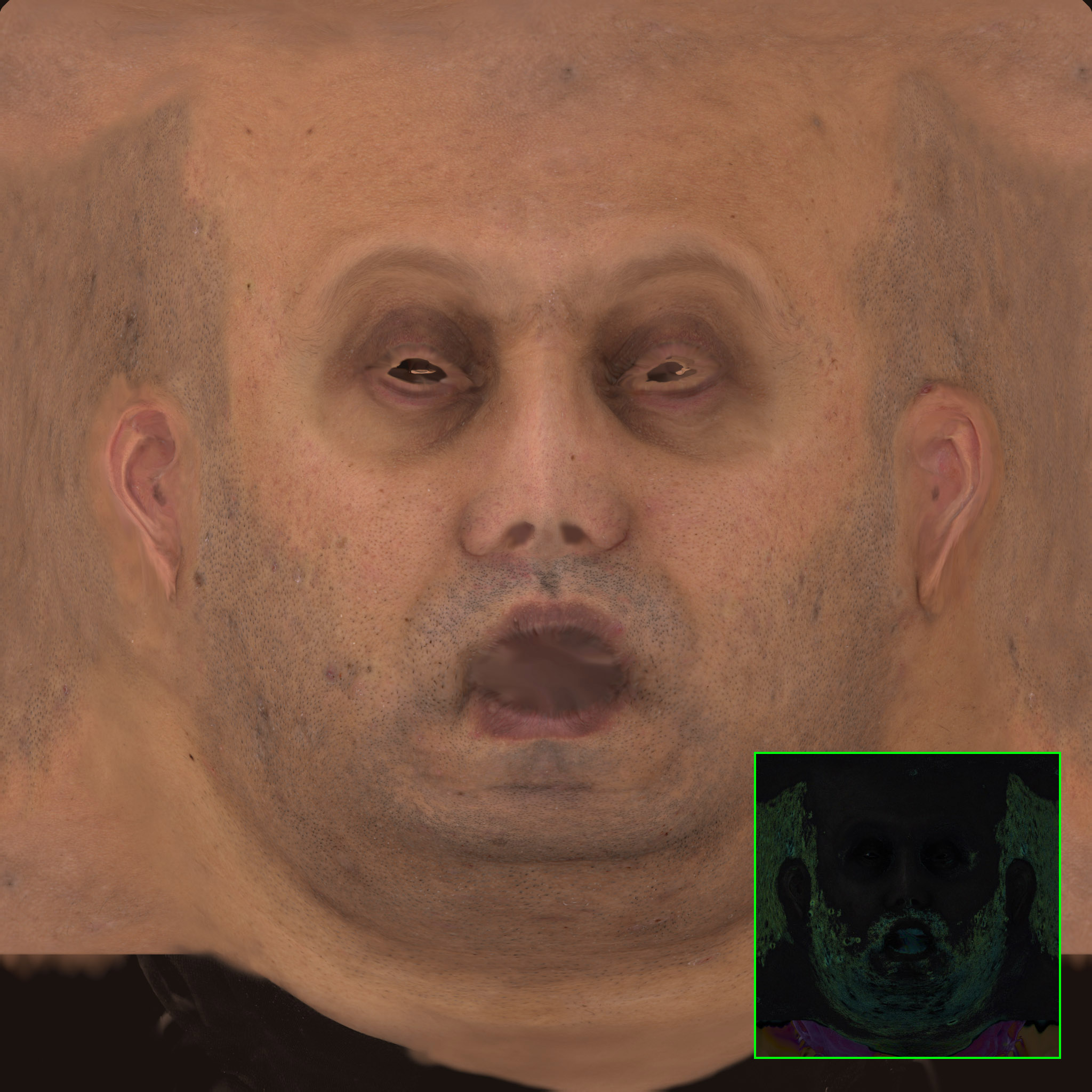}   
  &
   \includegraphics[width=0.195\textwidth]{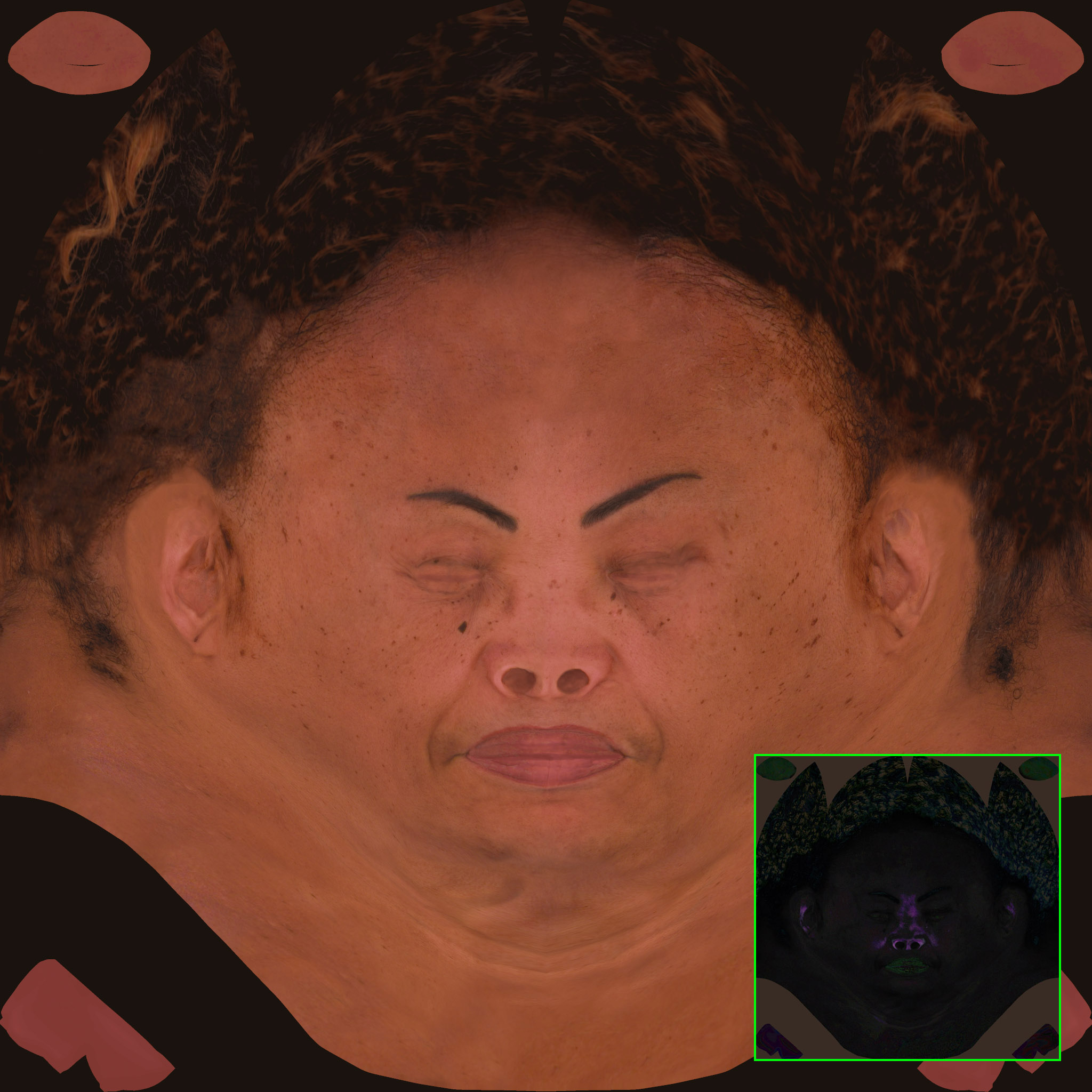}

\vspace*{3mm}
 \end{tabular}
 \vspace{-7mm}
  \caption{From top to bottom, captured albedos and reconstructions our Neural and LUT approaches for five subjects of different skin type (x4 absolute errors as insets). The learned mapping exhibits slightly higher reconstruction error than the discrete LUT approach. This was expected due to the inductive bias of the network, forcing smoothness in the recovered maps and thus enabling robust manipulation of the estimated properties.}
  \label{fig:albedoreconstruction}
  \vspace{-3mm}
\end{figure*}

\begin{figure*}[t!]
 \contourlength{0.1em}%
 \centering
 \hspace*{-4.5mm}%
  \begin{tabular}{l@{\;}c@{\;}c@{\;}c@{\;}c@{\;}c}
 & \textsc{Melanin} & \textsc{Hemoglobin} & \textsc{Thickness} & \textsc{Melanin ratio} & \textsc{Blood Oxygenation}
  \\
    \begin{sideways}\hspace{0.6cm}\textsc{Subject A}\end{sideways}
  &
  \includegraphics[trim= 132 60 73 38, clip, width =  0.2\textwidth]{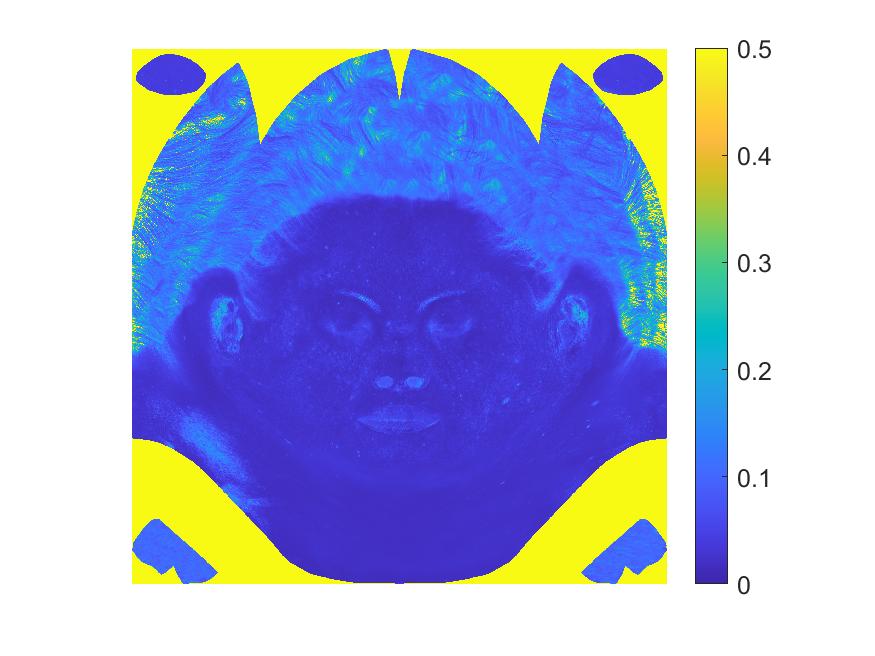}
  &
  \includegraphics[trim= 132 60 73 38, clip, width =  0.2\textwidth]{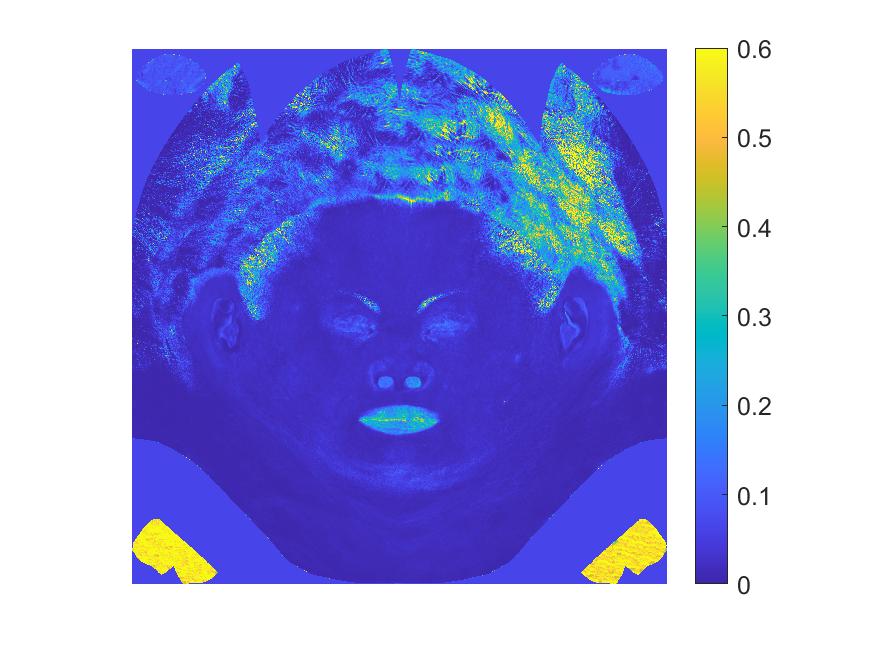}
    &
  \includegraphics[trim= 132 60 73 38, clip, width =  0.2\textwidth]{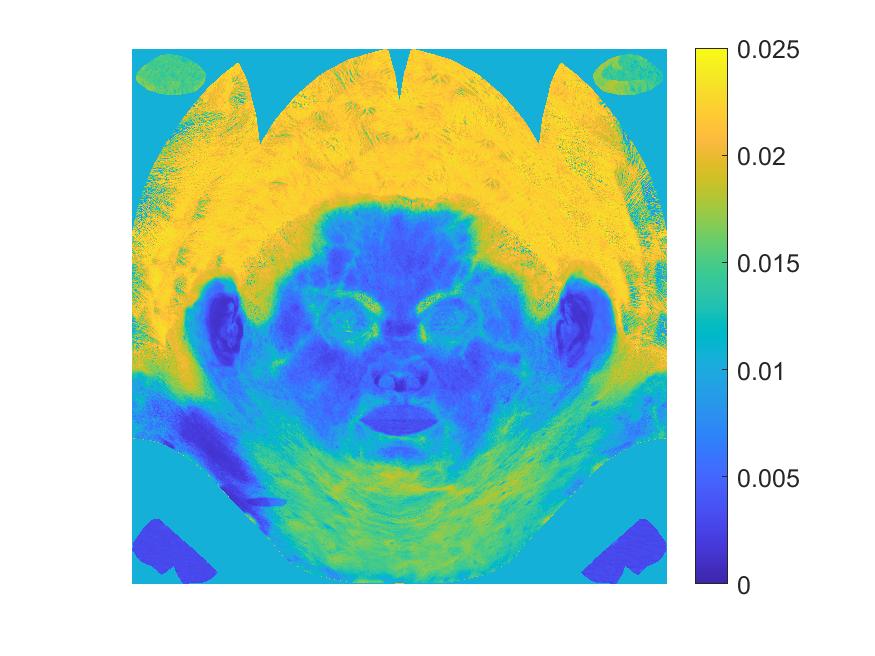}
    &
  \includegraphics[trim= 132 60 73 38, clip, width =  0.2\textwidth]{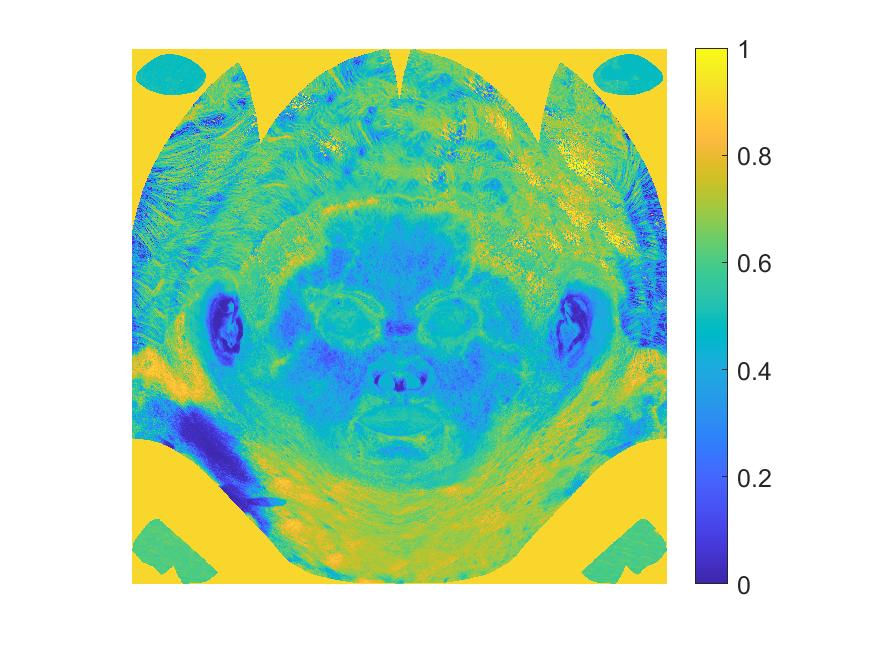}
    &
  \includegraphics[trim= 132 60 73 38, clip, width =  0.2\textwidth]{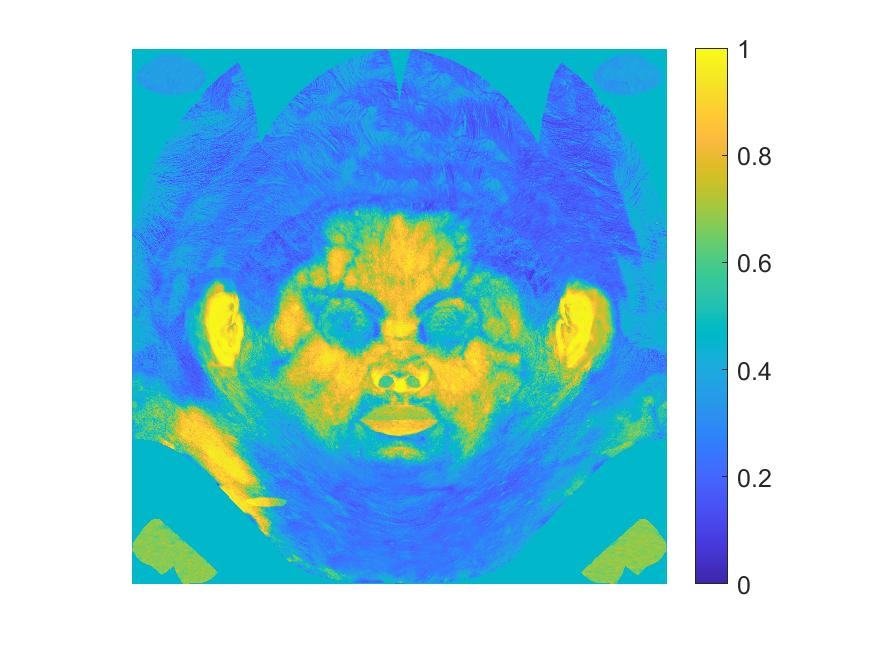}
  \\
    \begin{sideways}\hspace{0.6cm}\textsc{Subject C}\end{sideways}
  &
  \includegraphics[trim= 132 60 73 38, clip, width =  0.2\textwidth]{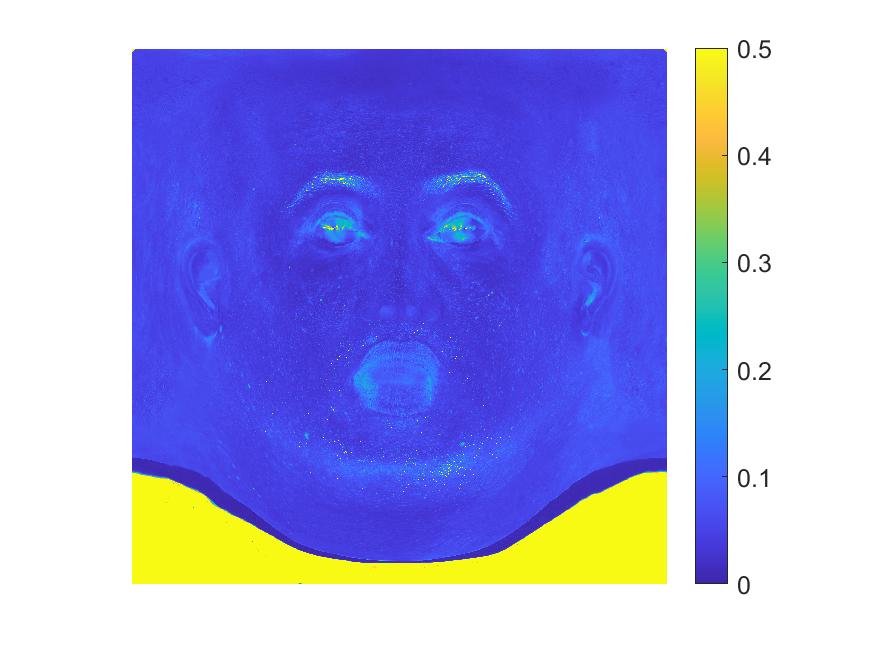}
  &
  \includegraphics[trim= 132 60 73 38, clip, width =  0.2\textwidth]{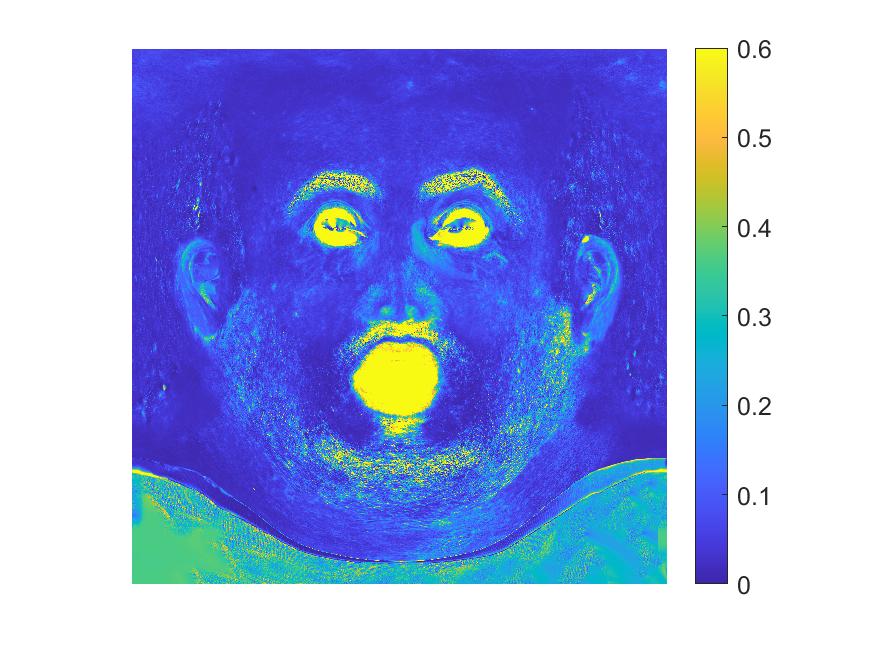}
    &
  \includegraphics[trim= 132 60 73 38, clip, width =  0.2\textwidth]{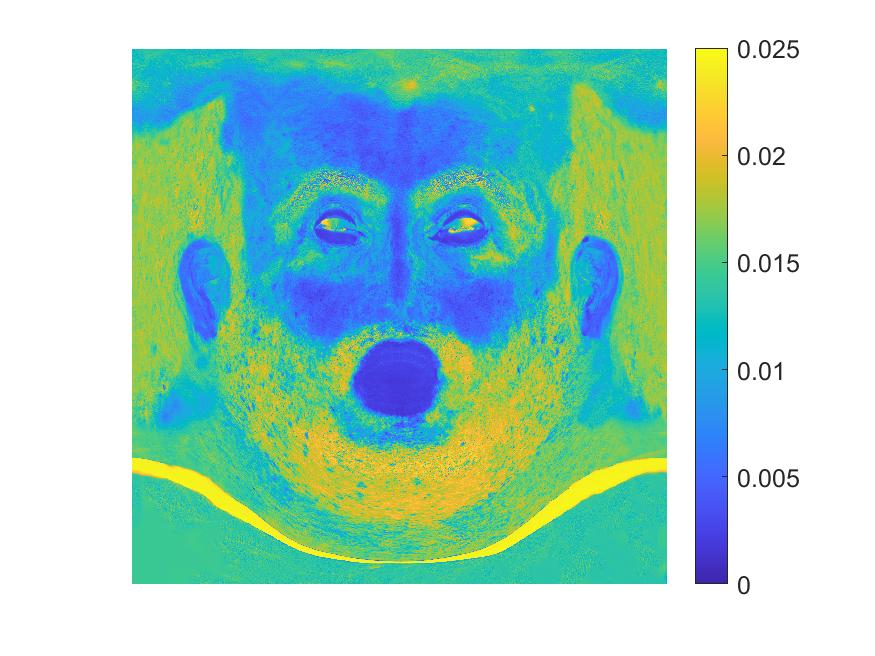}
    &
  \includegraphics[trim= 132 60 73 38, clip, width =  0.2\textwidth]{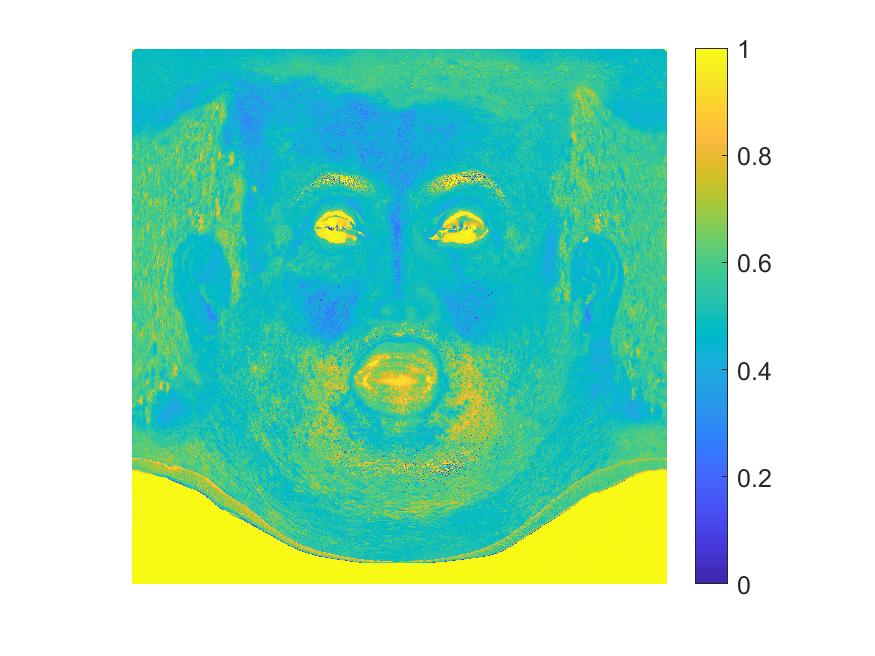}
    &
  \includegraphics[trim= 132 60 73 38, clip, width =  0.2\textwidth]{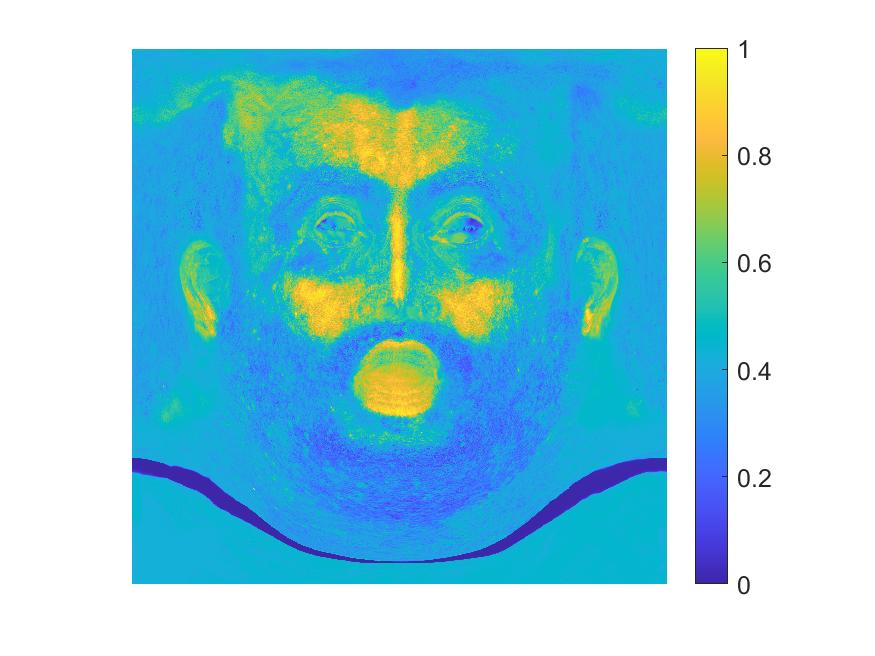}
  \\
    \begin{sideways}\hspace{0.6cm}\textsc{Subject D}\end{sideways}
  &
  \includegraphics[trim= 132 60 73 38, clip, width =  0.2\textwidth]{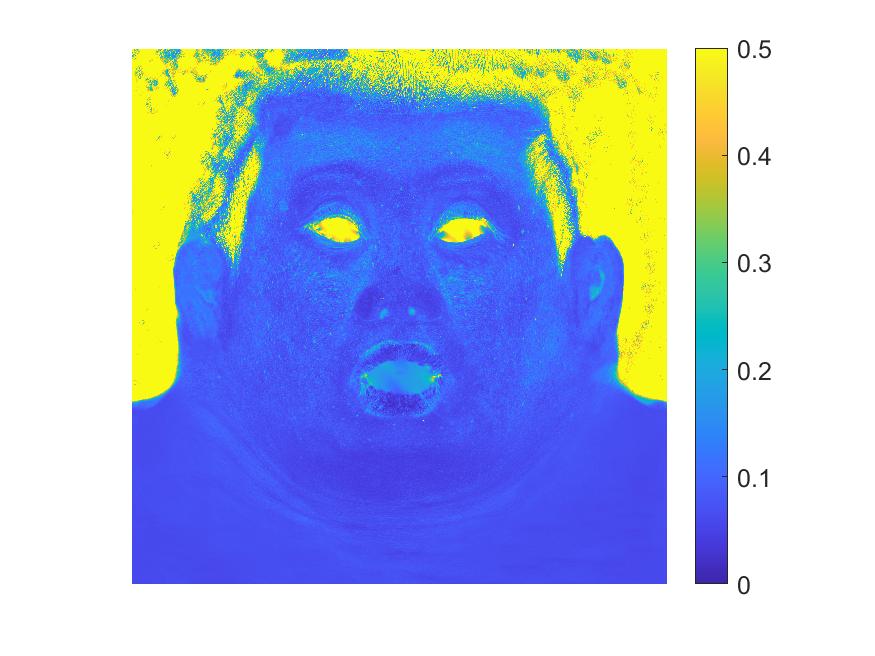}
  &
  \includegraphics[trim= 132 60 73 38, clip, width =  0.2\textwidth]{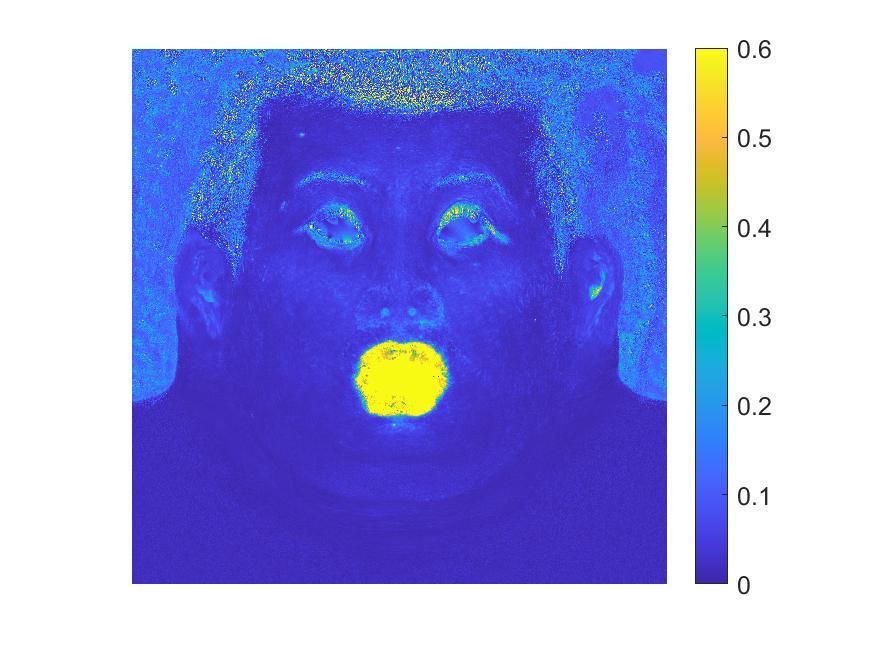}
    &
  \includegraphics[trim= 132 60 73 38, clip, width =  0.2\textwidth]{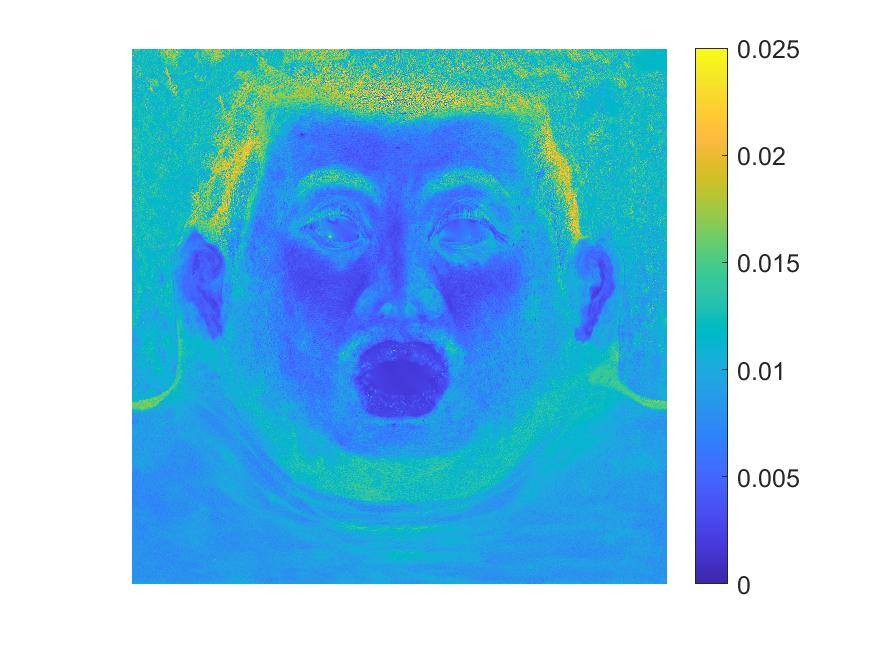}
    &
  \includegraphics[trim= 132 60 73 38, clip, width =  0.2\textwidth]{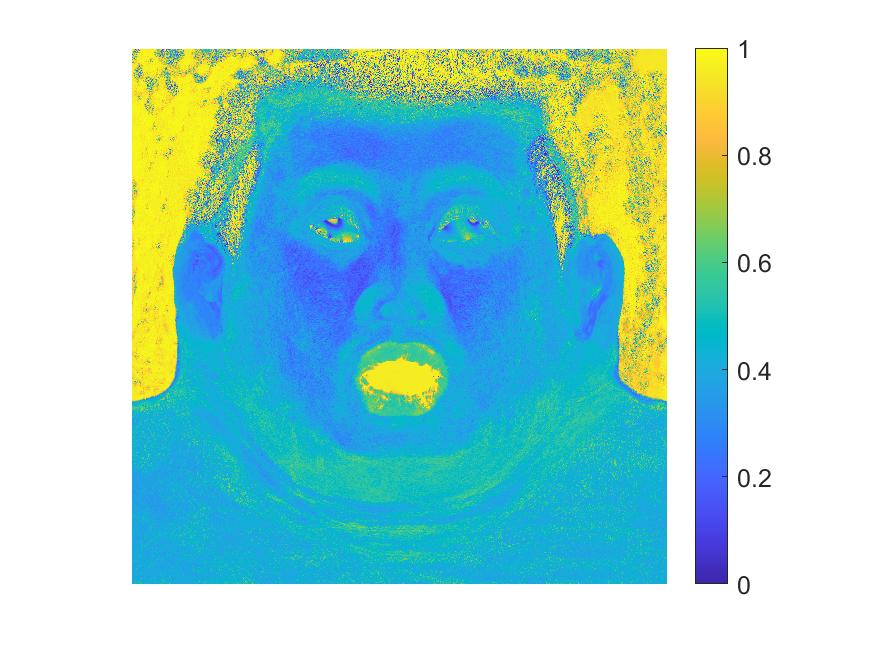}
    &
  \includegraphics[trim= 132 60 73 38, clip, width =  0.2\textwidth]{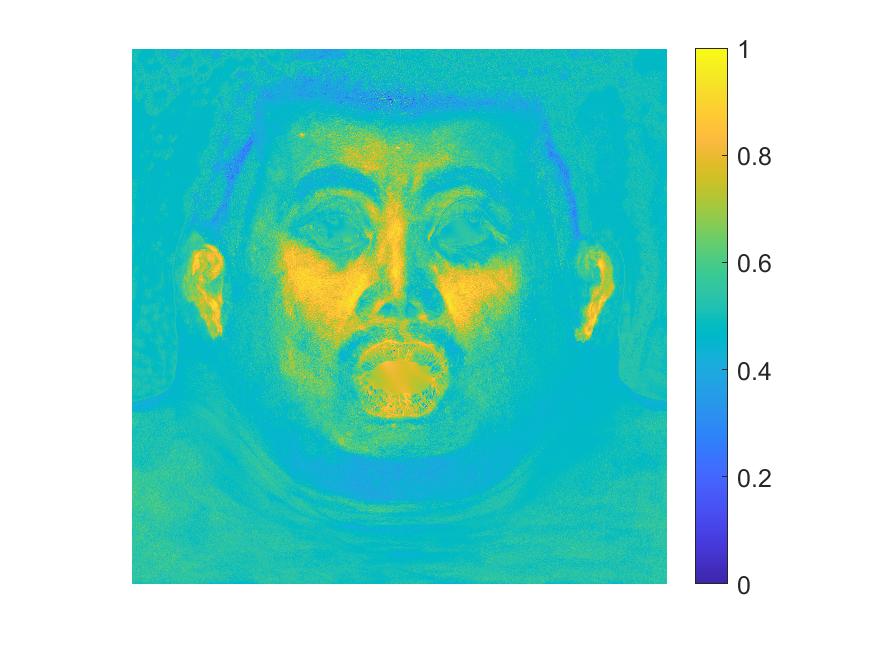}
    \\
    \begin{sideways}\hspace{0.6cm}\textsc{Subject E}\end{sideways}
  &
  \includegraphics[trim= 132 60 73 38, clip, width =  0.2\textwidth]{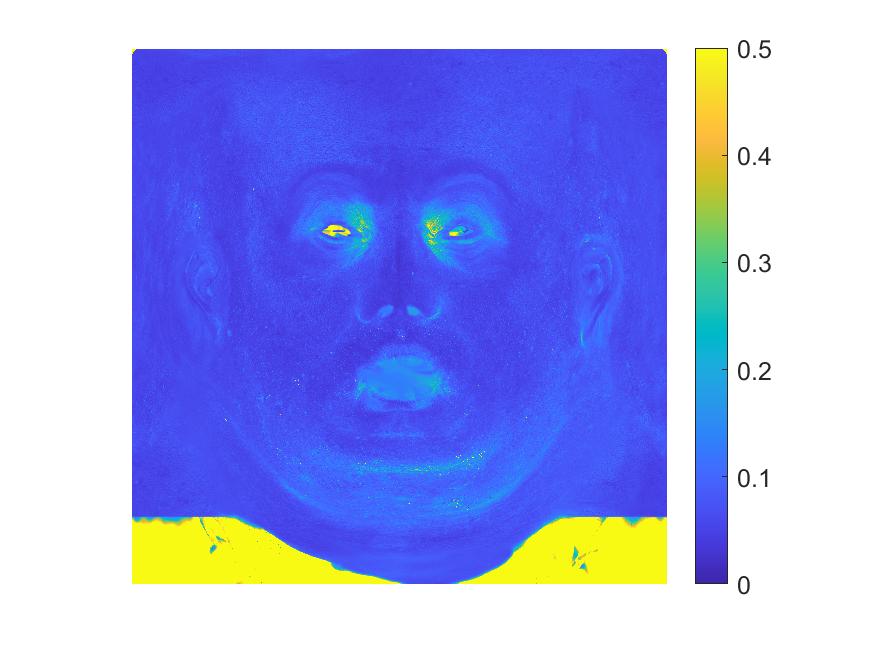}
  &
  \includegraphics[trim= 132 60 73 38, clip, width =  0.2\textwidth]{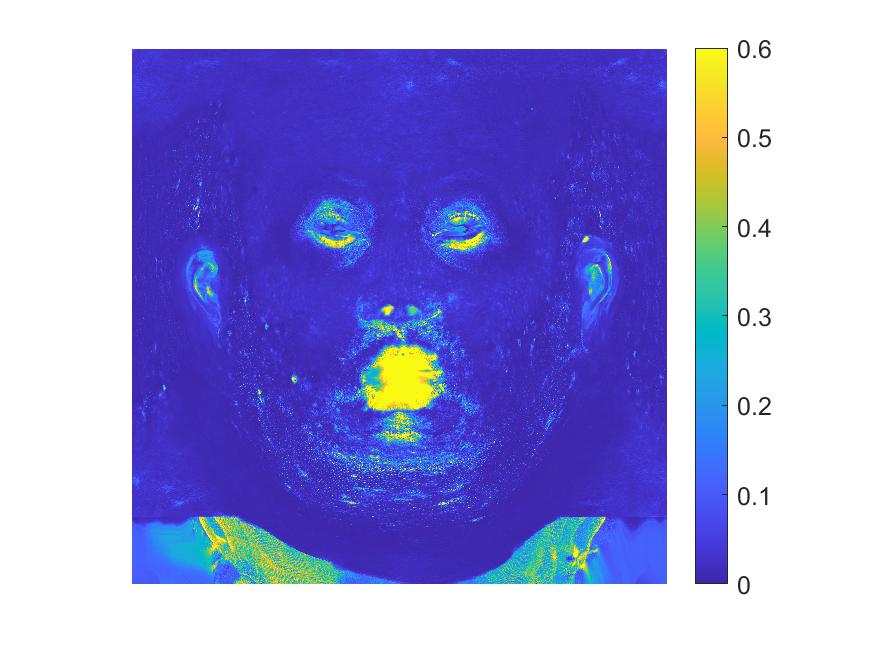}
    &
  \includegraphics[trim= 132 60 73 38, clip, width =  0.2\textwidth]{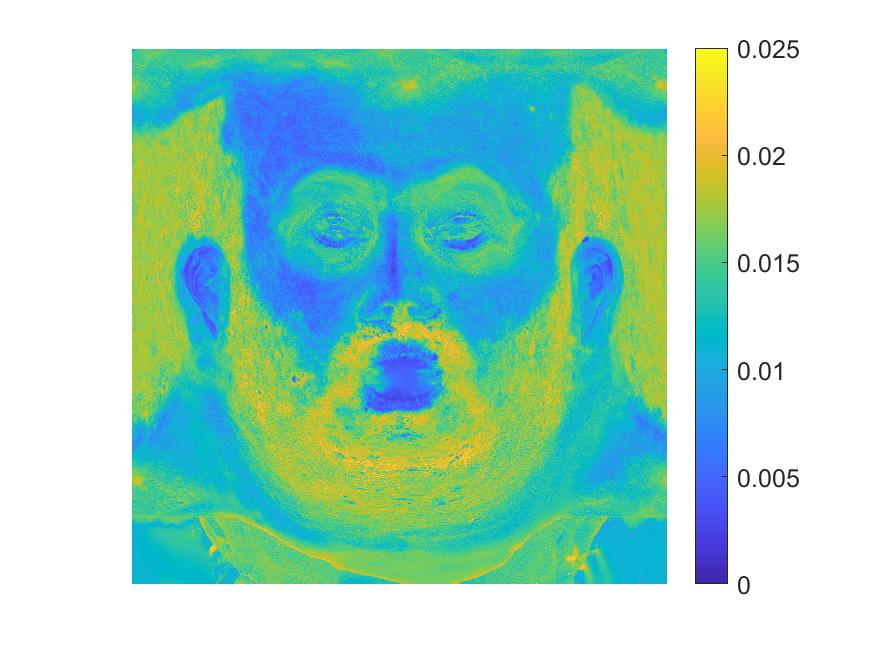}
    &
  \includegraphics[trim= 132 60 73 38, clip, width =  0.2\textwidth]{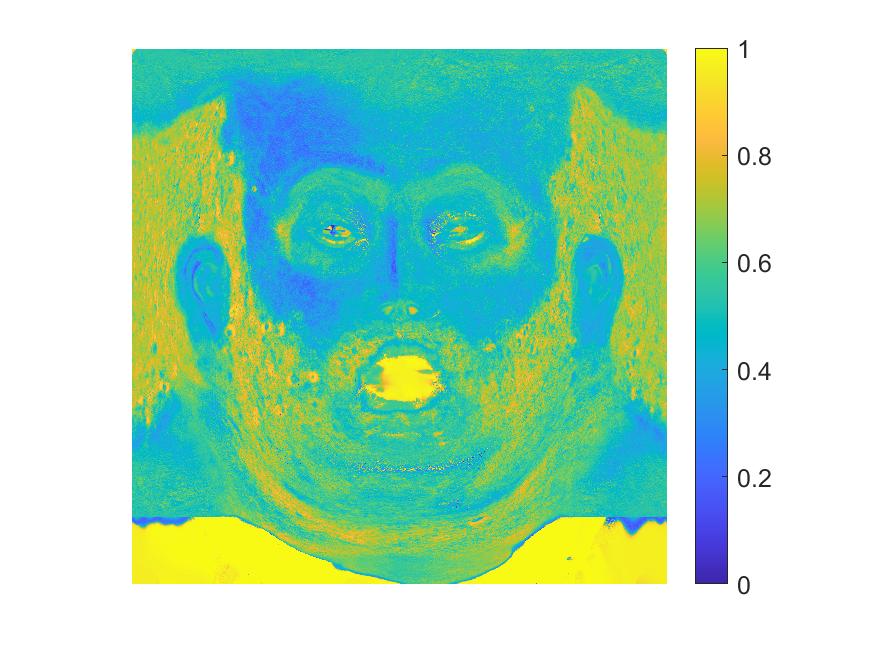}
    &
  \includegraphics[trim= 132 60 73 38, clip, width =  0.2\textwidth]{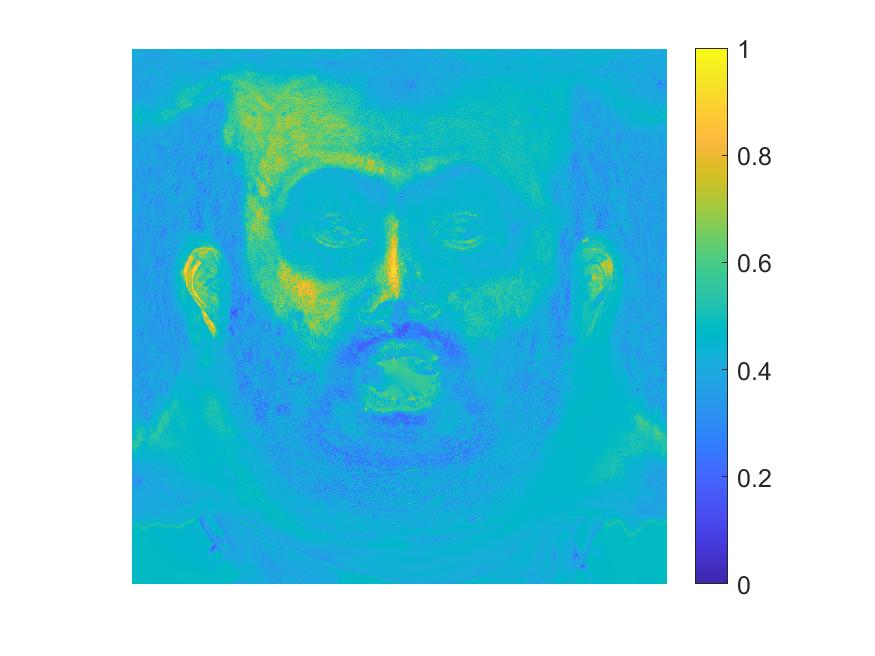}
      \\
    \begin{sideways}\hspace{0.6cm}\textsc{Subject F}\end{sideways}
  &
  \includegraphics[trim= 132 60 73 38, clip, width =  0.2\textwidth]{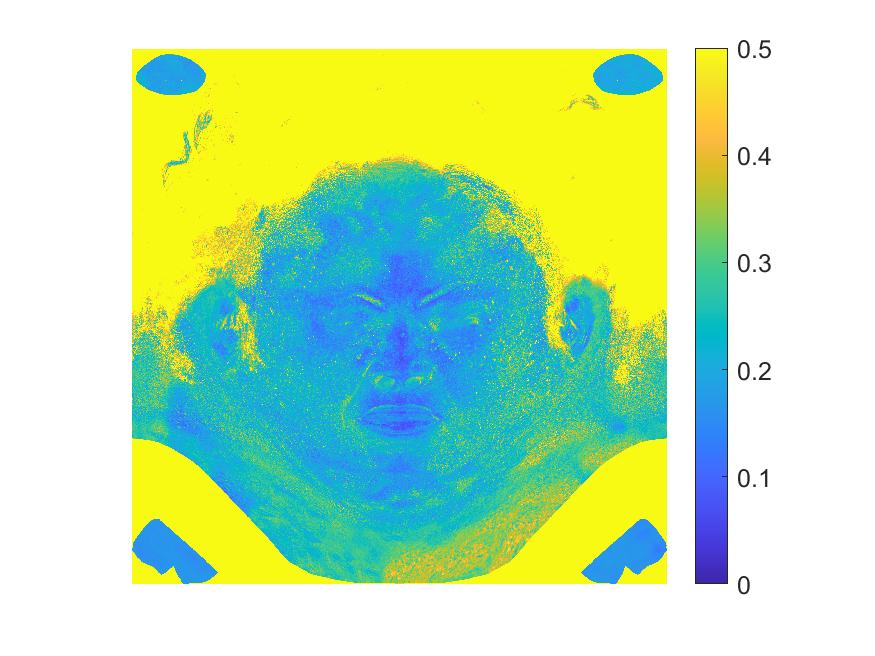}
  &
  \includegraphics[trim= 132 60 73 38, clip, width =  0.2\textwidth]{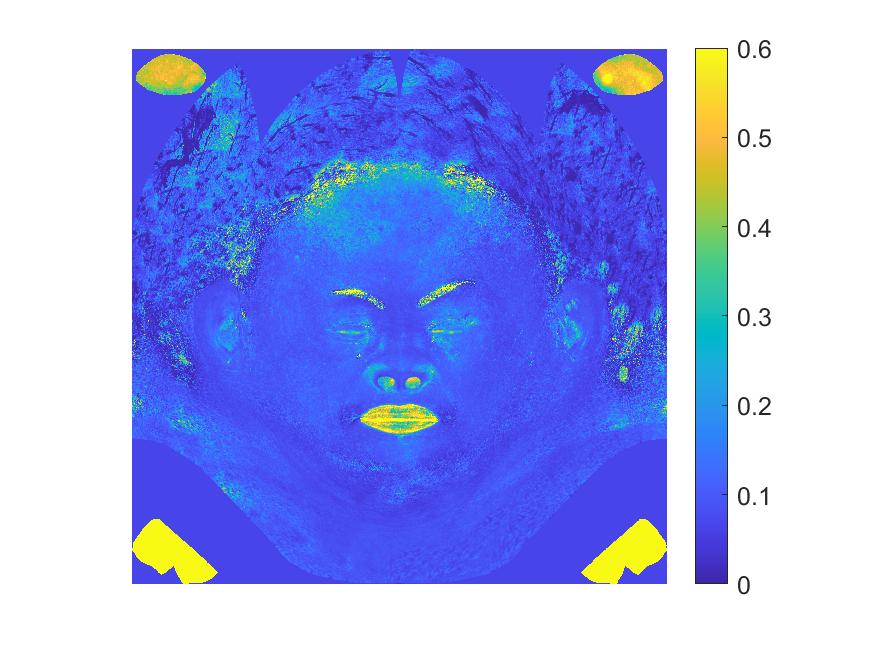}
    &
  \includegraphics[trim= 132 60 73 38, clip, width =  0.2\textwidth]{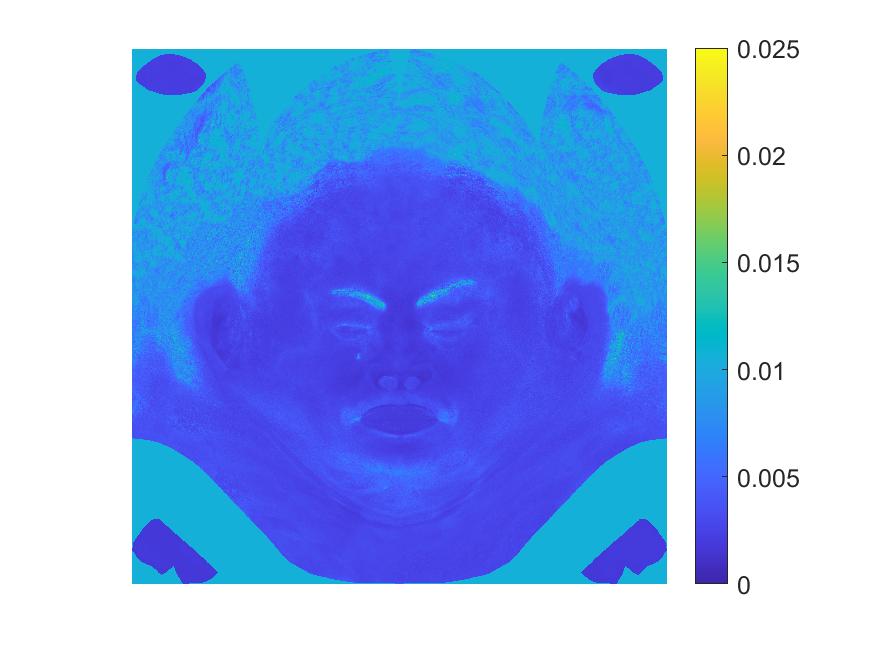}
    &
  \includegraphics[trim= 132 60 73 38, clip, width =  0.2\textwidth]{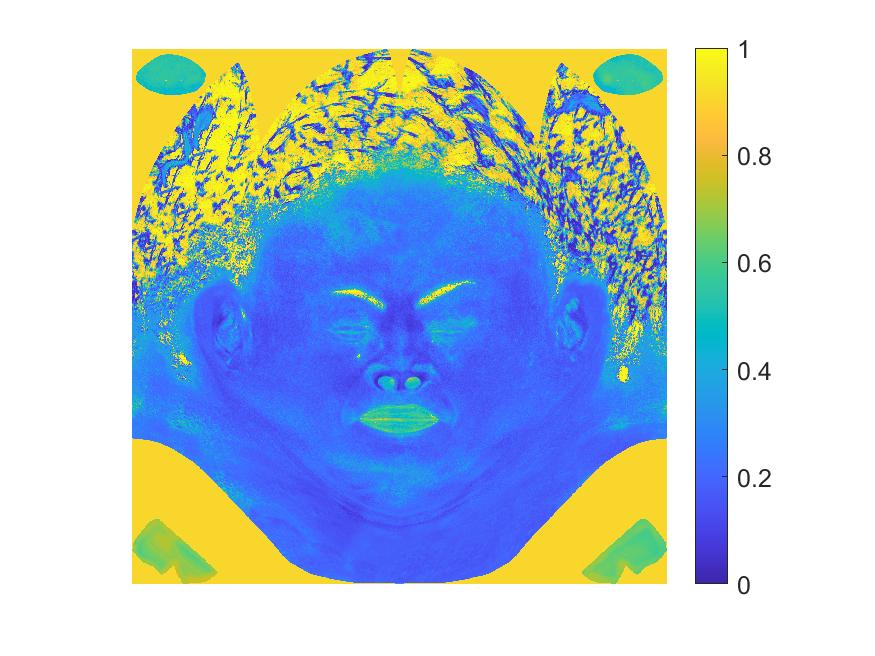}
    &
  \includegraphics[trim= 132 60 73 38, clip, width =  0.2\textwidth]{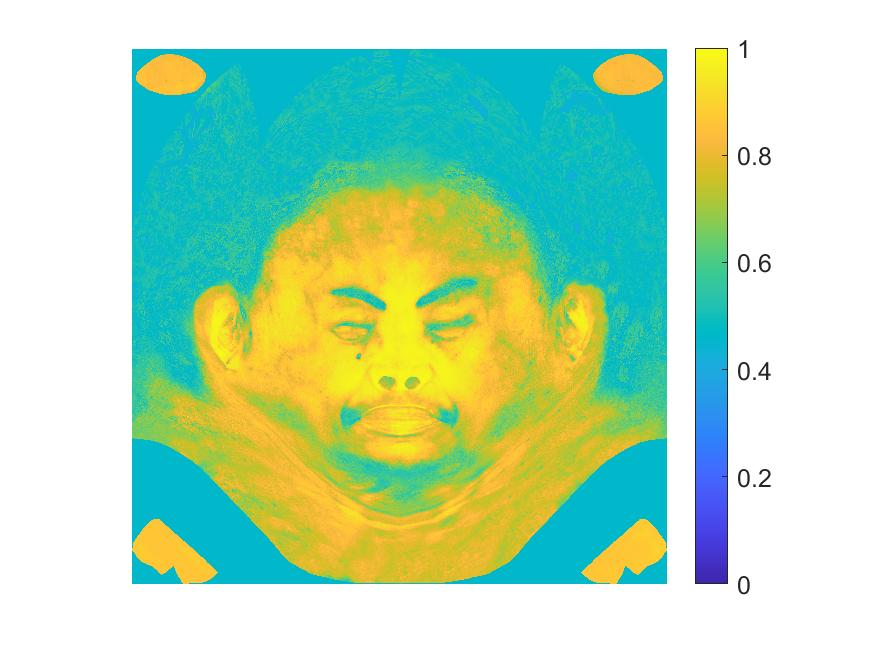}
 \end{tabular}
 \vspace{-3mm}
  \caption{\emph{Estimated skin parameters}. The overall melanin concentration is coherent to the skin type of each subject. Note how melanin spots are nicely isolated from small capillars, veins or reddish imperfections. Also, the color of the lips is mostly due to the high concentration of blood and the relatively thin epidermis in all cases.}
  \label{fig:neuralparam}
  \vspace{-6mm}
\end{figure*}

\begin{figure*}[t!]
 \contourlength{0.1em}%
 \centering
 \hspace*{-4.5mm}%
  \begin{tabular}{l@{\;}c@{\;}c@{\;}c@{\;}c@{\;}c@{\;}c@{\;}c@{\;}c}
  
 & & \textsc{0.125x blood } & \textsc{0.5x blood} & \textsc{3x blood} & \textsc{1.5x melanin} & \textsc{3x melanin} & \textsc{5x melanin} 
  \\
  \begin{sideways}\hspace{0.2cm}\textsc{Type I (Subject A)}\end{sideways}
 &
  \includegraphics[trim= 0 340 0 0, clip, width =0.14\textwidth]{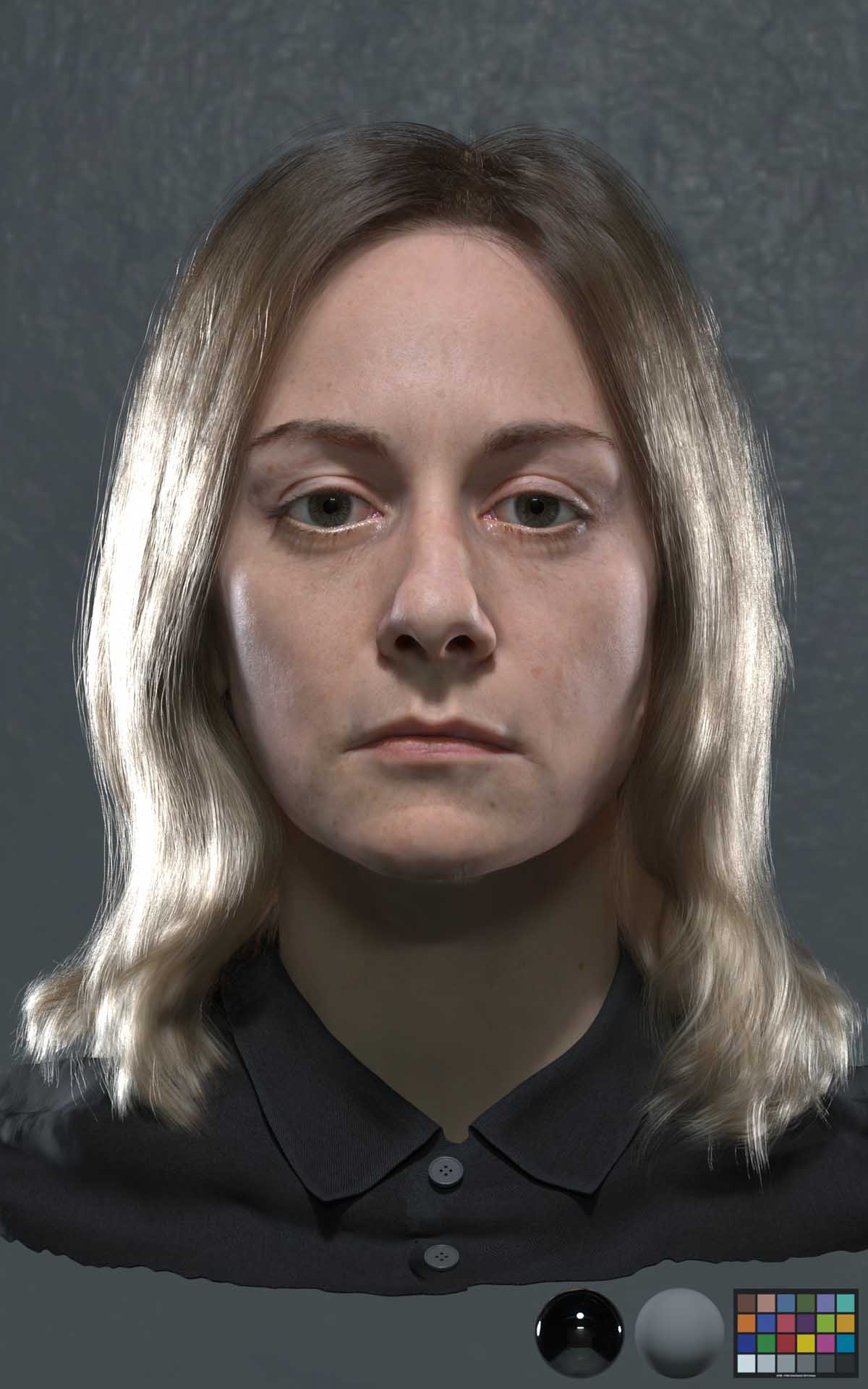}
 &
   \includegraphics[trim= 0 340 0 0, clip, width =0.14\textwidth]{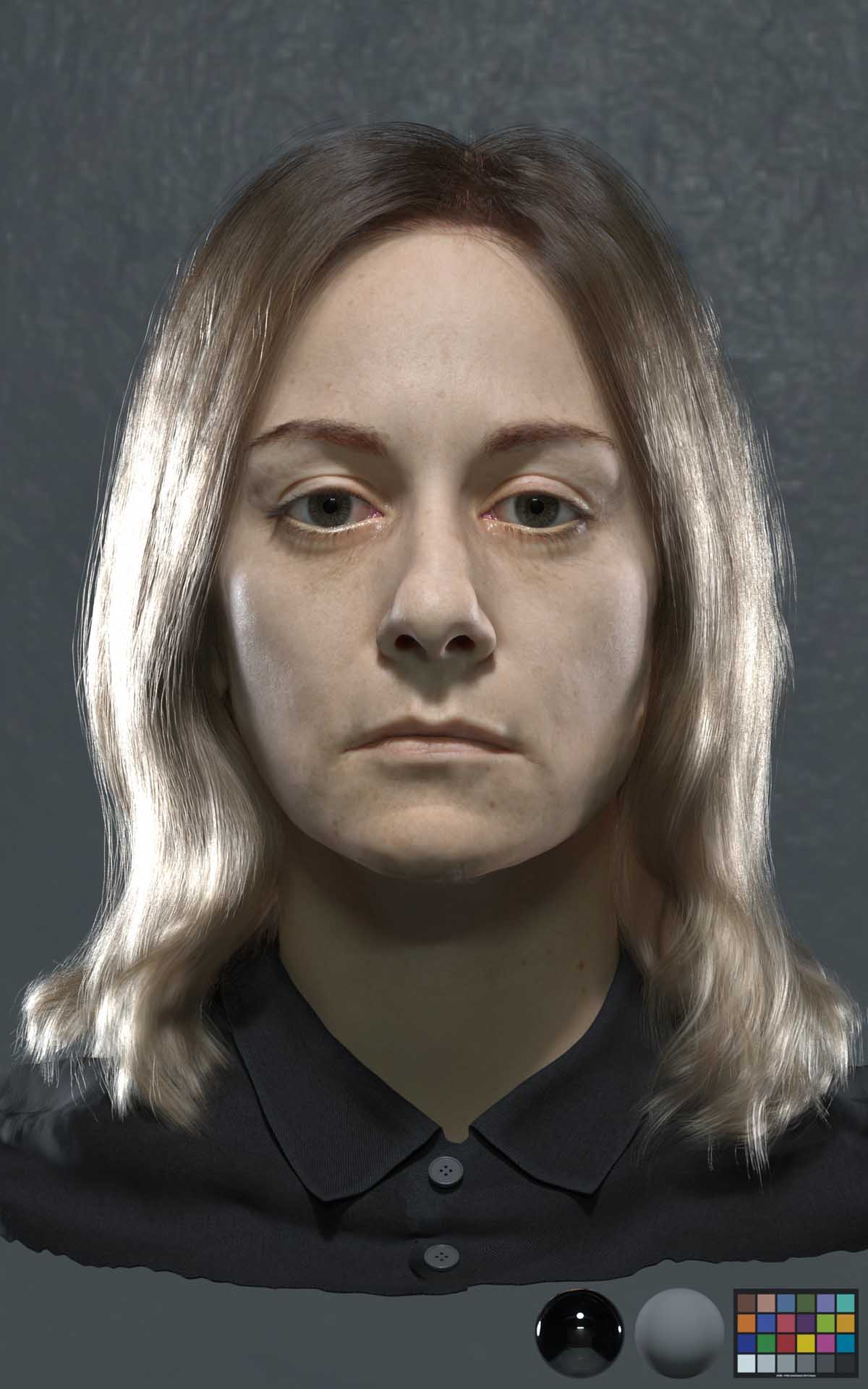}
 &
   \includegraphics[trim= 0 340 0 0, clip, width =0.14\textwidth]{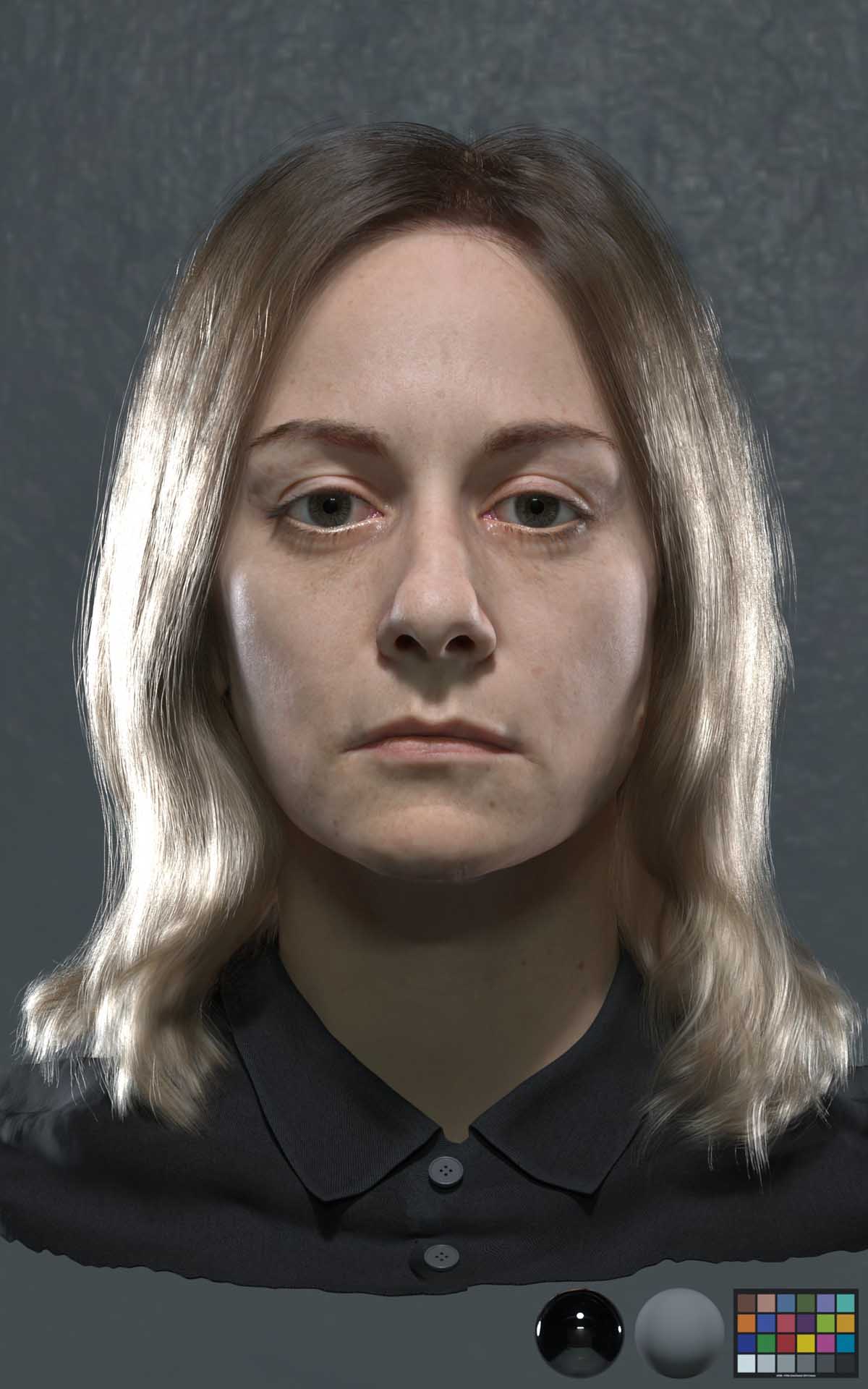}
 &
   \includegraphics[trim= 0 340 0 0, clip, width =0.14\textwidth]{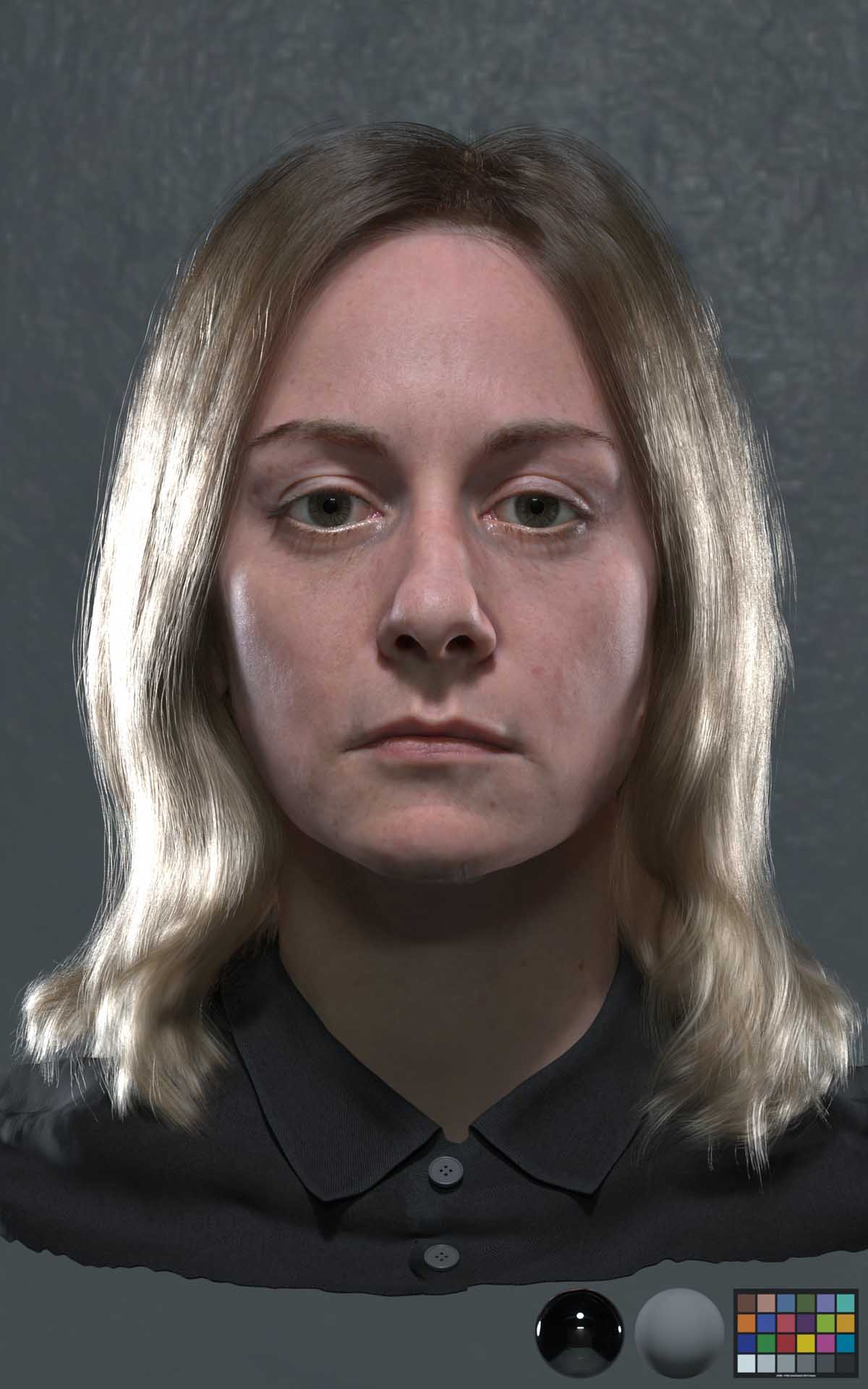}
 &
   \includegraphics[trim= 0 340 0 0, clip, width =0.14\textwidth]{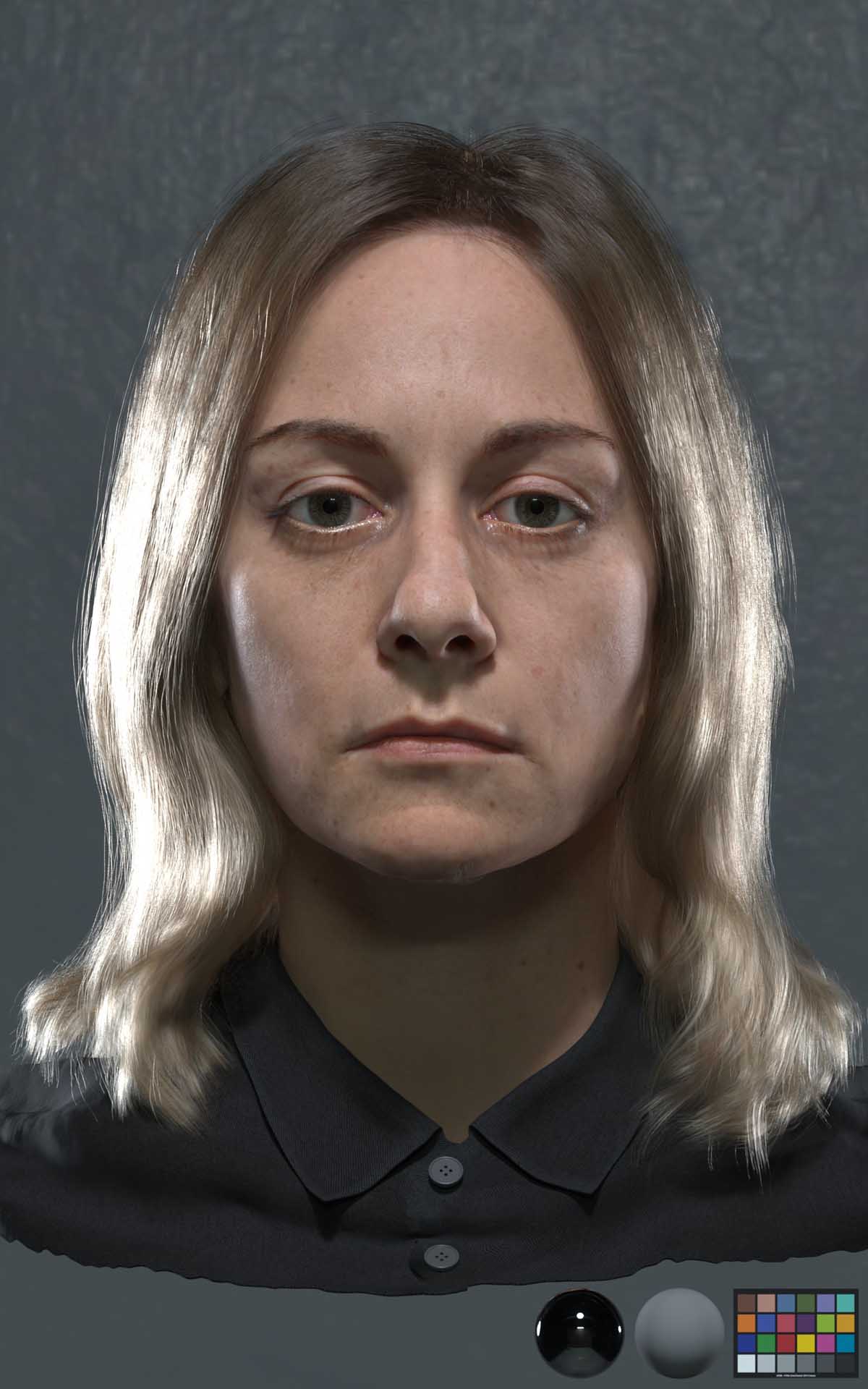}
 &
   \includegraphics[trim= 0 340 0 0, clip, width =0.14\textwidth]{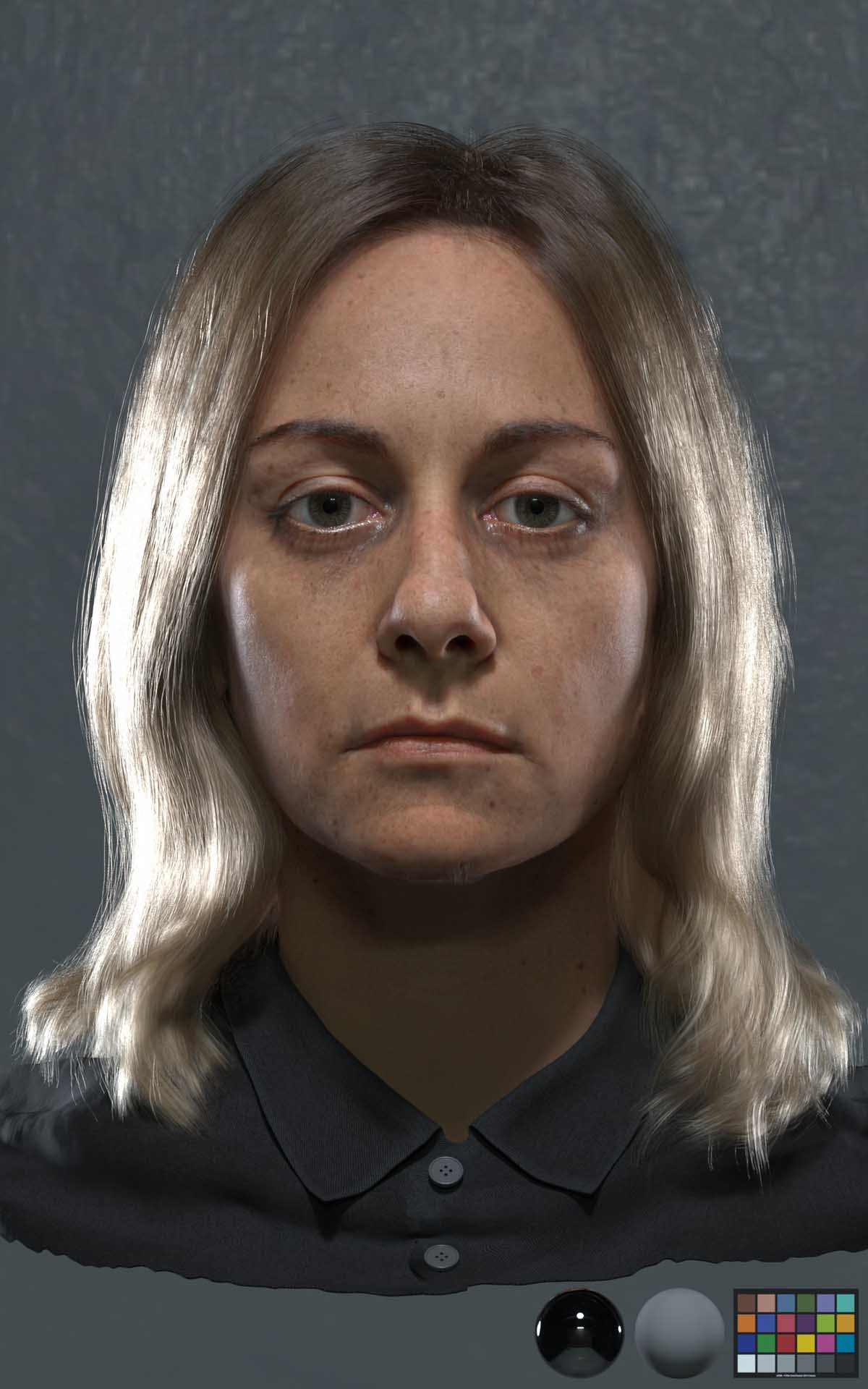}
 &
   \includegraphics[trim= 0 340 0 0, clip, width =0.14\textwidth]{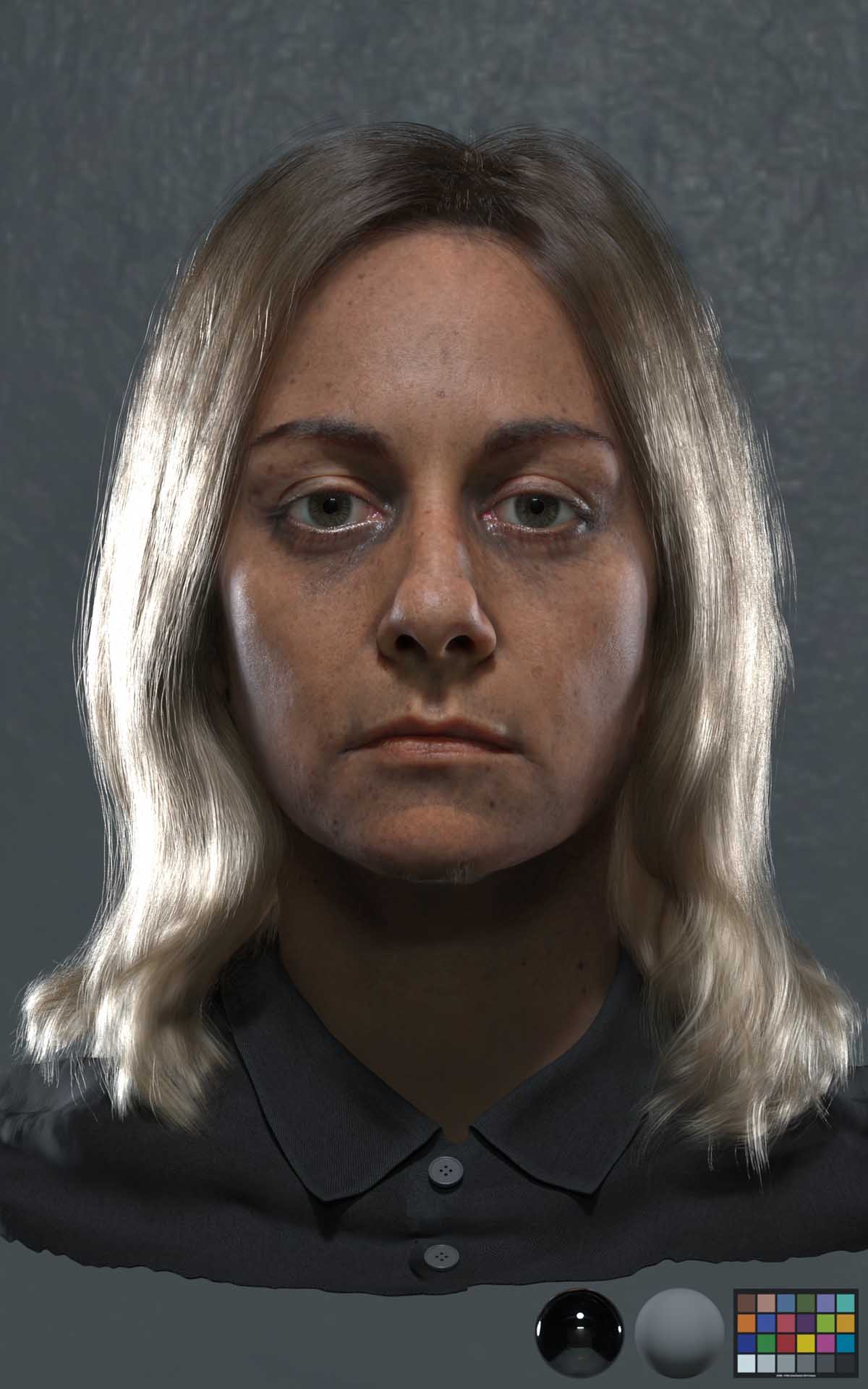}
%
  
  \\
 & & \textsc{0.125x blood } & \textsc{0.5x blood} & \textsc{3x blood} & \textsc{1.5x melanin} & \textsc{3x melanin} & \textsc{5x melanin} 
  \\
  \begin{sideways}\hspace{0.2cm}\textsc{Type II (Subject C)}\end{sideways}
 &
  \includegraphics[trim= 0 340 0 0, clip, width =0.14\textwidth]{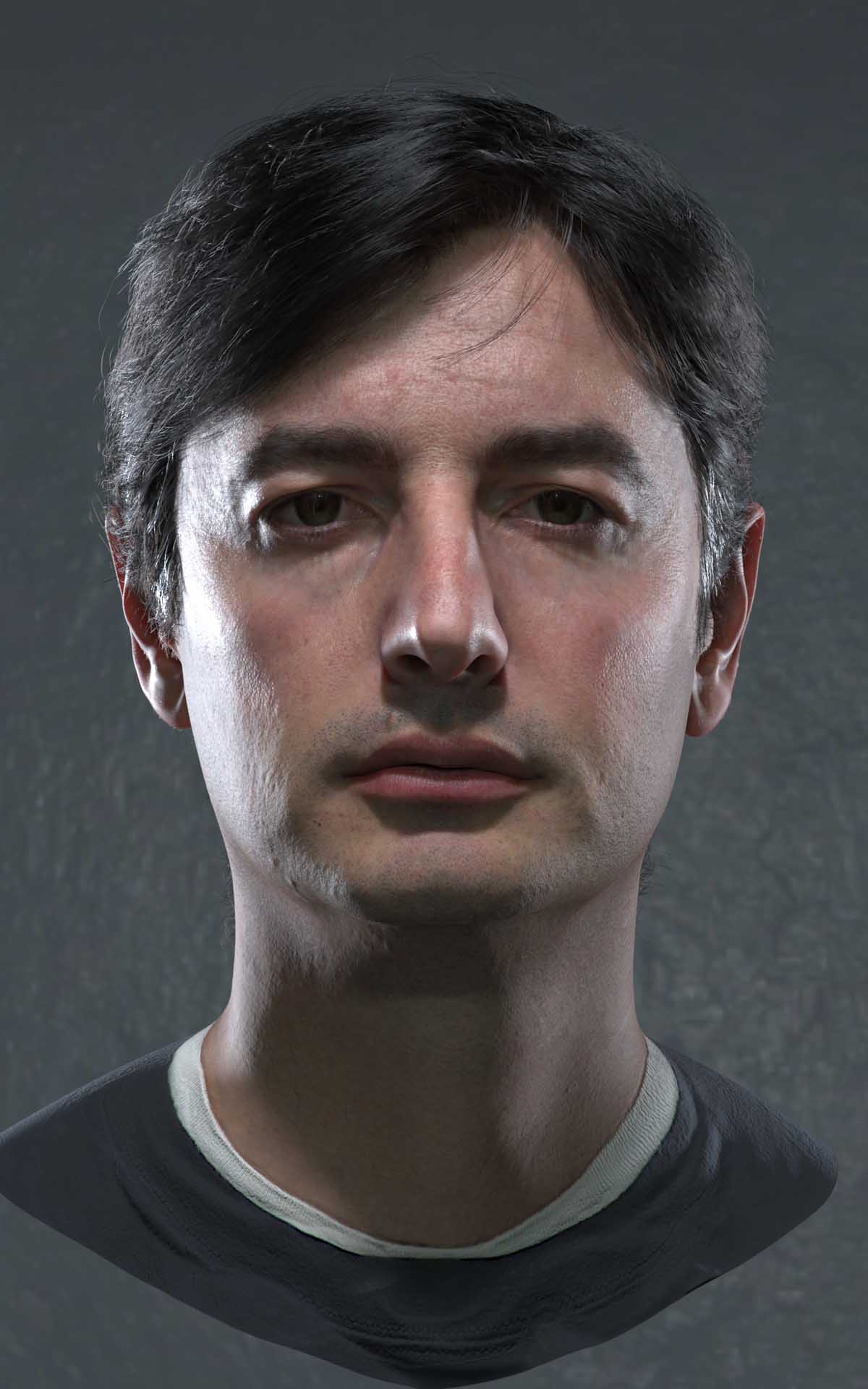}
 &
   \includegraphics[trim= 0 340 0 0, clip, width =0.14\textwidth]{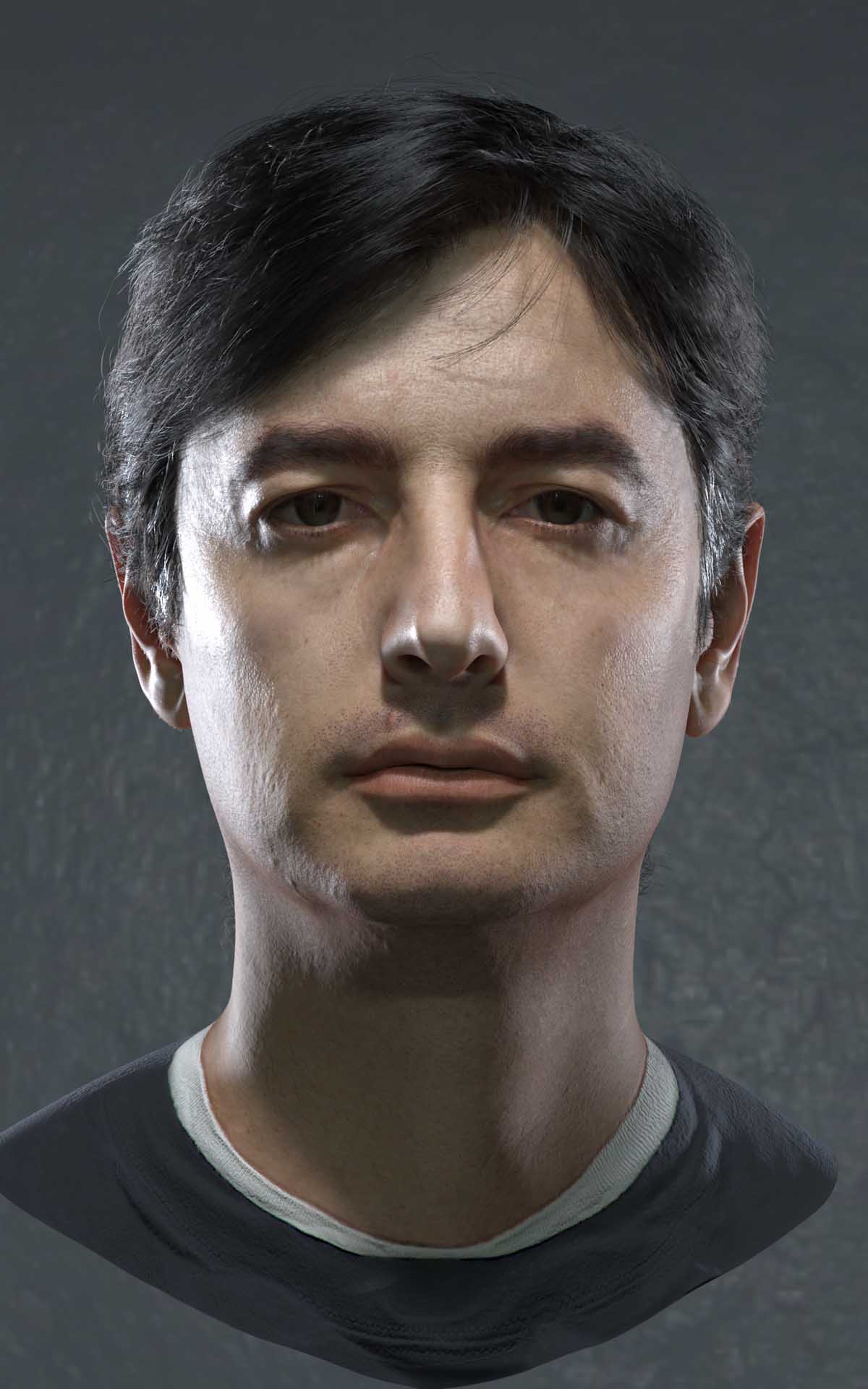}
 &
   \includegraphics[trim= 0 340 0 0, clip, width =0.14\textwidth]{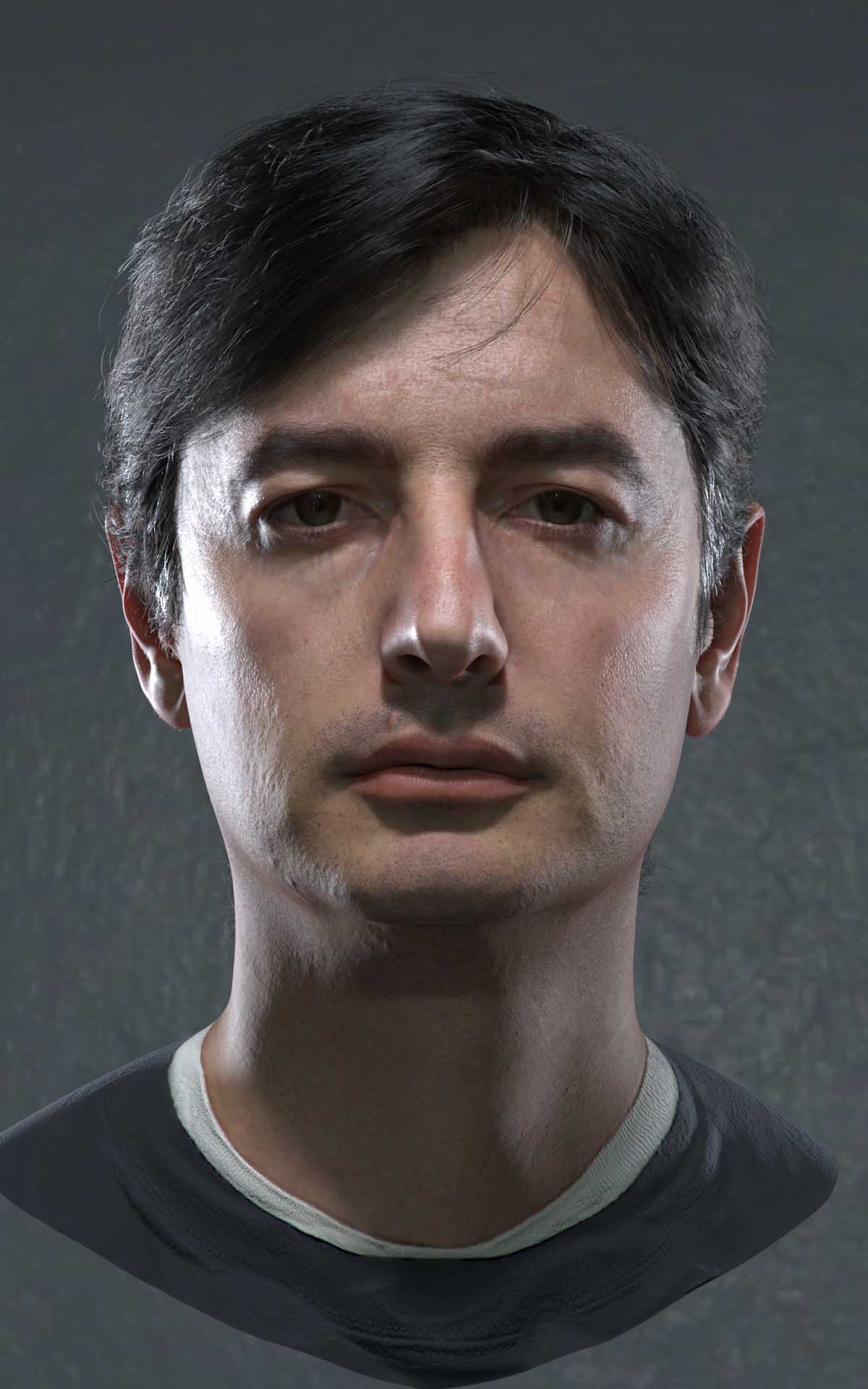}
 &
   \includegraphics[trim= 0 340 0 0, clip, width =0.14\textwidth]{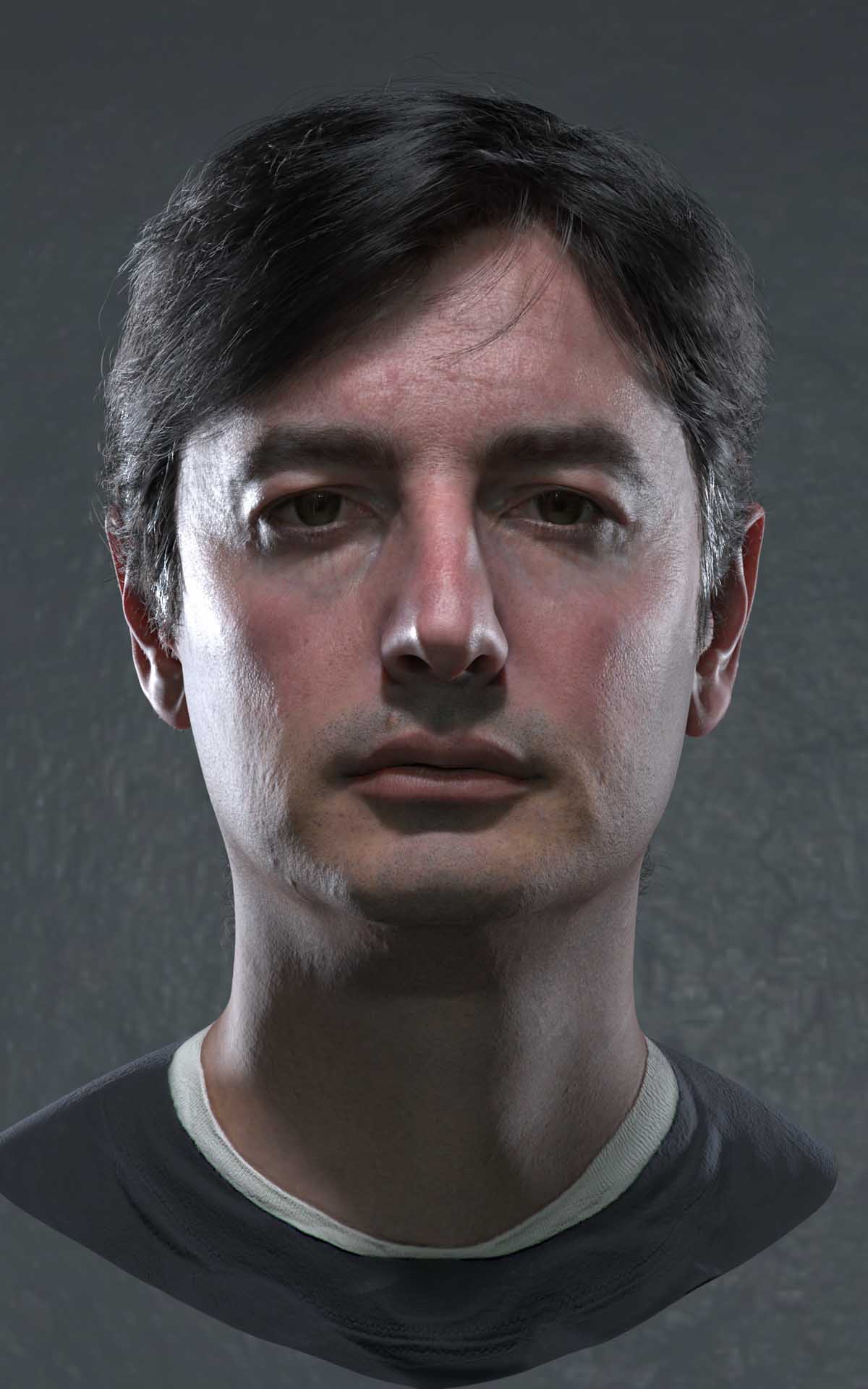}
 &
   \includegraphics[trim= 0 340 0 0, clip, width =0.14\textwidth]{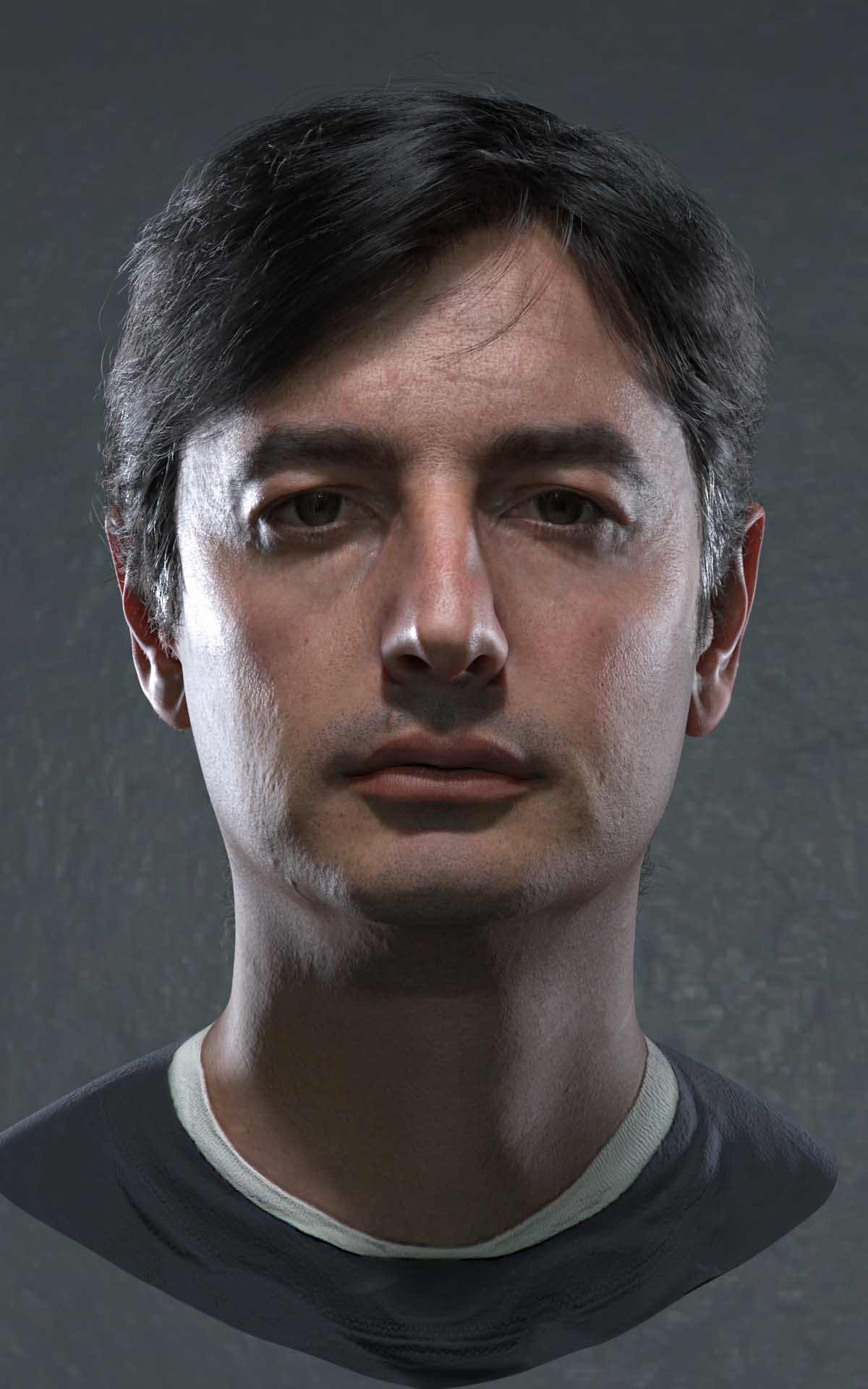}
 &
   \includegraphics[trim= 0 340 0 0, clip, width =0.14\textwidth]{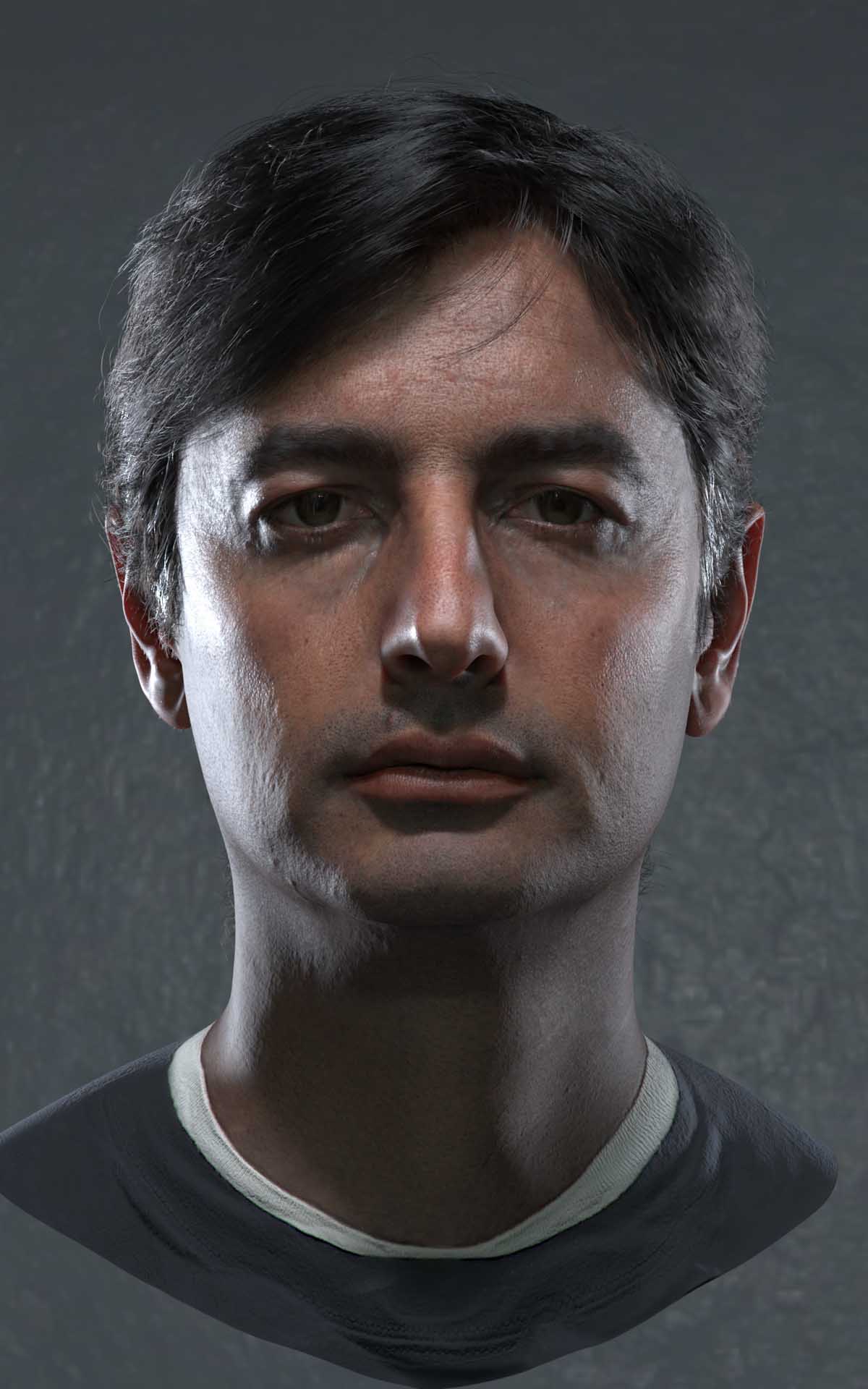}
 &
   \includegraphics[trim= 0 340 0 0, clip, width =0.14\textwidth]{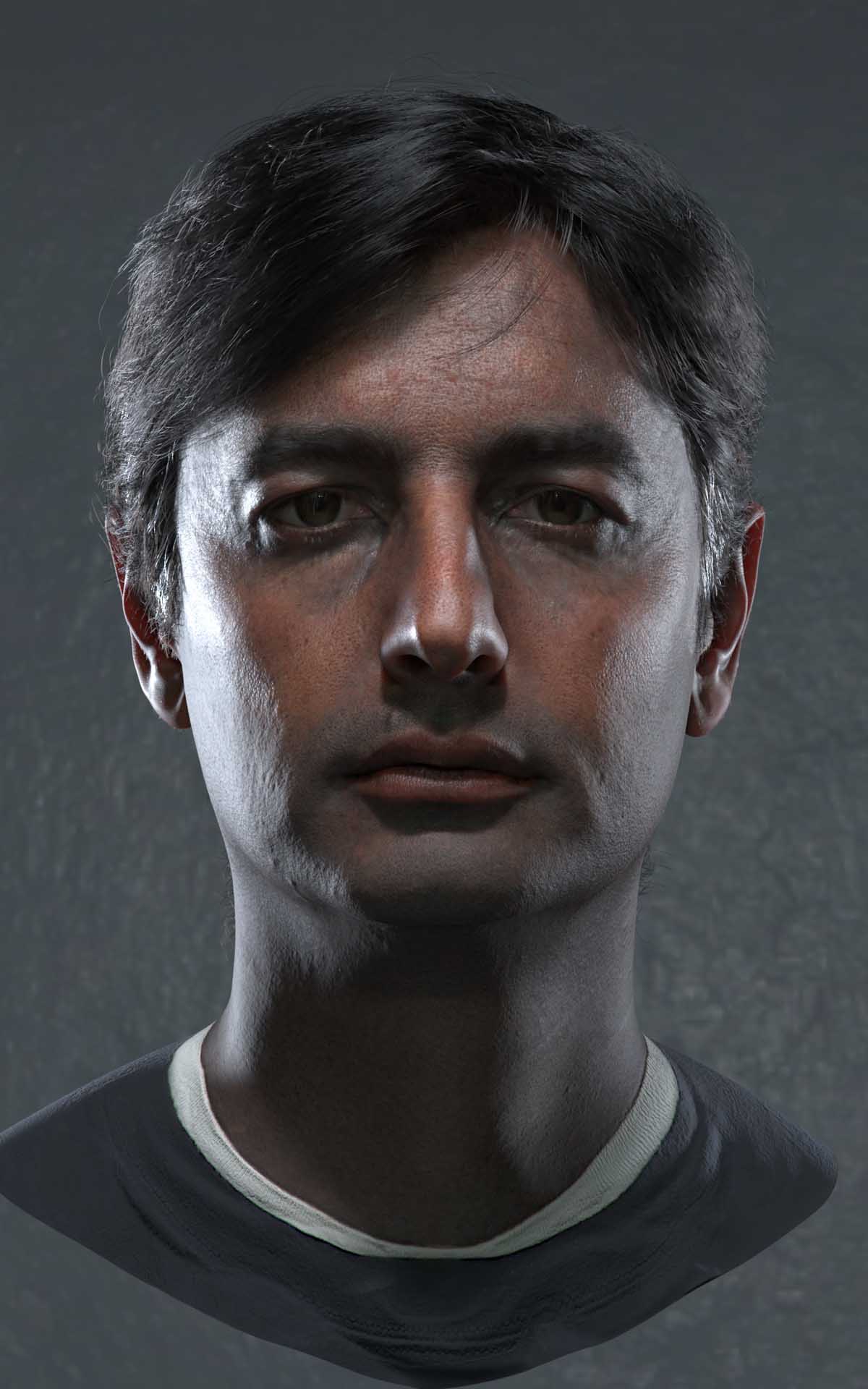}
%
  
  \\
 & & \textsc{0.125x blood } & \textsc{0.5x blood} & \textsc{3x blood} & \textsc{0.5x melanin} & \textsc{1.5x melanin} & \textsc{3x melanin} 
  \\
  \begin{sideways}\hspace{0.2cm}\textsc{Type III (Subject D)}\end{sideways}
 &
  \includegraphics[trim= 0 340 0 0, clip, width =0.14\textwidth]{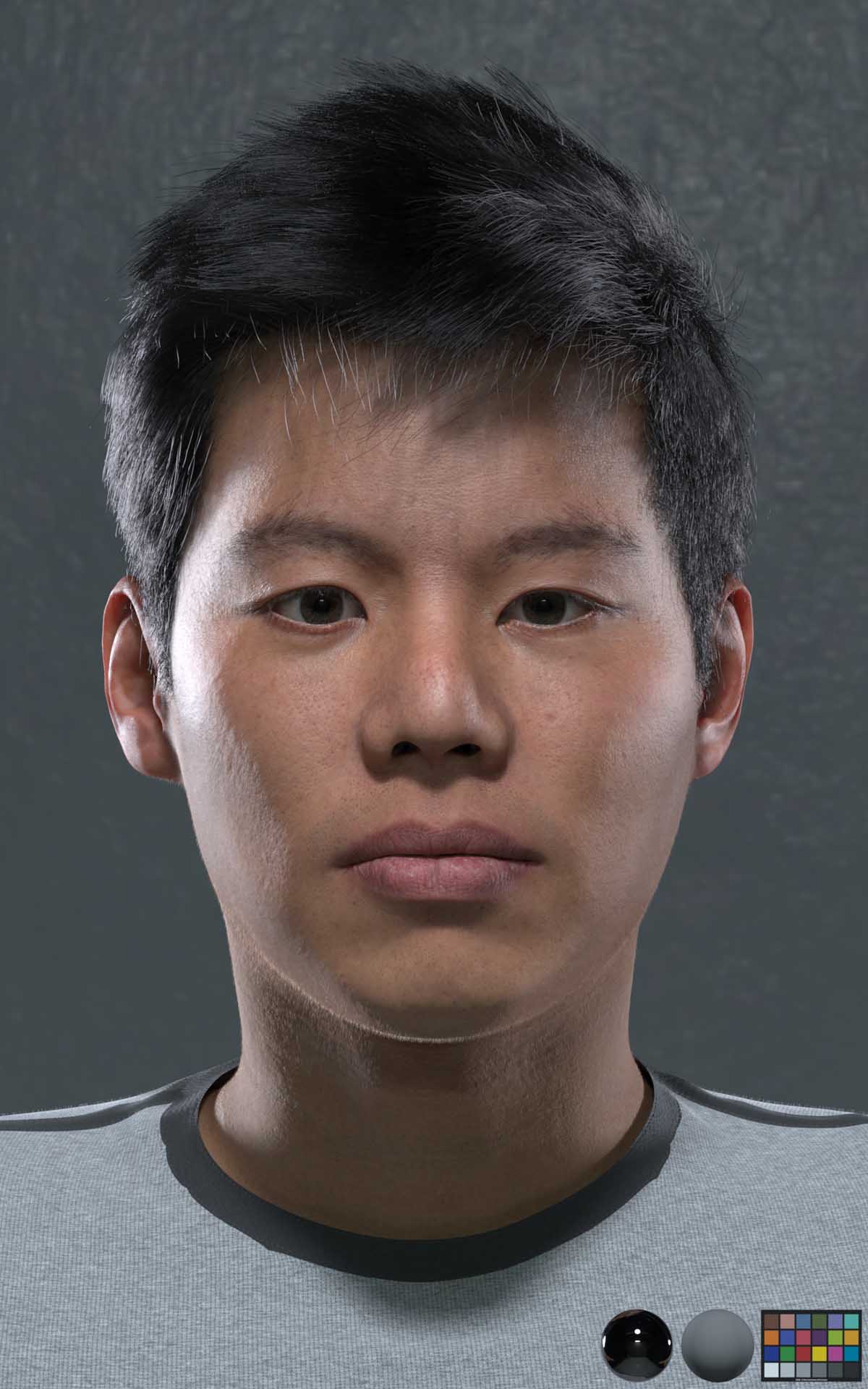}
 &
   \includegraphics[trim= 0 340 0 0, clip, width =0.14\textwidth]{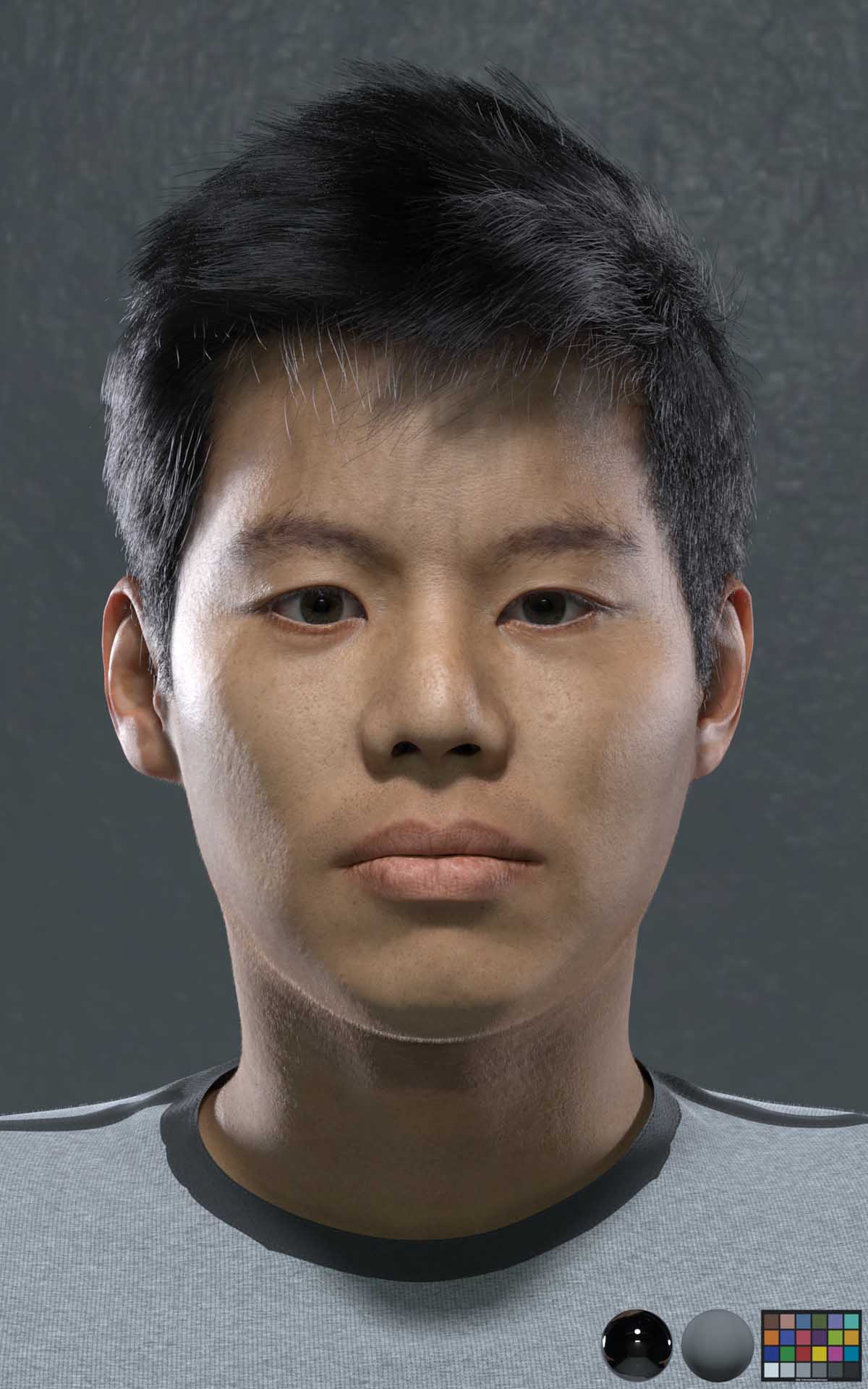}
 &
   \includegraphics[trim= 0 340 0 0, clip, width =0.14\textwidth]{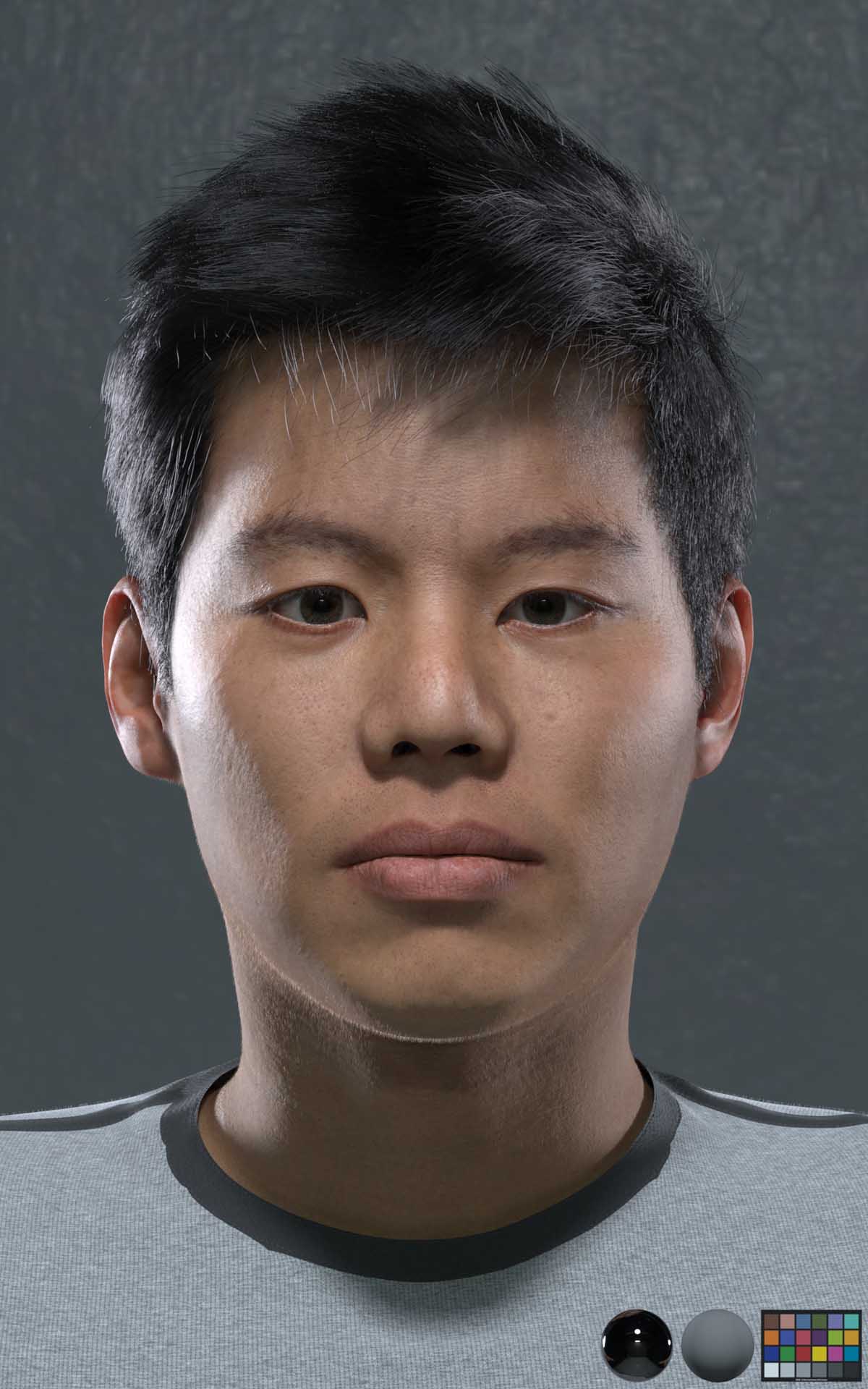}
 &
   \includegraphics[trim= 0 340 0 0, clip, width =0.14\textwidth]{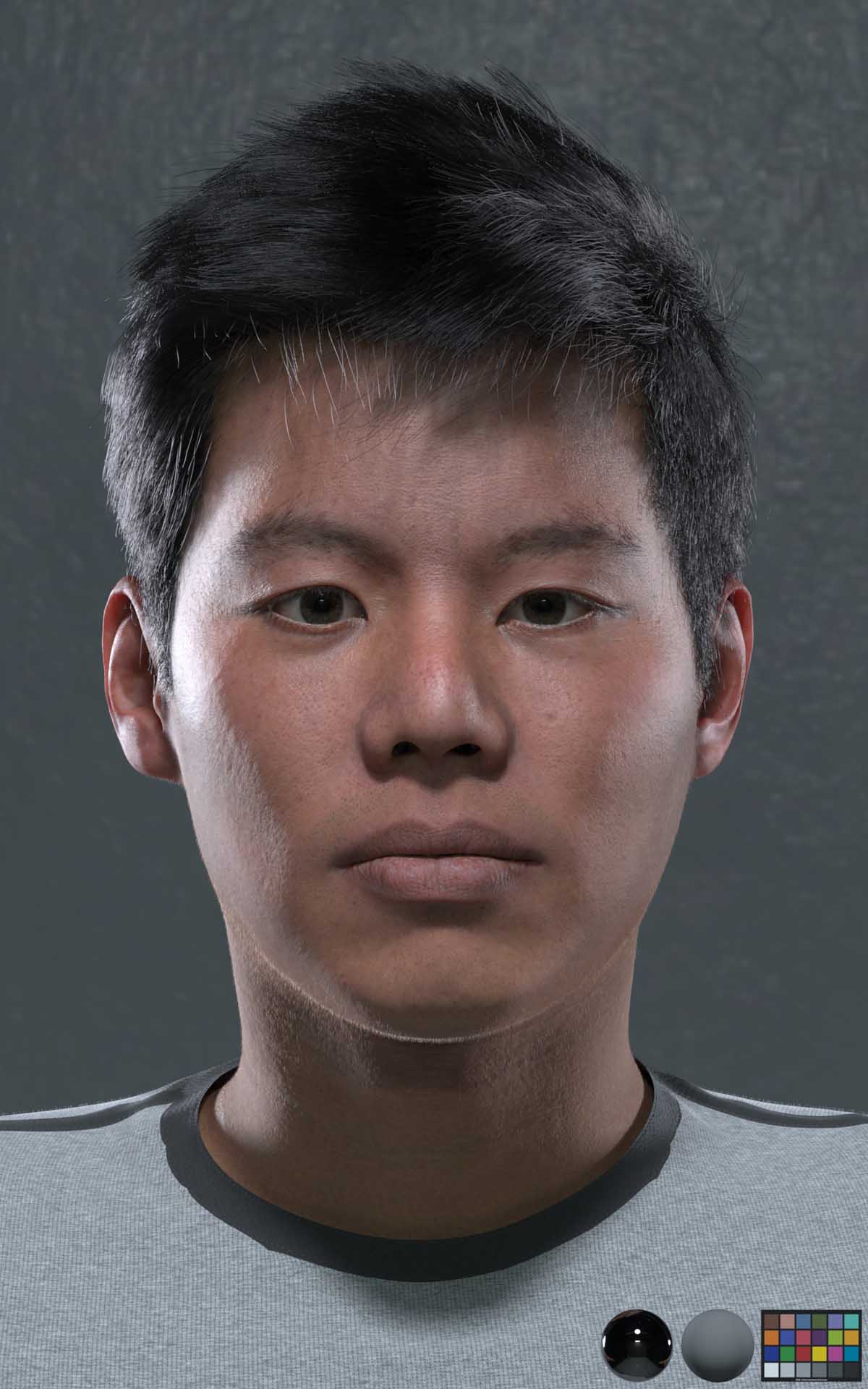}
 &
   \includegraphics[trim= 0 340 0 0, clip, width =0.14\textwidth]{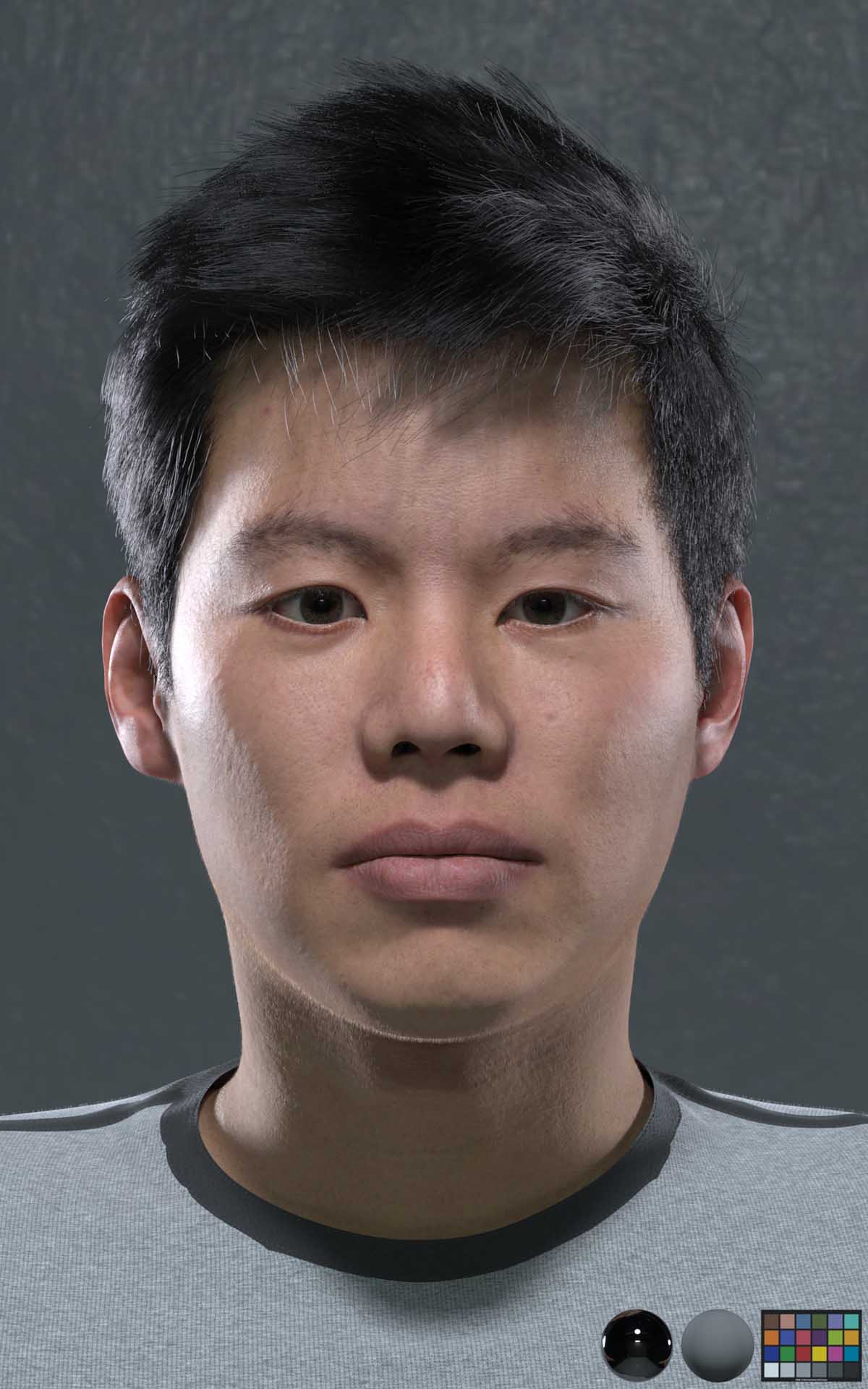}
 &
   \includegraphics[trim= 0 340 0 0, clip, width =0.14\textwidth]{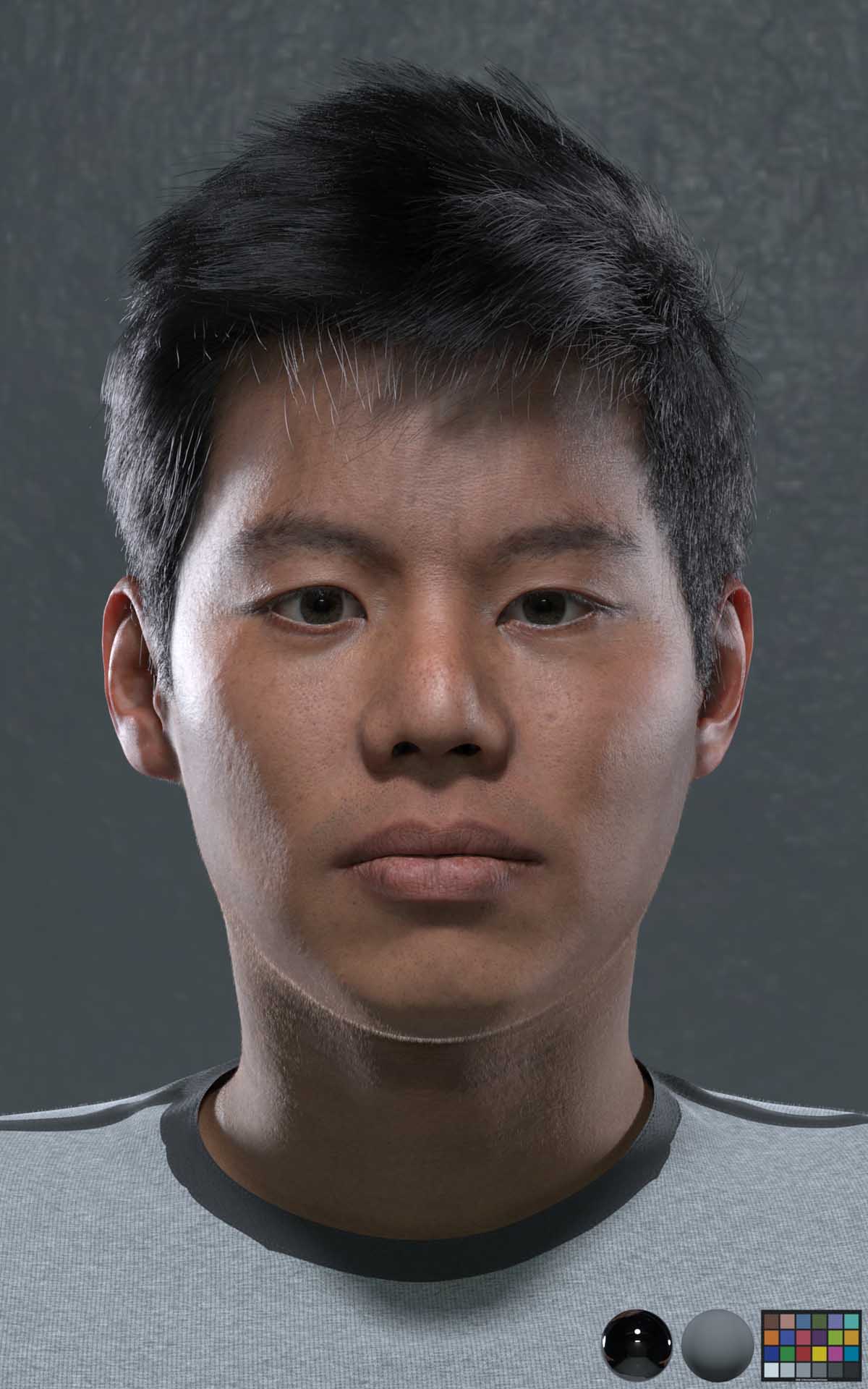}
 &
   \includegraphics[trim= 0 340 0 0, clip, width =0.14\textwidth]{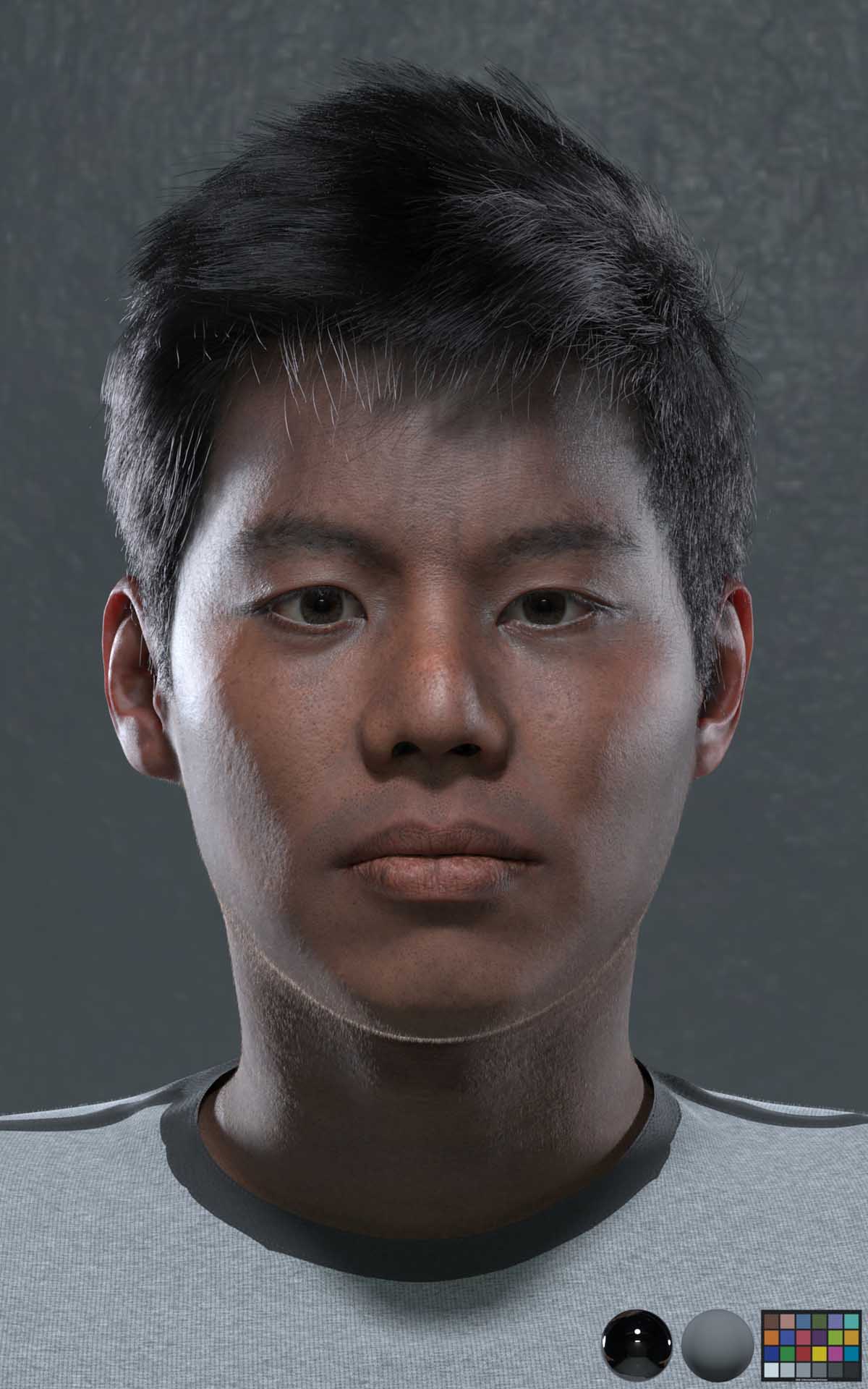}
%

  \\
 & & \textsc{0.125x blood } & \textsc{0.5x blood} & \textsc{3x blood} & \textsc{0.5x melanin} & \textsc{1.5x melanin} & \textsc{3x melanin} 
  \\
  \begin{sideways}\hspace{0.2cm}\textsc{Type IV (Subject E)}\end{sideways}
 &
  \includegraphics[trim= 0 340 0 0, clip, width =0.14\textwidth]{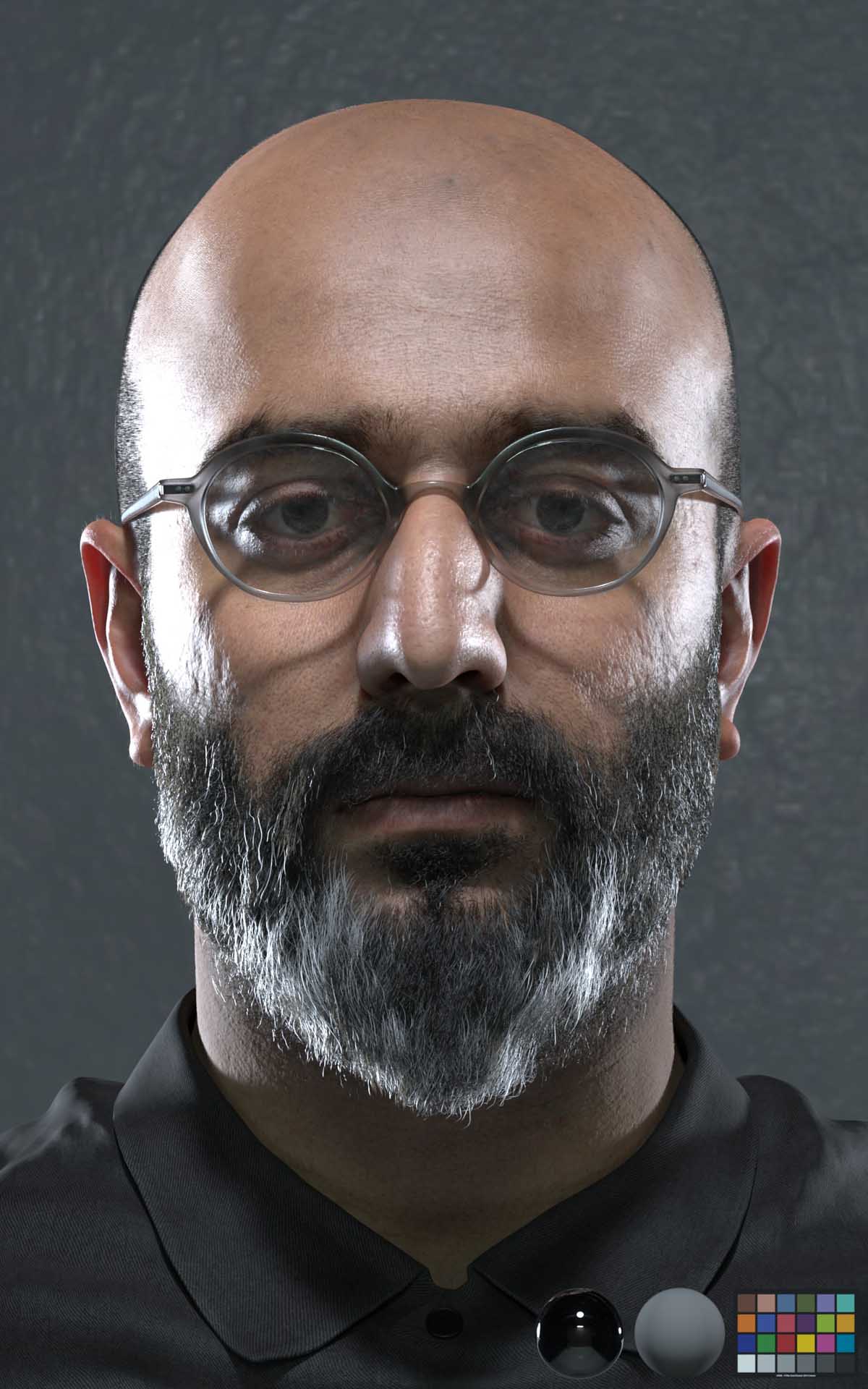}
 &
   \includegraphics[trim= 0 340 0 0, clip, width =0.14\textwidth]{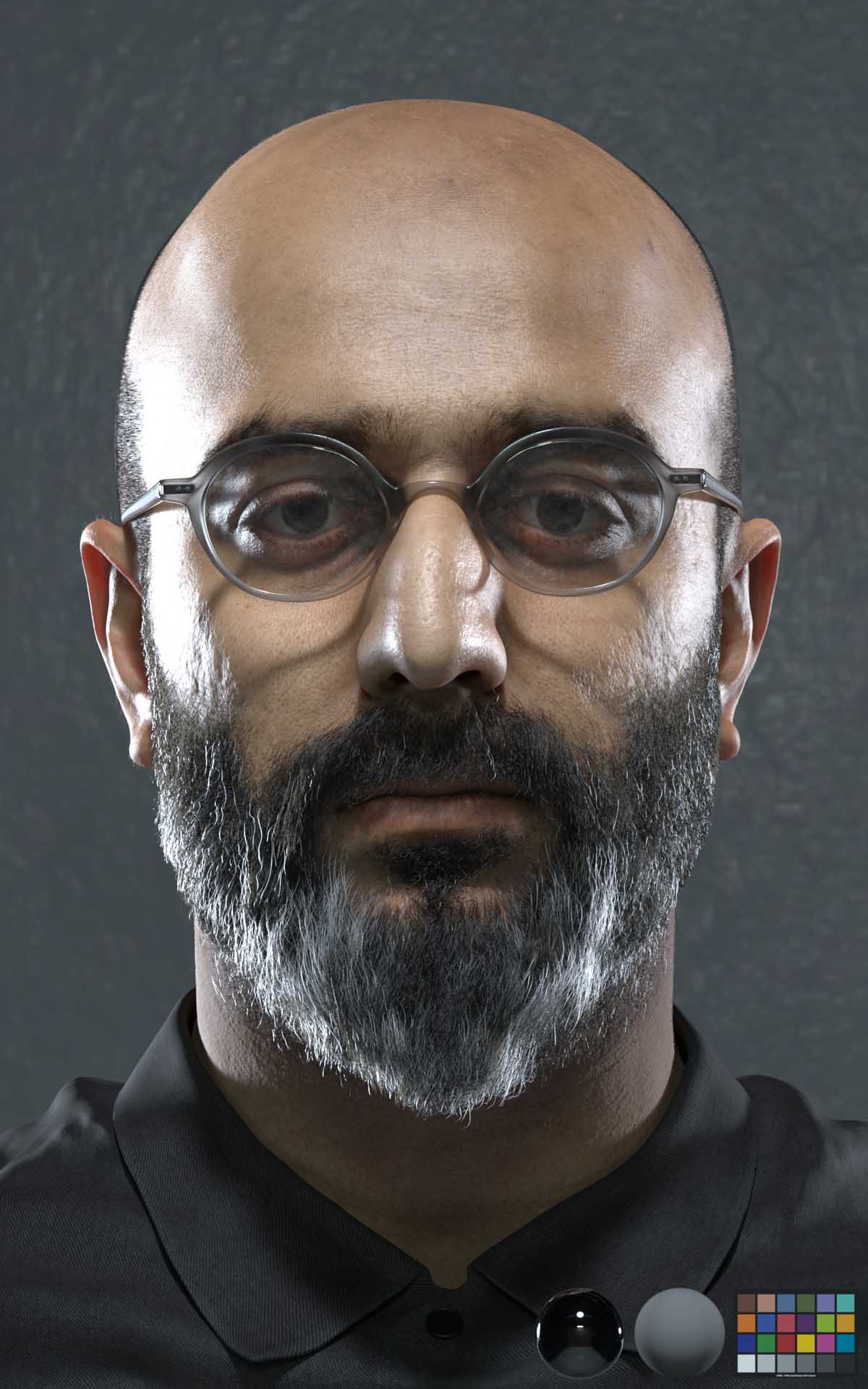}
 &
   \includegraphics[trim= 0 340 0 0, clip, width =0.14\textwidth]{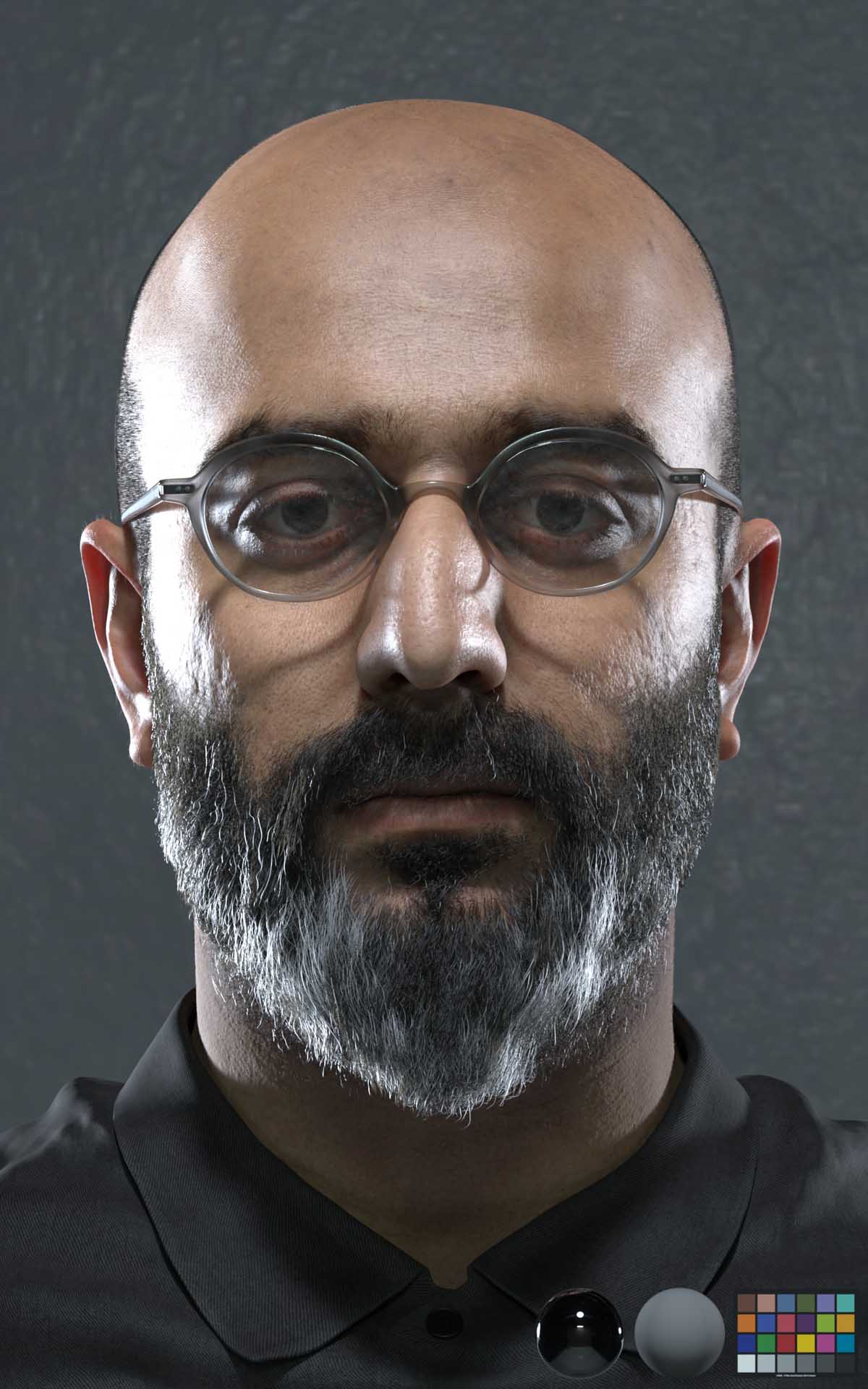}
 &
   \includegraphics[trim= 0 340 0 0, clip, width =0.14\textwidth]{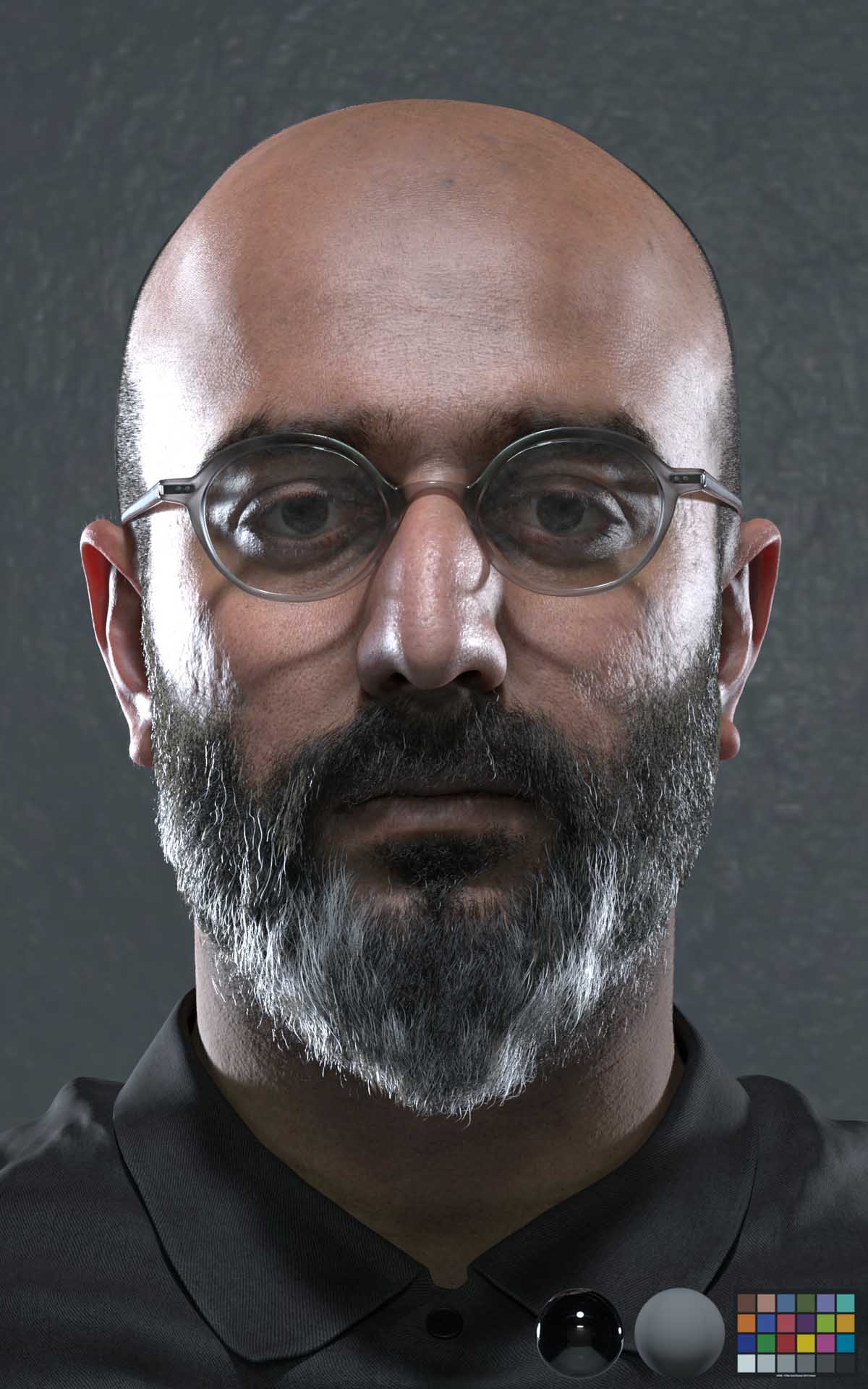}
 &
   \includegraphics[trim= 0 340 0 0, clip, width =0.14\textwidth]{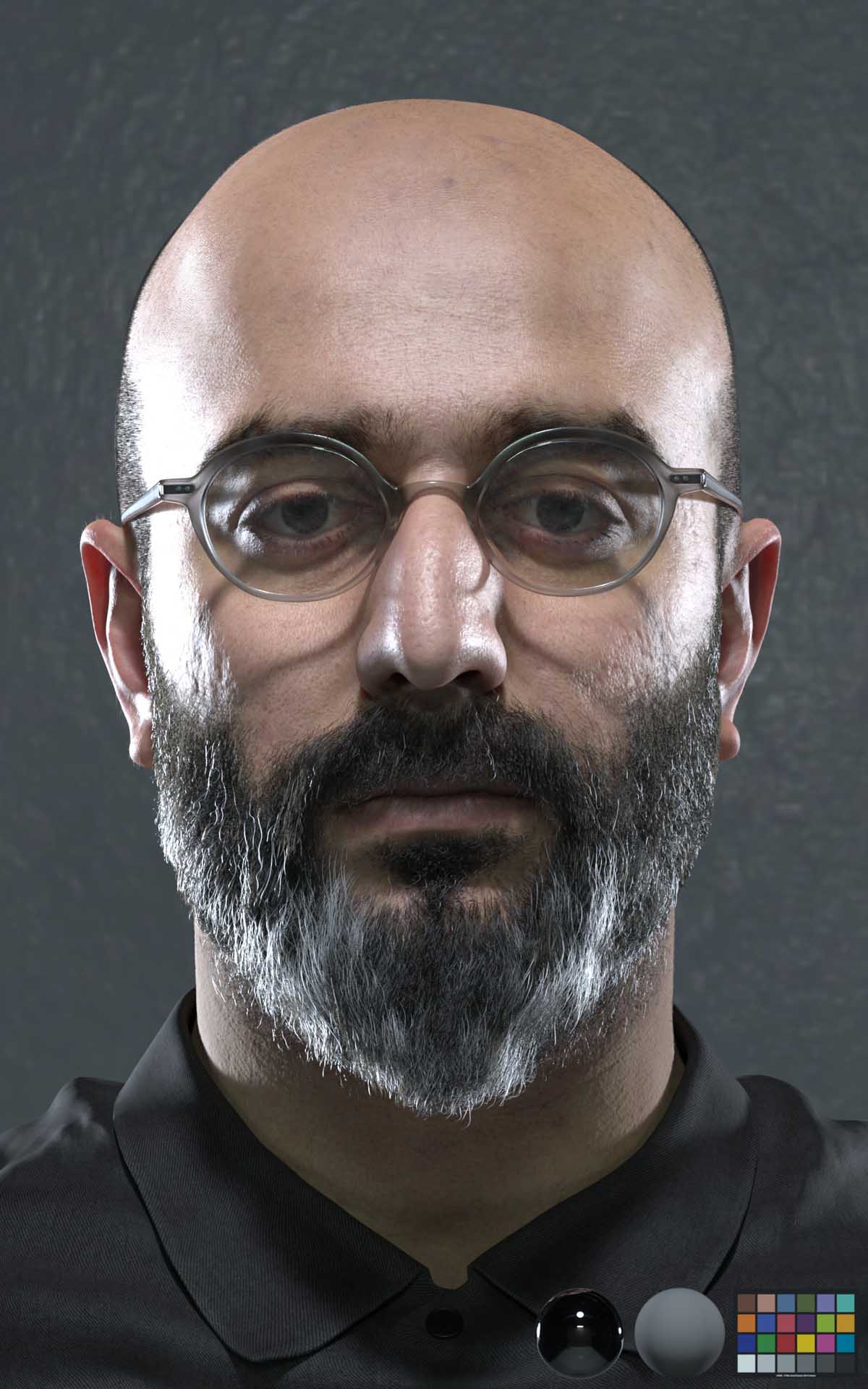}
 &
   \includegraphics[trim= 0 340 0 0, clip, width =0.14\textwidth]{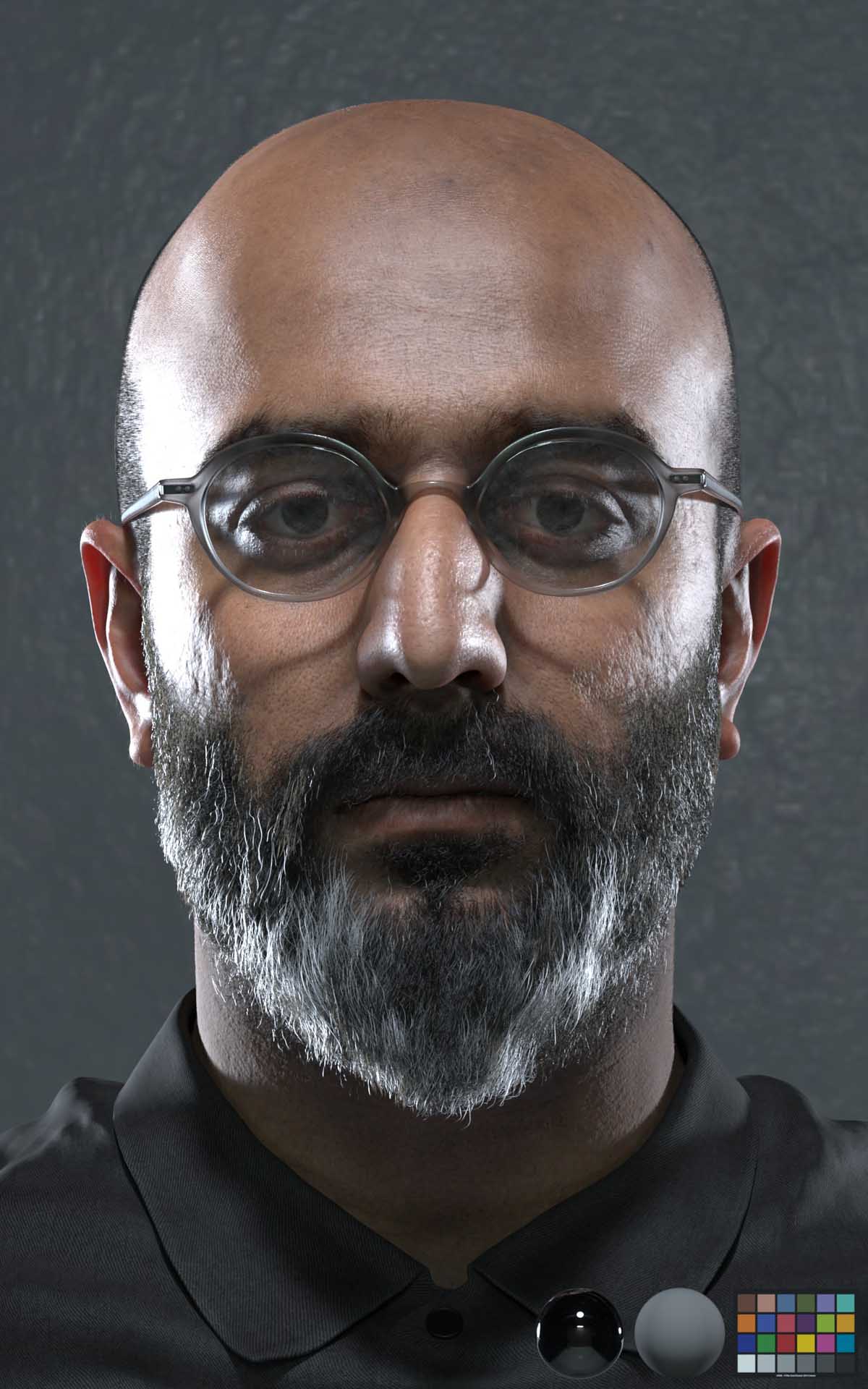}
 &
   \includegraphics[trim= 0 340 0 0, clip, width =0.14\textwidth]{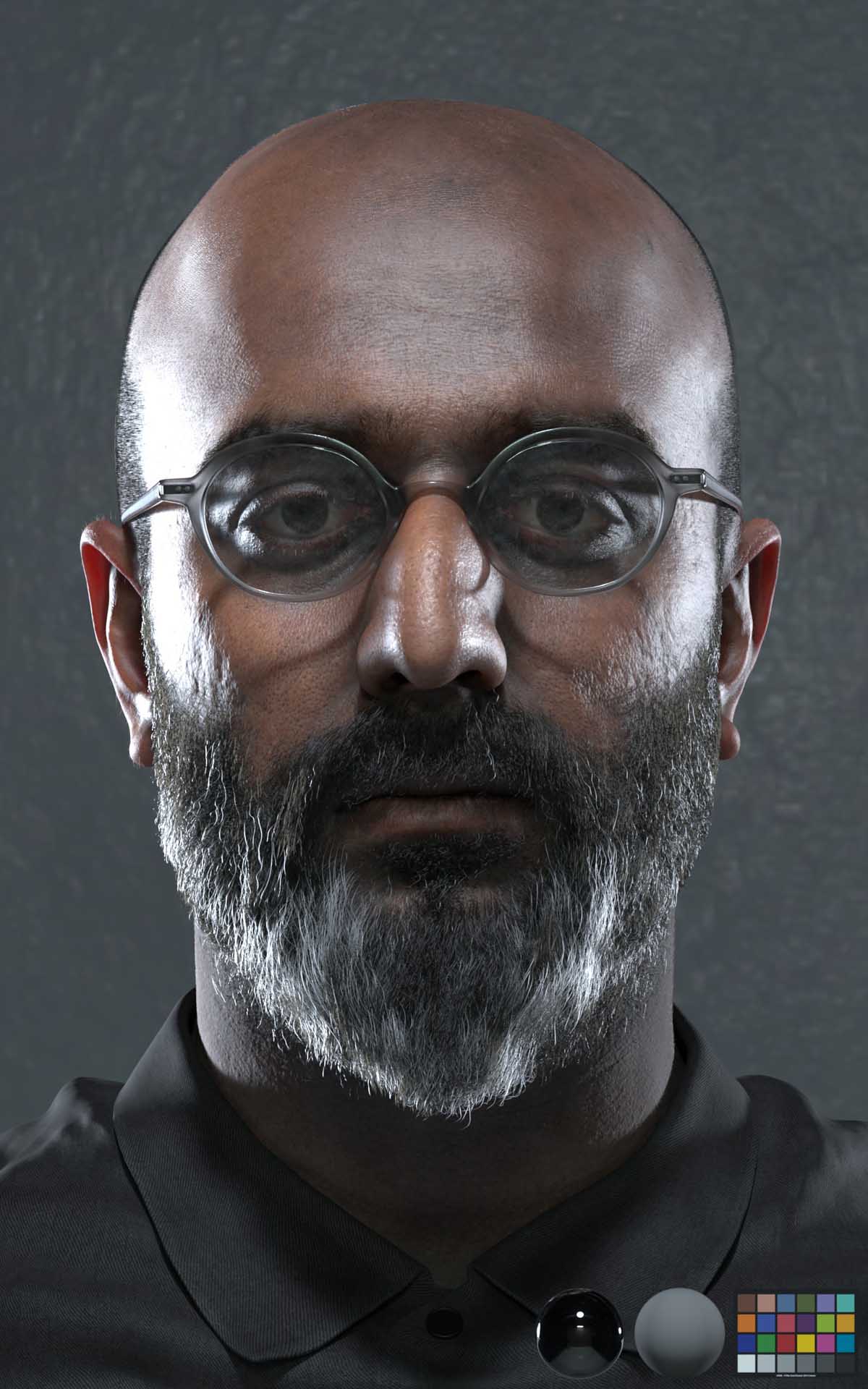}
%
  
  \\
 & & \textsc{0.125x blood } & \textsc{0.5x blood} & \textsc{3x blood} & \textsc{0.5x melanin} & \textsc{1.5x melanin} & \textsc{3x melanin} 
  \\
  \begin{sideways}\hspace{0.2cm}\textsc{Type V (Subject F)}\end{sideways}
 &
  \includegraphics[trim= 0 340 0 0, clip, width =0.14\textwidth]{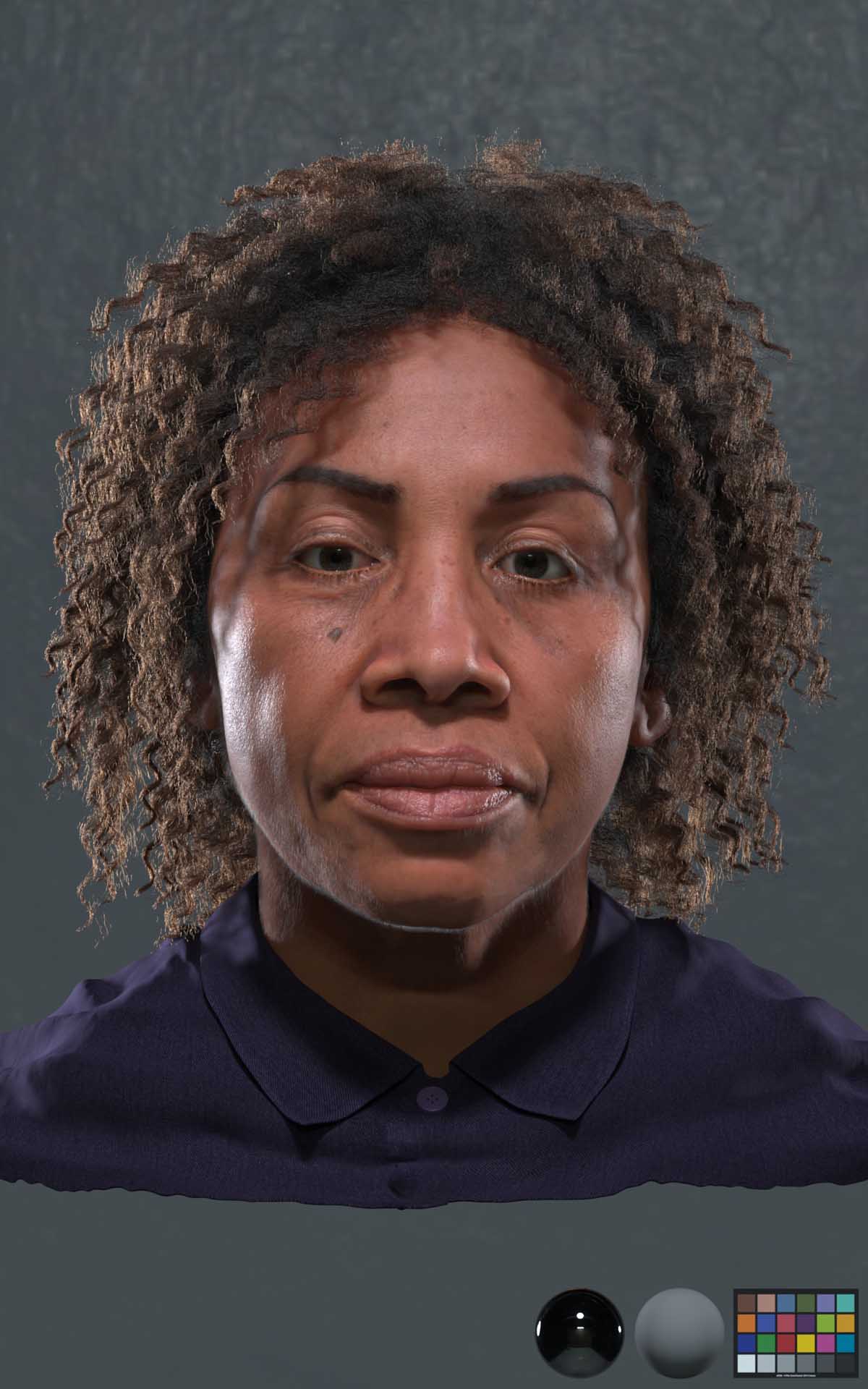}
 &
   \includegraphics[trim= 0 340 0 0, clip, width =0.14\textwidth]{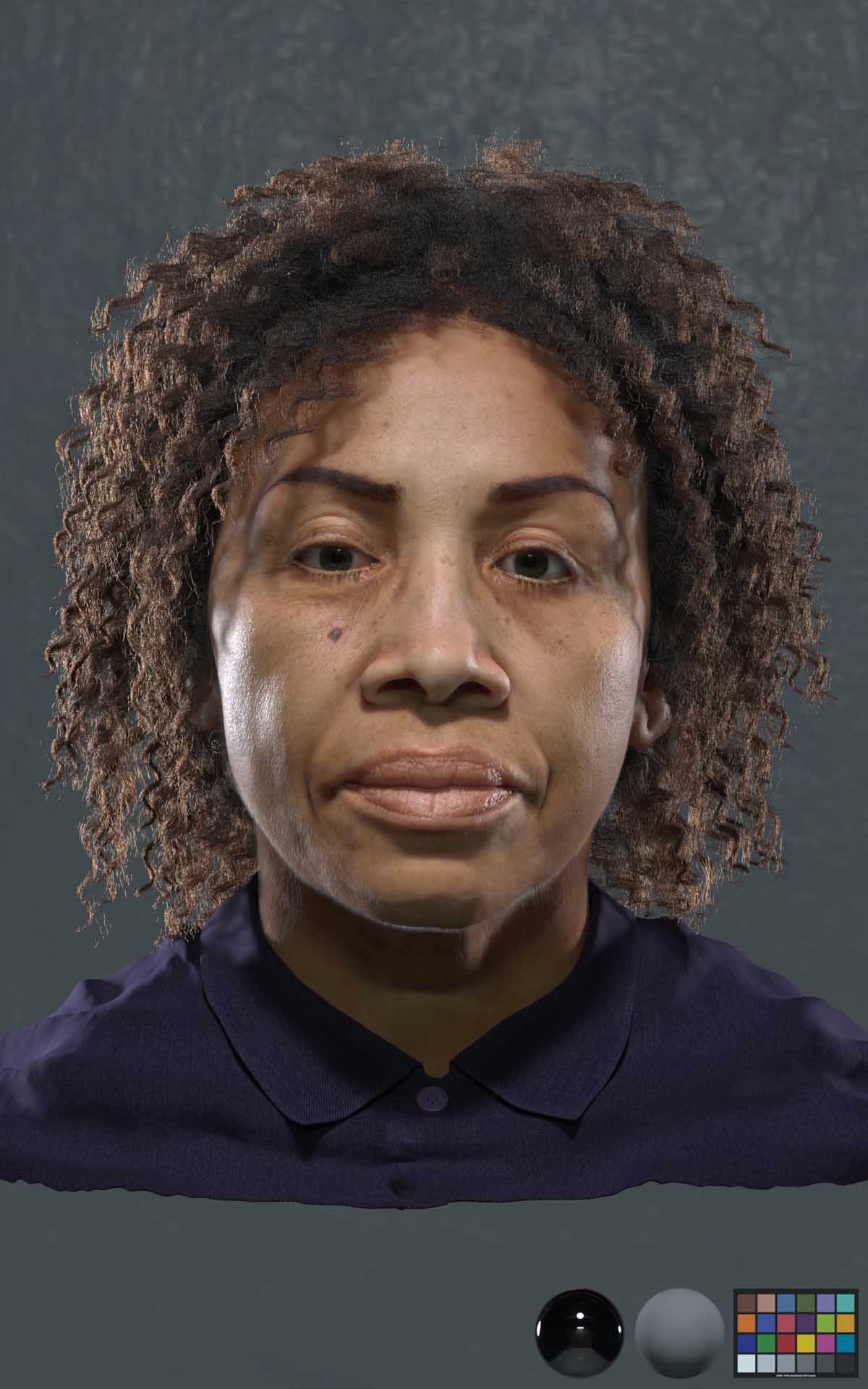}
 &
   \includegraphics[trim= 0 340 0 0, clip, width =0.14\textwidth]{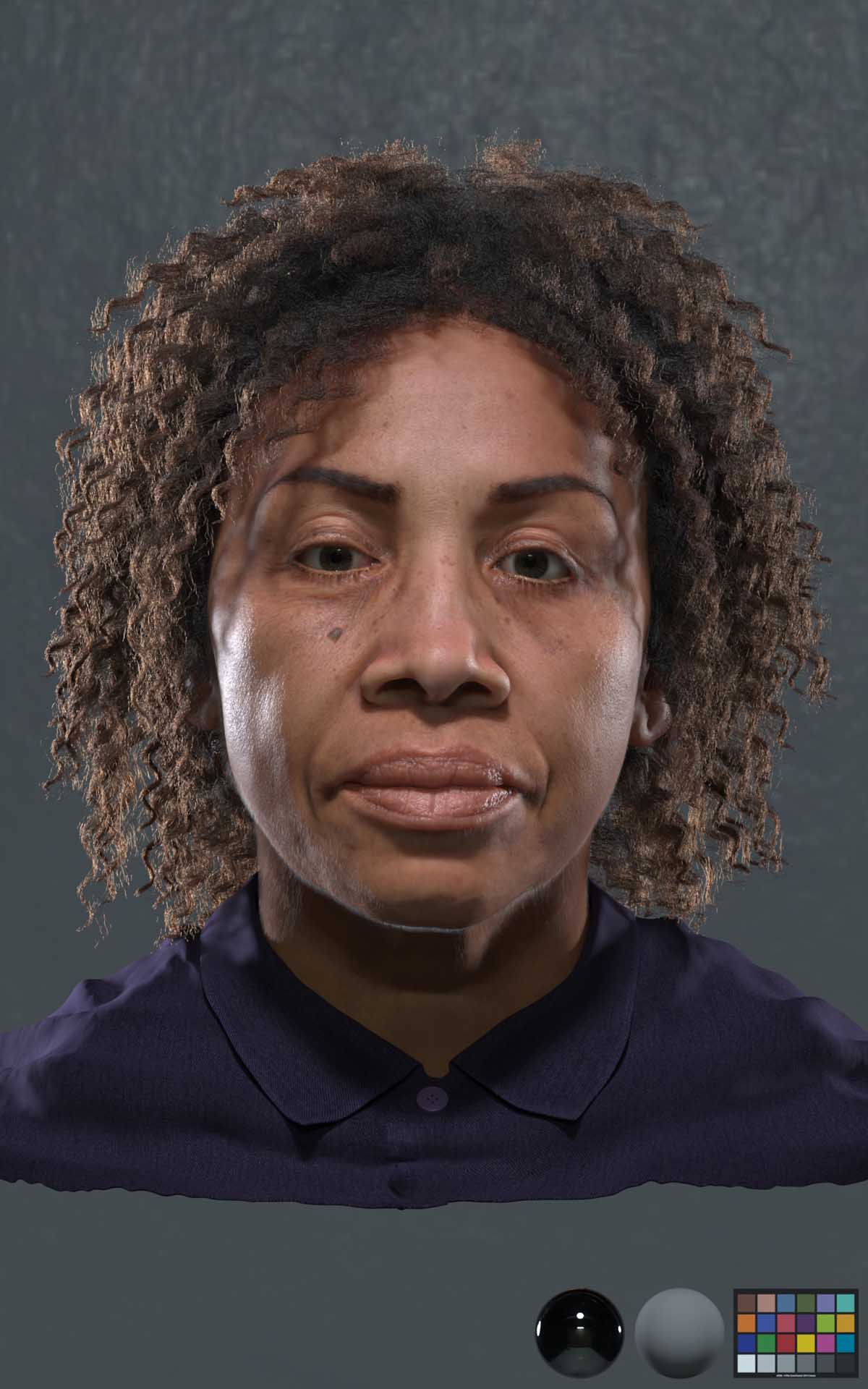}
 &
   \includegraphics[trim= 0 340 0 0, clip, width =0.14\textwidth]{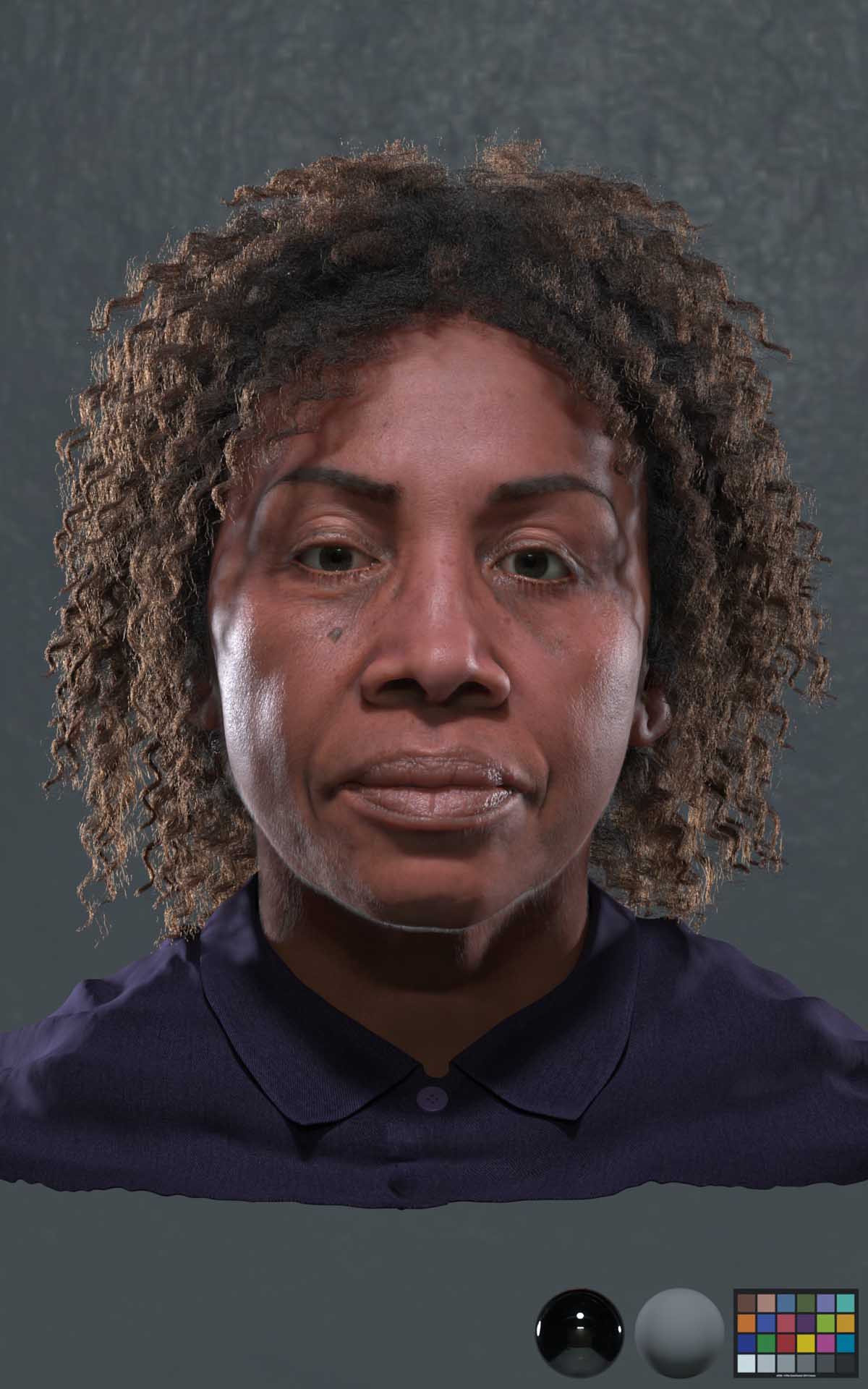}
 &
   \includegraphics[trim= 0 340 0 0, clip, width =0.14\textwidth]{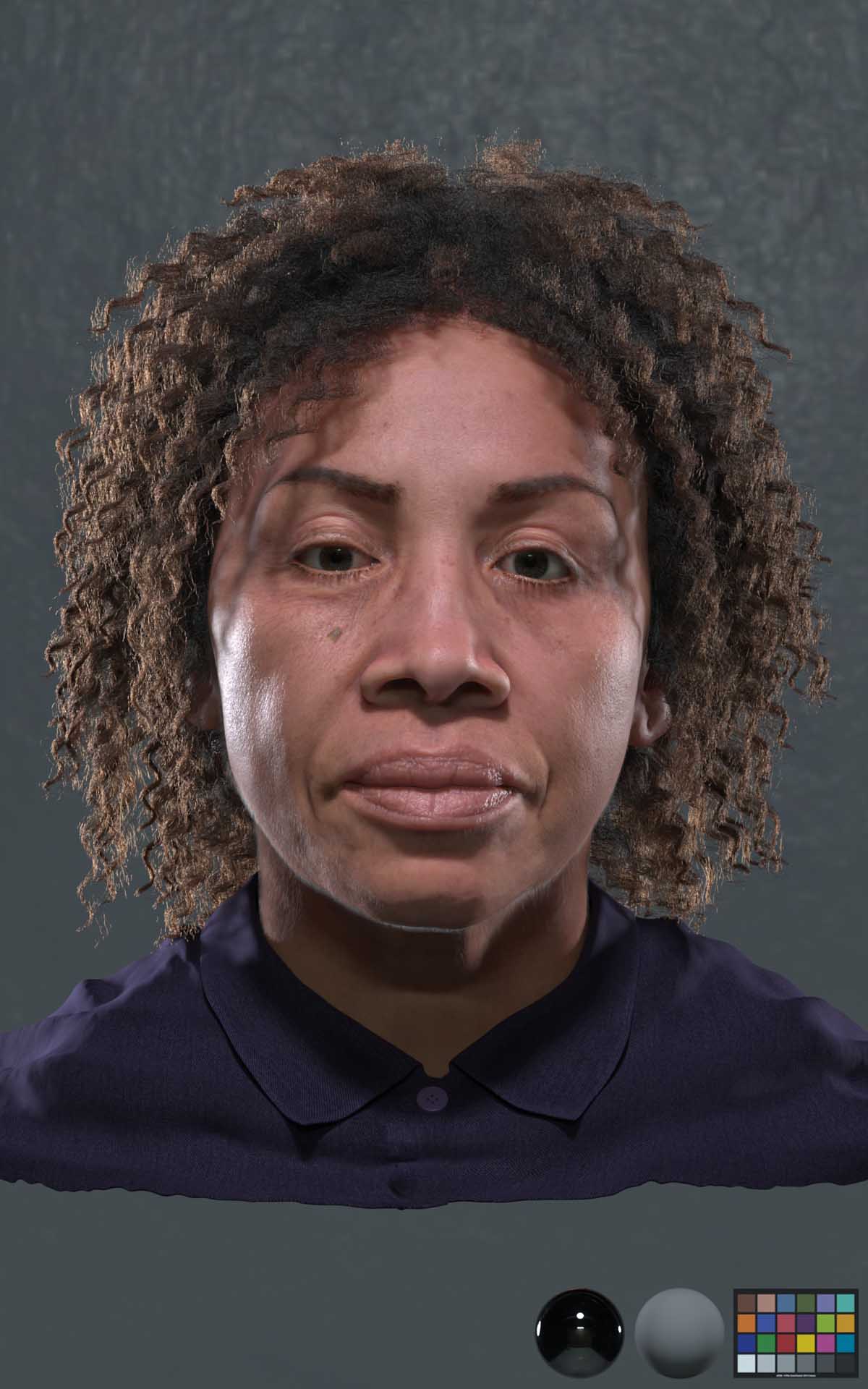}
 &
   \includegraphics[trim= 0 340 0 0, clip, width =0.14\textwidth]{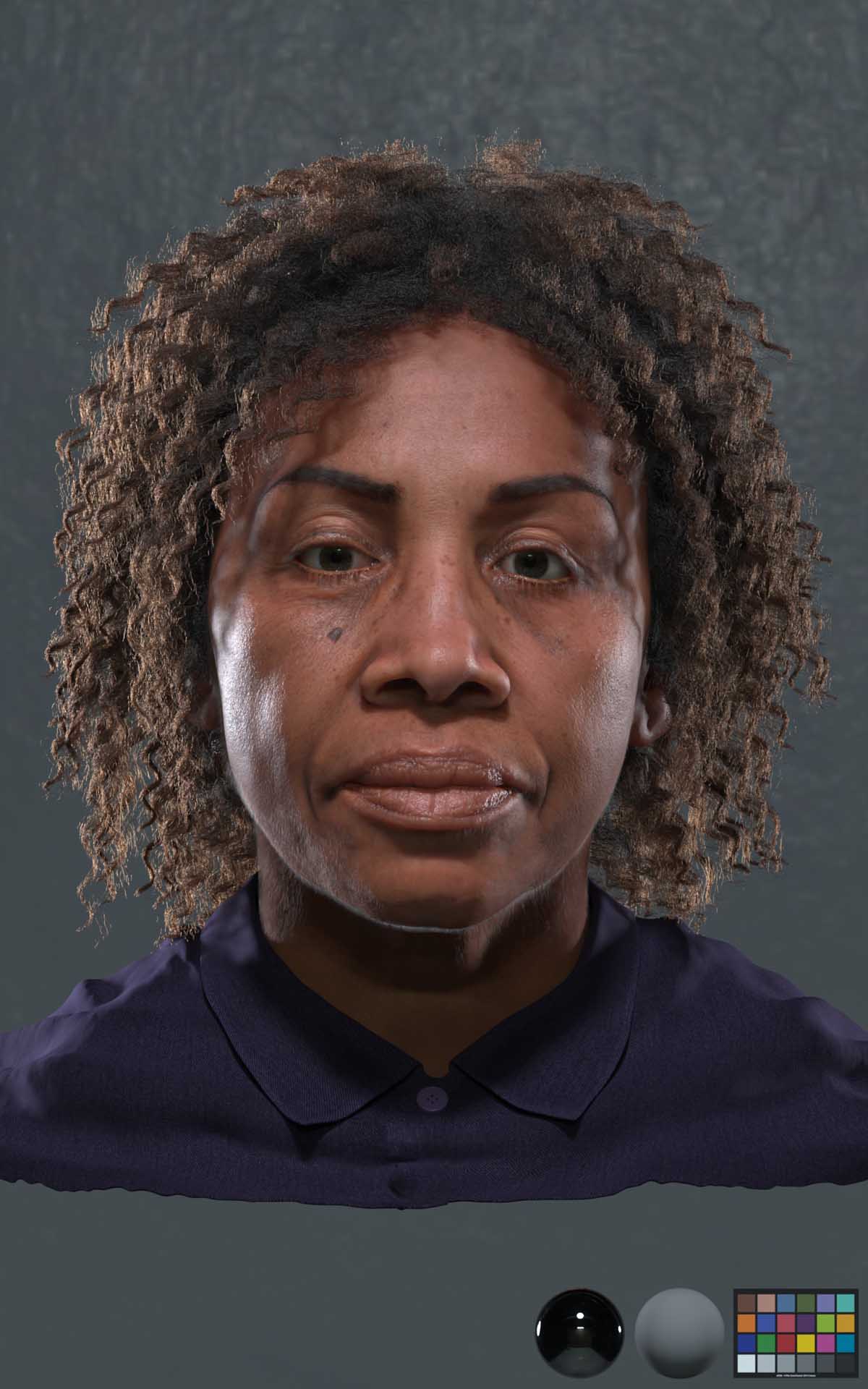}
 &
   \includegraphics[trim= 0 340 0 0, clip, width =0.14\textwidth]{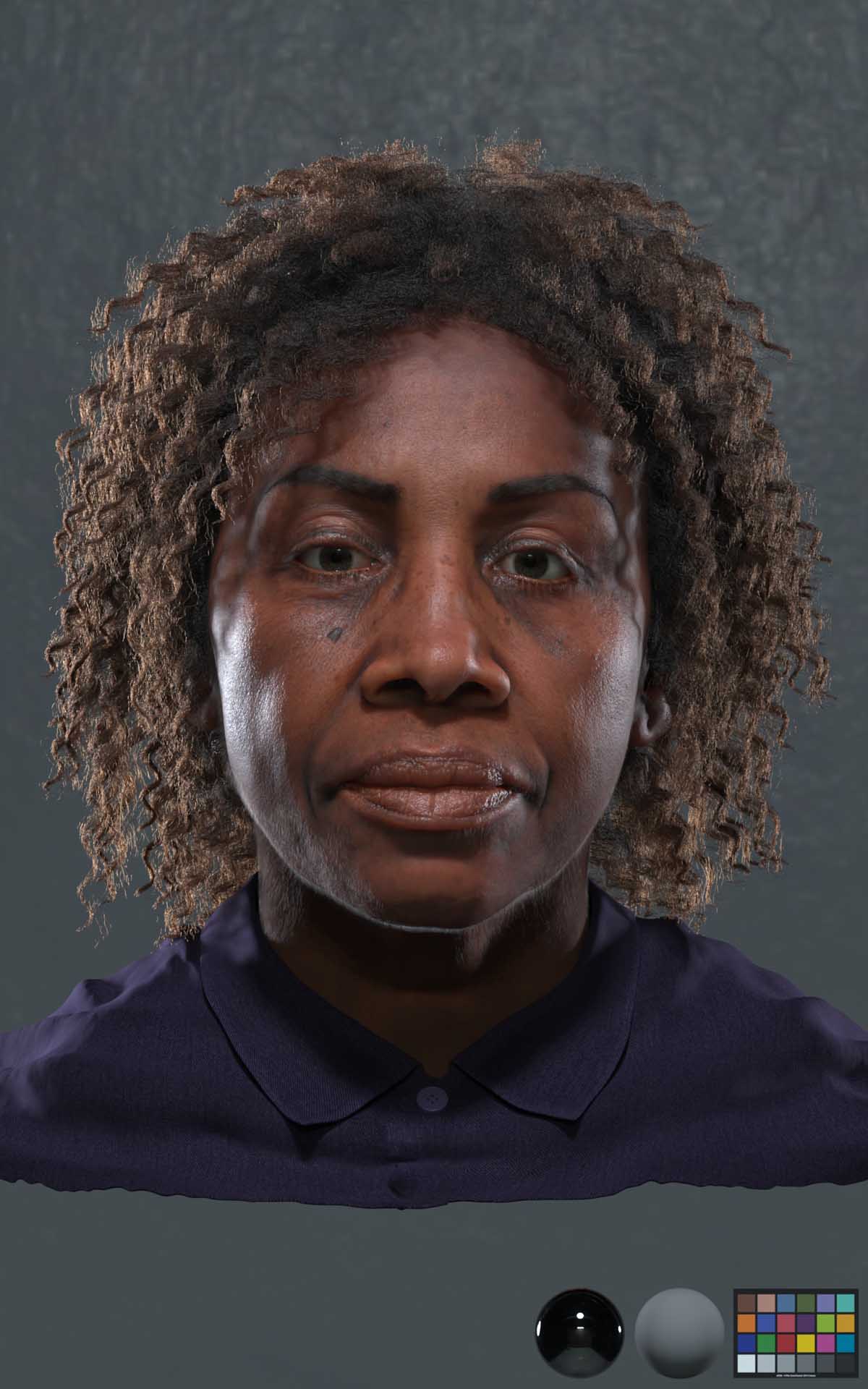}
%

 \end{tabular}
 \vspace{3mm}
  \caption{Rendering results under our most neutral lighting scenario for edited skin parameters inferred by our model over different skin types (classified according to the Fitzpatrick scale). From left to right, original, followed by different manipulations per subject, three in blood concentration and three in melanin concentration (see similar renders under 2 additional lighting environments in Supplementary Figures 7 and 8). Note that these are straight algebraic edits on the recovered skin components, with no additional artistic tweaks or touch-ups involved. Though they demonstrate that our skin model describes an expressive latent space, these naive edits can result sometimes in semi non natural skins, and we suggest maybe more complex edits, some of them shown in Figure~\ref{fig:editing2}.}
  \label{fig:editing}
  \vspace{-6mm}
\end{figure*}

We perform estimations and manipulations of skin parameters over several skin types covering the Fitzpatrick scale~\cite{fitzpatrick1988validity}. With the LUT approach, using the largest tensor (256k skintones) resulted in varying times from 2 to 5 hours for 2k by 2k images, or more than 7 hours for 4k by 4k images, using brute force multi threaded (12) search on the tensor in an Intel Xeon W-2135 at 3.70GHz. Instead, the learned mapping has the advantages of little memory consumption and efficient computation (less than 2 seconds on average for 2k by 2k images, on the same Intel Xeon CPU mentioned above). We refer the readers to Figure~\ref{fig:quantization}, Figure~\ref{fig:albedoreconstruction},  Figure~\ref{fig:neuralparam}, Figure~\ref{fig:editing} and the Supplemental Material to see the neural results. In Figure~\ref{fig:albedoreconstruction}, we observe that we trade a tiny bit of reconstruction correctness (imperceptible in final renders), in favor of spatial regularization and performance.

\begin{figure*}[t!]
 \centering
   \includegraphics[width=\textwidth]{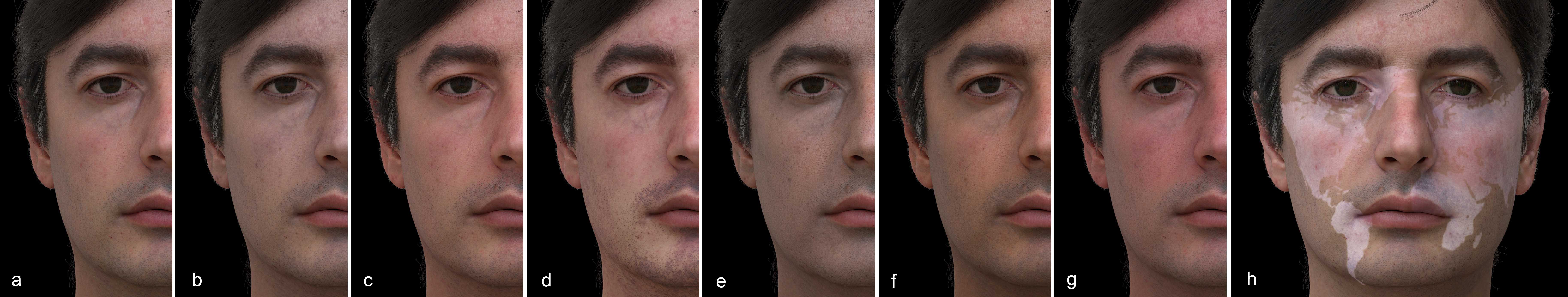}
 \vspace{-5mm}
  \caption{\emph{Further examples of parameter manipulation}. From left to right, a) original albedo, b) fully deoxygenated and c) oxygenated blood, epidermal thinning d) and thickening e) (minimum and maximum respectively), f) tanning (40\% melanin increase, full pheomelanin), g) flushing (70\% increase blood, fully oxygenated), and h) simulated vitiligo through edited melanin concentration. }
  \label{fig:editing2}
  \vspace{-6mm}
\end{figure*}

\emph{Editing the Skin Parameters} 
We show how we can manipulate directly in this space of inferred skin properties, scaling some of them up or down in an intuitive and predictable manner. We run the neural decoder on these modified quantities to reconstruct biophysical albedos, and finally render them on 3D faces. For skin types ranging from I to V, we perform large edits in haemoglobin and melanin content, with details explained in Figure~\ref{fig:editing}. Note the edits are naive in order to cover similar ranges for all skin types. Figure~\ref{fig:editing2} shows edits over the rest of the parameters of the model. The level of blood oxygenation translates into paler or more saturated skin colors. The thinning of the epidermis, which typically occurs with aging, translates into a more translucent look, revealing the heterogeneities of the underlying layers (e.g. capillary and veins), while a thicker epidermis results in a more opaque and rough appearance. Last, we vary several components to simulate tanning and  flushing. While it is hard to fully validate the correctness of the recovered skin properties, we find the parametrization adequate to produce plausible human skin albedos.

\begin{figure}[t!]
 \contourlength{0.1em}%
 \centering
  \begin{tabular}{c@{\;}c@{\;}c}
  \textsc{Input} & \textsc{Gitlina'20} & \textsc{Ours} 

  \\
  \includegraphics[width = 0.3\columnwidth]{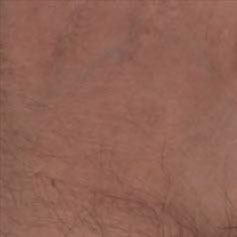}
  &
  \includegraphics[width = 0.3\columnwidth]{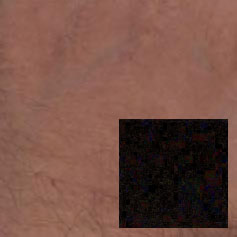}
  &
  \includegraphics[width = 0.3\columnwidth]{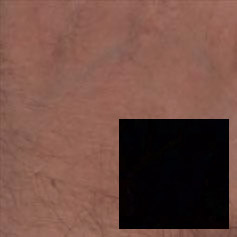}

  \\
  \includegraphics[width = 0.3\columnwidth]{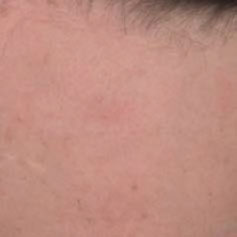}
  &
  \includegraphics[width = 0.3\columnwidth]{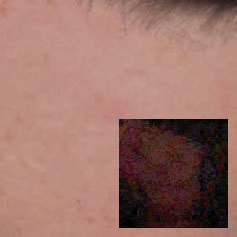}
  &
  \includegraphics[width = 0.3\columnwidth]{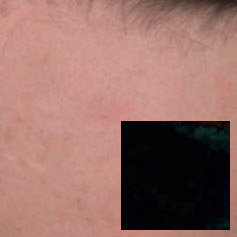}

  \\
  \includegraphics[width = 0.3\columnwidth]{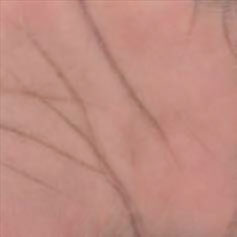}
  &
  \includegraphics[width = 0.3\columnwidth]{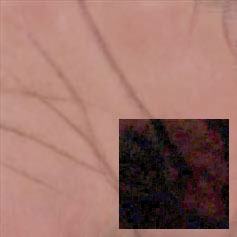}
  &
  \includegraphics[width = 0.3\columnwidth]{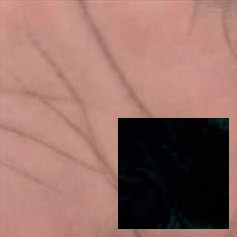}

  \\
  \includegraphics[width = 0.3\columnwidth]{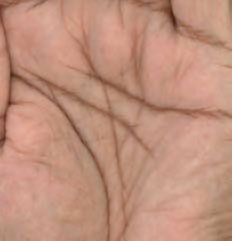}
  &
  \includegraphics[width = 0.3\columnwidth]{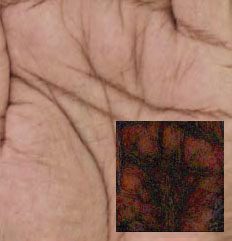}
  &
  \includegraphics[width = 0.3\columnwidth]{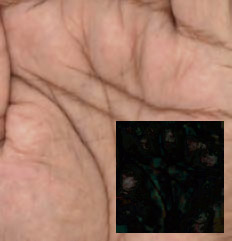}

 \end{tabular}
   \vspace{-2mm}
  \caption{\emph{Antera device comparisons}. Our reconstructions consistently exhibit less errors than in the previous work: MSE (\cite{gitlina2020practical}, ours) of the skin patches from top to bottom: $(0.0142, 1.65x10^{-6}), (0.1016, 2.31x10^{-5}), (0.0585, 1.08x10^{-5}),$ $(0.2239, 8.93x10^{-5})$. }
  \vspace{-3mm}
  \label{fig:comp_gitlina_albedos2}
\end{figure}

\begin{figure*}[t!]
 \contourlength{0.1em}%
 \centering
 \hspace*{-4.5mm}%
  \begin{tabular}{l@{\;}c@{\;}c@{\;}c@{\;}c@{\;}c@{\;}c@{\;}c}
  
  \\
   & \textsc{Melanin} & \textsc{Hemoglobin(d)} & \textsc{Hemoglobin(e)} & \textsc{Melanin ratio} & \textsc{Input D65} & \textsc{Rec. (Ours)} \\

  \begin{sideways}\hspace{0.05cm}\textsc{Gitlina'20}\end{sideways}
  &
  \includegraphics[width = 0.16\textwidth]{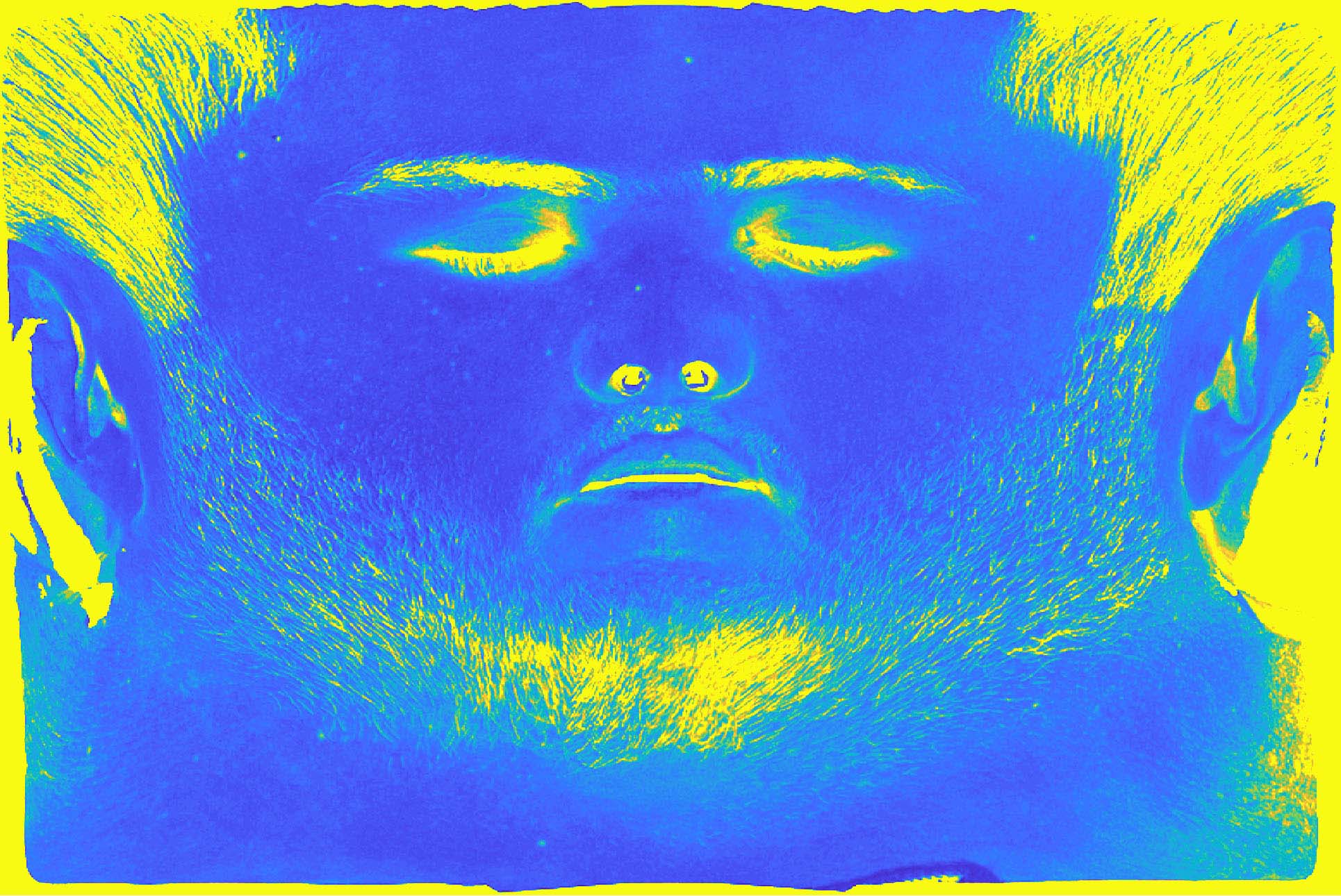}
  &
  \includegraphics[width =  0.16\textwidth]{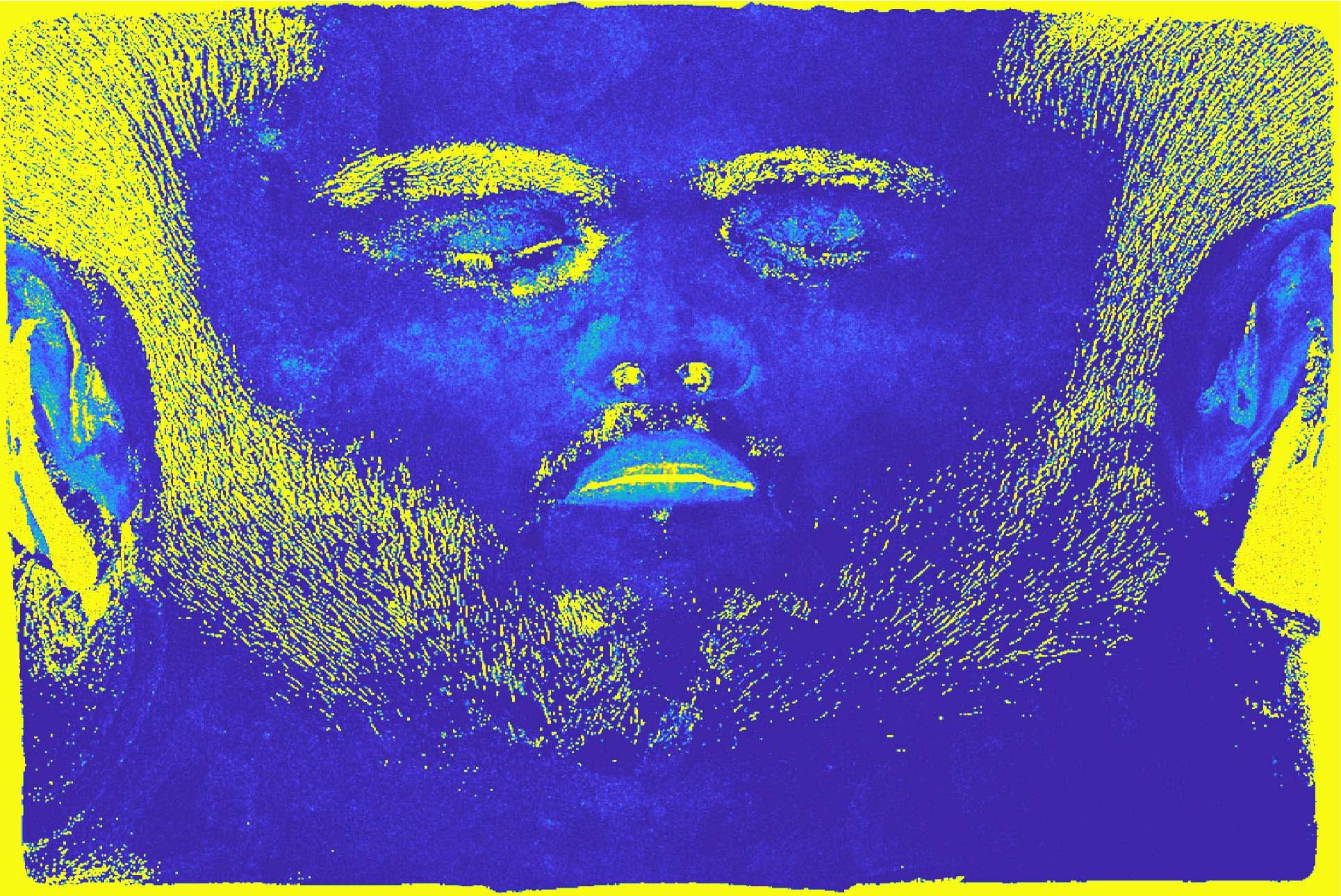}
  &
  \includegraphics[width = 0.16\textwidth]{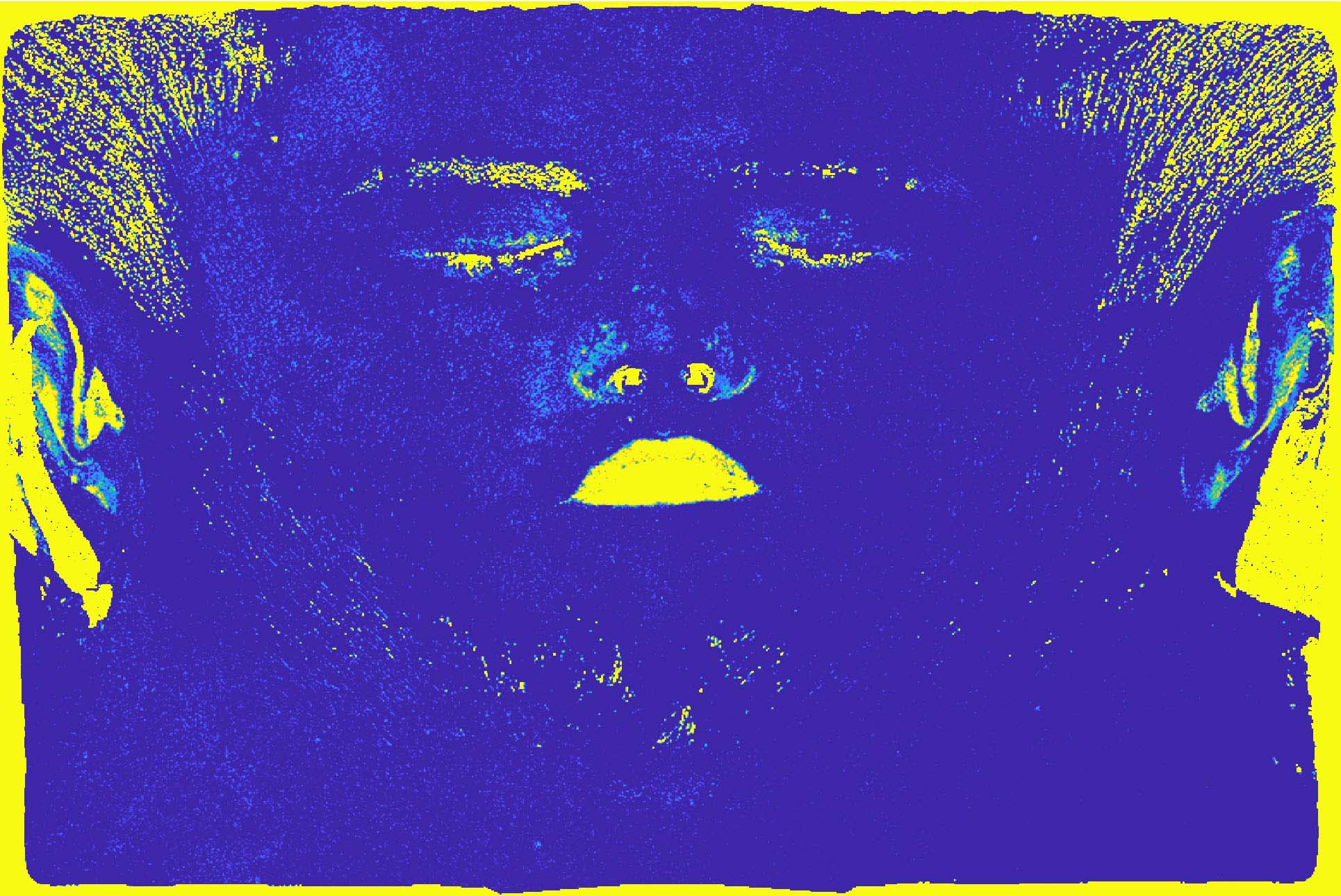}
  &
  \includegraphics[width = 0.16\textwidth]{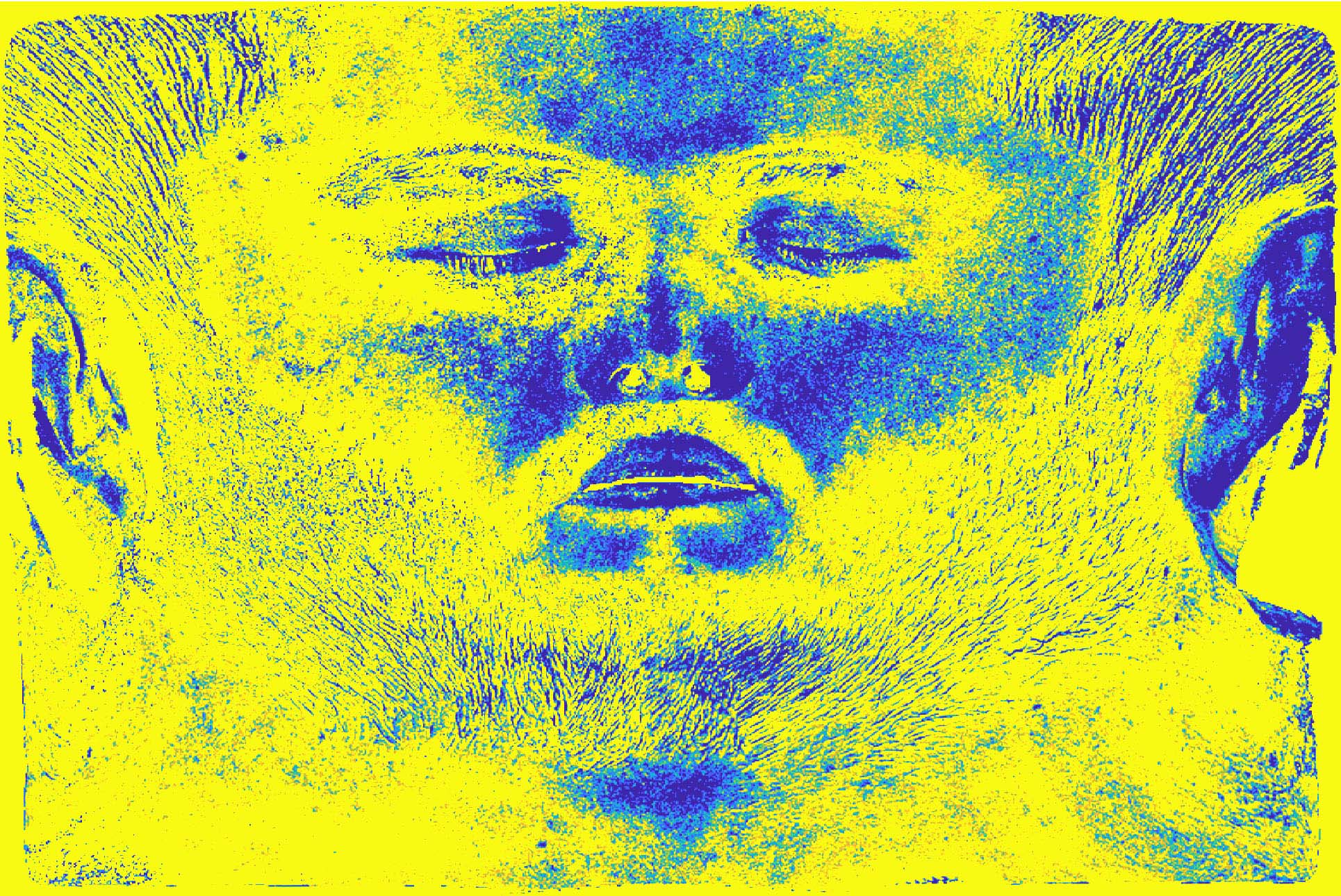}
  &
  \includegraphics[width =  0.16\textwidth]{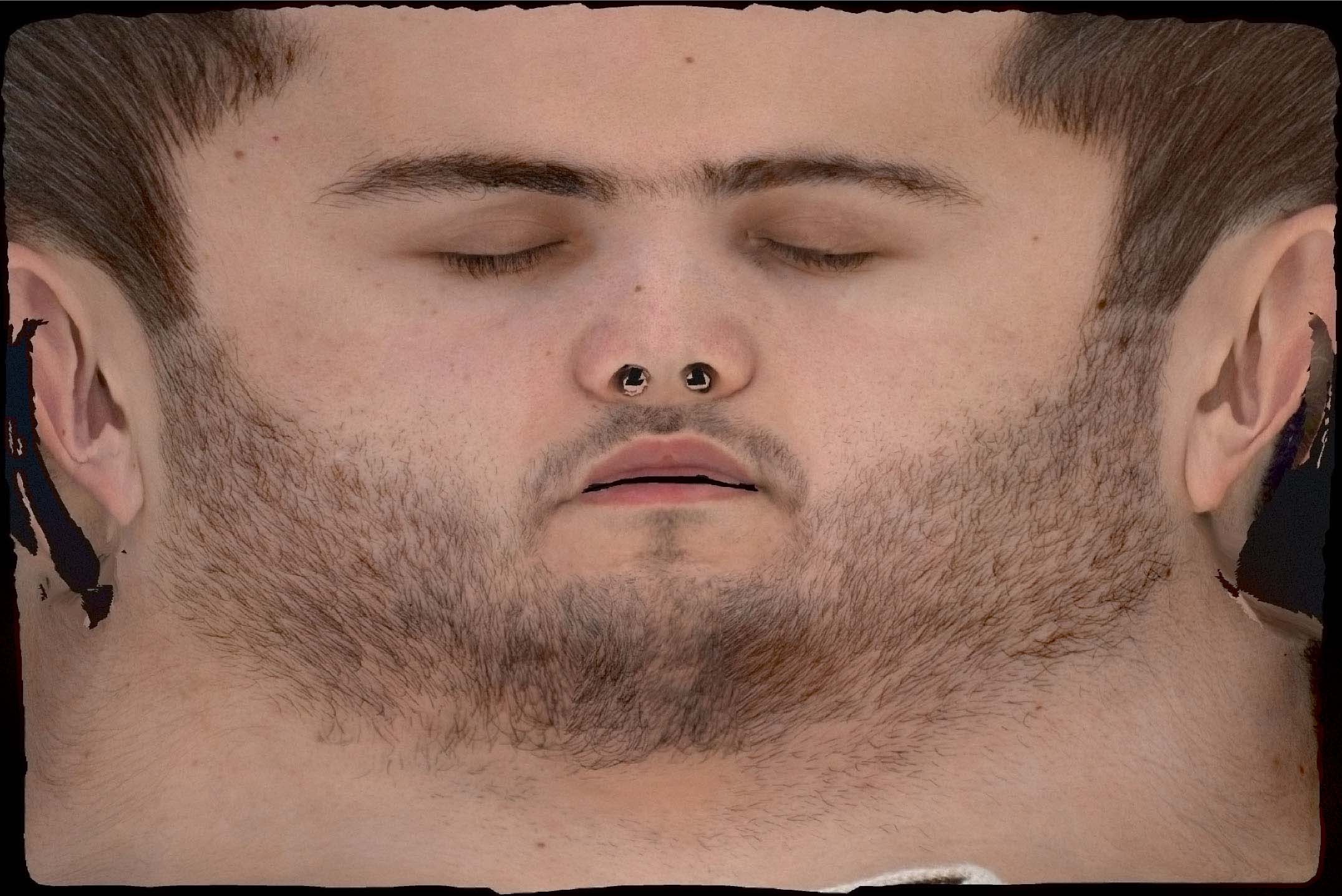}
  &
  \includegraphics[width =  0.16\textwidth]{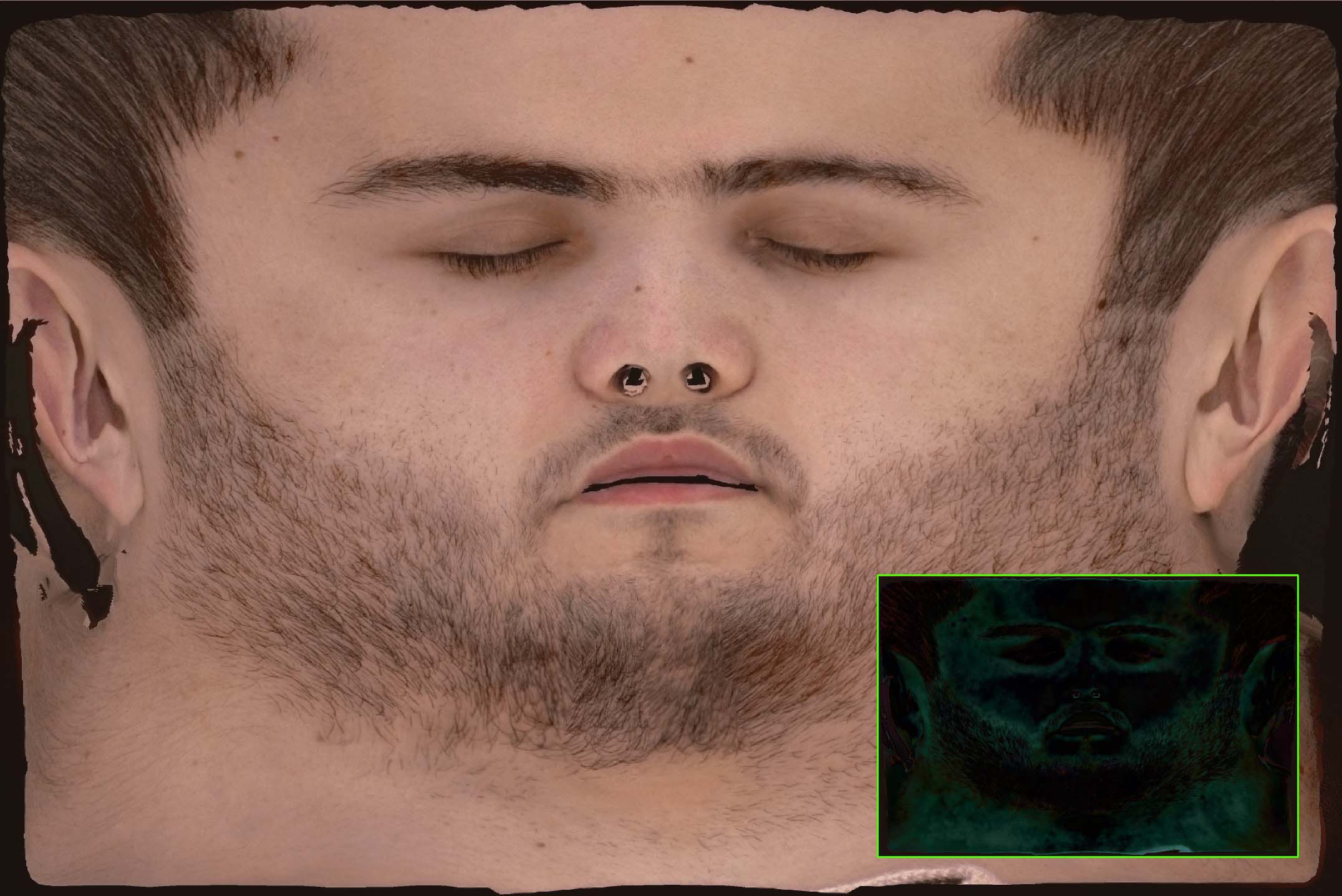} \\
  
   & \textsc{Melanin} & \multicolumn{2}{c}{\textsc{Hemoglobin}} & \textsc{Melanin ratio} & \textsc{Hemoglobin ratio} & \textsc{Thickness} \\
   
  \begin{sideways}\hspace{0.5cm}\textsc{Ours}\end{sideways}
  &
  \includegraphics[trim= 407 179 0 120, clip, width = 0.16\textwidth]{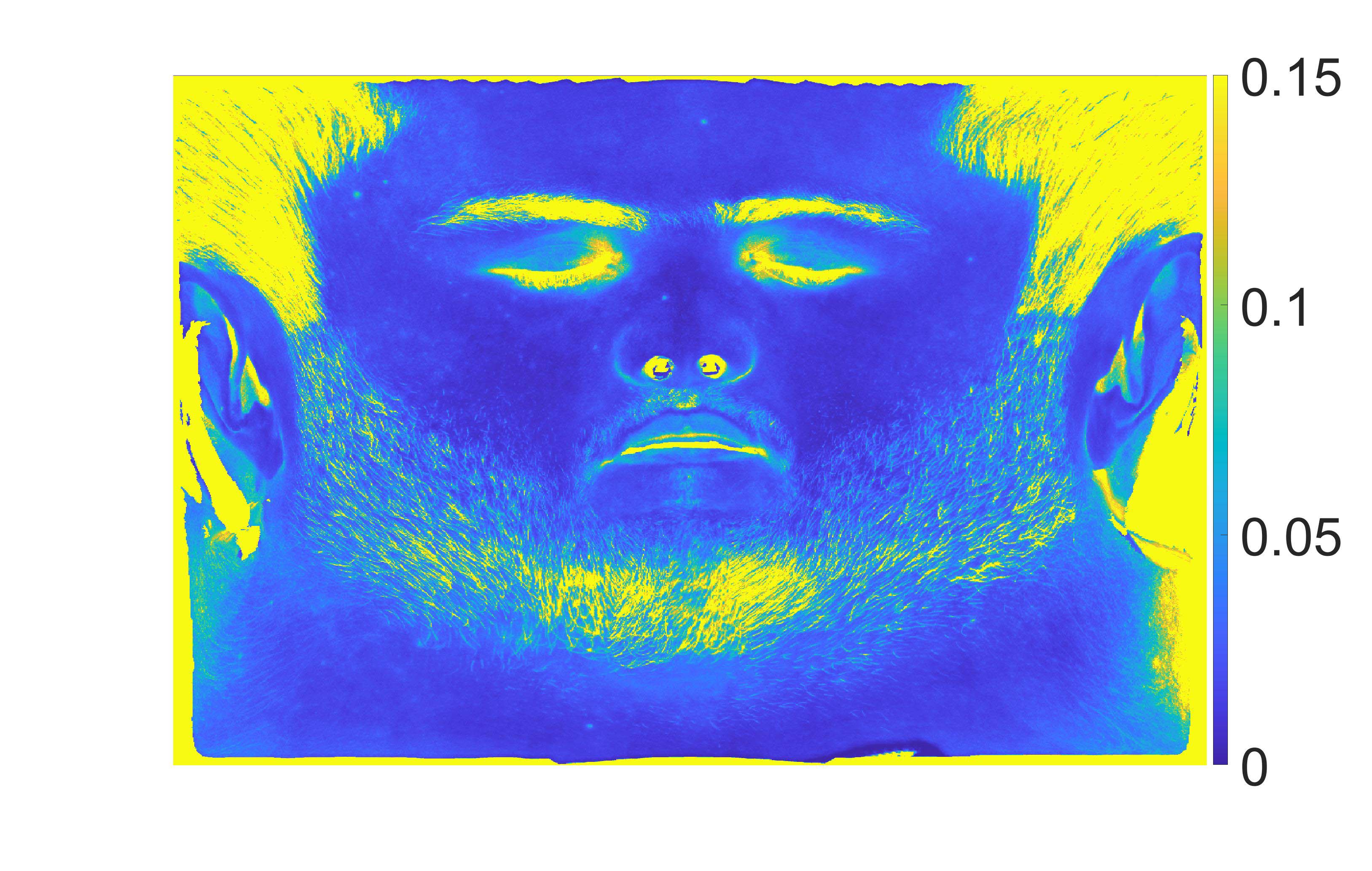}
  &
  \multicolumn{2}{c}{\includegraphics[trim=  407 179 0 120, clip, width =  0.16\textwidth]{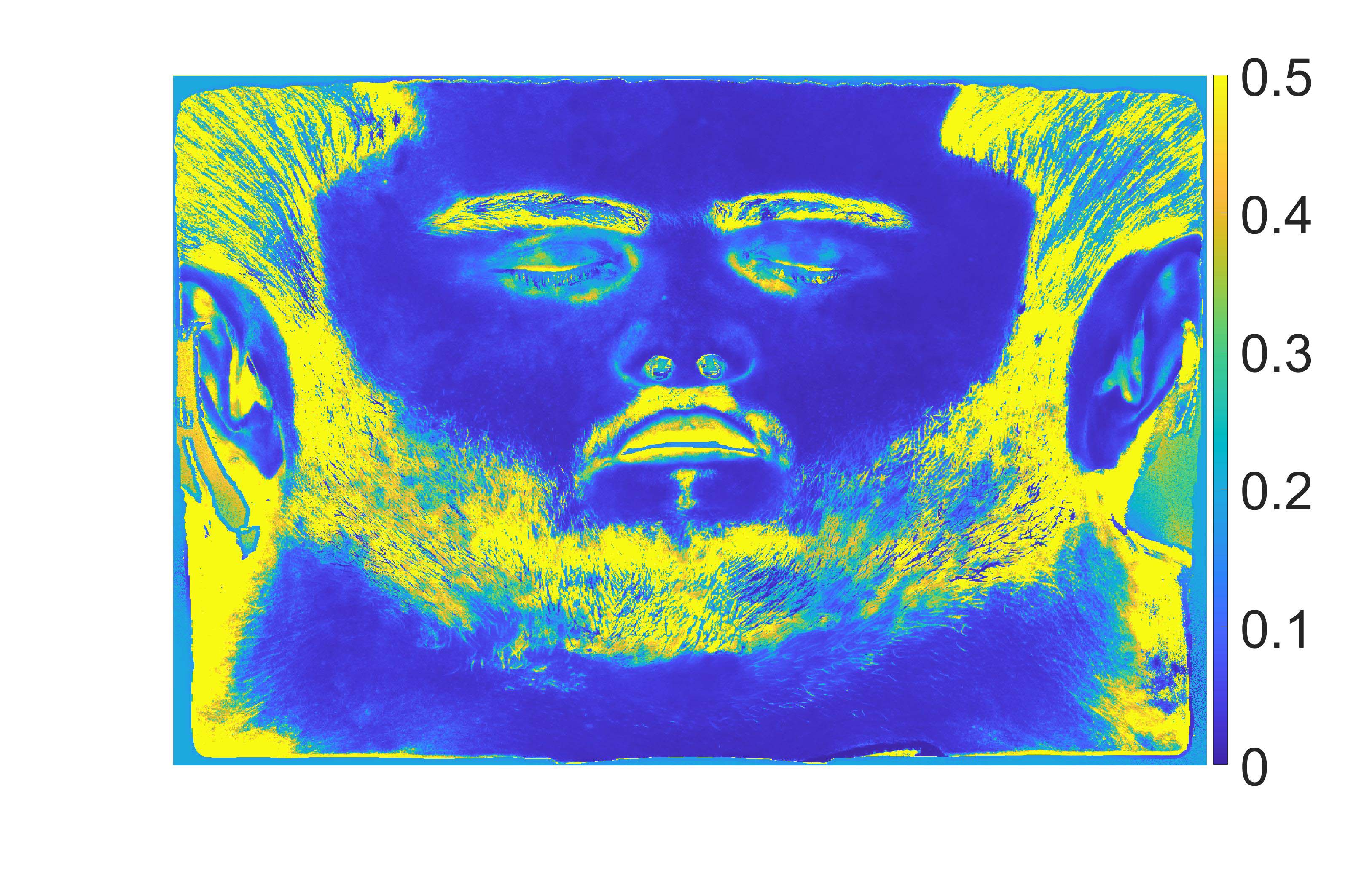}}
  &
  \includegraphics[trim= 407 179 0 120, clip, width = 0.16\textwidth]{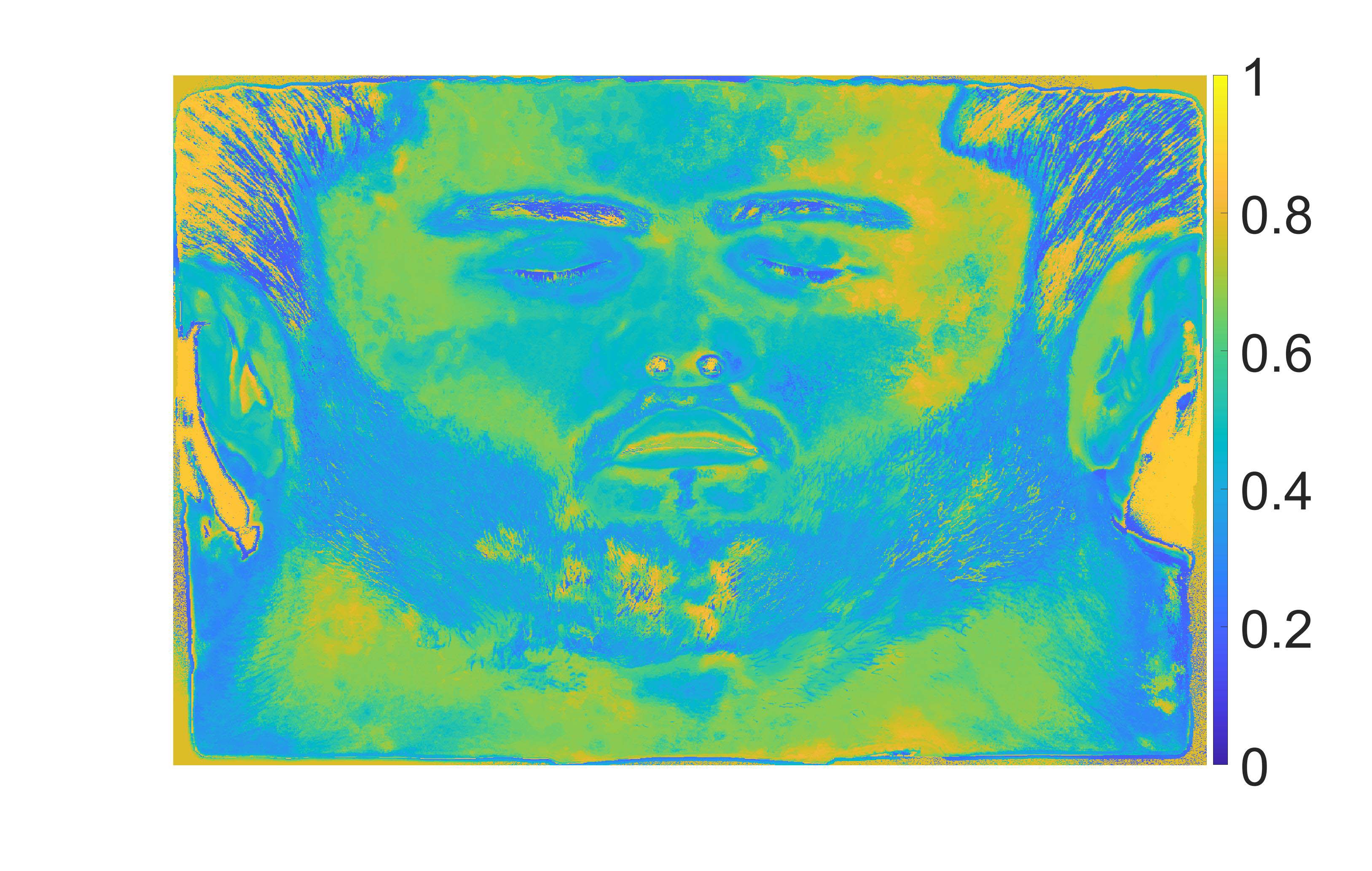}
  &
  \includegraphics[trim=  407 179 0 120, clip, width = 0.16\textwidth]{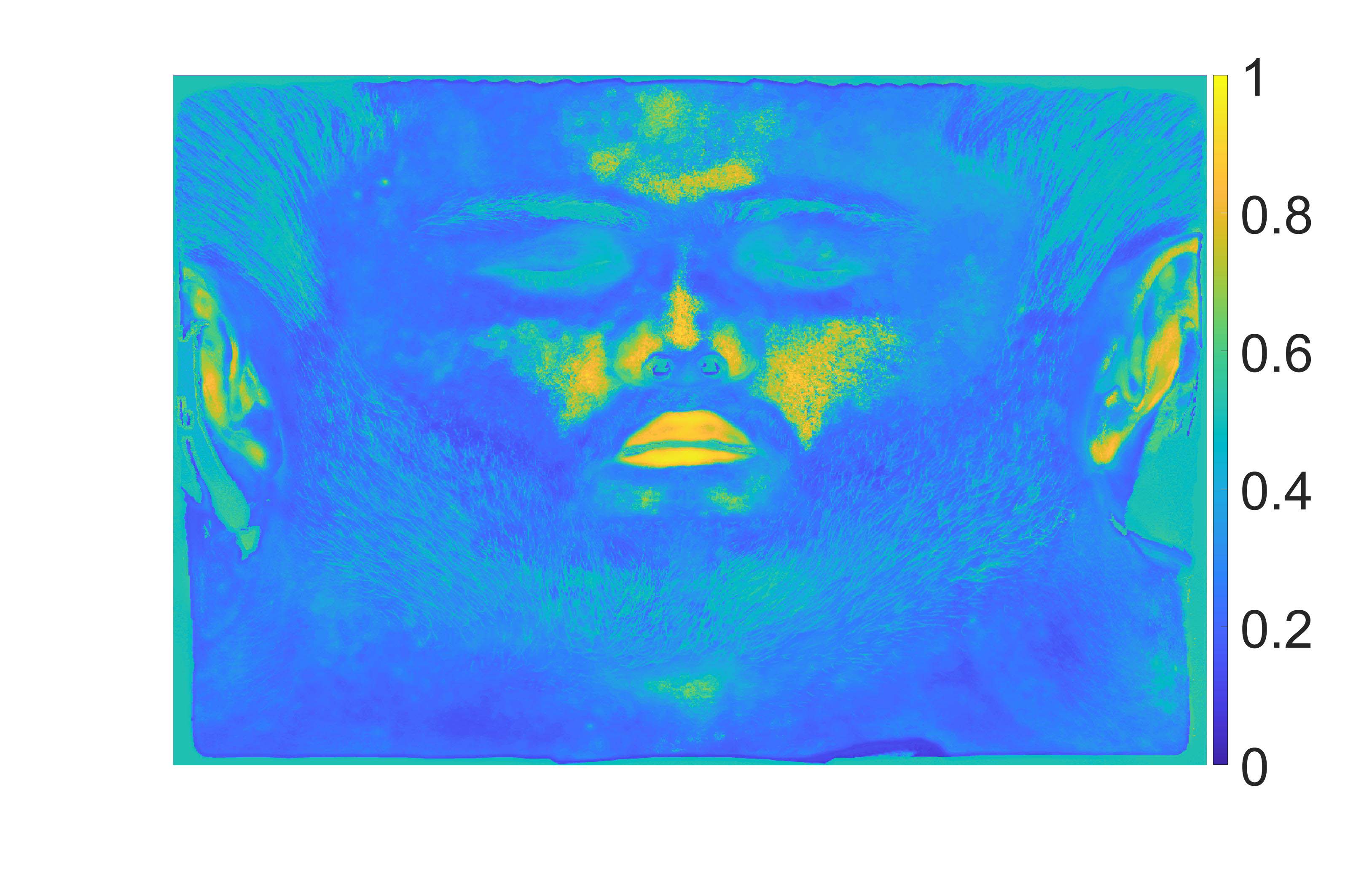}
  &
  \includegraphics[trim=  407 179 0 120, clip, width = 0.16\textwidth]{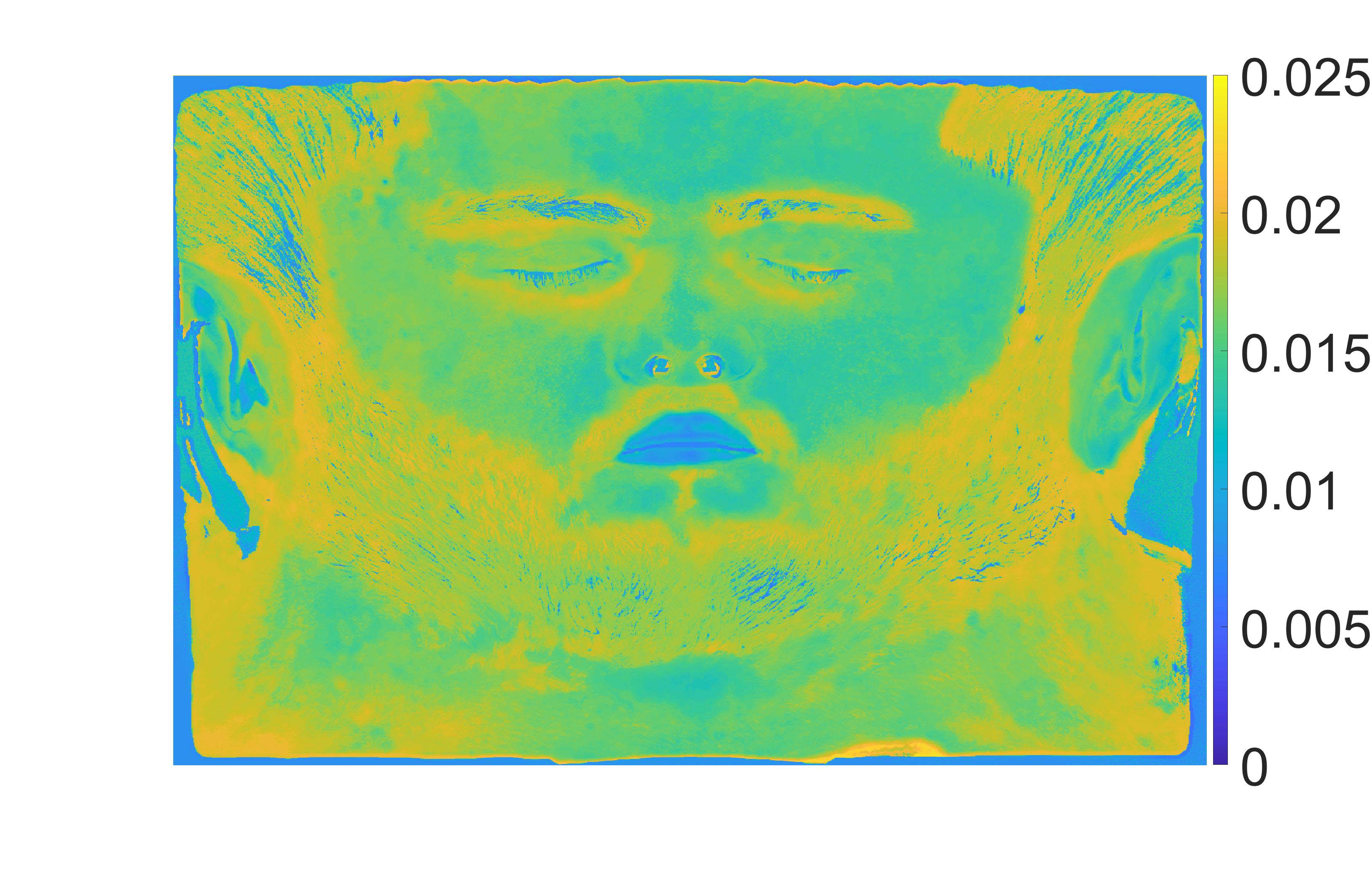}

 \end{tabular}
 \vspace{0mm}
  \caption{Albedo map from~\cite{gitlina2020practical}, comparison of their estimated parameters (top row) and ours (bottom row). On top right, we converted their albedo map from D65’ illuminant to standard D65, and removed some baked lighting (see Supplemental, Section 1). Our model is able to reconstruct the albedo with MSE = 0.1608 (abs. error in the inset). MSE = 0.0687 when a mask over hair regions is applied as in Figure~\ref{fig:albedoreconstruction}. While the common parameters with related work are similar, our new parameters predict well features like the reddish areas due to blood oxygenation, removing the need of extra blood in the epidermis. Or the epidermal thickness, where lips are identified as the thinnest area. The extra parameters available in our model might also explain the smoother melanin type ratio compared to \cite{gitlina2020practical}, where it is the only extra parameter their model heavily relies on to enable a closer matching to the original albedo than previous works.}
  \label{fig:comp_params2}
  \vspace{4mm}
\end{figure*}

\section{Comparisons with Previous Work}
\label{sec:comparison}
We conduct a series of comparisons with the related and recent work from~\cite{gitlina2020practical}. Not having access to the training data, we decided to reproduce the authors' results from the examples in the paper with our approach.  Figure~\ref{fig:comp_gitlina_albedos2} is a comparison on more true albedos from~\cite{gitlina2020practical}, which were obtained via the Antera device (under D65 illuminant). In Figure~\ref{fig:comp_params2} we process one of the D65' faces. This was a bit more complex, because of the conversion of the color spaces involved in some of their captures (which we outline in Supplemental). The general observation is that our reconstruction error is much lower, and we tested the model across more skin types. See the Supplemental Section 1 for additional tests.
\section{Discussion and Future Work}
In this paper, we presented our framework to recover and robustly manipulate the key biophysical properties of human skin from a given RGB albedo. This is accomplished through the combination of an expressive biophysically-inspired skin model, together with an encoder-decoder structure that maps albedo to skin properties and back. We demonstrate how such learned mapping overcomes the limitations of a Look-Up Tensor approach, proving how the latter becomes impractical and leads to quantization artifacts and shifts in colors. On the other hand, the expressiveness and robustness of our method is assessed through estimating and manipulating biophysical parameters of a variety of skin shades covering the Fitzpatrick scale.

As future work, although we consider all the chromophores in our current model, we aim to widen the albedo space to handle uncommon skin shades resulting from rare chromophore concentrations, such as the excess of bilirubin, beta-carotene, or weird pathologies related to methemoglobin. Incorporating the effect of hair roots on the light transport inside the skin would be interesting to better handle short beards or shaved skulls.

Another direction to further improve our results would be to train the neural network to learn the mapping between skin parameters and spectral albedos directly, instead of RGB. Finally, we want to look at adding the ability to automatically remove or compensate the baked lighting that exists in some captures, through analyzing the parts of the image that show incoherence in the estimated skin parameters, with the goal of reconstructing the skin properties under uncontrolled lighting scenarios like normal pictures.

\section*{Acknowledgments}
We thank Olivier Maury and Akira Orikasa for general lookdev support, Christoph Lassner for conversations about the neural model, and Tony Tung, Ronald Mallet and Yaser Sheikh for their face captures. Scans and actors for Subjects A and F were provided by Eisko~\cite{eisko}.
\printbibliography   

\appendix
\section{Specifications of Skin Chromophores}
\label{appendix1}
\begin{table}[h]
\footnotesize
\begin{tabular}{lll}
 \textbf{Parameter} &  \textbf{Description} &  \textbf{Value} \\
 \hline
 $\mu_{a_{hbO2}}$ & Oxy-Haemoglobin absorption & 2.303 $\frac{p_{hb}\varepsilon_{hbO2}}{w_{hb}}$ \\
 
  $\mu_{a_{hb}}$ & Deoxy-Haemoglobin absorption & 2.303 $\frac{p_{hb}\varepsilon_{hb}}{w_{hb}}$ \\
  
 $\varepsilon_{hbO2}$ & Oxy-Haemoglobin Extinction &  \cite{jacques2013optical} \\
 
 $\varepsilon_{hb}$ & Deoxy-Haemoglobin Extinction &  \cite{jacques2013optical} \\
 
 $p_{hb}$ & Haemoglobin Concentration &  150 \\
 $w_{hb}$ & Molar weight of Haemoglobin  &  64500 \\
 \hline
 $\mu_{a_{eu}}$ & Eumelanin absorption & $6.6 x 10^{11}$ $\lambda^{-3.33}$ \\
 $\mu_{a_{pheo}}$ & Pheomelanin absorption & $2.9 x 10^{15}$ $\lambda^{-4.75}$ \\
 \hline
 $p_{bil}$ & Bilirubin Concentration &  0.05 \\
 $w_{bil}$ &  Molar weight of bilirubin & 584.66\\
 \hline
 $p_{\beta-c_{e}}$ & $\beta$-carotene Concentration (Epidermis) &  $2.1 x 10^{-4}$ \\
 $p_{\beta-c_{d}}$ & $\beta$-carotene Concentration (Dermis) &  $7 x 10^{-5}$ \\
 $w_{\beta-c}$ & Molar weight of $\beta$-carotene &  536.8726\\
\end{tabular}
\caption{Chromophore specifications. The absorption coefficient is defined in $cm^{-1}$; the extinction coefficient $\varepsilon_{c}$ in $\frac{cm^{-1}}{moles / liter}$; the concentration of the chromophore $p_{c}$ in g/L; and the molar weight $w_{c}$ in g/mol. The absorption of melanins $\mu_{a_{eu}}$ and $\mu_{a_{pheo}}$ is defined through a fit from~\cite{donner2006spectral} to the measurements from~\cite{jacques1991melanosome} and~\cite{sarna2006physical} respectively. The 2.303 coefficient in $\mu_{a_{hb}}$ comes from deriving a factor of $ln\left(0\right)$, since $\varepsilon$ has been historically recorded in such base 10 nomenclature from measurements of old spectrometers in literature. Finally, oxy and deoxy haemoglobin extinction can be found tabulated in~\cite{jacquesData}.}
\label{table:chromos}
\end{table}

\end{document}


\maketitle
\begin{abstract}
This is the supplemental material for the paper Estimation of Spectral Biophysical Skin Properties from Captured RGB Albedo. It includes further comparisons with previous work, implementation details and some decisions on the model, like an analysis on Kubelka-Munk layering approaches and why we abandoned them, or details on the albedo inversion and the usage of the albedo maps during rendering. Finally we show more rendering results for our edits of the extracted chromophores.

\ccsdesc[500]{Computing methodologies~Reflectance modeling}
\ccsdesc[500]{Computing methodologies~Reconstruction}

\printccsdesc   
\end{abstract}  

\vspace{-15mm}

\definecolor{gray}{rgb}{0.5,0.5,0.5}
\definecolor{purple}{rgb}{0.7,0.3,0.7}
\definecolor{blue}{rgb}{0.3,0.5,1}
\definecolor{darkblue}{rgb}{0,0,0.6}
\definecolor{orange}{rgb}{1,0.5,0} 
\definecolor{red}{rgb}{1,0,0}
\definecolor{green}{rgb}{0,0.5,0}
\definecolor{purple}{rgb}{0.5,0,1}

\newcommand{\new}[1]{{\textcolor{darkblue}{#1}}}


\newcommand{\IGNORE}[1]{}

\definecolor{MyDarkBlue}{rgb}{0,0.08,1}
\definecolor{MyDarkGreen}{rgb}{0.02,0.6,0.02}
\definecolor{MyDarkRed}{rgb}{0.8,0.02,0.02}
\definecolor{MyDarkOrange}{rgb}{0.70,0.35,0.02}
\definecolor{MyPurple}{rgb}{0.43,0,1.}
\definecolor{MyRed}{rgb}{1.0,0.0,0.0}
\definecolor{MyGold}{rgb}{0.75,0.6,0.12}
\definecolor{MyDarkgray}{rgb}{0.66, 0.66, 0.66}

\newcommand{\chery}[1]{{\color{green}\emph{Chery: #1}}}
\newcommand{\carlos}[1]{{\color{blue}\emph{Carlos: #1}}}
\newcommand{\mandy}[1]{{\color{purple}\emph{Mandy: #1}}}


\section{Further Comparisons with Previous Work}
\begin{figure}[h!]
 \contourlength{0.1em}%
 \centering
 \vspace{5mm}
 \hspace*{-4.5mm}%
  \begin{tabular}{c@{\;}c@{\;}c@{\;}c}
 \textsc{Input D65'} & \textsc{D65} & \textsc{D65*} & \textsc{|D65*-D65|}
  
  \\
  \includegraphics[width = 0.25\columnwidth]{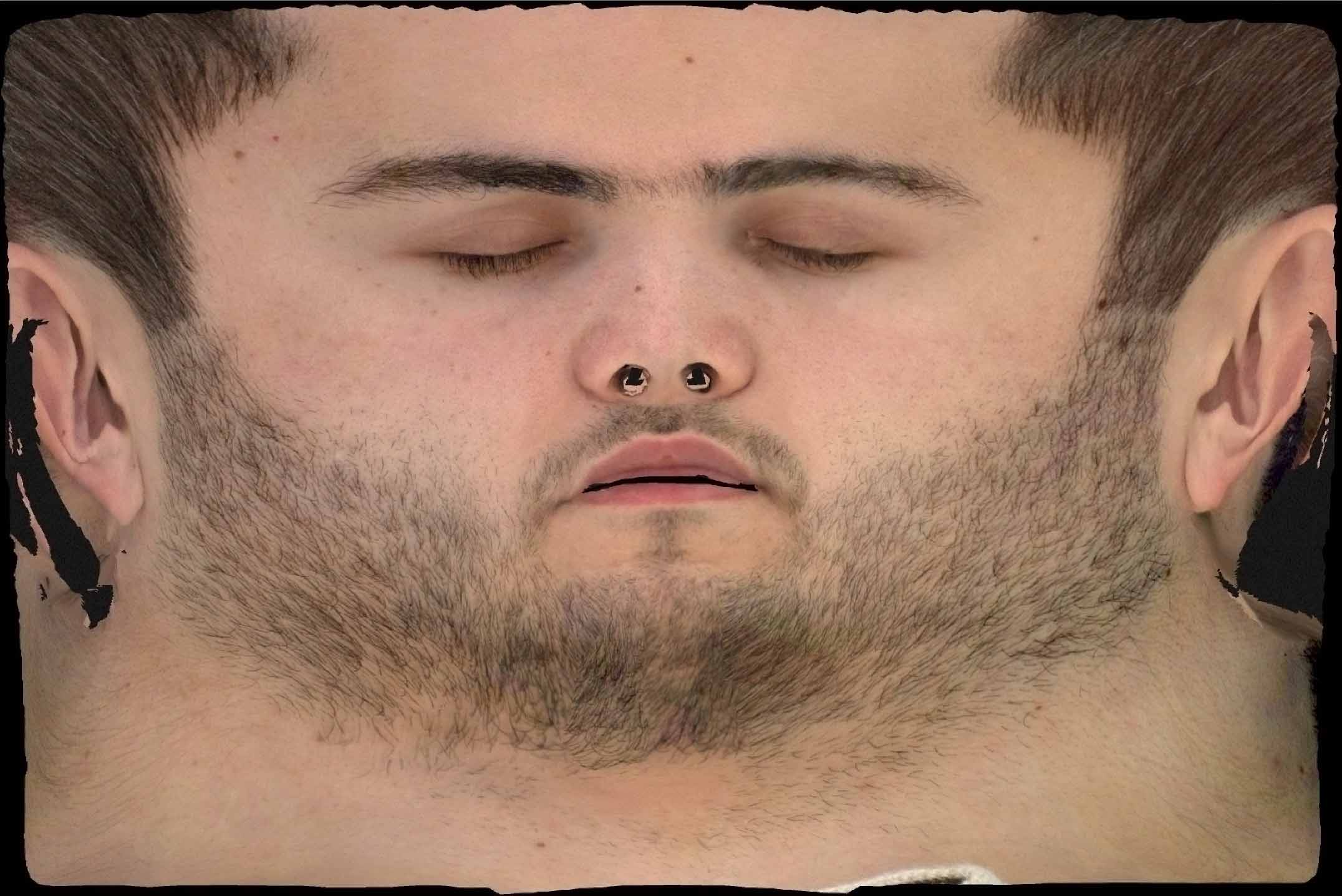}
  &
  \includegraphics[width = 0.25\columnwidth]{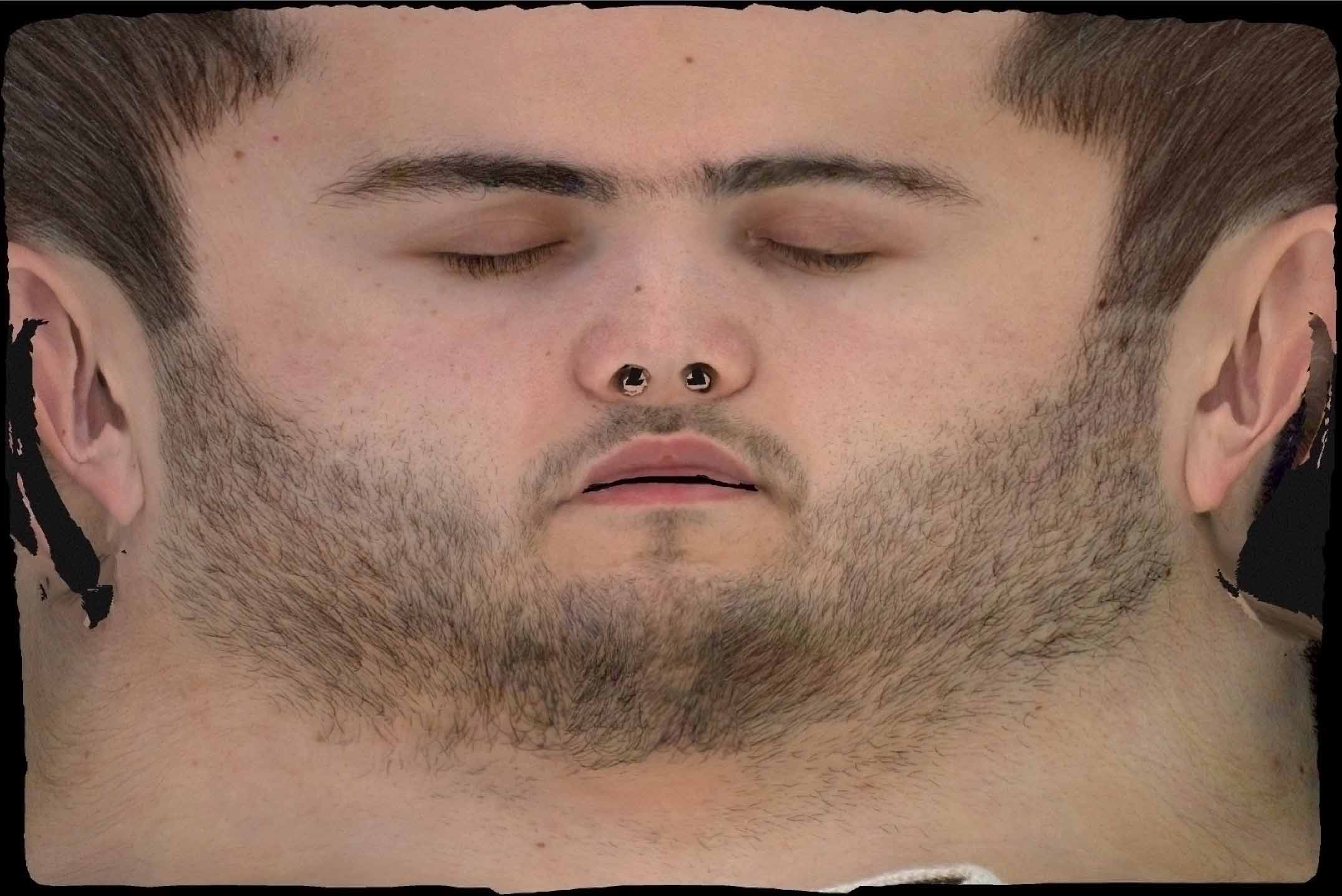}
  &
  \includegraphics[width = 0.25\columnwidth]{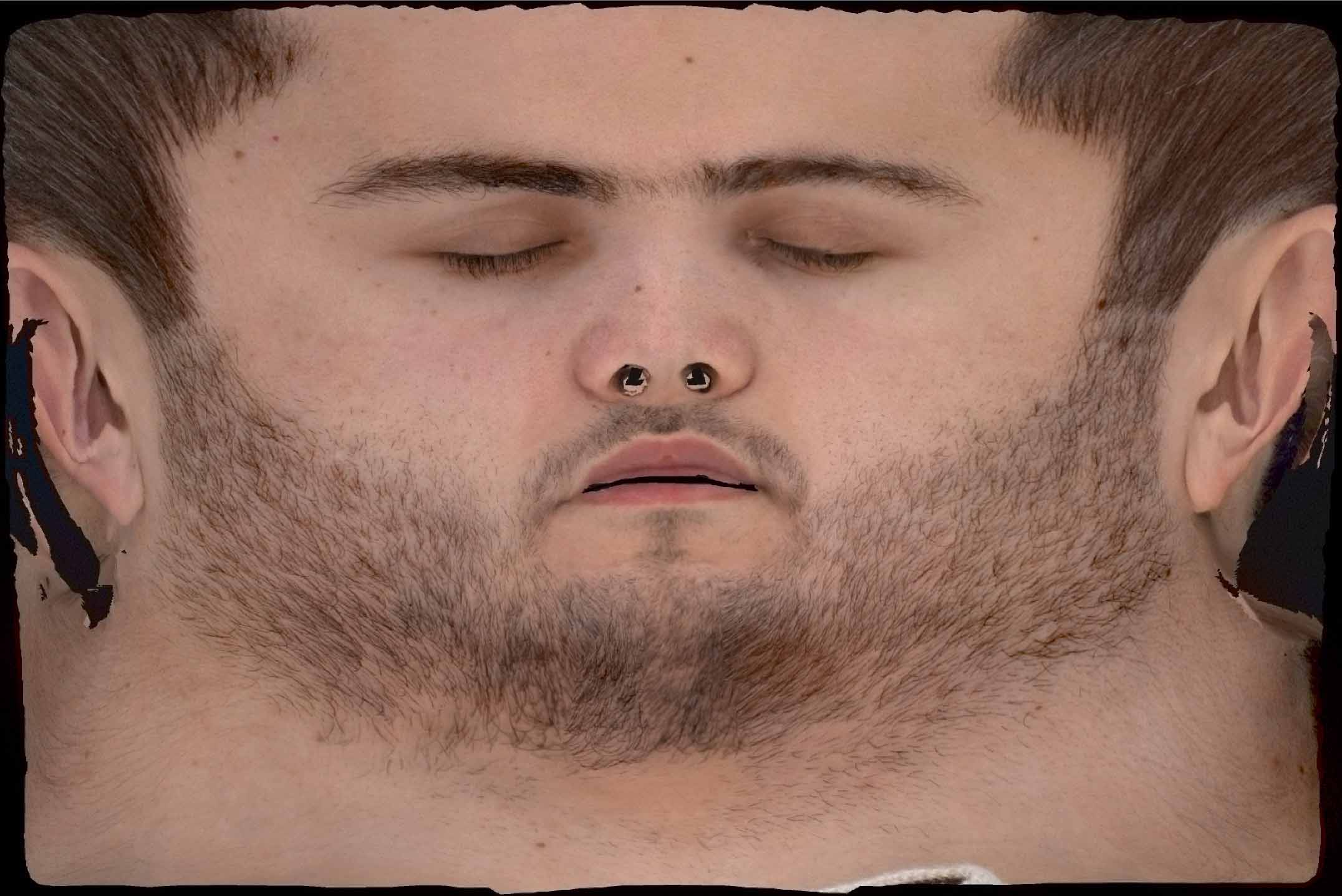}
  &
  \includegraphics[width = 0.25\columnwidth]{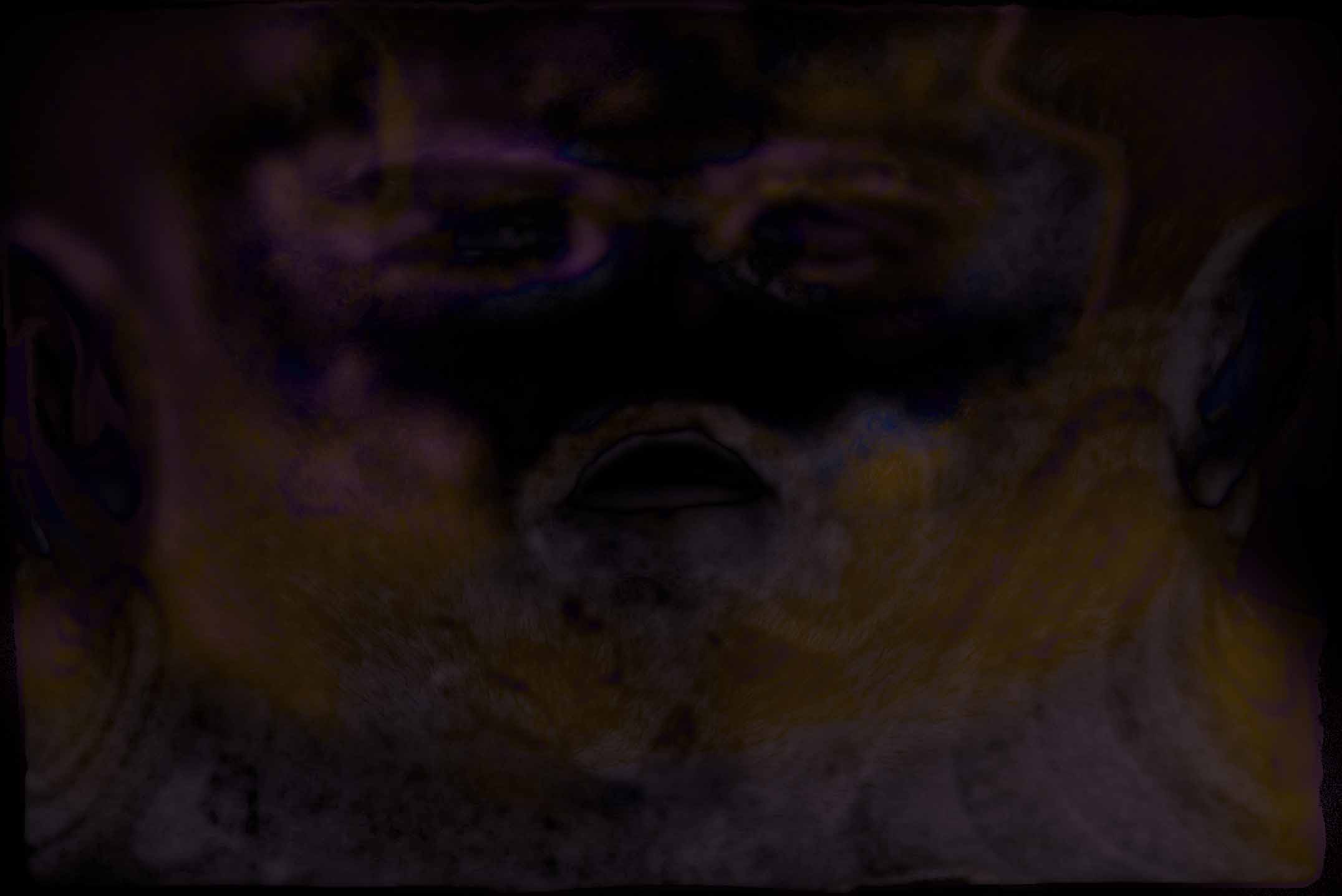}

  \\
  \includegraphics[width = 0.25\columnwidth]{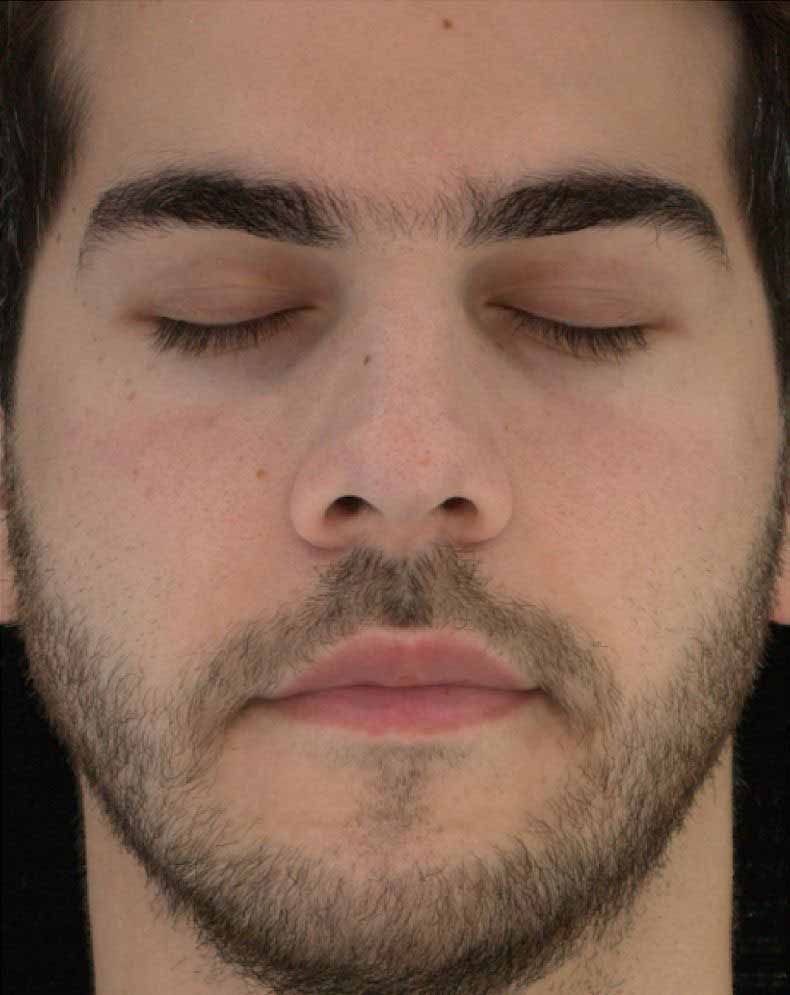}
  &
  \includegraphics[width = 0.25\columnwidth]{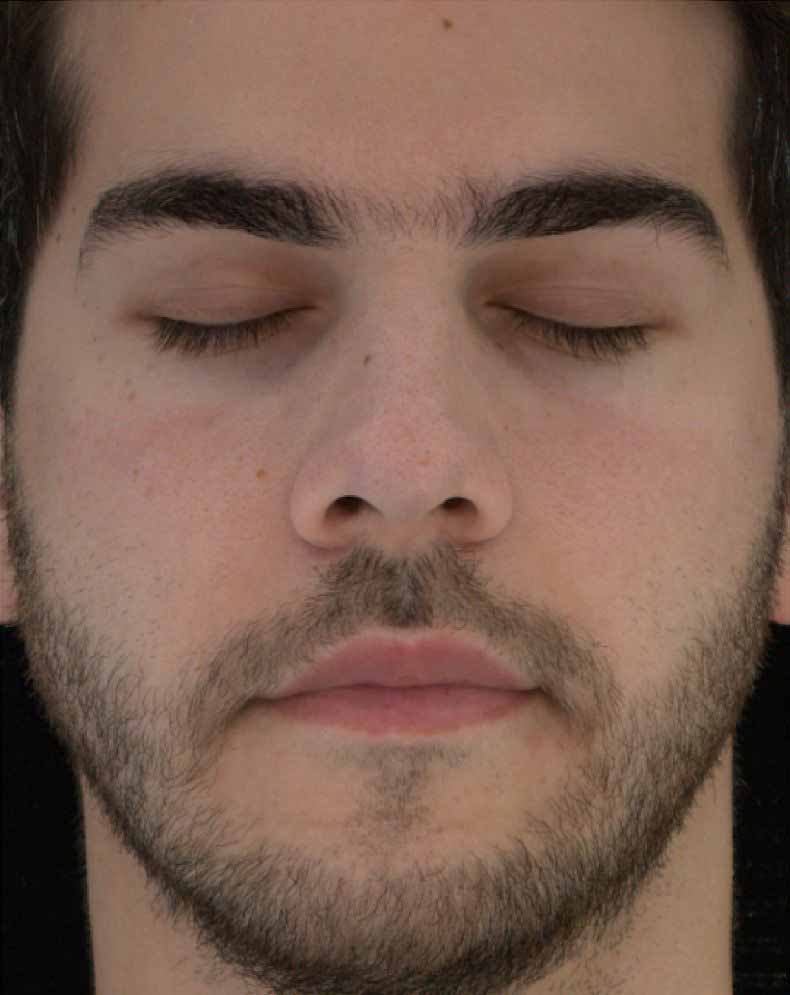}
  &
  \includegraphics[width = 0.25\columnwidth]{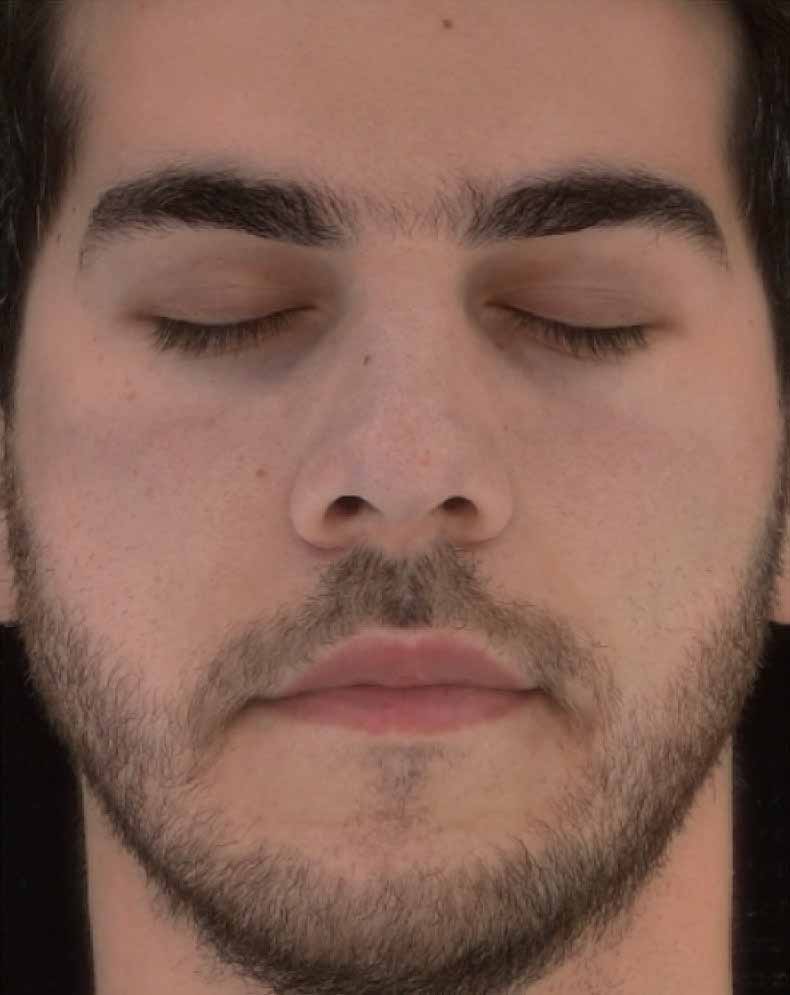}
  &
  \includegraphics[width = 0.25\columnwidth]{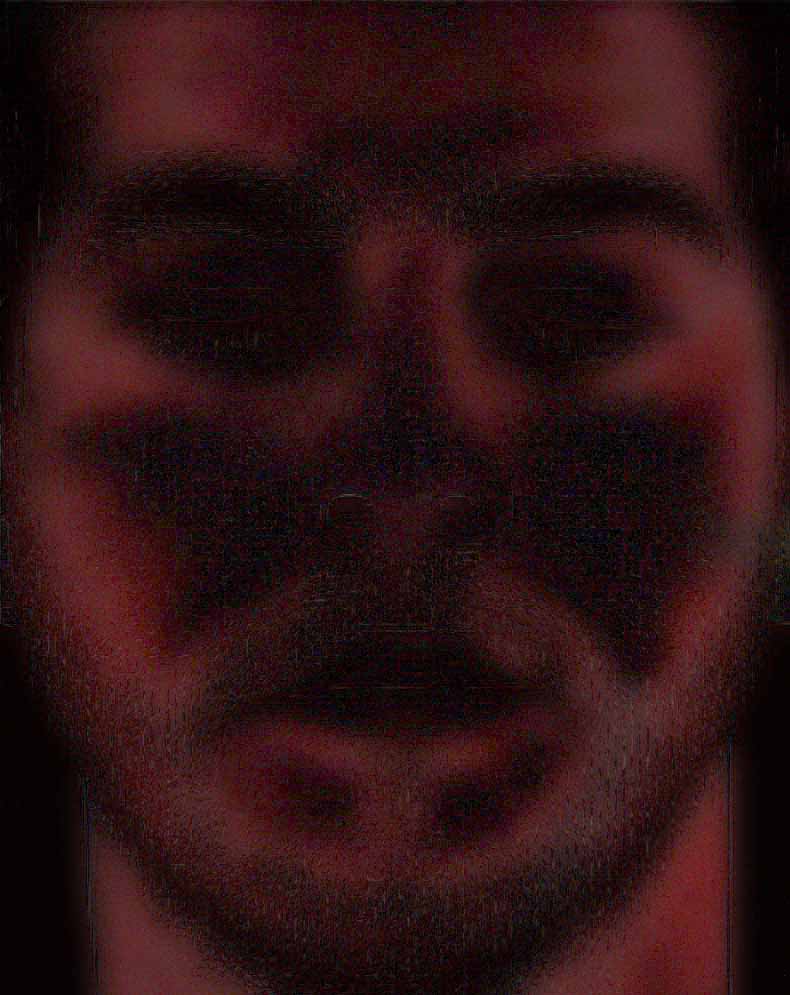}

  \\
  \includegraphics[width = 0.25\columnwidth]{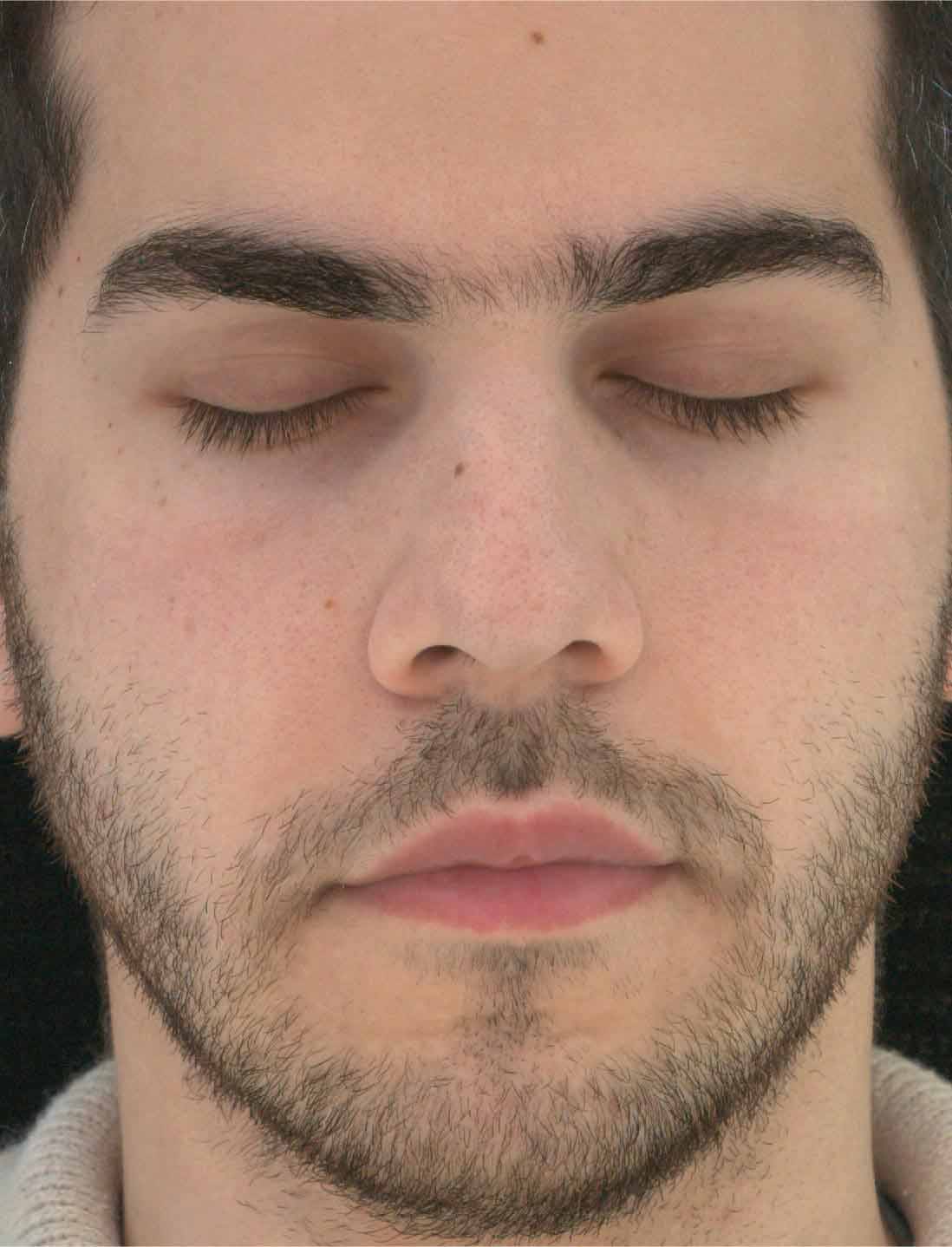}
  &
  \includegraphics[width = 0.25\columnwidth]{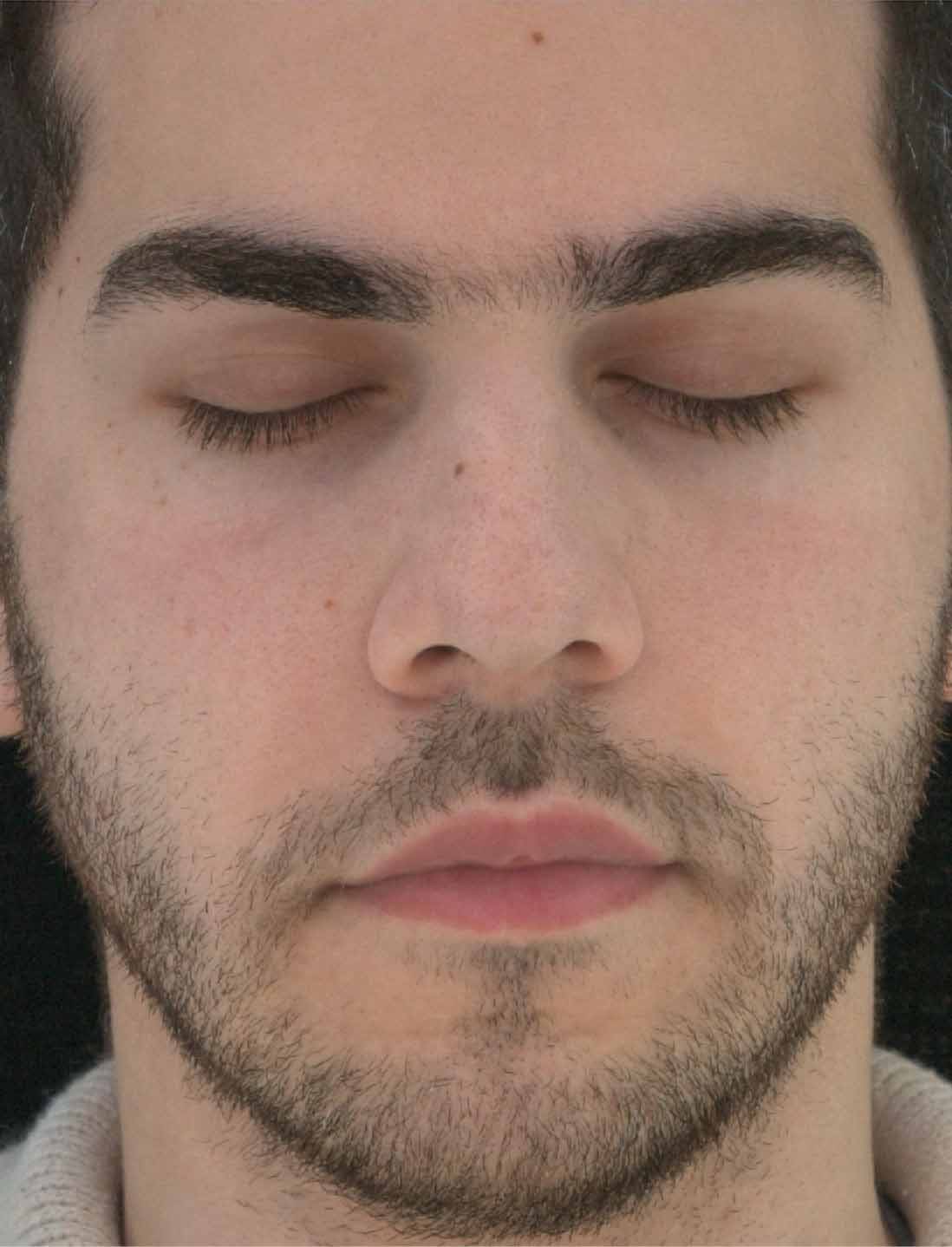}
  &
  \includegraphics[width = 0.25\columnwidth]{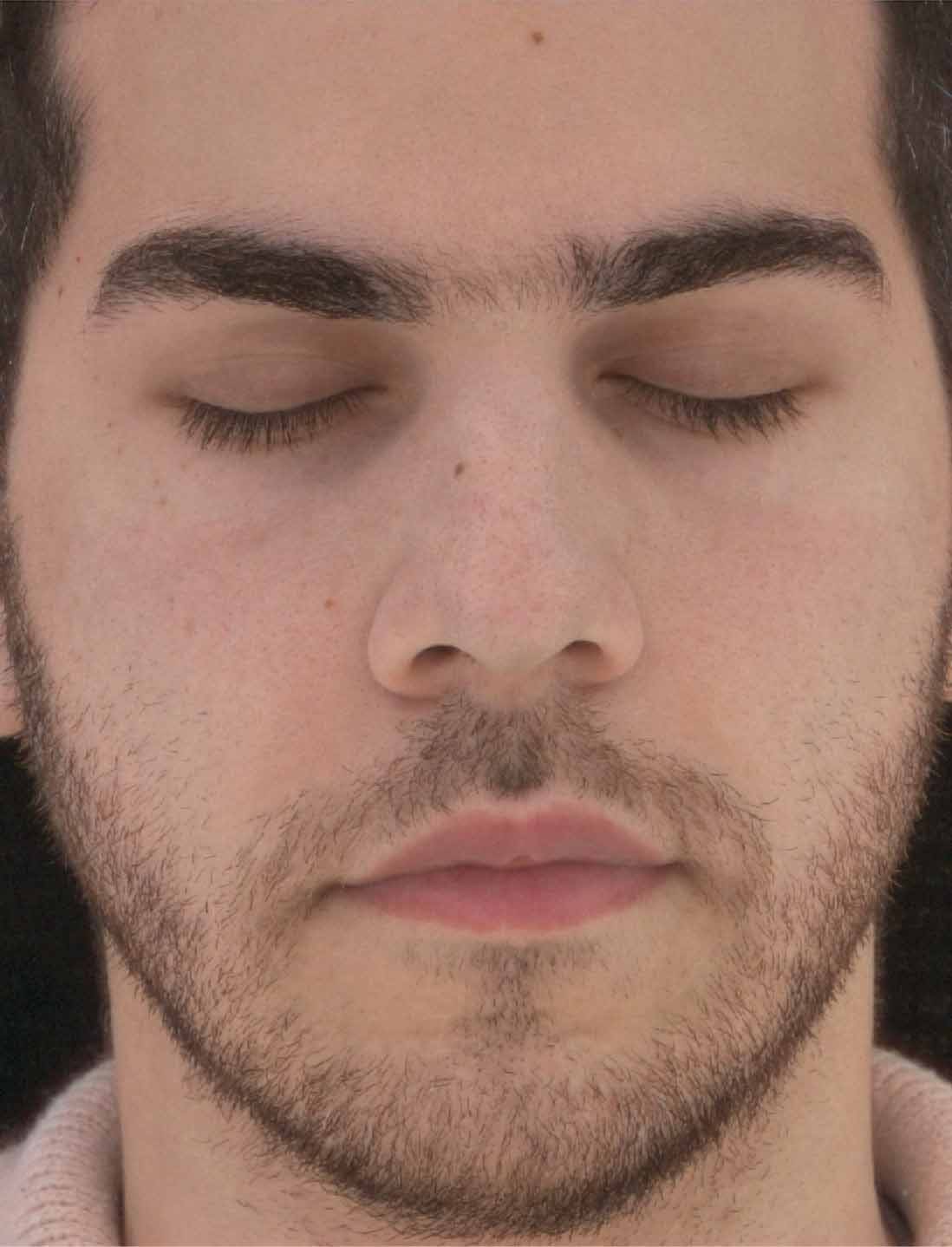}
  &
  \includegraphics[width = 0.25\columnwidth]{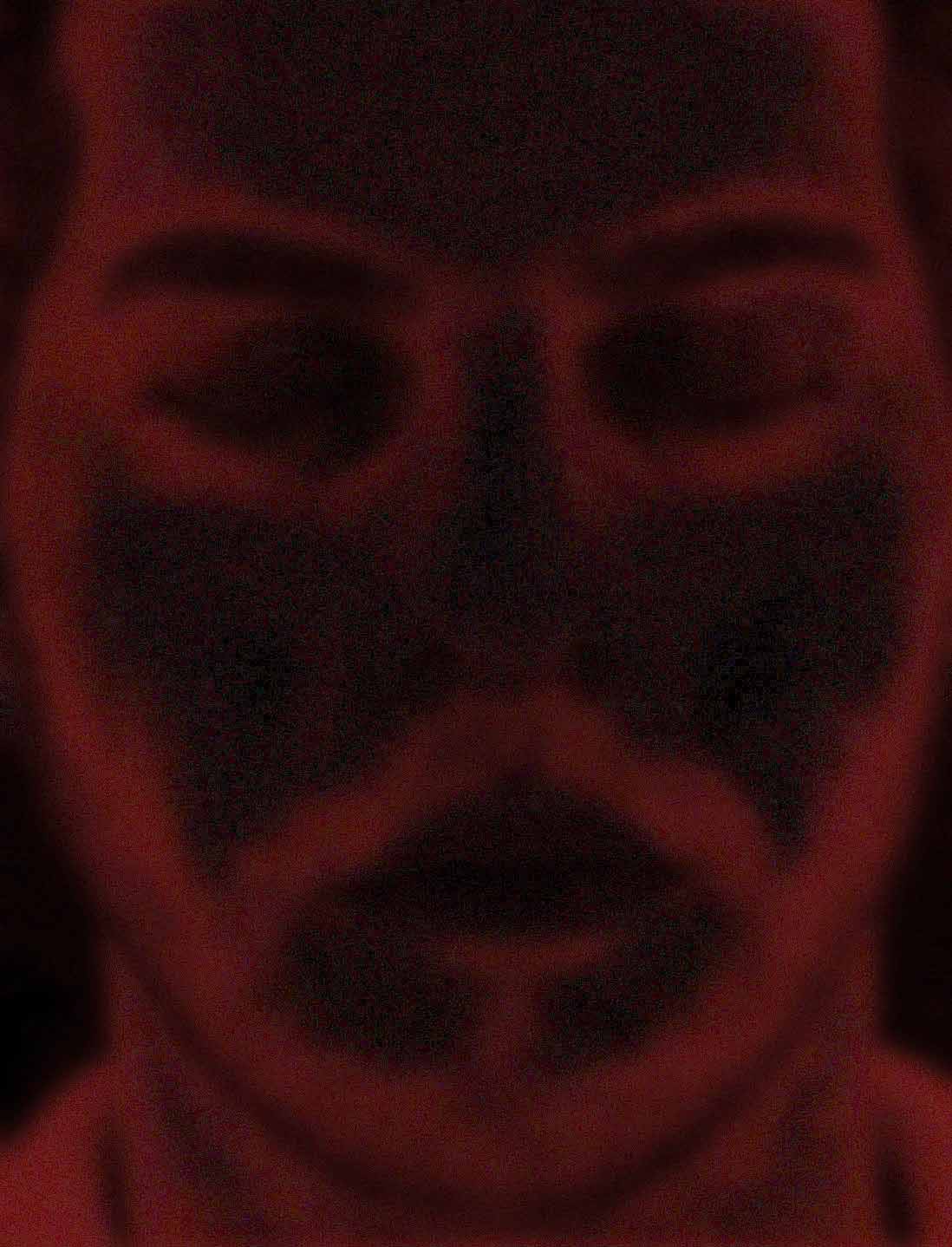}

 \end{tabular}
  \caption{\emph{D65' to D65 conversion and baked lighting removal}. Each row is a different capture. From left to right: some of the original data from~\cite{gitlina2020practical} were obtained in a specially designed color space (Input D65'), and we bring them via spectral upsampling and downsampling to our D65 illuminant, for later processing. Lastly,  we manually remove the still present baked lighting (D65*, absolute difference shown last). }
  \label{fig:illuminant_baked_lighting}
  \vspace{-6mm}
\end{figure}

The work we compare directly~\cite{gitlina2020practical} relies partly on the creation of a new illuminant, that the authors call D65', intending to represent the response of their capturing LEDs. We assume D65. In order to process their input data with our model, we thus first need to convert from that D65' color space into the regular D65. The paper provides the spectral curve of this D65' in the form of a plot. We can then upsample the D65' RGB inputs into spectral domain~\cite{Mallett2019}, then downsample back to RGB with the D65 illuminant. This process is shown in Figure~\ref{fig:illuminant_baked_lighting}. Note that, for some tests, we go further, by also editing out the baked lighting (noticeable at grazing angles and through a general green cast) still present in the images provided, since this is not desirable for the chromophore extraction and the reconstruction.

In Figure~\ref{fig:comp_params} we show our reconstructions of the skin patches captured by previous work~\cite{gitlina2020practical}. Figures~\ref{fig:comp_gitlina_albedos} and~\ref{fig:comp_gitlina_params} demonstrate our reconstruction and chromophore contents, on the main subjects from~\cite{gitlina2020practical}. These particular input images are actually photographs (in D65', so we resort to our conversion approach from above), not albedos. As a consequence, the overall parameter estimations are questionable. Nonetheless, they compare pretty well to the results from the original paper.

\begin{figure}[t]
  \centering
  \includegraphics[width=\linewidth]{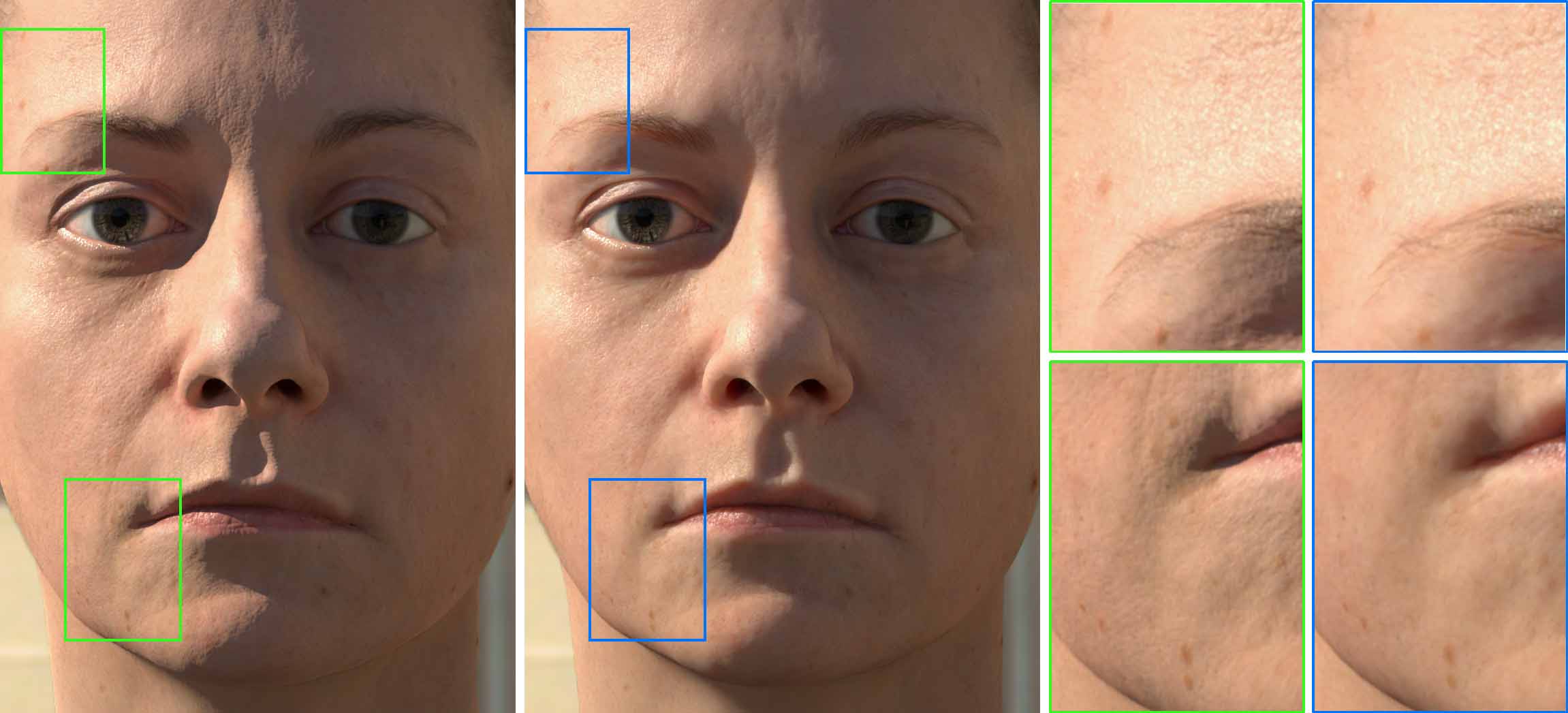}
  \caption{\emph{Lambertian vs Random Walk Subsurface Scattering in final rendering stage.} The numerical albedo inversion allows to drive the attenuation by the albedo map, ensuring skin details are preserved and not over blurred by the 3D SSS simulation.}
  \label{fig:sss_blur}
  \vspace{-6mm}
\end{figure}

\section{Implementation Details and Decisions on the Model}
\subsection{Using the model In Rendering}
In the spirit of a well known technique in production~\cite{wrenninge2017path}, to preserve the skin details coming from the albedo texture, we make use of the reconstructed and edited albedo maps and compute a 3D random walk subsurface scattering solution that relies on a numerical albedo inversion around the mean free path and accounting for the anisotropy factor $g$ (see Figure~\ref{fig:sss_blur}).


\subsection{Kubelka-Munk}
For computing the diffuse reflectance of a skin patch, we first started by using a Kubelka-Munk (KM)~\cite{kubelka1931article} layering model, following previous work~\cite{alotaibi2017biophysical}.

We experimented through different variations of the KM-based model, all of them suffering from lack of expressiveness and requiring many ad-hoc parameters to tune the layering quantities. These issues with KM approaches stem mostly from the fact that the theory was initially derived for pigments, and it is known to exhibit some limitations, like inaccuracies in cases of dark shades or thin films~\cite{choudhury2014principles}, that can make this technique not adequate in our context.

\begin{figure}[t]
 \contourlength{0.1em}%
 \centering
 \hspace*{-4.5mm}%
  \begin{tabular}{l@{\;}c@{\;}c@{\;}c@{\;}}
 & \textsc{Input} & \textsc{Reconstructed} & \textsc{Abs. Error}
  \\
  \begin{sideways}\hspace{0.3cm}\textsc{Type I female}\end{sideways}
  &
  \begin{overpic}[width=0.32\columnwidth]{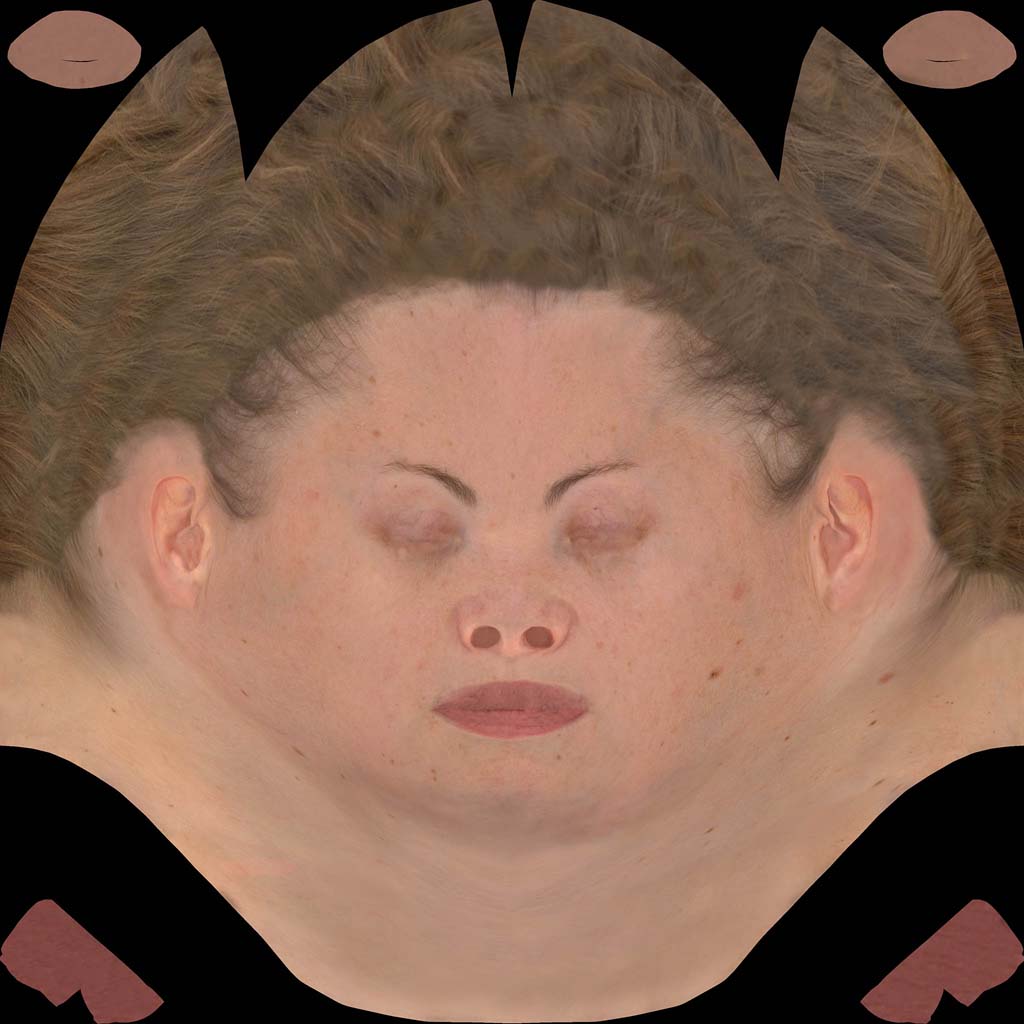}
  \end{overpic}
 &
  \begin{overpic}[width=0.32\columnwidth]{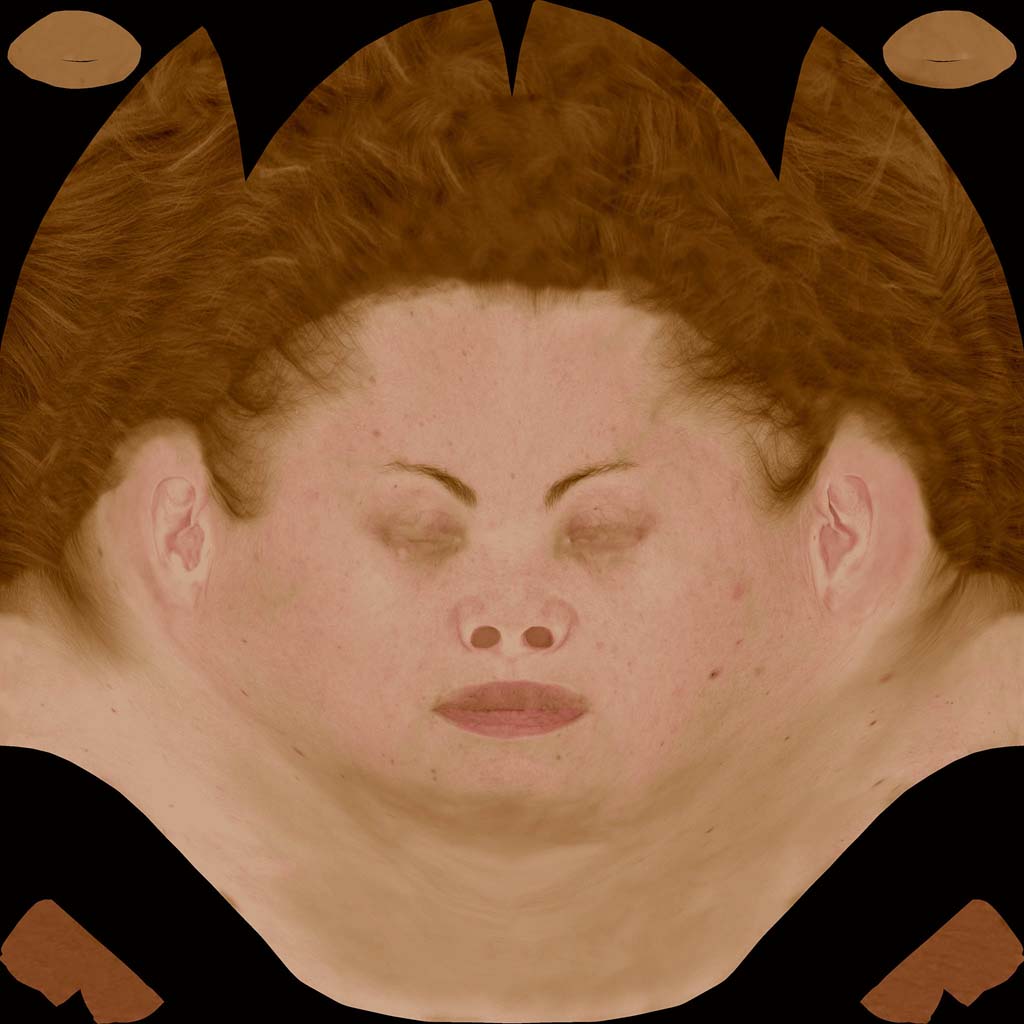}
  \end{overpic}
  &
  \begin{overpic}[width=0.32\columnwidth]{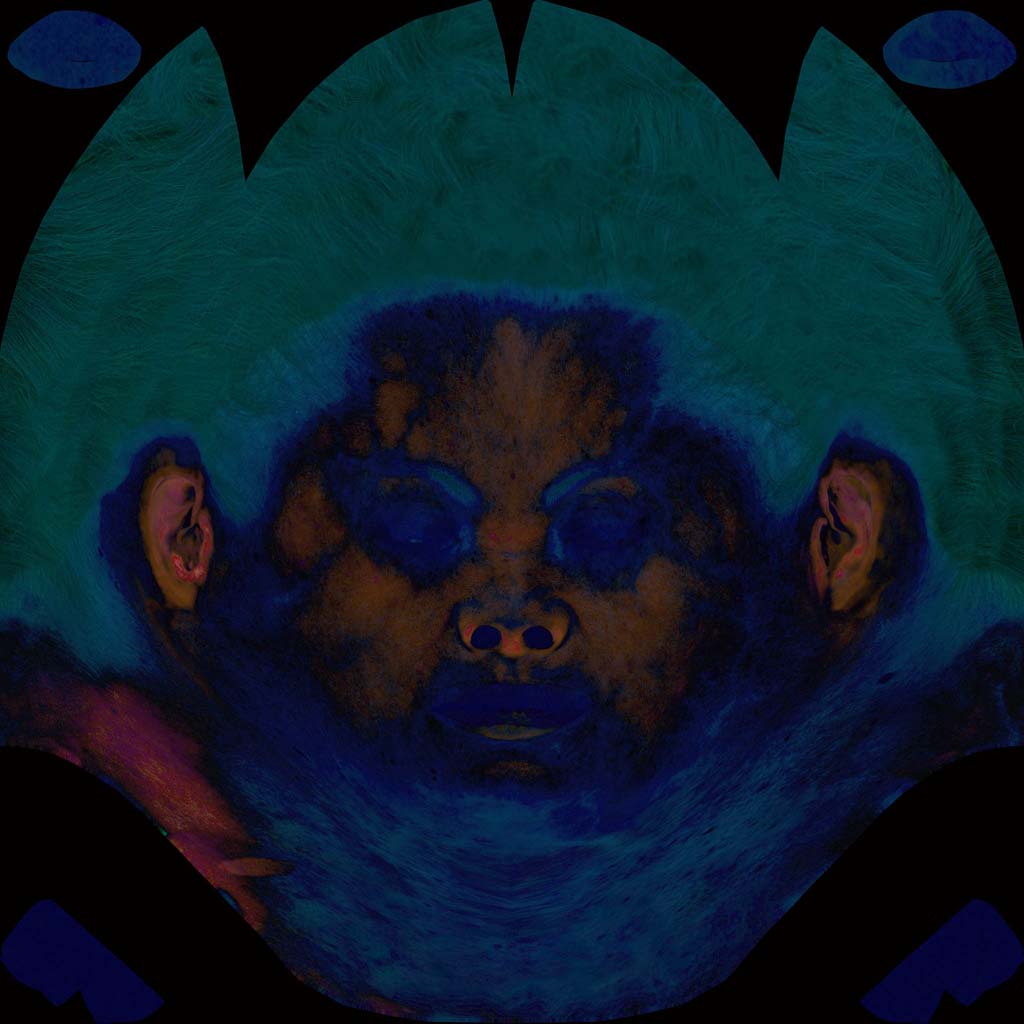}
  \end{overpic}
  \\
  \begin{sideways}\hspace{0.3cm}\textsc{Type IV male}\end{sideways}
  &
  \begin{overpic}[width=0.32\columnwidth]{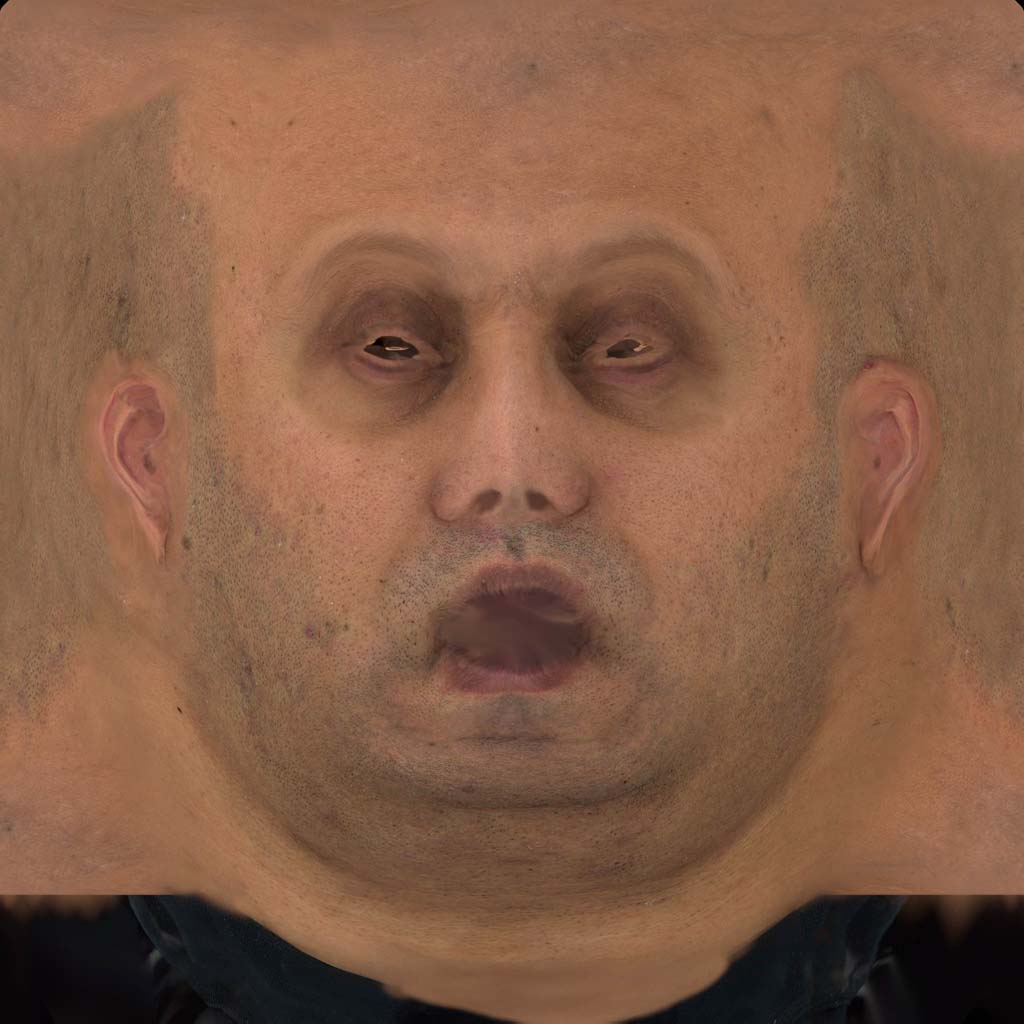}
  \end{overpic}
 &
  \begin{overpic}[width=0.32\columnwidth]{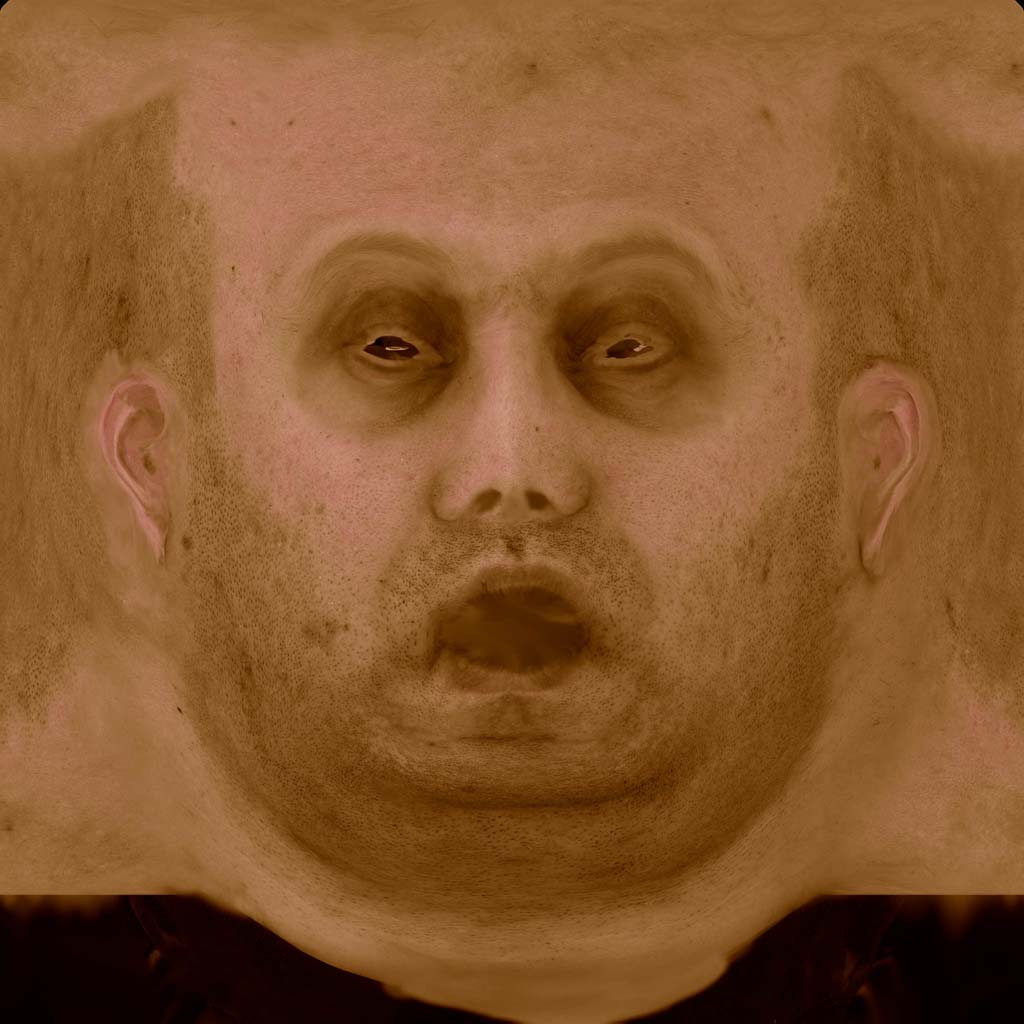}
  \end{overpic}
  &
  \begin{overpic}[width=0.32\columnwidth]{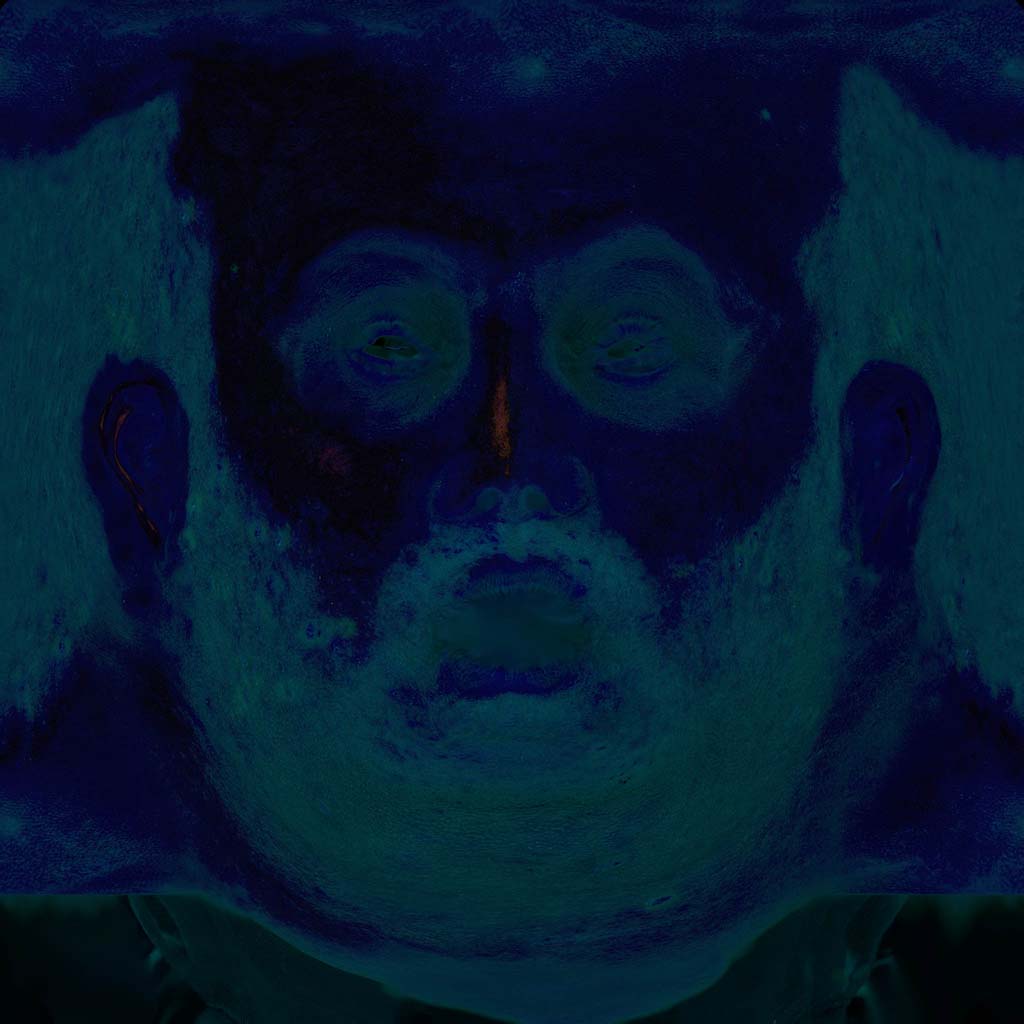}
  \end{overpic}
  \\
 \end{tabular}
  \caption{Reconstruction results using a Kubelka-Munk based model~\cite{alotaibi2017biophysical} for different types of skin. The model is unable to recover the original albedo from the inferred properties (only melanin and haemoglobin volume fractions in this case), leading to noticeable shifts in the skin shades.}
  \vspace{-6mm}
  \label{fig:failure_cases_KM}
\end{figure}

\begin{figure*}[h]
 \contourlength{0.1em}%
 \centering
 \hspace*{-4.5mm}%
  \begin{tabular}{l@{\;}c@{\;}c@{\;}c@{\;}c@{\;}c@{\;}c@{\;}c@{\;}c@{\;}c}
  
 & \textsc{Input} & \textsc{Reconstruction} & \textsc{Melanin} & \textsc{Melanin ratio} & \textsc{Hemoglobin(d)} & \textsc{Hemoglobin(e)} &

  \\
  \begin{sideways}\hspace{0.4cm}\textsc{Antera}\end{sideways}
  &
  \includegraphics[width =  0.120\textwidth]{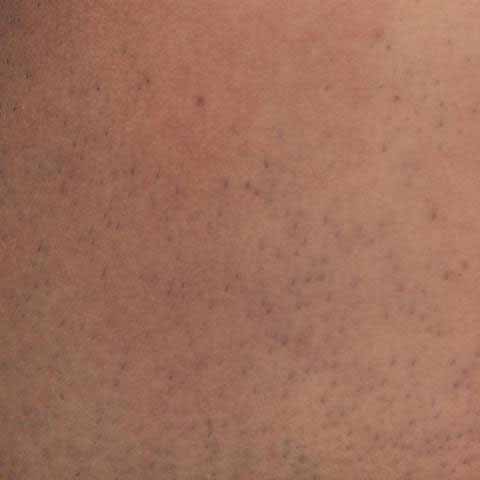}
  &
  &
  \includegraphics[trim= 113 111 0 83, clip, width = 0.14\textwidth]{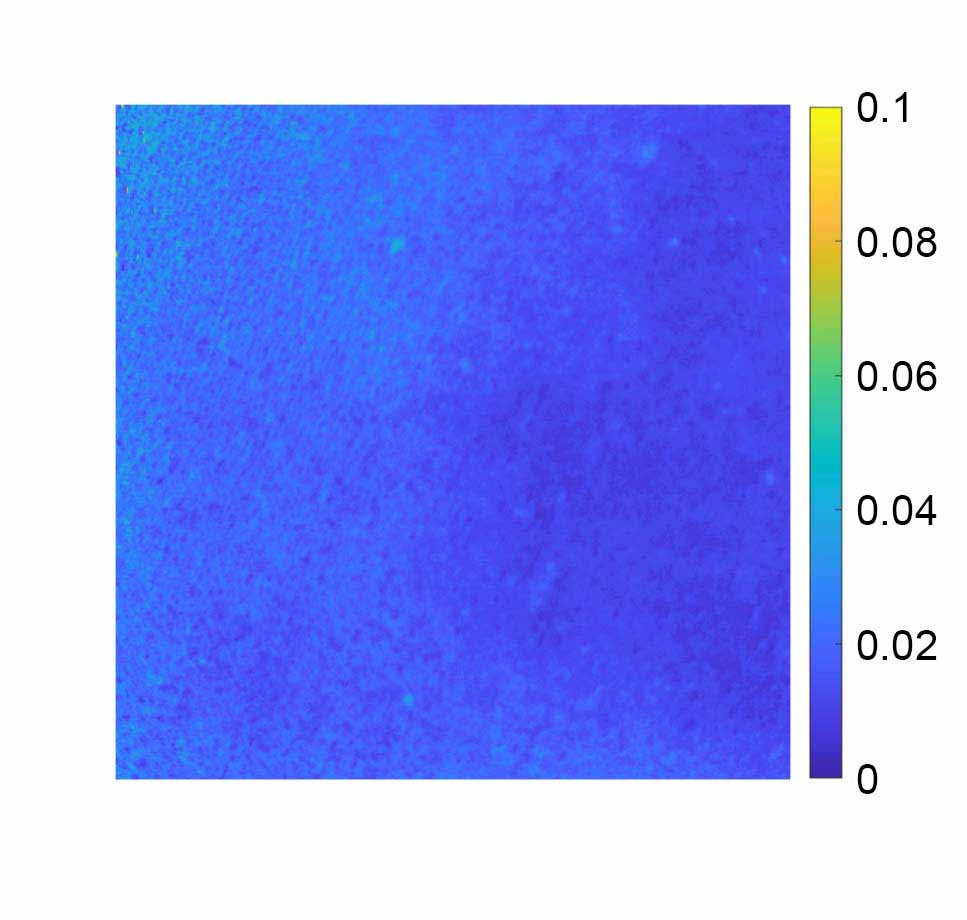}
  &
  &
  \includegraphics[trim= 113 111 0 83, clip, width =  0.14\textwidth]{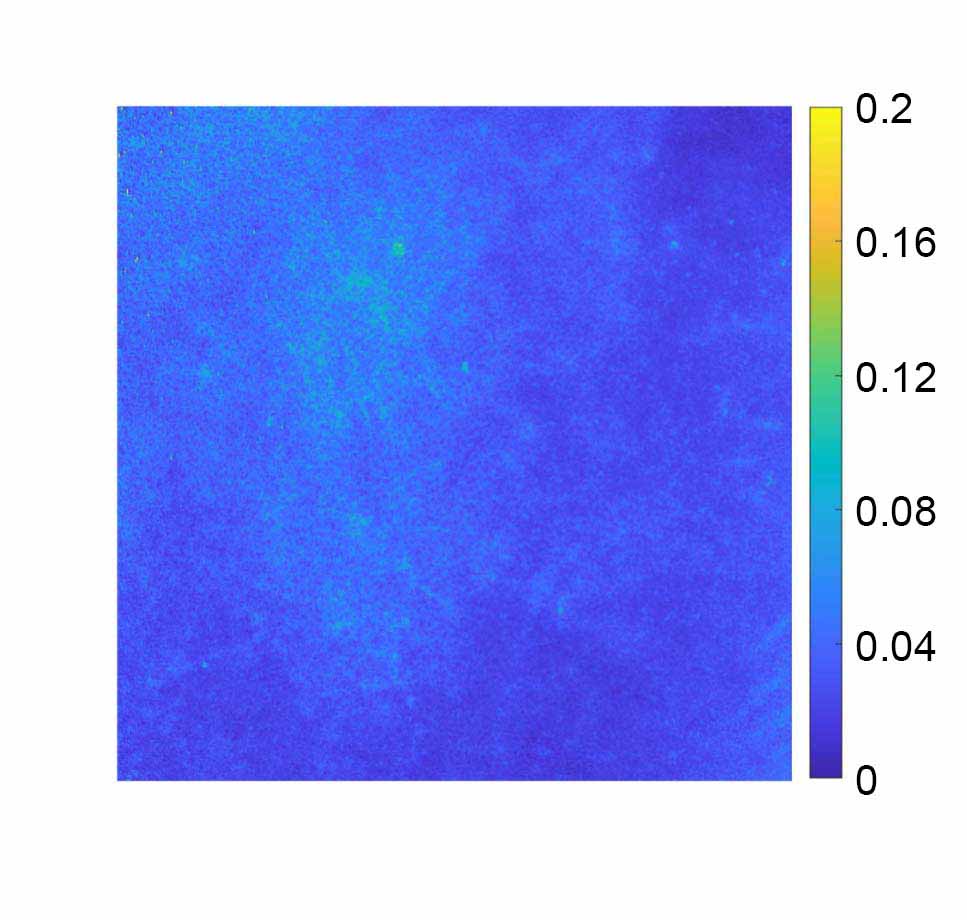}
  &
  &
  &

  \\
  \begin{sideways}\hspace{0.4cm}\textsc{Gitlina'20}\end{sideways}
  &
  &
  \includegraphics[width =  0.120\textwidth]{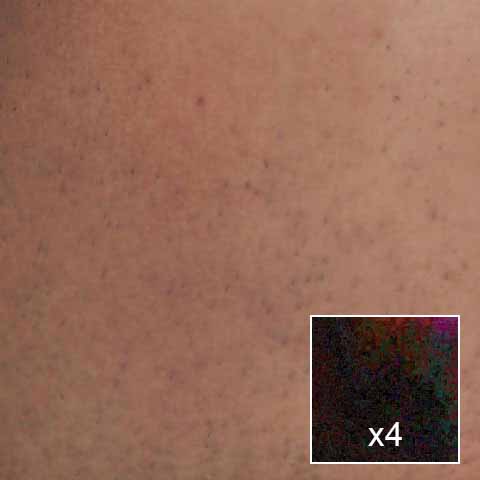}
  &
  \includegraphics[trim= 113 111 0 83, clip, width = 0.14\textwidth]{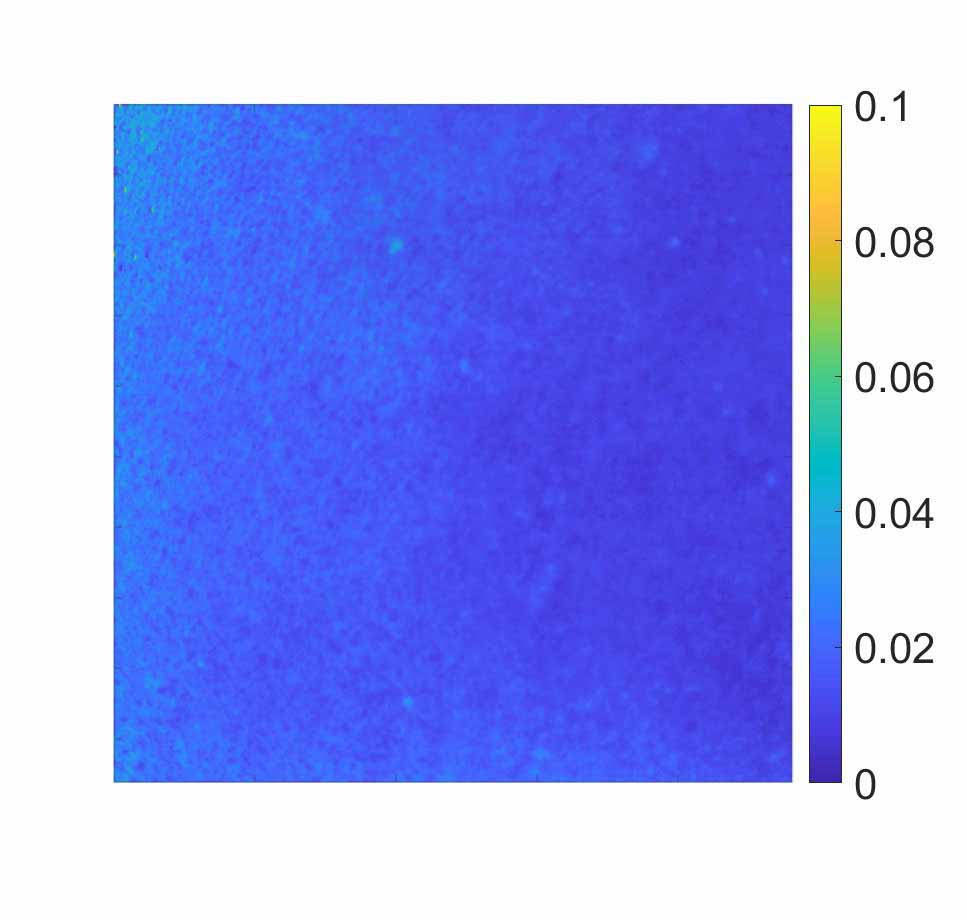}
  &
  \includegraphics[trim= 113 111 0 83, clip, width =  0.14\textwidth]{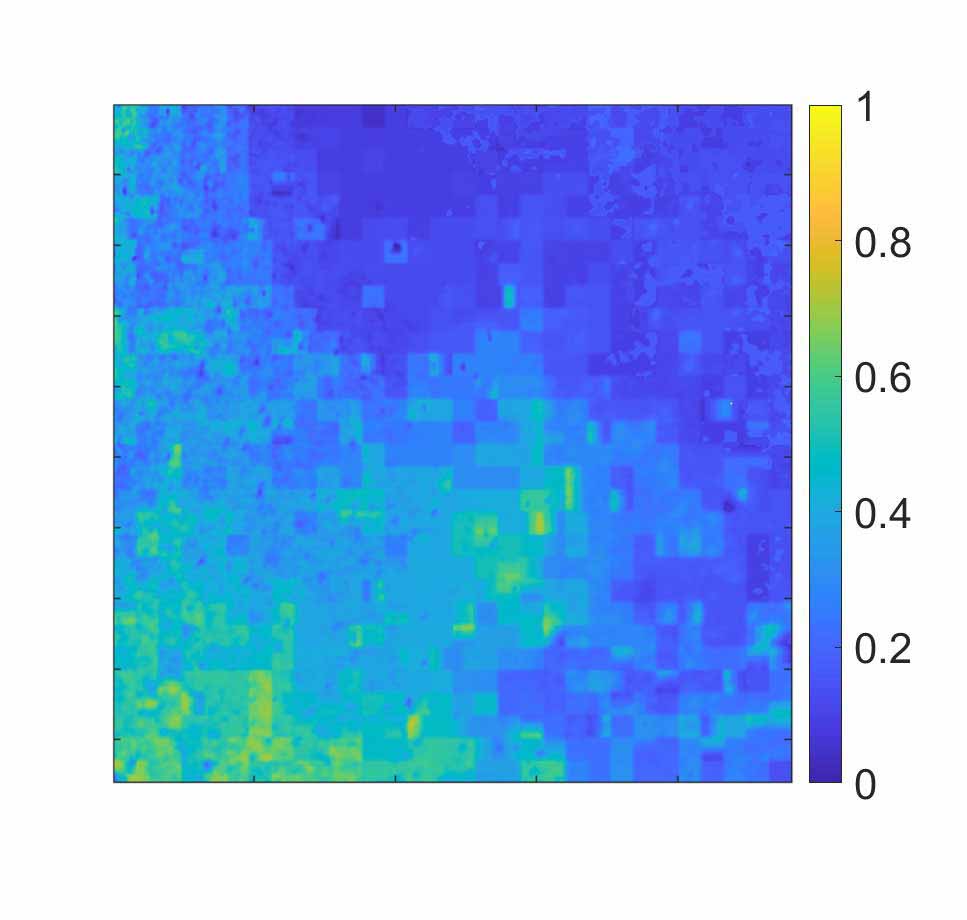}
  &
  \includegraphics[trim= 113 111 0 83, clip, width = 0.14\textwidth]{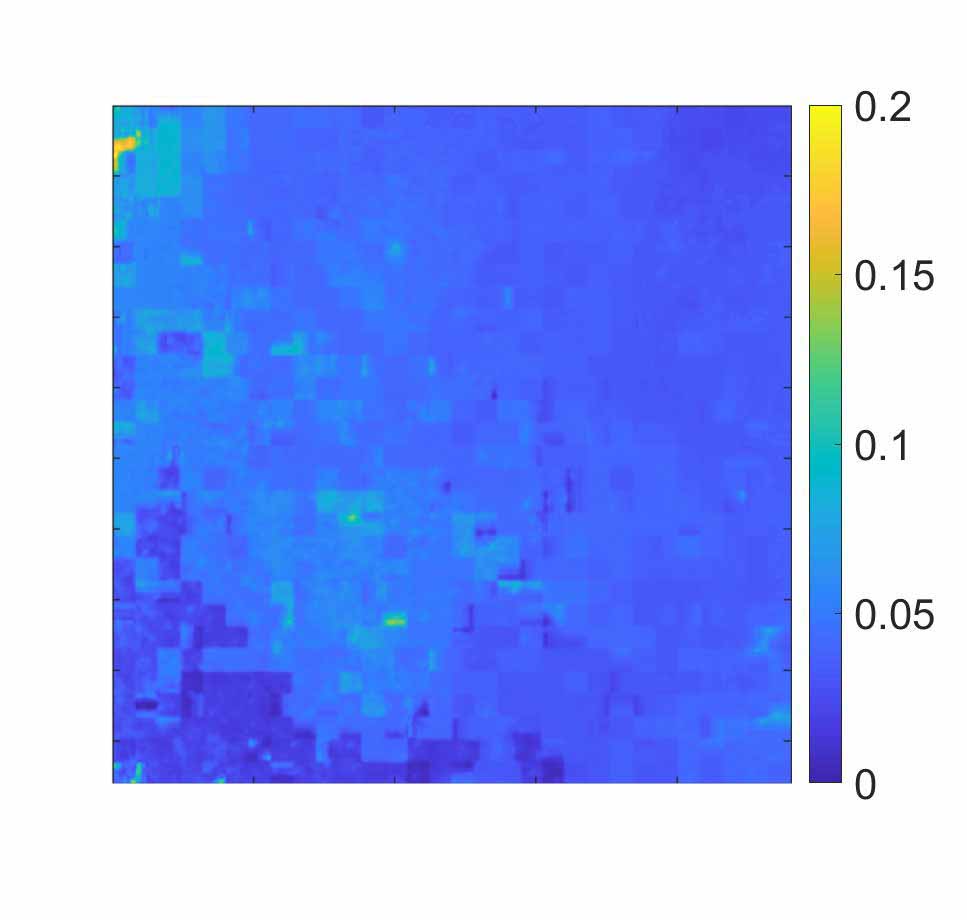}
  &
  \includegraphics[trim= 113 111 0 83, clip, width = 0.14\textwidth]{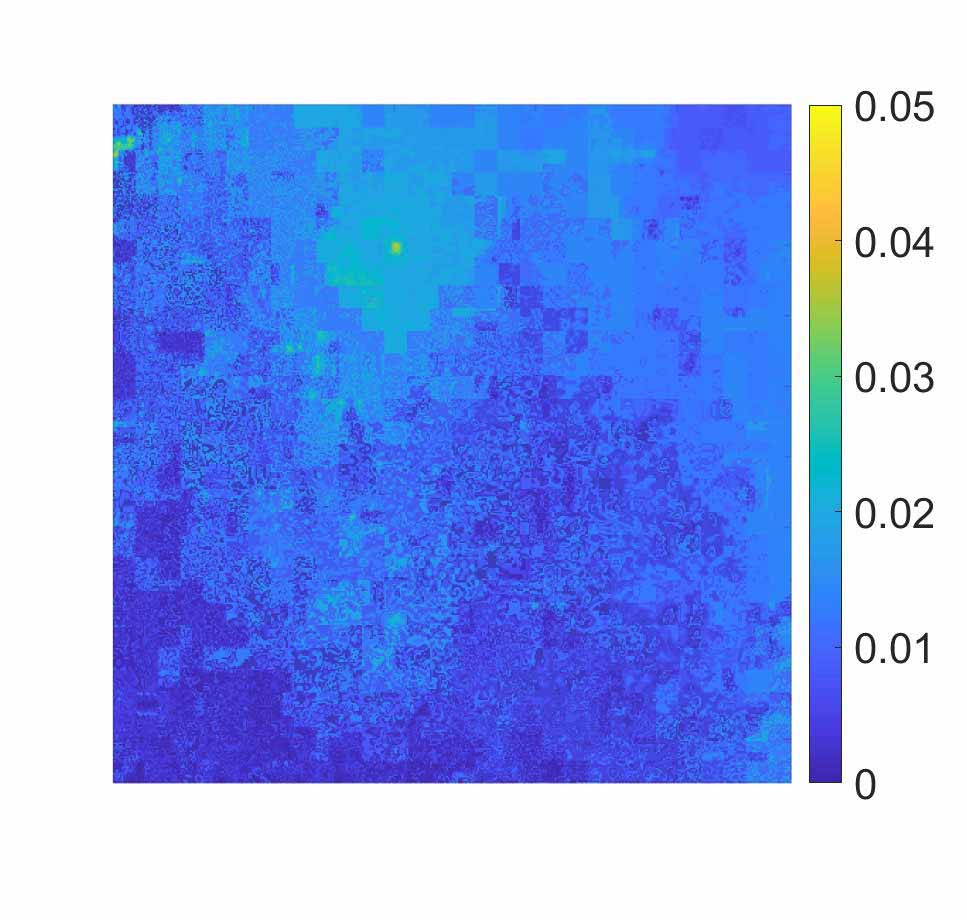}
  &
  &

      \\
  & &
     & &  & \textsc{Hemoglobin} & \textsc{Hemoglobin ratio} & \textsc{Thickness}\\
  
  \begin{sideways}\hspace{0.8cm}\textsc{Ours}\end{sideways}
  &
  &
  \includegraphics[width =  0.120\textwidth]{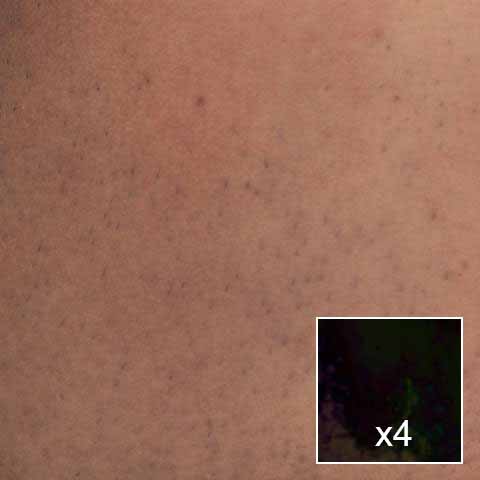}
  &
  \includegraphics[trim= 136 51 55 30, clip, width =  0.14\textwidth]{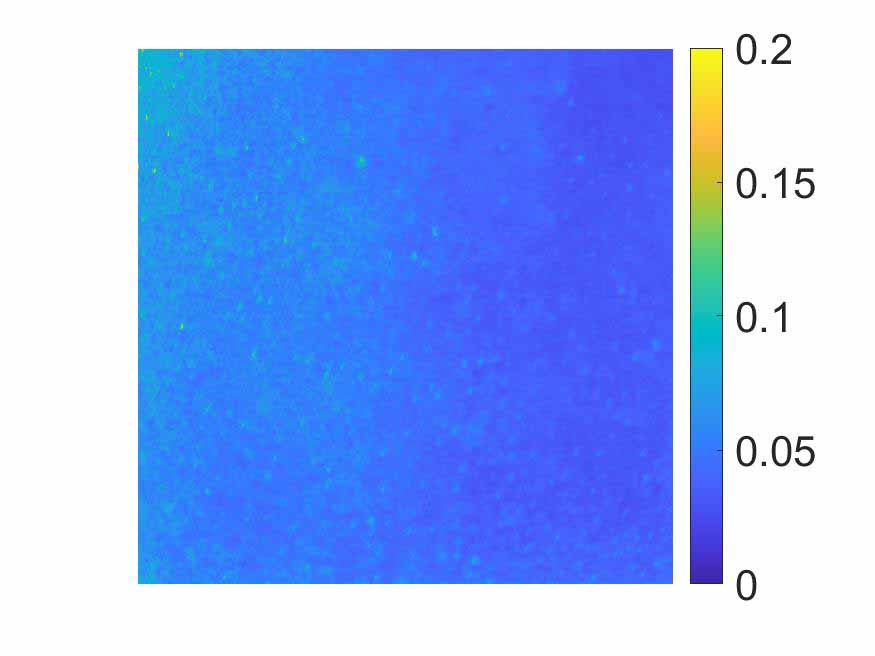}
  &
  \includegraphics[trim= 136 51 55 30, clip, width = 0.14\textwidth]{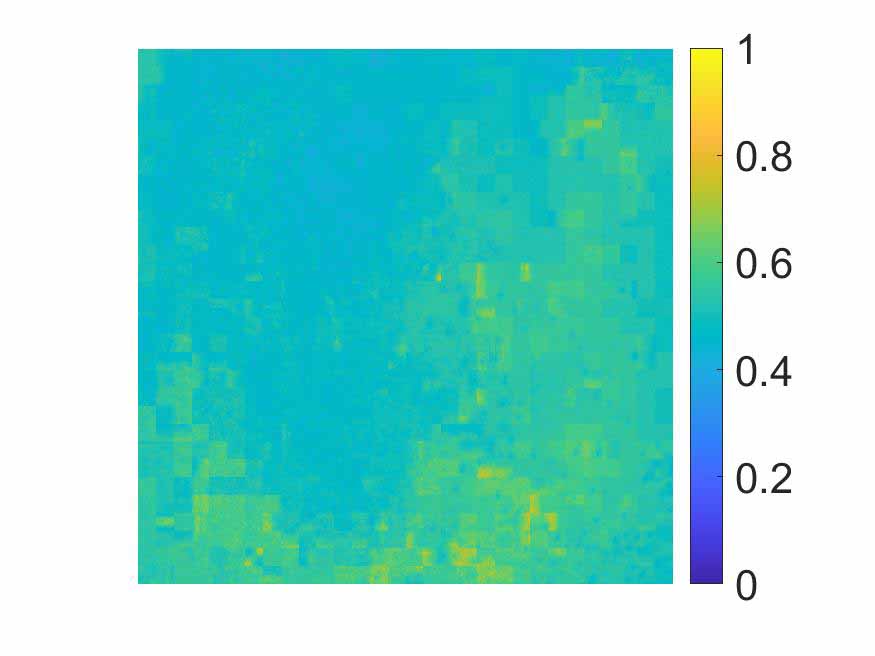}
  &
  \includegraphics[trim= 136 51 55 30, clip, width = 0.14\textwidth]{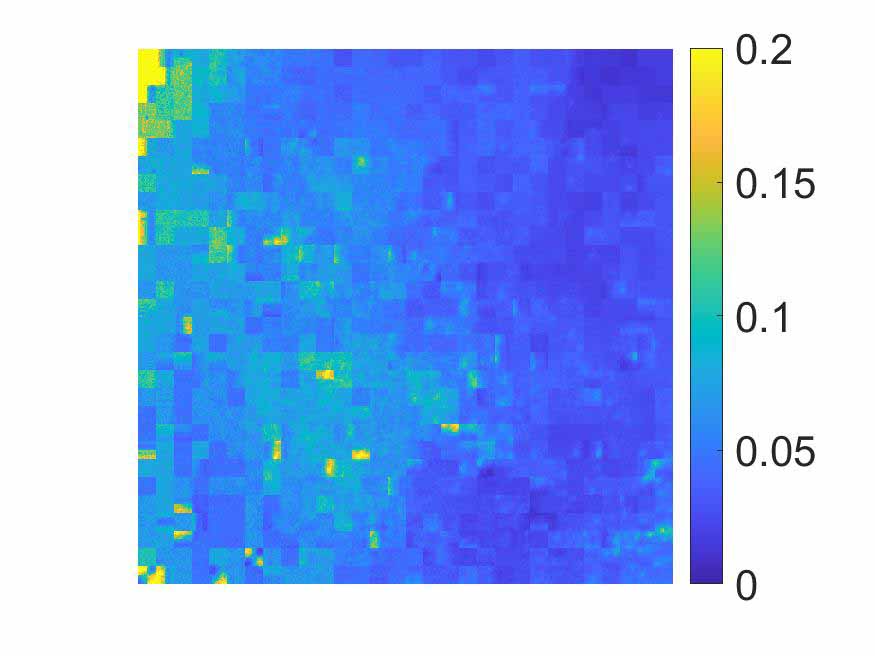}
  &
  \includegraphics[trim= 136 51 55 30, clip, width = 0.14\textwidth]{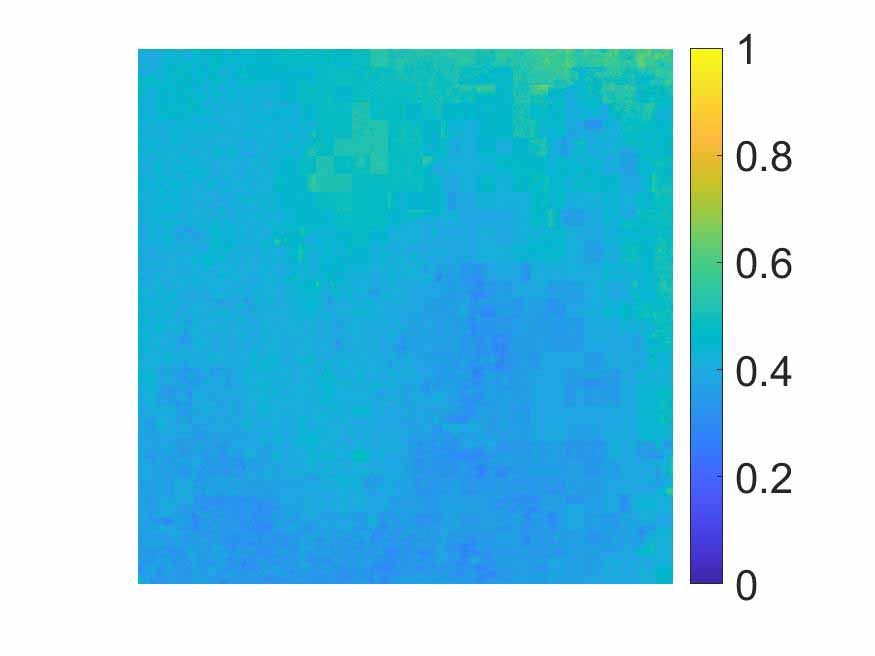}
  &
    \includegraphics[trim= 136 51 55 30, clip, width = 0.14\textwidth]{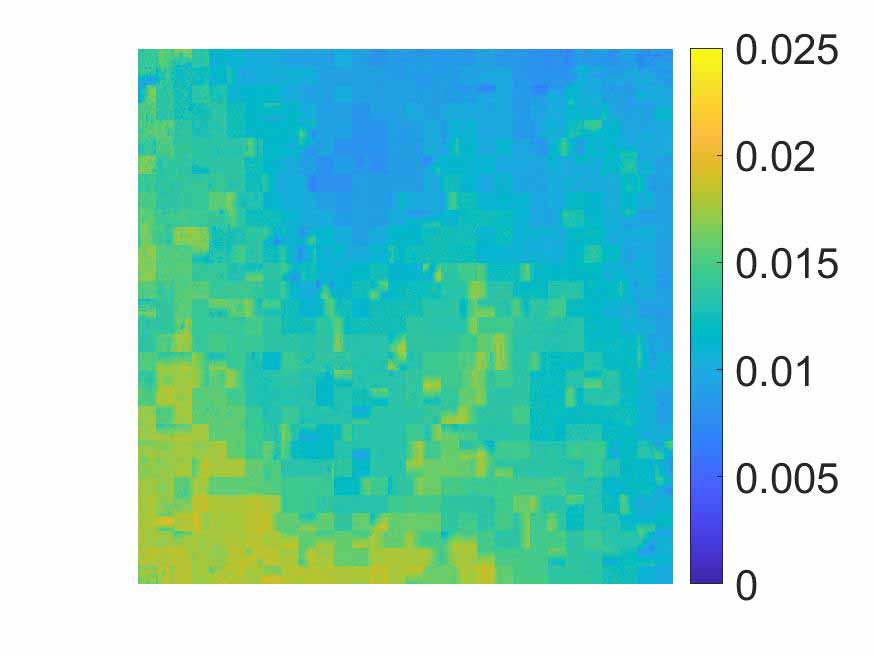}

    \\
  & &
     &  &  &  &  & \\
  
  \begin{sideways}\hspace{0.8cm}\textsc{Ours (F.)}\end{sideways}
  &
  &
  \includegraphics[width =  0.120\textwidth]{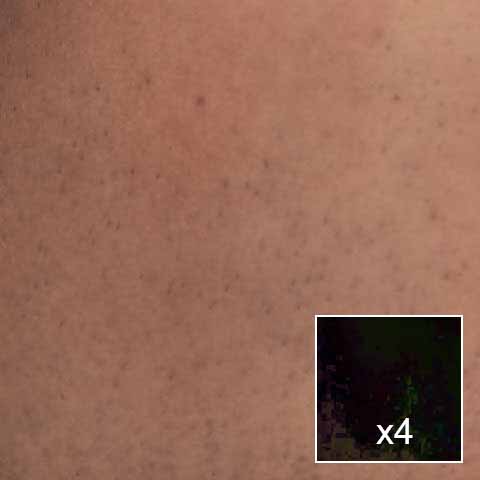}
  &
  \includegraphics[trim= 136 51 55 30, clip, width =  0.14\textwidth]{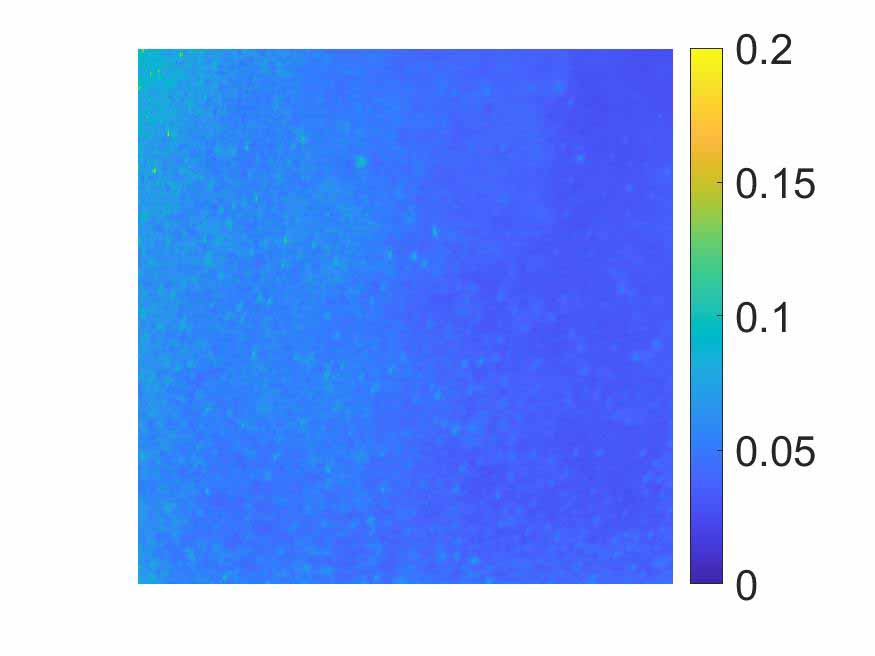}
  &
  \includegraphics[trim= 136 51 55 30, clip, width = 0.14\textwidth]{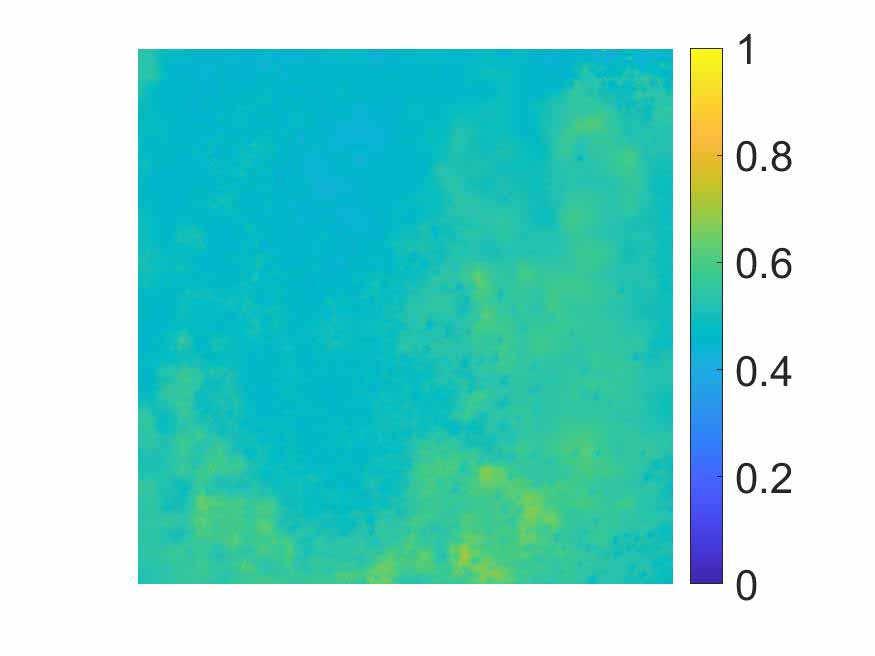}
  &
  \includegraphics[trim= 136 51 55 30, clip, width = 0.14\textwidth]{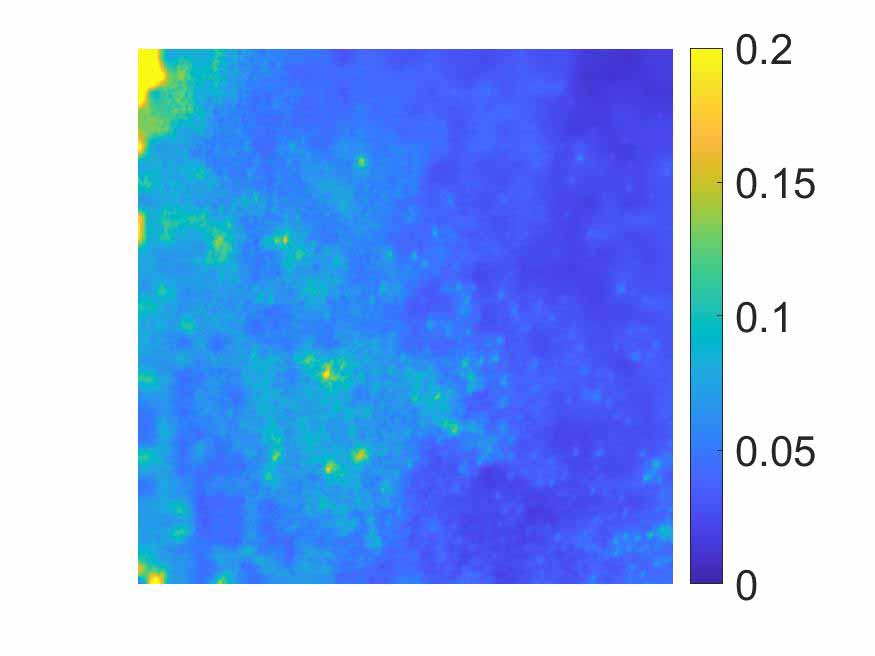}
  &
  \includegraphics[trim= 136 51 55 30, clip, width = 0.14\textwidth]{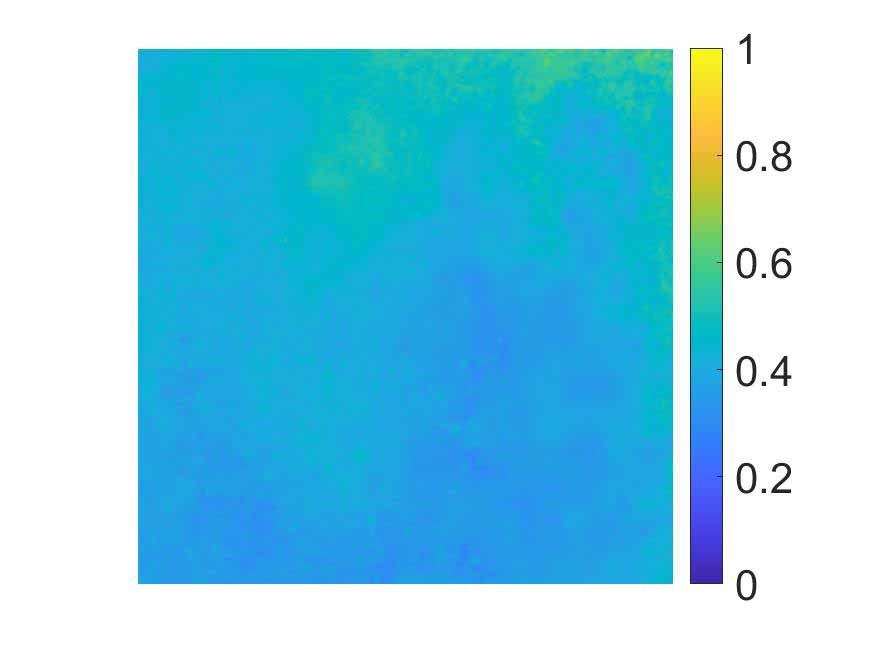}
  &
    \includegraphics[trim= 136 51 55 30, clip, width = 0.14\textwidth]{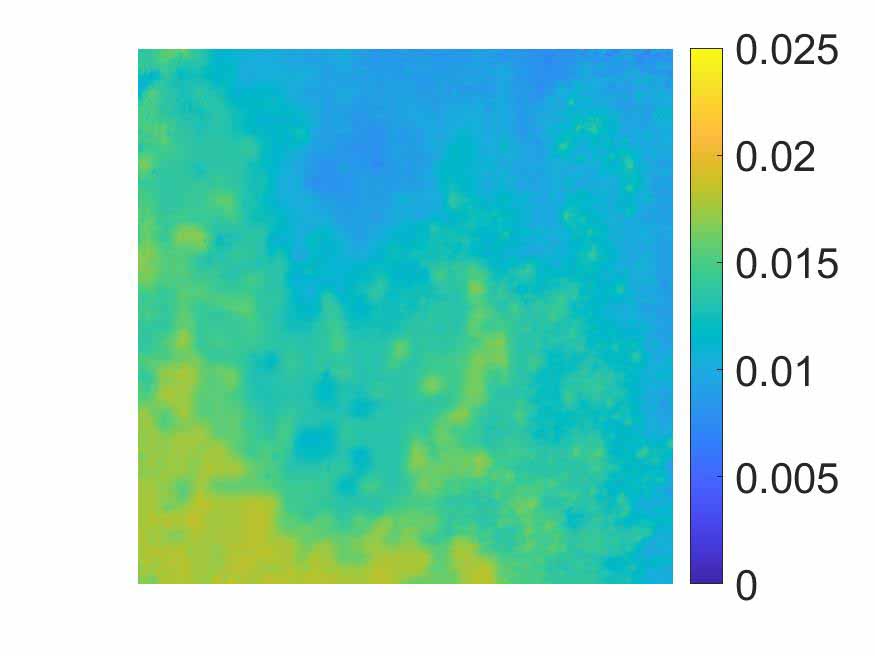}

 \end{tabular}
 \vspace{-3mm}
  \caption{\emph{Comparison with the skin albedo patch from~\cite{gitlina2020practical}}. \emph{Top:} their reconstruction (MSE = 0.0130) and estimated parameters. \emph{Middle:} ours (MSE = 0.0126). Note that the image contains compression and color artifacts, and does not seem to have an homogeneous lighting. This causes an excess of estimated melanin concentration (higher even in our reconstruction), in reference to what seems appropriate for that type of skin. Interestingly, the level of oxygenation appears to correlate with the extended blood concentration in the epidermis from their work. \emph{Bottom:} our reconstruction over a filtered input (after removal of spatial and color JPEG artifacts).}
  \label{fig:comp_params}
  \vspace{-5mm}
\end{figure*}

\begin{figure*}[h]
 \contourlength{0.1em}%
 \centering
 \vspace{10mm}
  \begin{tabular}{l@{\;}c@{\;}c@{\;}c@{\;}c@{\;}c@{\;}c@{\;}c@{\;}c@{\;}c}
  
    &
    \multicolumn{3}{c}{\textsc{[GGD*20]} } & 
    \multicolumn{3}{c}{\textsc{Ours} } & 
      \multicolumn{3}{c}{\textsc{Ours*} }  \\

 & \textsc{Input D65'} & \textsc{Rec. D65'} & \textsc{ Err. D65'} & \textsc{Input D65} & \textsc{Rec. D65} & \textsc{Err. D65}  & \textsc{Input D65* }  & \textsc{Rec. D65*}  & \textsc{Err. D65*}

  \\
  \begin{sideways}\hspace{0.5cm}\textsc{Capture 1}\end{sideways}
  &
  \includegraphics[width = 0.105\textwidth]{fig_supp/comparison_supp/1_albedo_input_gitlina.jpg}
  &
  \includegraphics[width = 0.105\textwidth]{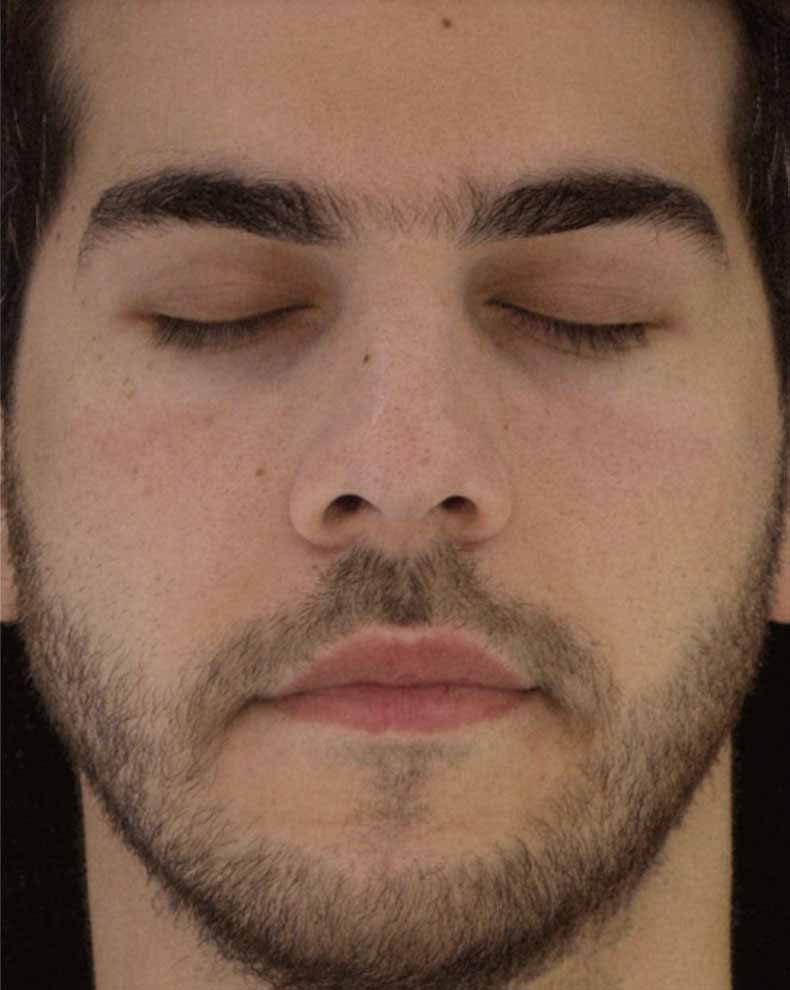}
  &
  \includegraphics[width = 0.105\textwidth]{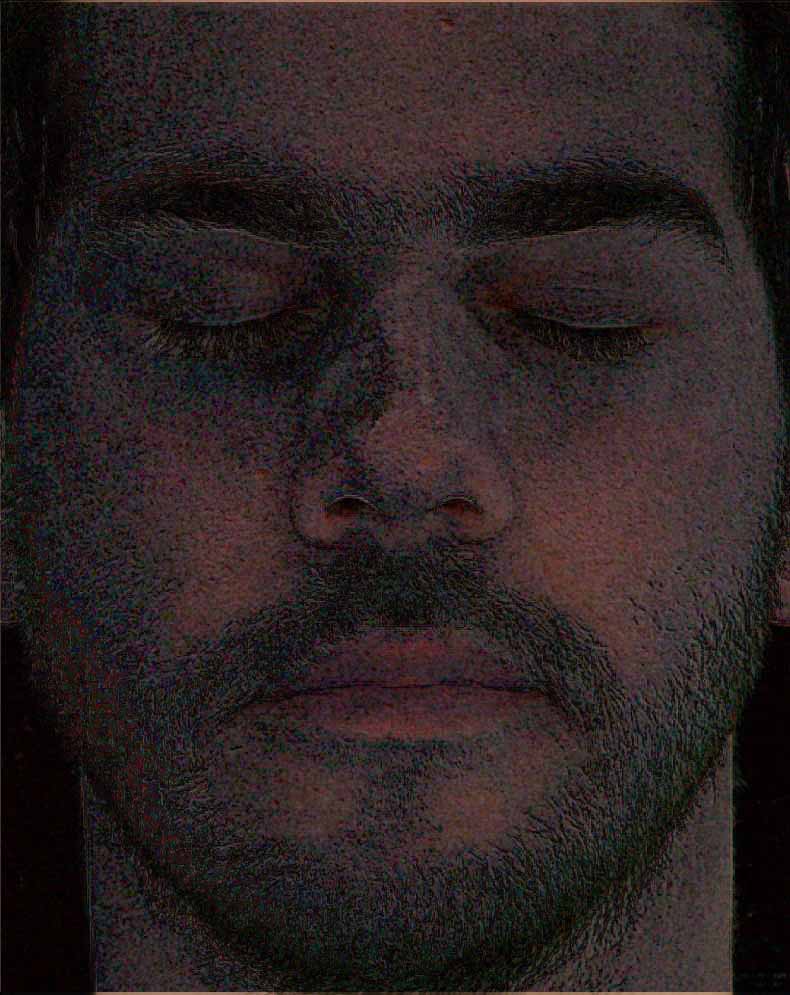}
  &
  \includegraphics[width = 0.105\textwidth]{fig_supp/comparison_supp/1_d65_input.jpg}
  &
  \includegraphics[width = 0.105\textwidth]{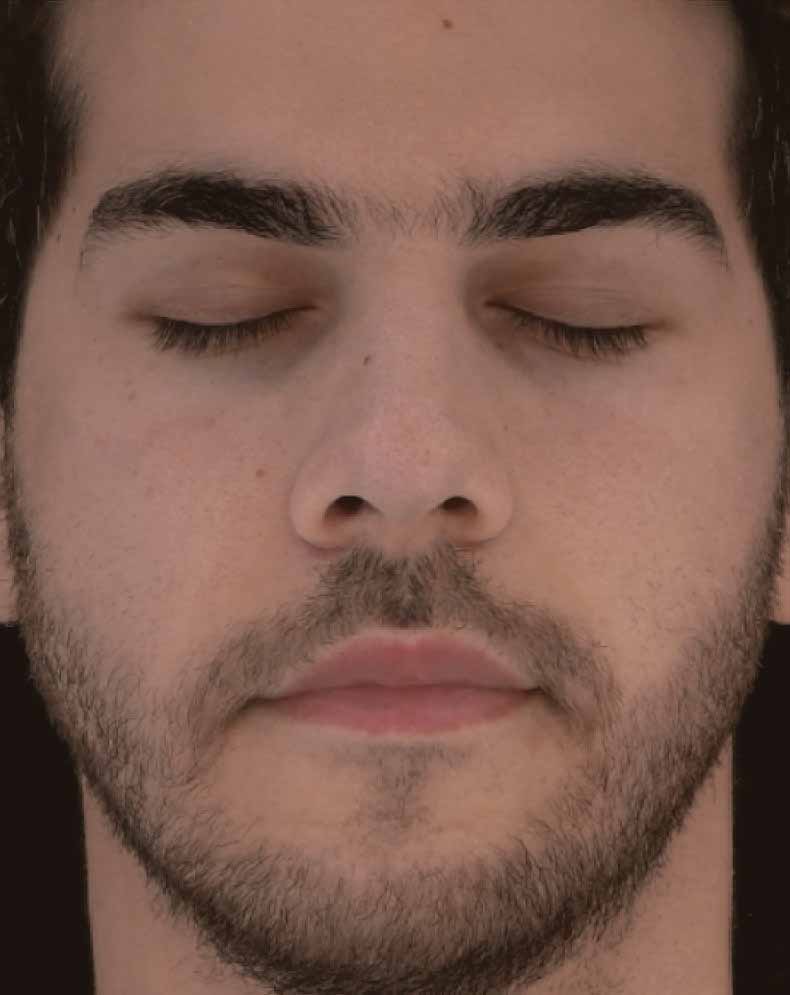}
  &
  \includegraphics[width = 0.105\textwidth]{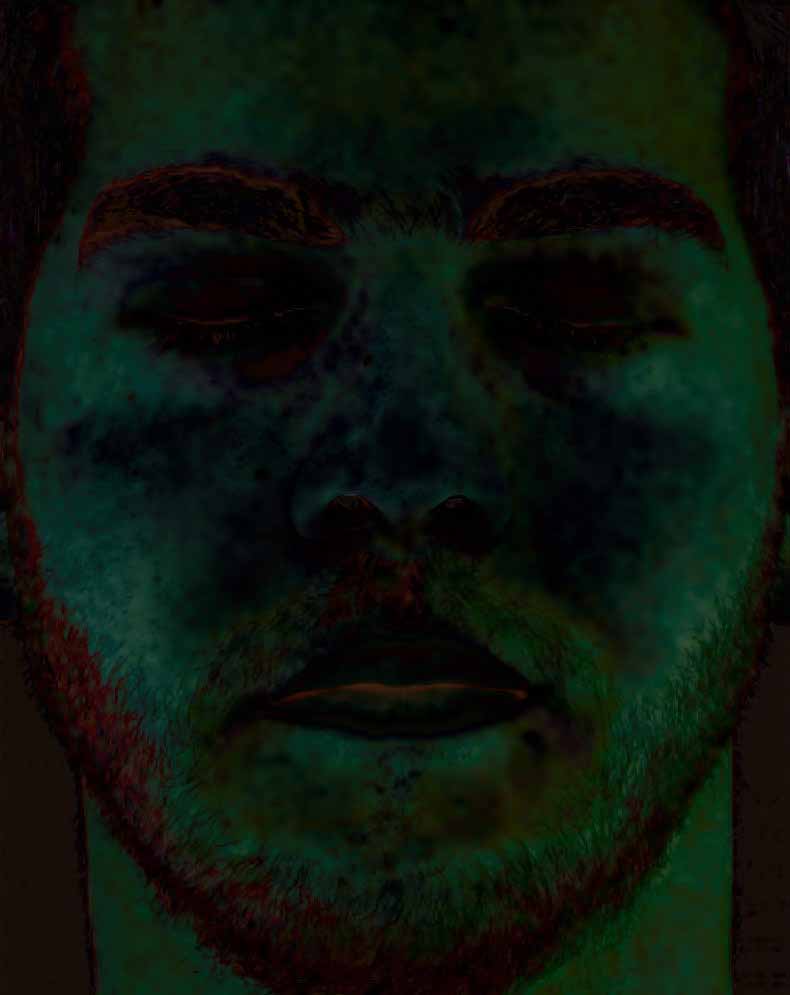}
  &
  \includegraphics[width = 0.105\textwidth]{fig_supp/comparison_supp/1_d65_delight_input.jpg}
  &
  \includegraphics[width = 0.105\textwidth]{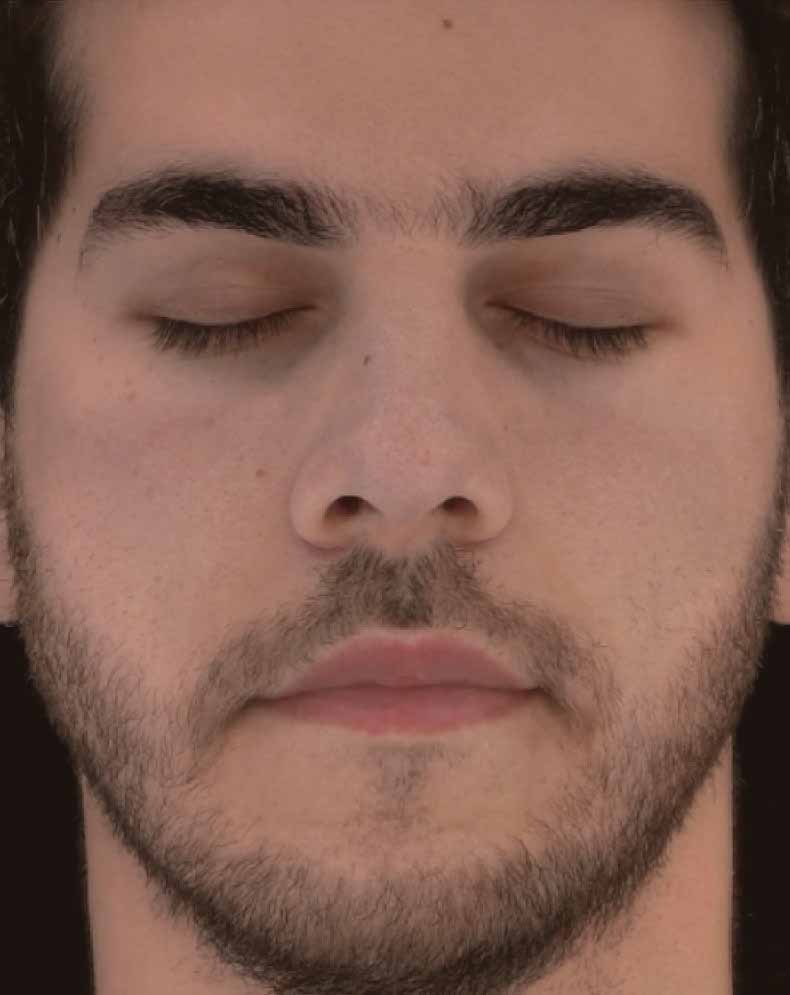}
  &
  \includegraphics[width = 0.105\textwidth]{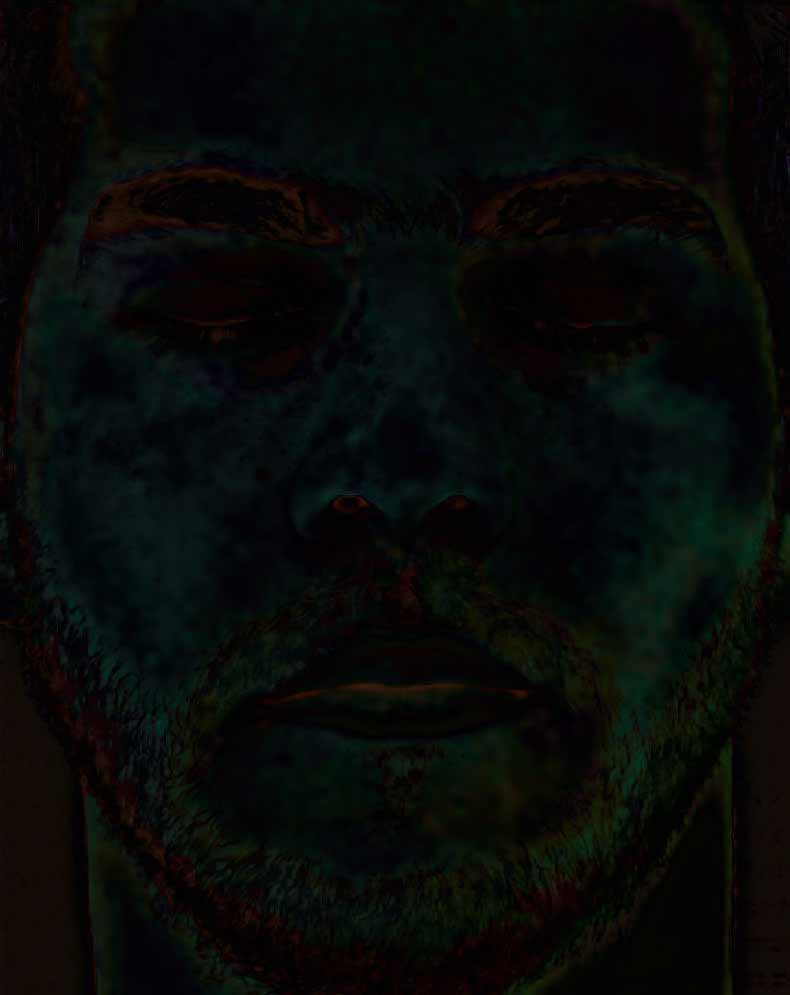}
  
  \\
  \begin{sideways}\hspace{0.5cm}\textsc{Capture 2}\end{sideways}
  &
  \includegraphics[width = 0.105\textwidth]{fig_supp/comparison_supp/0_albedo_input_gitlina.jpg}
  &
  \includegraphics[width = 0.105\textwidth]{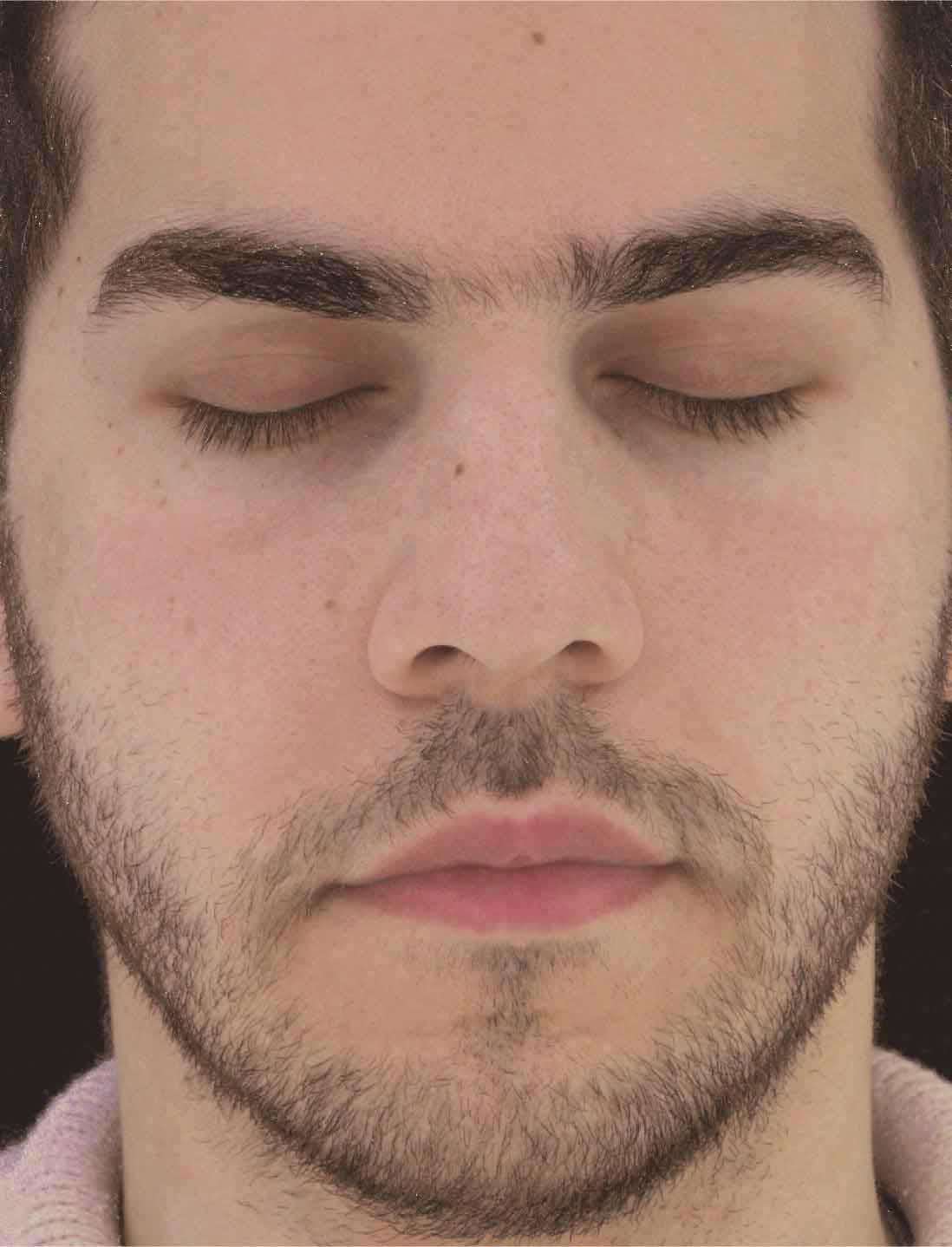}
  &
  \includegraphics[width = 0.105\textwidth]{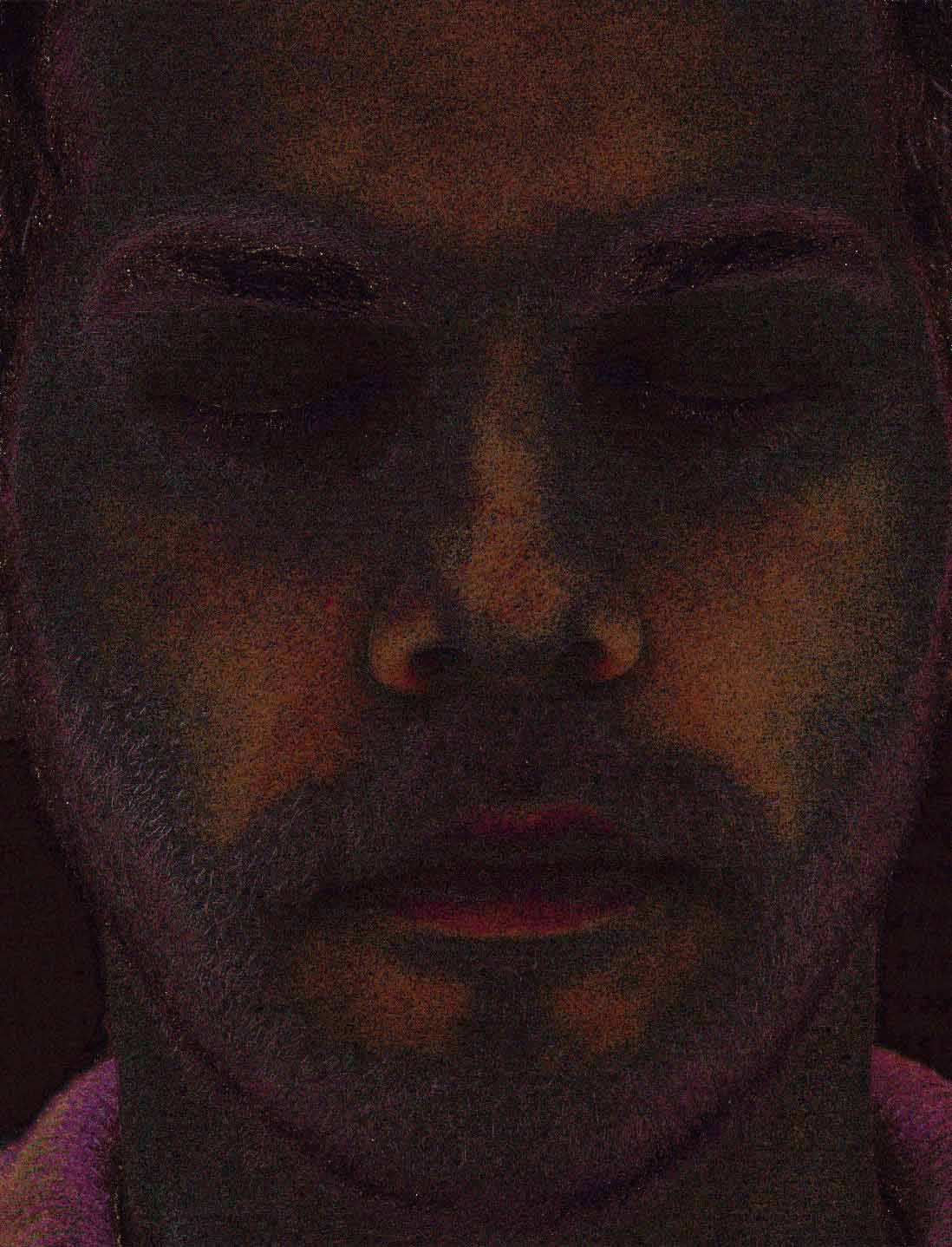}
  &
  \includegraphics[width = 0.105\textwidth]{fig_supp/comparison_supp/0_d65_input.jpg}
  &
  \includegraphics[width = 0.105\textwidth]{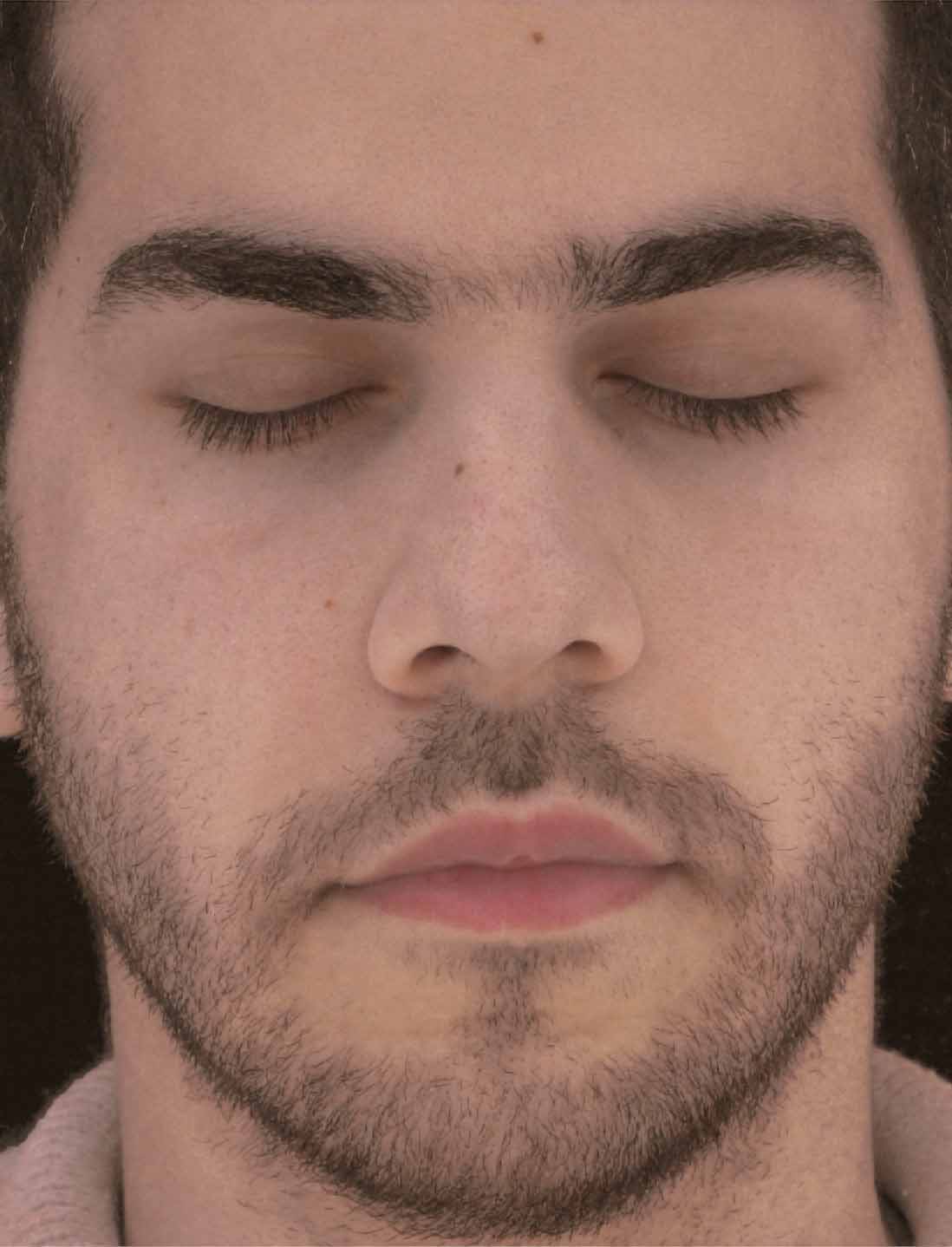}
  &
  \includegraphics[width = 0.105\textwidth]{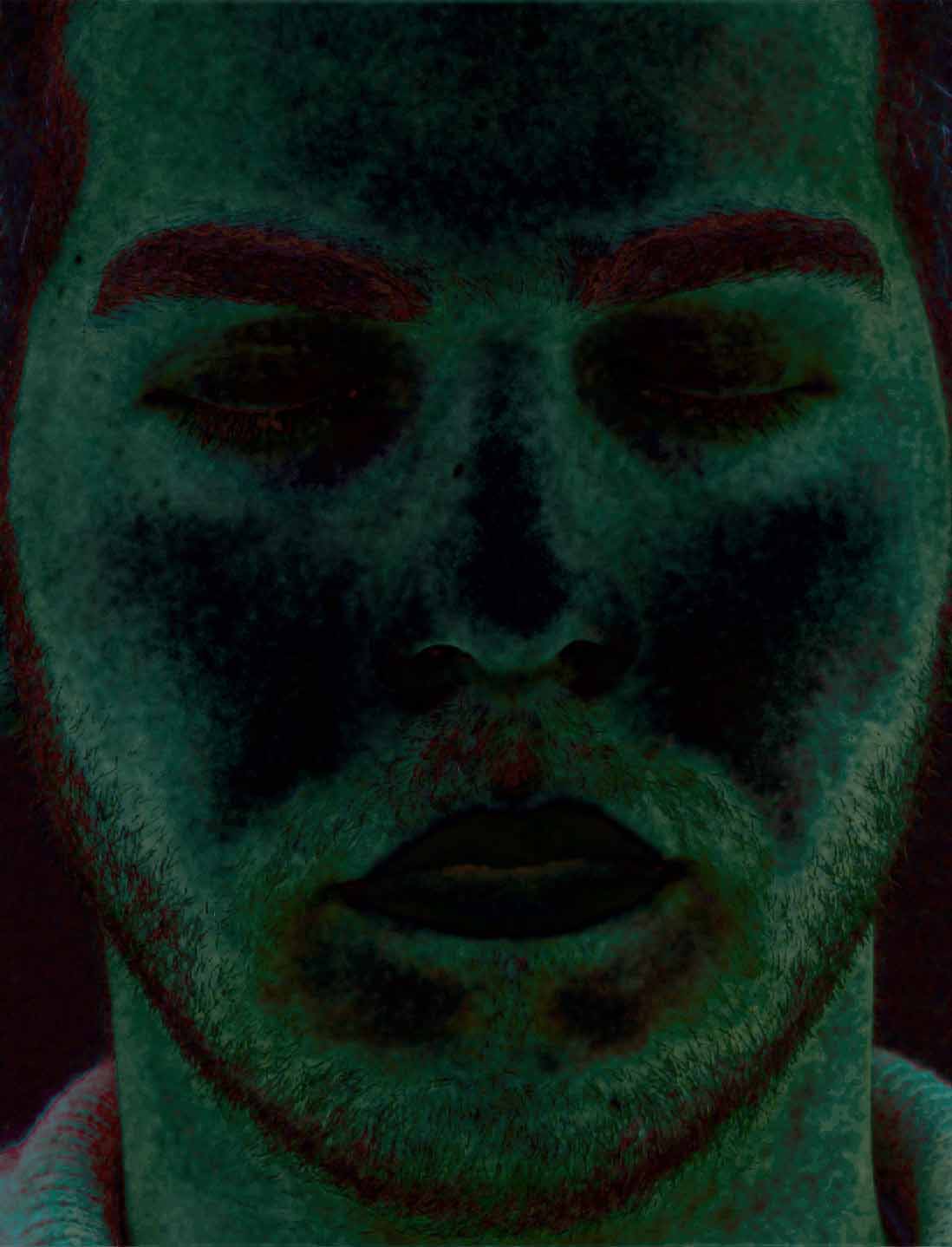}
  &
  \includegraphics[width = 0.105\textwidth]{fig_supp/comparison_supp/0_d65_delight_input.jpg}
  &
  \includegraphics[width = 0.105\textwidth]{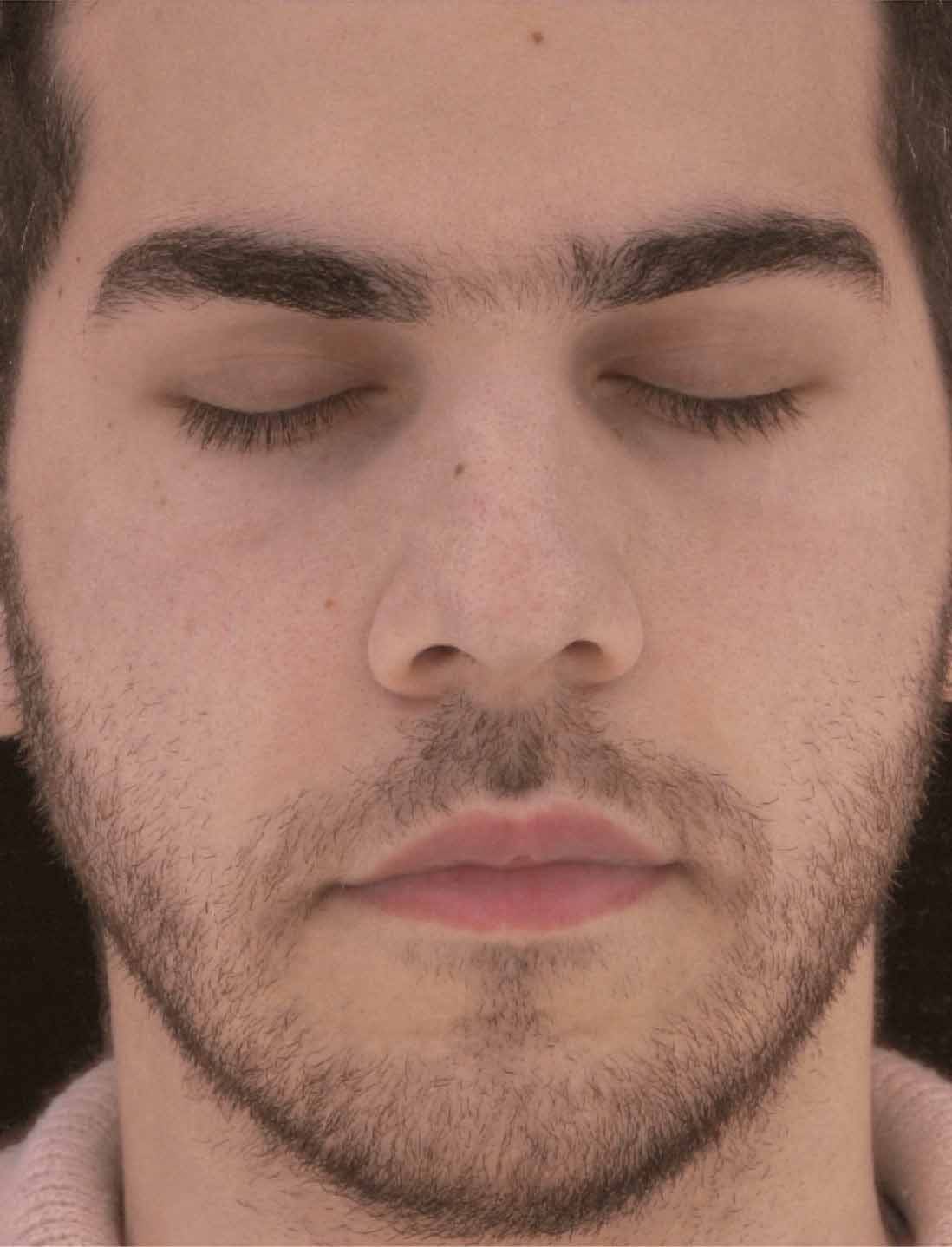}
  &
  \includegraphics[width = 0.105\textwidth]{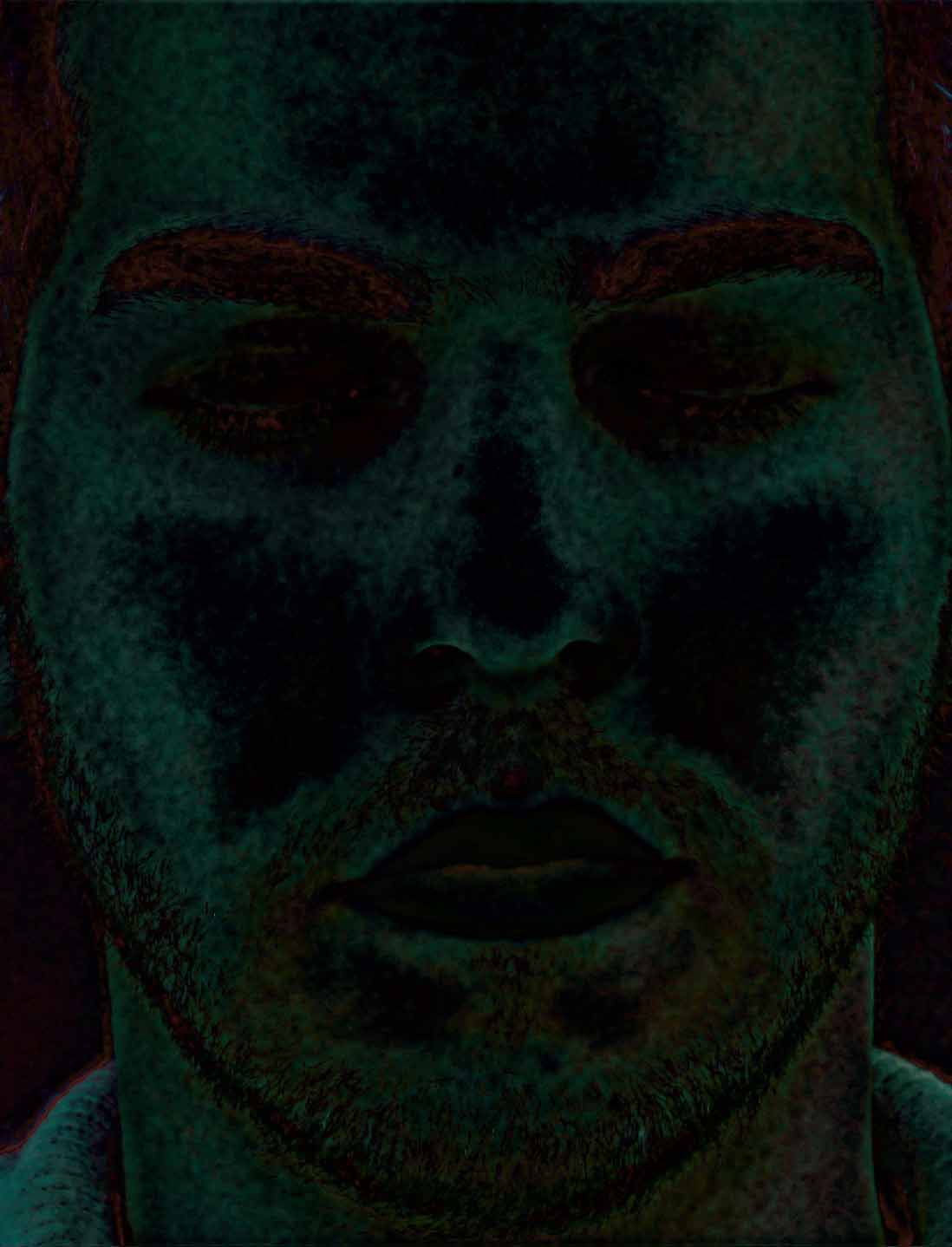}

  

 \end{tabular}
  \caption{\emph{Comparisons on photographs~\cite{gitlina2020practical}}. Since these images exhibit baked in lighting effects, the reconstruction error comes stronger than on our usual results for true albedos (Ours). This effect is greatly attenuated when correcting the areas with more baked in light (Ours*). Nonetheless our model performs well in these conditions. In fact, the error can be used as a mask revealing the regions of still significant lighting, which in turn could be fed to a semi-automatic process of delighting (future work).}
  \label{fig:comp_gitlina_albedos}
\end{figure*}

\begin{figure*}[h]
 \contourlength{0.1em}%
 \centering
 \vspace{10mm}
 \hspace*{-4.5mm}%
  \begin{tabular}{l@{\;}c@{\;}c@{\;}c@{\;}c@{\;}c@{\;}c}
  
 & \textsc{Melanin} & \textsc{Hemoglobin(d)} & \textsc{Hemoglobin(e)} & \textsc{Melanin ratio} & \textsc{Oxygenation} & \textsc{Thickness}
  
  \\
    \begin{sideways}\hspace{0.6cm}\textsc{C1 [GGD'20]}\end{sideways}
  &
  \includegraphics[width = 0.15\textwidth]{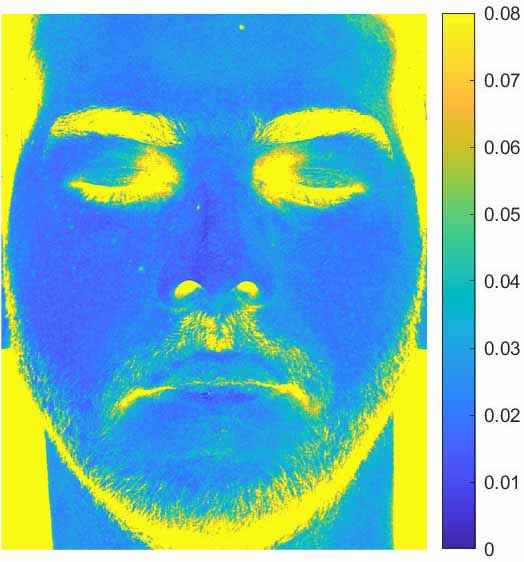}
  &
  \includegraphics[width = 0.15\textwidth]{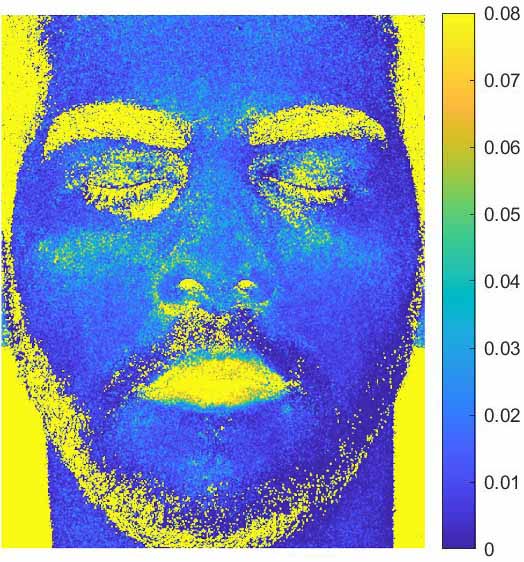}
  &
  \includegraphics[width = 0.15\textwidth]{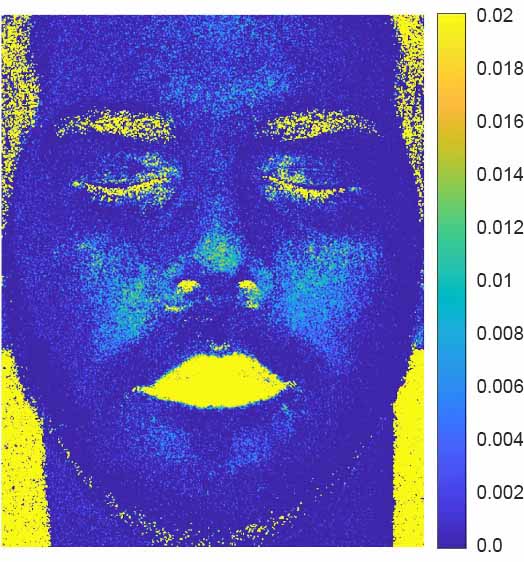}
  &
  \includegraphics[width = 0.15\textwidth]{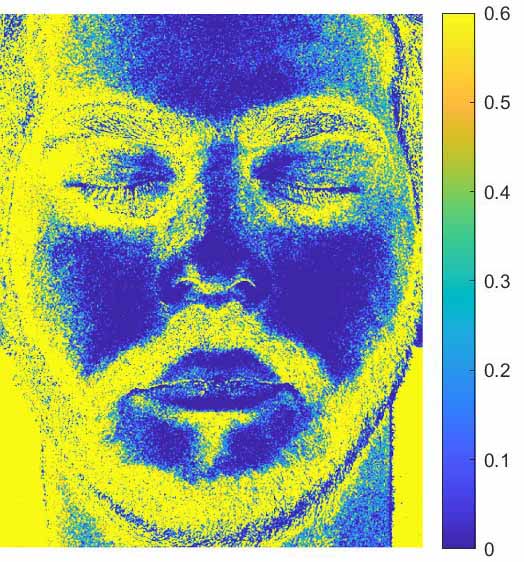}
  &
  &

  \\
  \begin{sideways}\hspace{0.5cm}\textsc{C1 Ours D65}\end{sideways}
  &
  \includegraphics[trim= 192 59 159 35, clip, width = 0.15\textwidth]{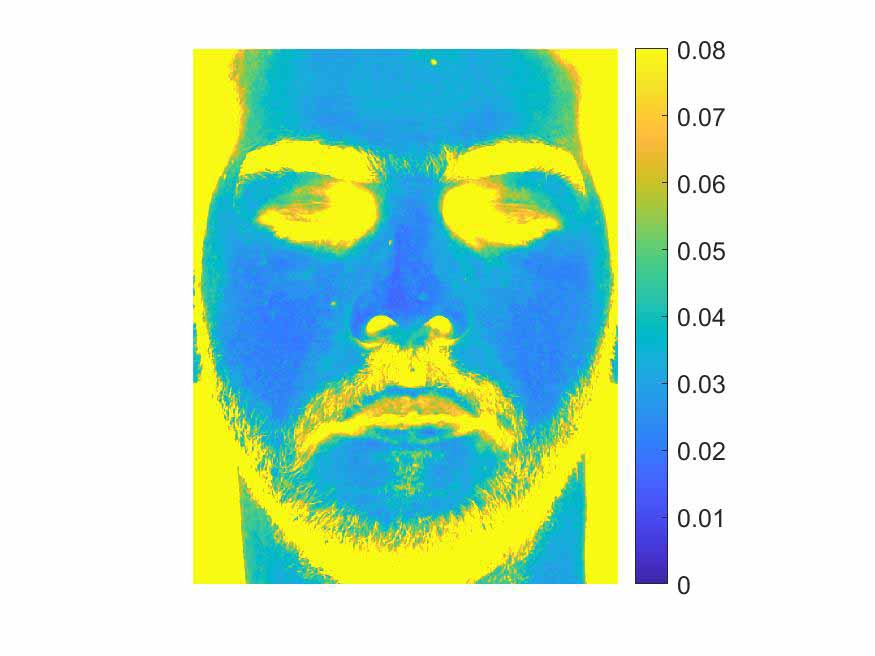}
  &
  \multicolumn{2}{c}{\includegraphics[trim= 192 59 159 35, clip, width = 0.15\textwidth]{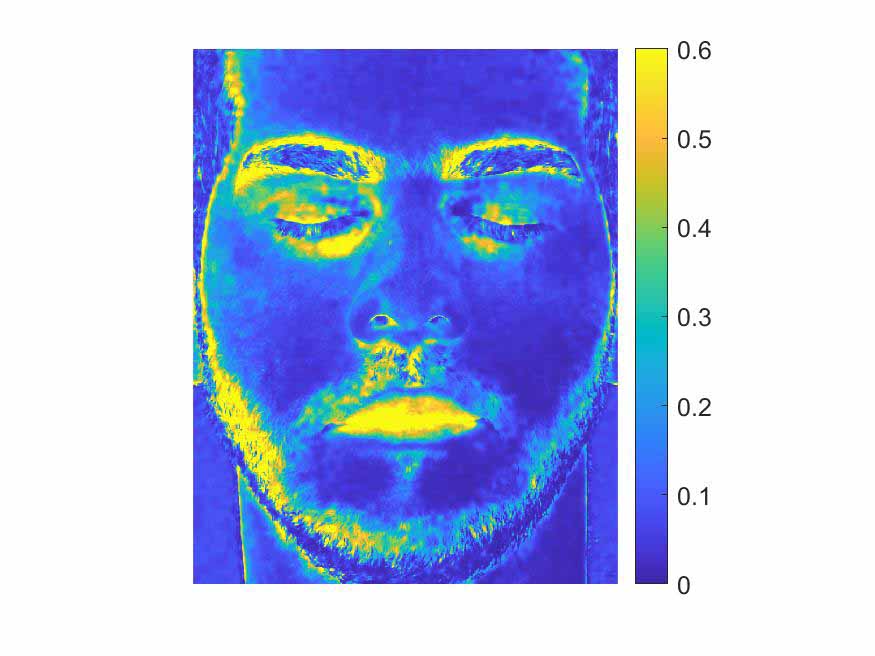}}
  &
  \includegraphics[trim= 192 59 159 35, clip, width = 0.15\textwidth]{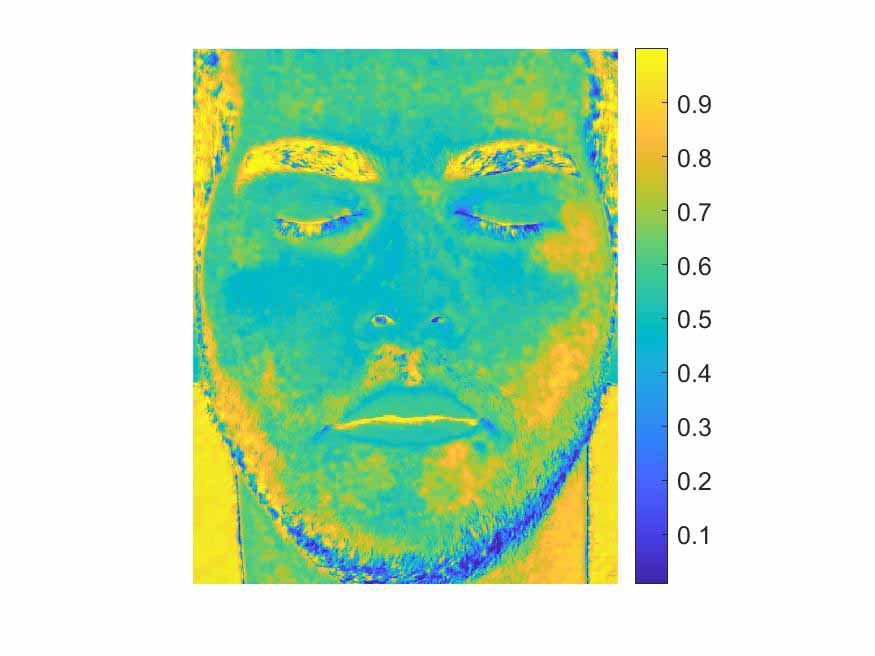}
  &
  \includegraphics[trim= 192 59 159 35, clip, width = 0.15\textwidth]{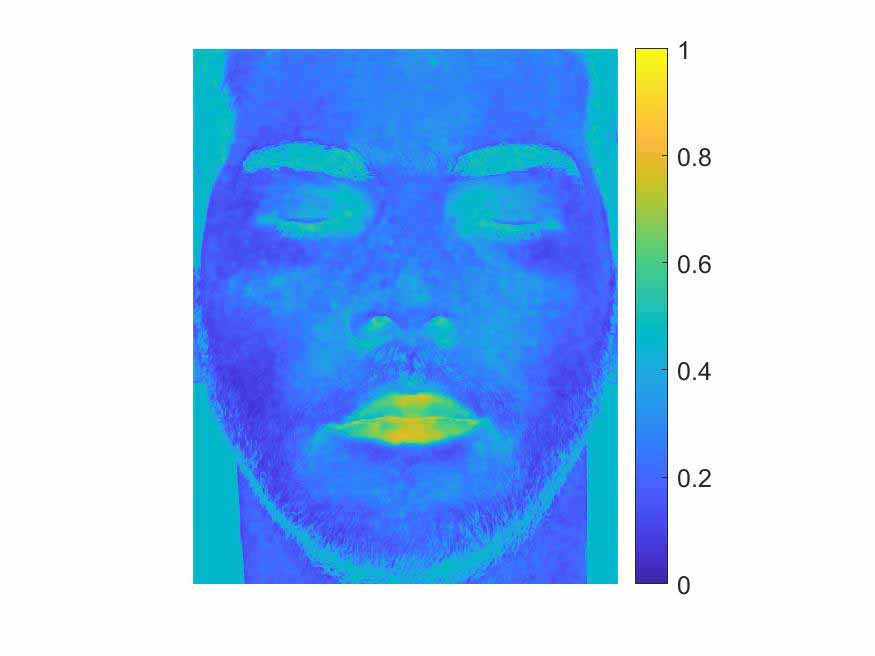}
  &
  \includegraphics[trim= 192 59 130 35, clip, width = 0.16\textwidth]{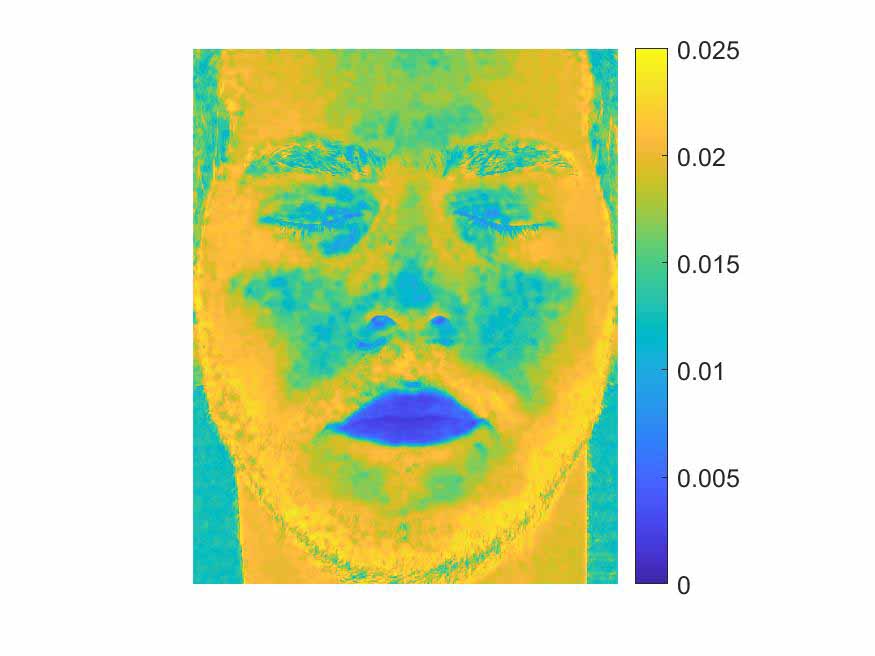}
  
  \\
  \begin{sideways}\hspace{0.5cm}\textsc{C1 Ours D65*}\end{sideways}
  &
  \includegraphics[trim= 192 59 159 35, clip, width = 0.15\textwidth]{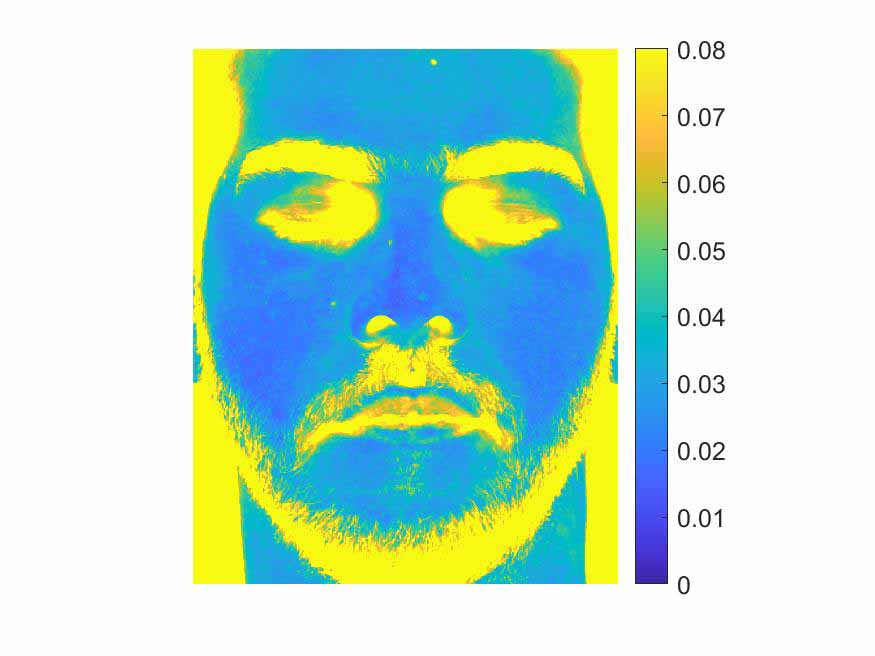}
  &
  \multicolumn{2}{c}{\includegraphics[trim= 192 59 159 35, clip, width = 0.15\textwidth]{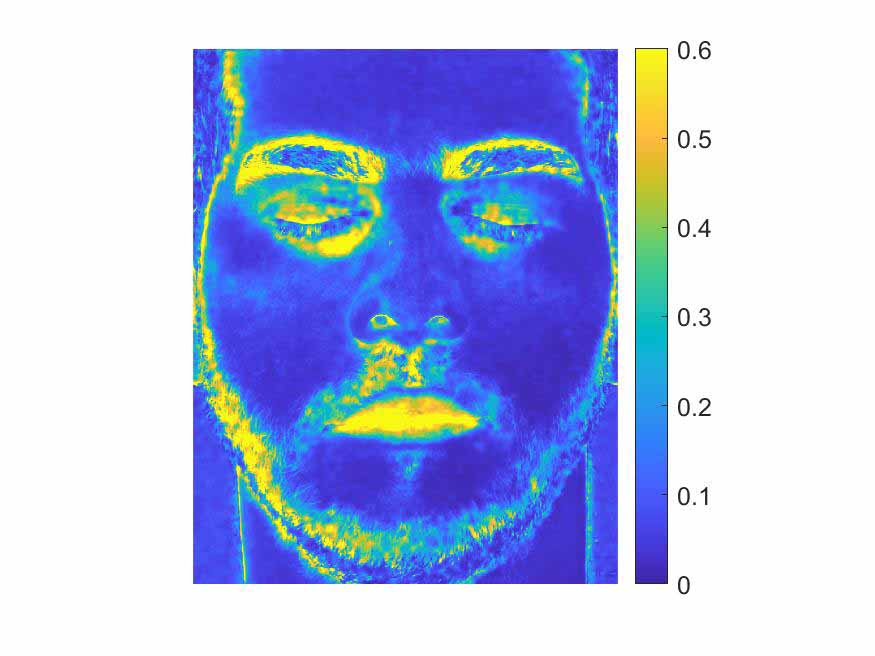}}
  &
  \includegraphics[trim= 192 59 159 35, clip, width = 0.15\textwidth]{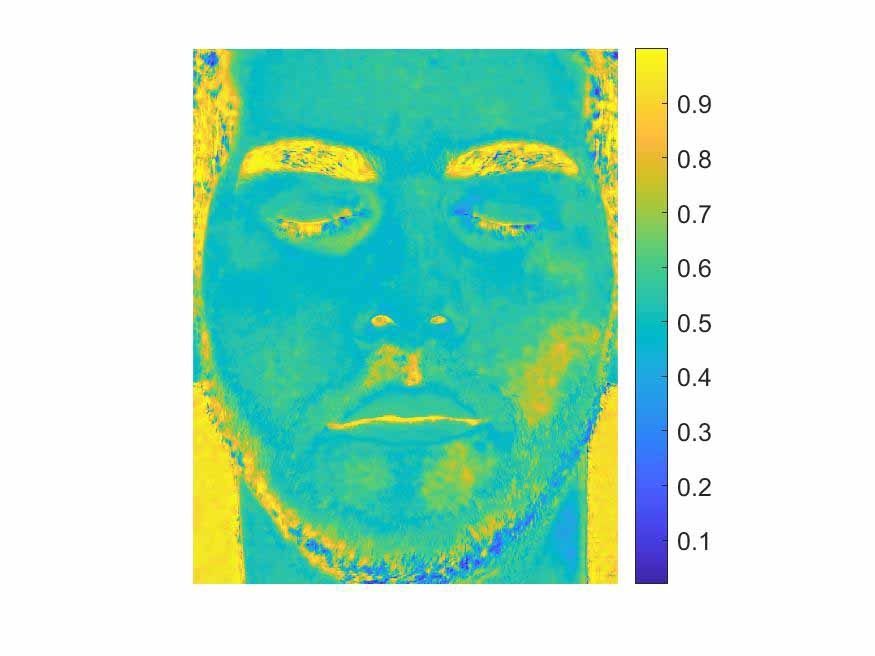}
  &
  \includegraphics[trim= 192 59 159 35, clip, width = 0.15\textwidth]{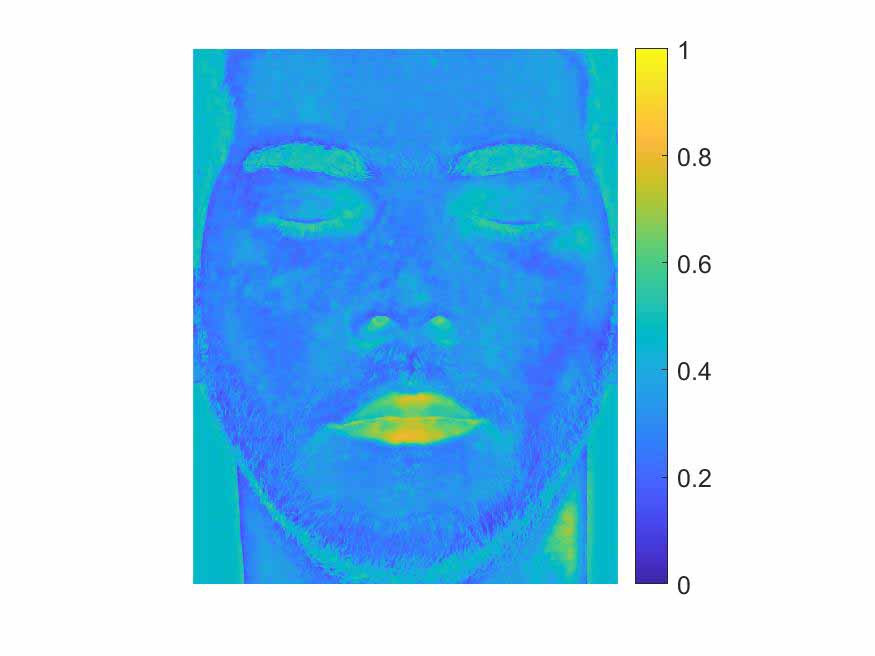}
  &
  \includegraphics[trim= 192 59 130 35, clip, width = 0.16\textwidth]{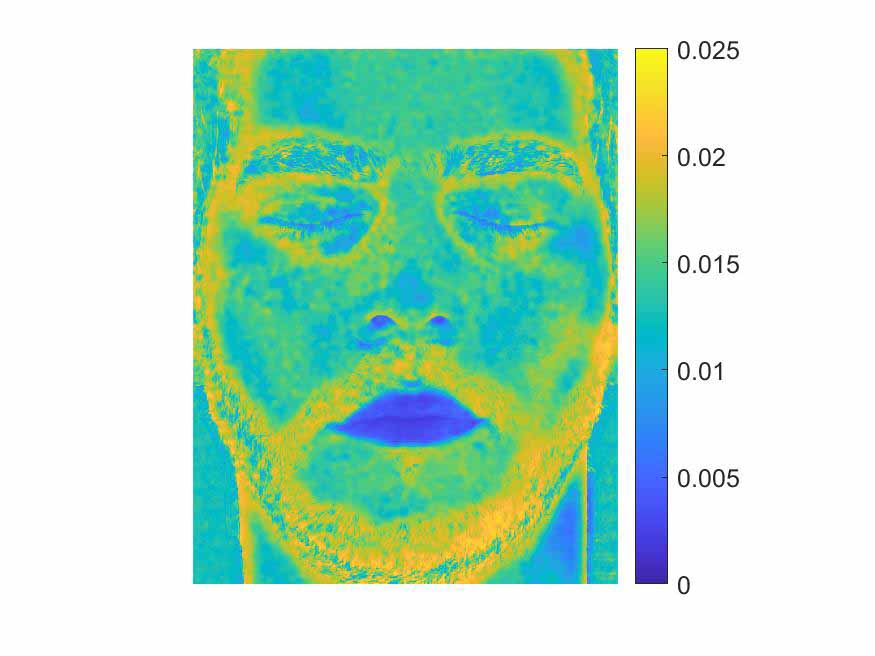}

  \\
    \begin{sideways}\hspace{0.6cm}\textsc{C2 [GGD'20]}\end{sideways}
  &
  \includegraphics[trim= 194 59 154 35, clip, width = 0.15\textwidth]{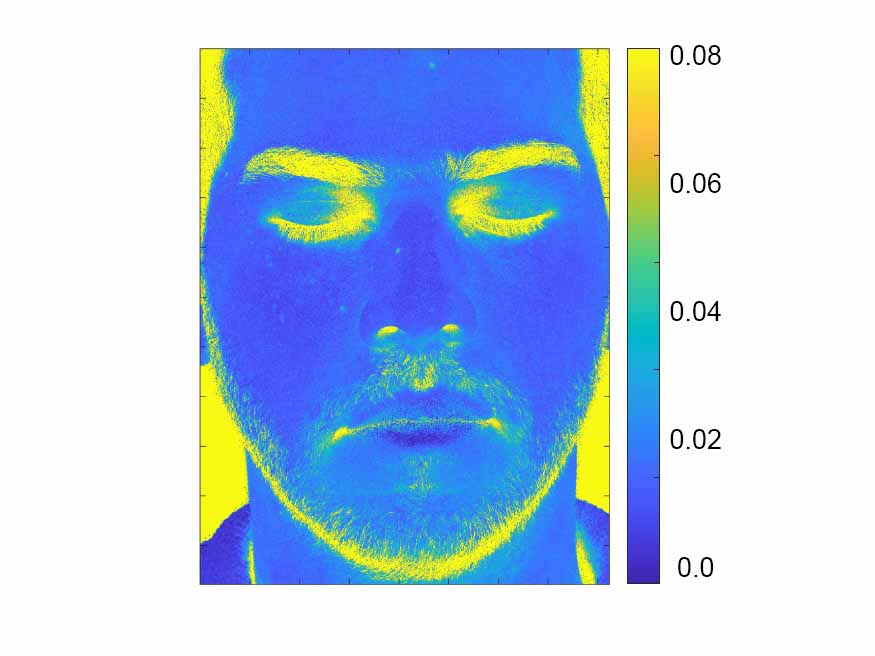}
  &
  \includegraphics[trim= 194 59 154 35, clip, width = 0.15\textwidth]{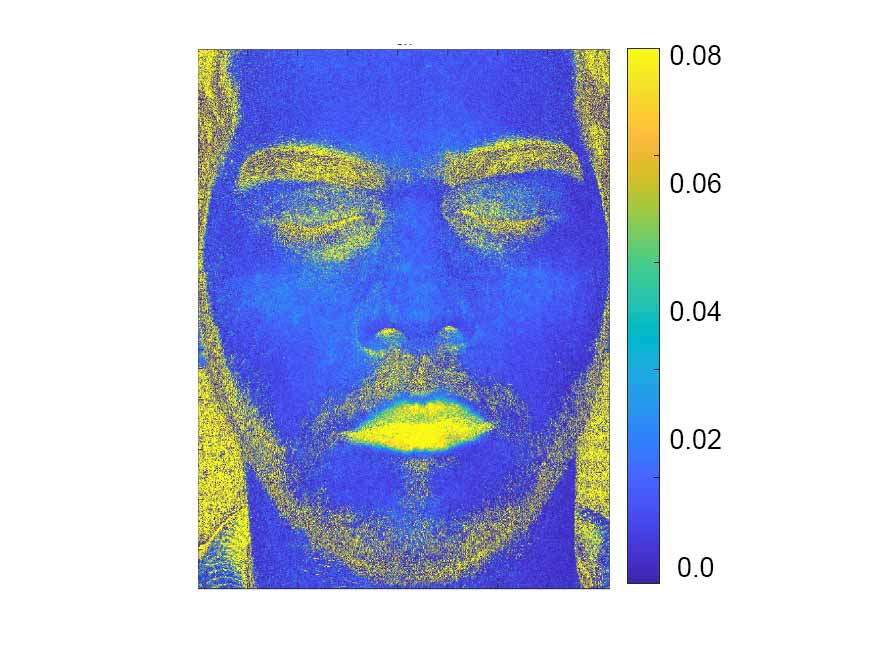}
  &
  \includegraphics[trim= 194 59 154 35, clip, width = 0.15\textwidth]{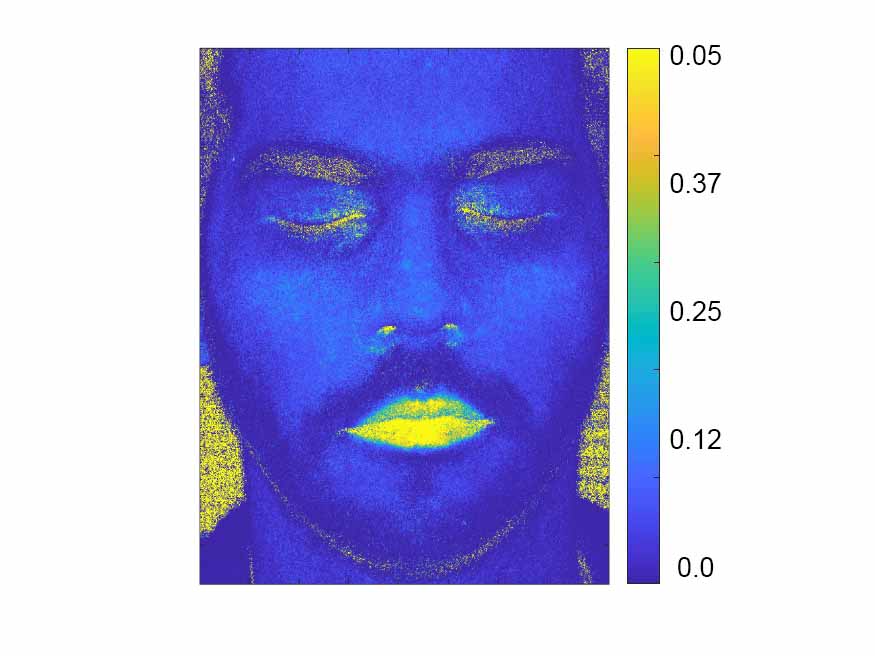}
  &
  \includegraphics[trim= 194 59 154 35, clip, width = 0.15\textwidth]{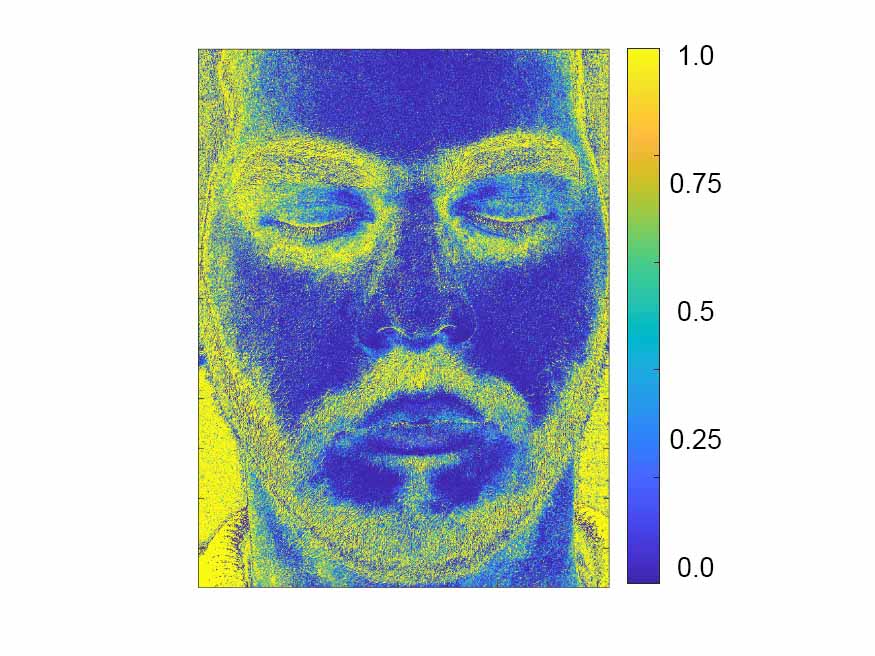}
  &
  &

  \\
  \begin{sideways}\hspace{0.5cm}\textsc{C2 Ours D65}\end{sideways}
  &
  \includegraphics[trim= 192 59 159 35, clip, width = 0.15\textwidth]{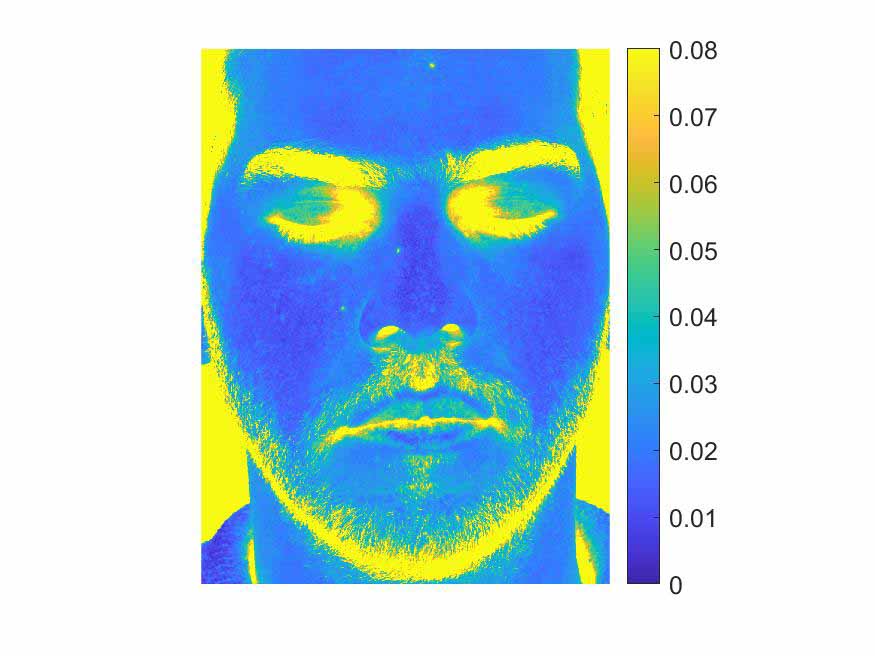}
  &
  \multicolumn{2}{c}{\includegraphics[trim= 192 59 159 35, clip, width = 0.15\textwidth]{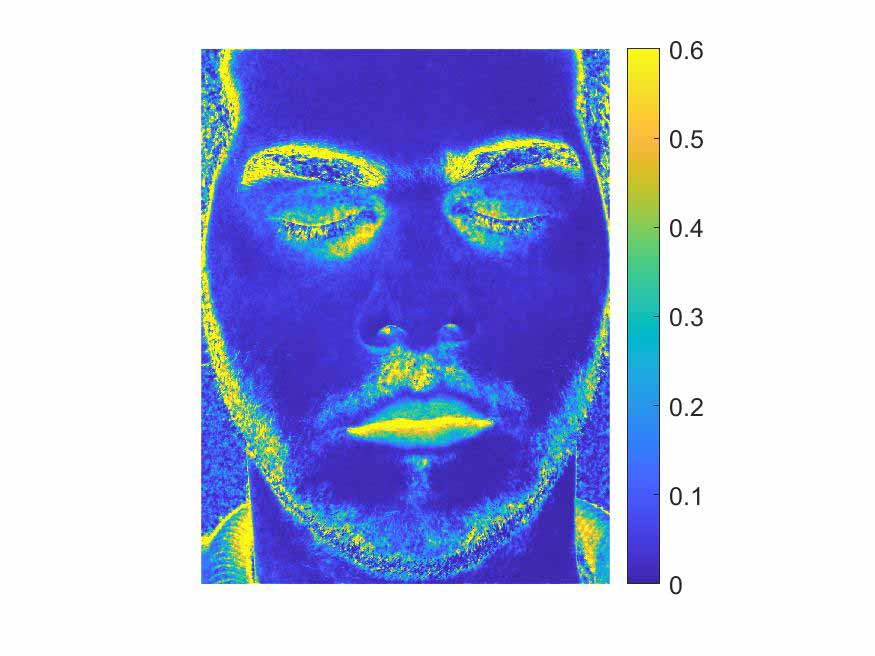}}
  &
  \includegraphics[trim= 192 59 159 35, clip, width = 0.15\textwidth]{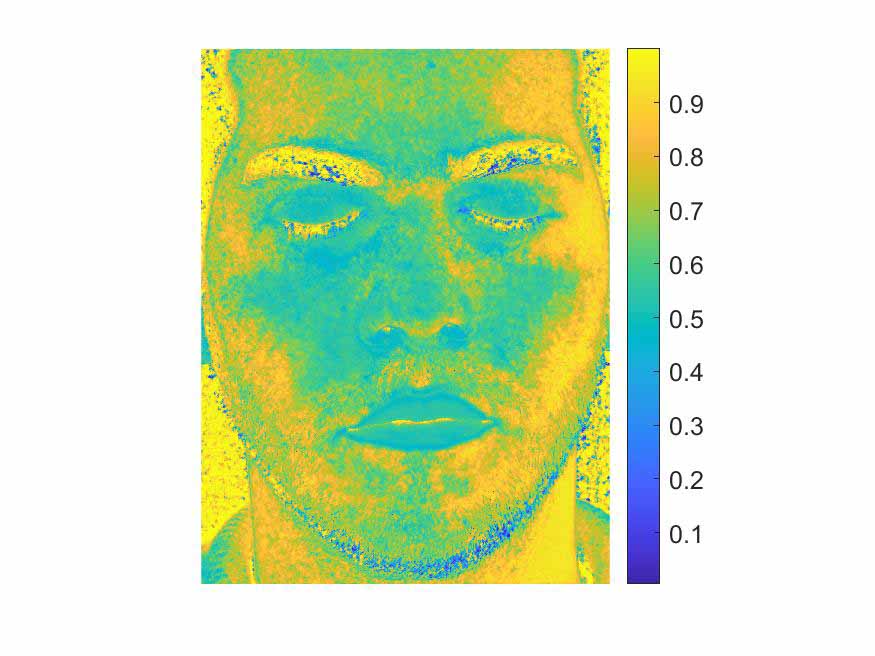}
  &
  \includegraphics[trim= 192 59 159 35, clip, width = 0.15\textwidth]{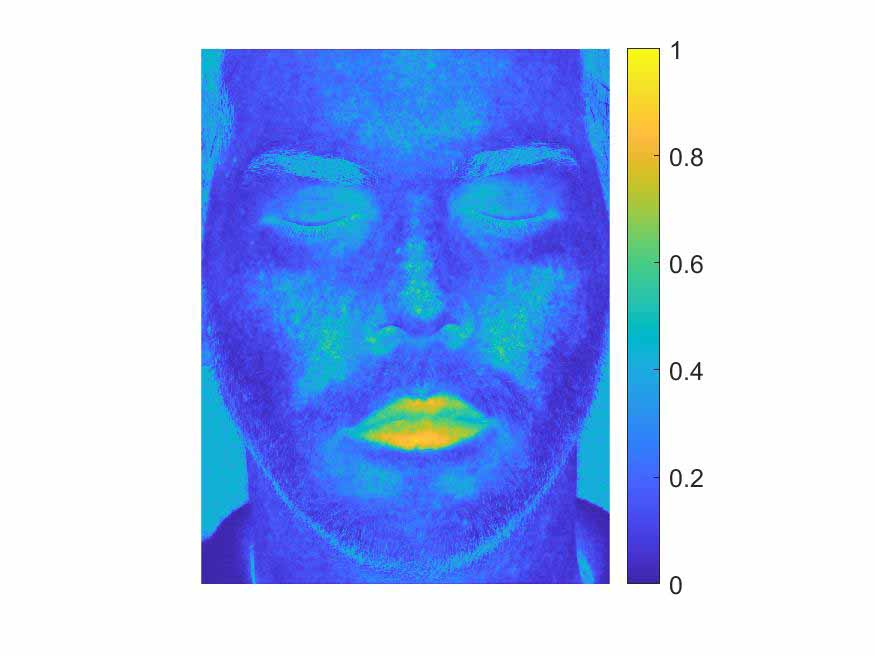}
  &
  \includegraphics[trim= 192 59 130 35, clip, width = 0.16\textwidth]{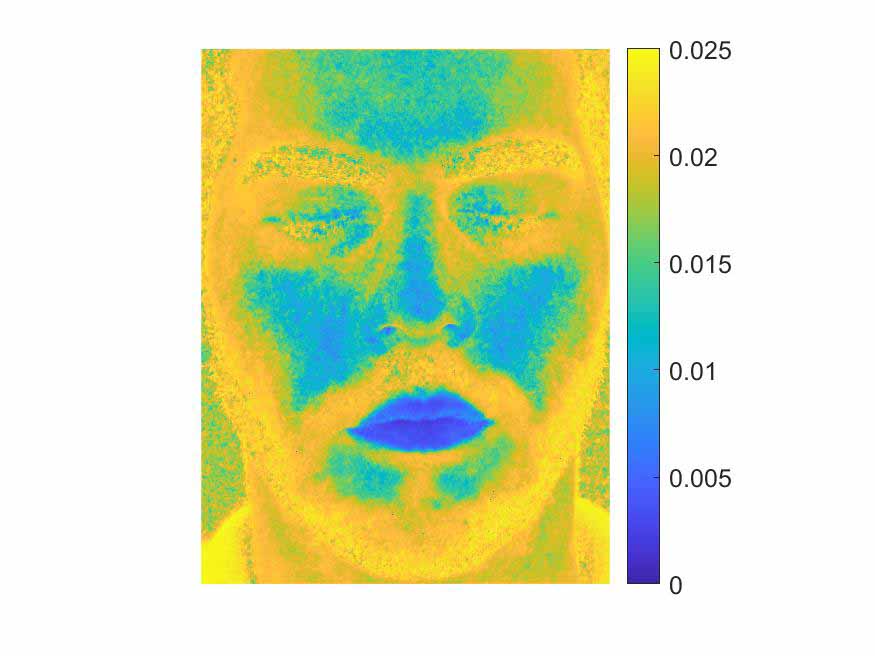}

    \\
  \begin{sideways}\hspace{0.5cm}\textsc{C2 Ours D65*}\end{sideways}
  &
  \includegraphics[trim= 192 59 159 35, clip, width = 0.15\textwidth]{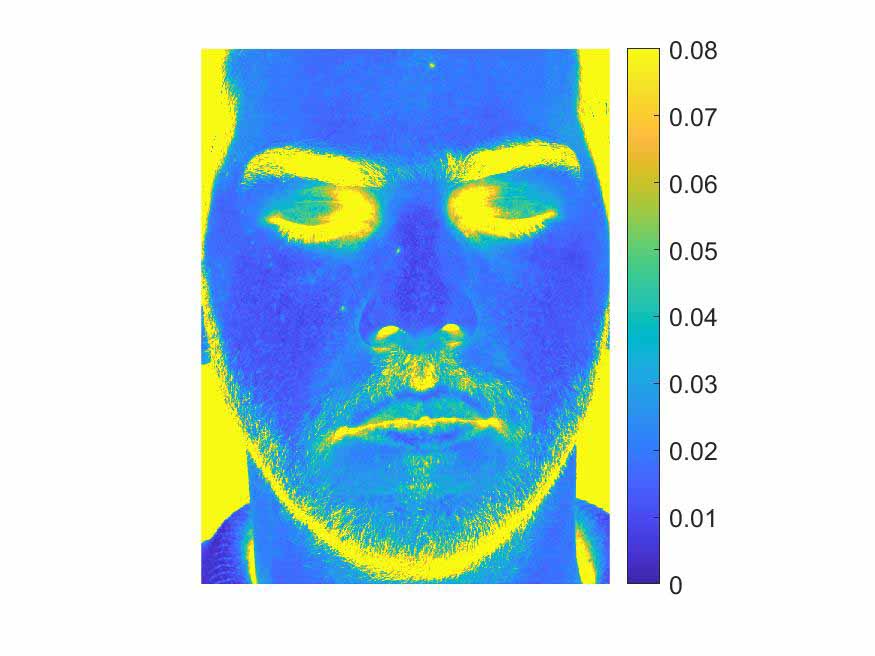}
  &
  \multicolumn{2}{c}{\includegraphics[trim= 192 59 159 35, clip, width = 0.15\textwidth]{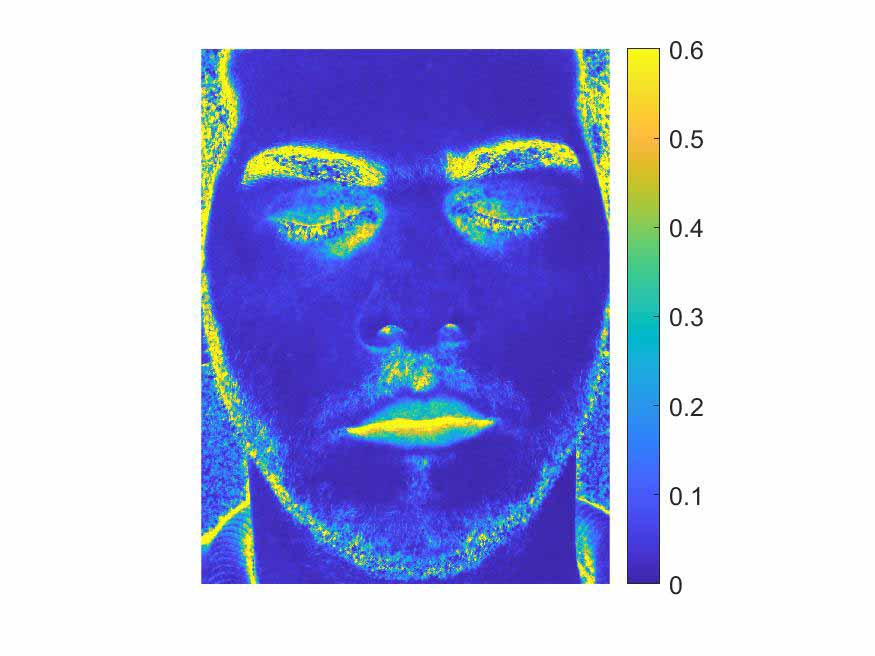}}
  &
  \includegraphics[trim= 192 59 159 35, clip, width = 0.15\textwidth]{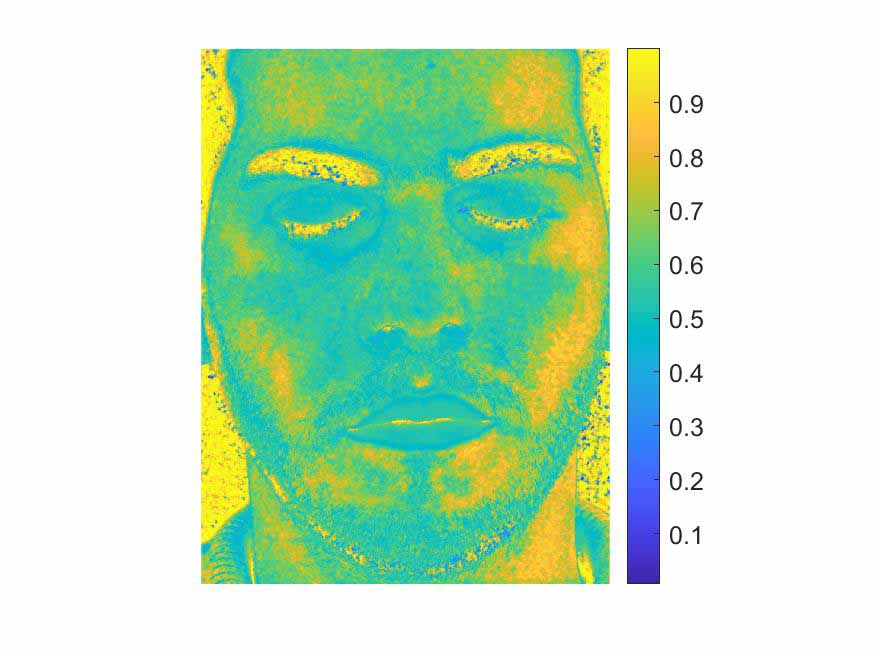}
  &
  \includegraphics[trim= 192 59 159 35, clip, width = 0.15\textwidth]{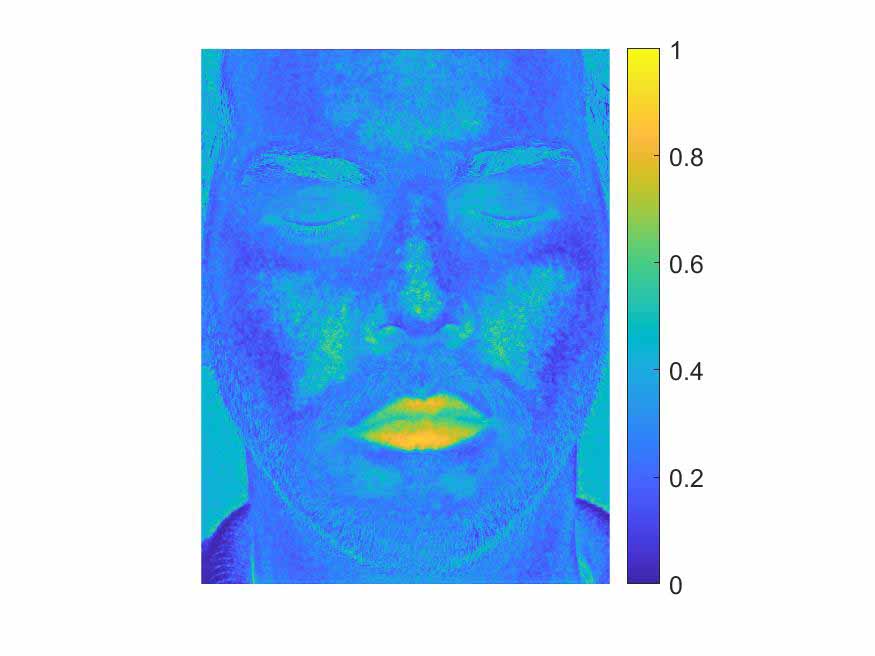}
  &
  \includegraphics[trim= 192 59 130 35, clip, width = 0.16\textwidth]{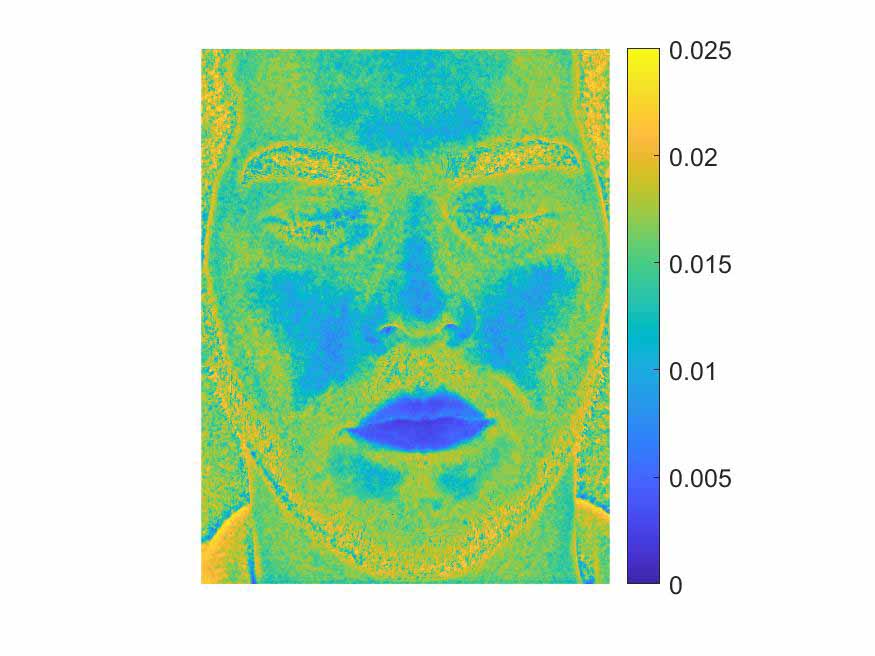}

 \end{tabular}
  \caption{\emph{Skin parameter estimation comparison}. For two captures C1 and C2, we compare our reconstructed parameters for both D65 and delight (D65*) versions (corresponding to Supplemental Figure~\ref{fig:comp_gitlina_albedos}), against~\cite{gitlina2020practical}. Similarly to the reconstruction error in Figure~\ref{fig:comp_gitlina_albedos} in the paper, from using input photographs as opposed to albedos, the extracted chromophore contents are not absolutely correct. Nonetheless our values are on par with what the previous work presented.}
  \label{fig:comp_gitlina_params}
\end{figure*}

Moreover, there is no clear conversion between the parameters of Radiative Transfer and those of KM theories. We tested existing empirical relationships~\cite{roy2012empirical} and more grounded derived ones from optics literature~\cite{sandoval2014deriving}. Unfortunately, the model is still unable to generalize to various skin types, lacks fidelity in certain areas such as the lips, and other heterogeneities (imperfections) are not really recovered (see Figure~\ref{fig:failure_cases_KM}).

\section{Renderings of Edited Albedos under Different Lighting}
\begin{figure*}[t!]
 \contourlength{0.1em}%
 \centering
 \hspace*{-4.5mm}%
  \begin{tabular}{l@{\;}c@{\;}c@{\;}c@{\;}c@{\;}c@{\;}c@{\;}c@{\;}c}
  
 & & \textsc{0.125x blood } & \textsc{0.5x blood} & \textsc{3x blood} & \textsc{1.5x melanin} & \textsc{3x melanin} & \textsc{5x melanin} 
  \\
  \begin{sideways}\hspace{0.2cm}\textsc{Type I (Subject A)}\end{sideways}
 &
   \includegraphics[trim= 0 340 0 0, clip, width =0.14\textwidth]{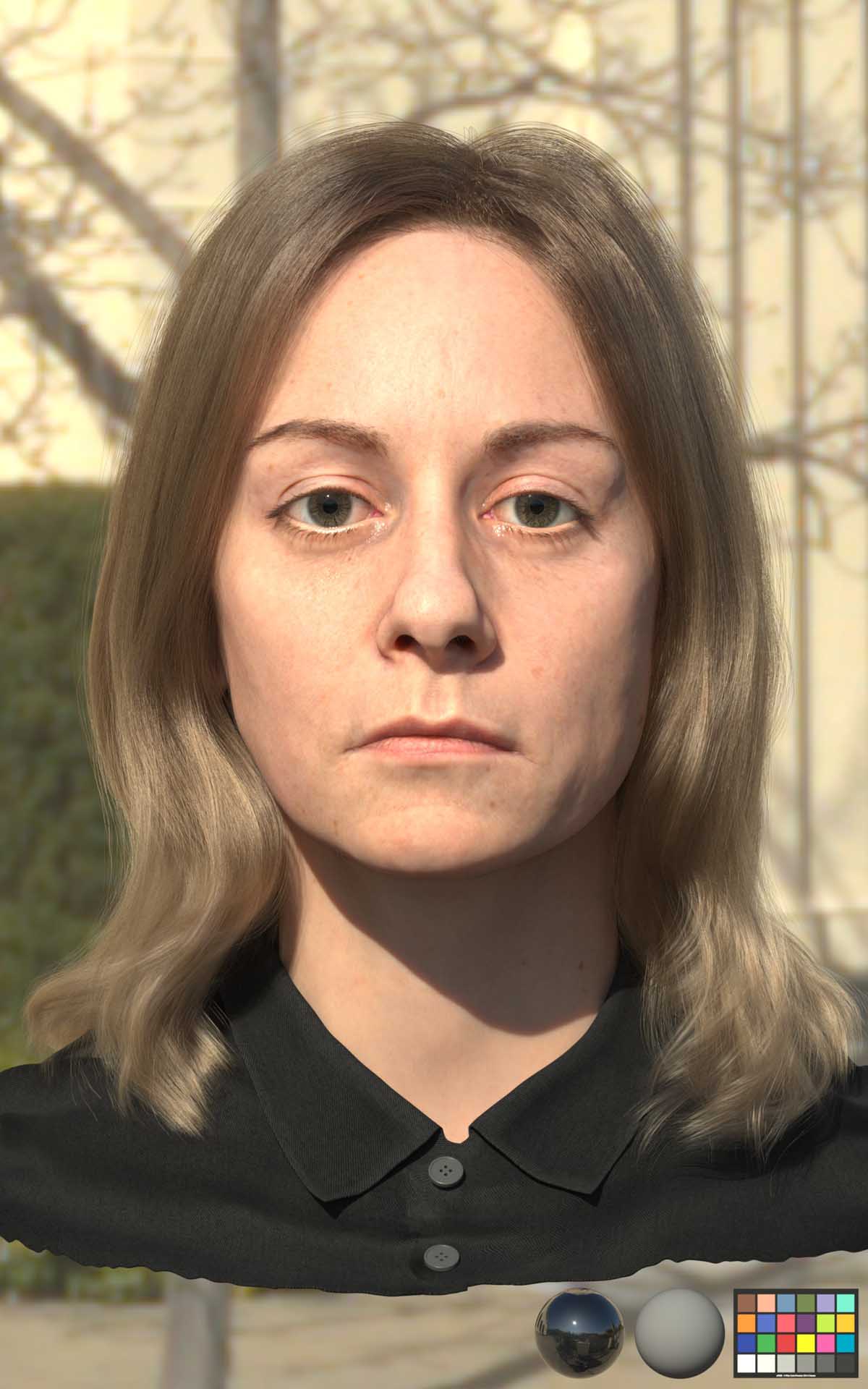}
 &
   \includegraphics[trim= 0 340 0 0, clip, width =0.14\textwidth]{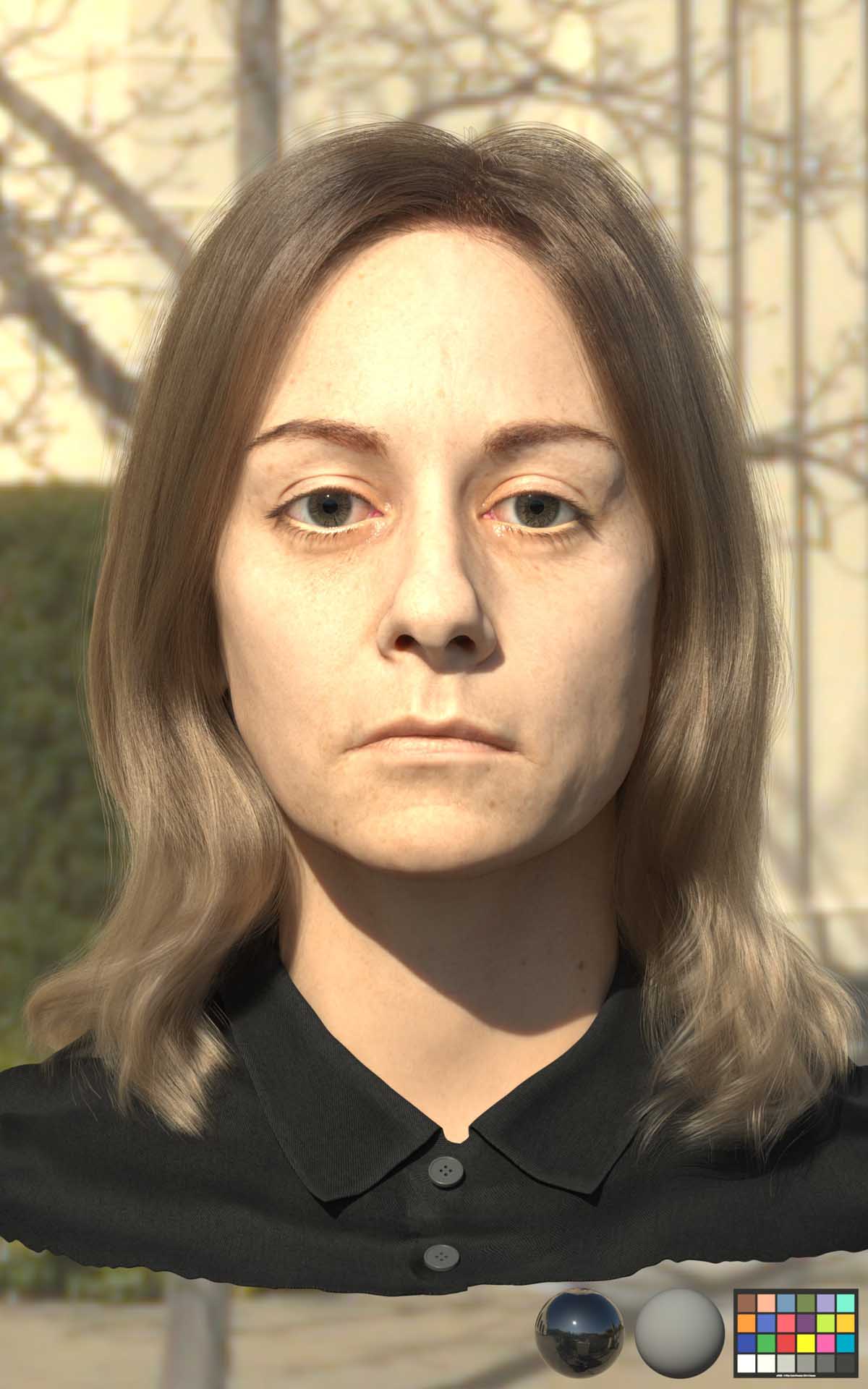}
 &
   \includegraphics[trim= 0 340 0 0, clip, width =0.14\textwidth]{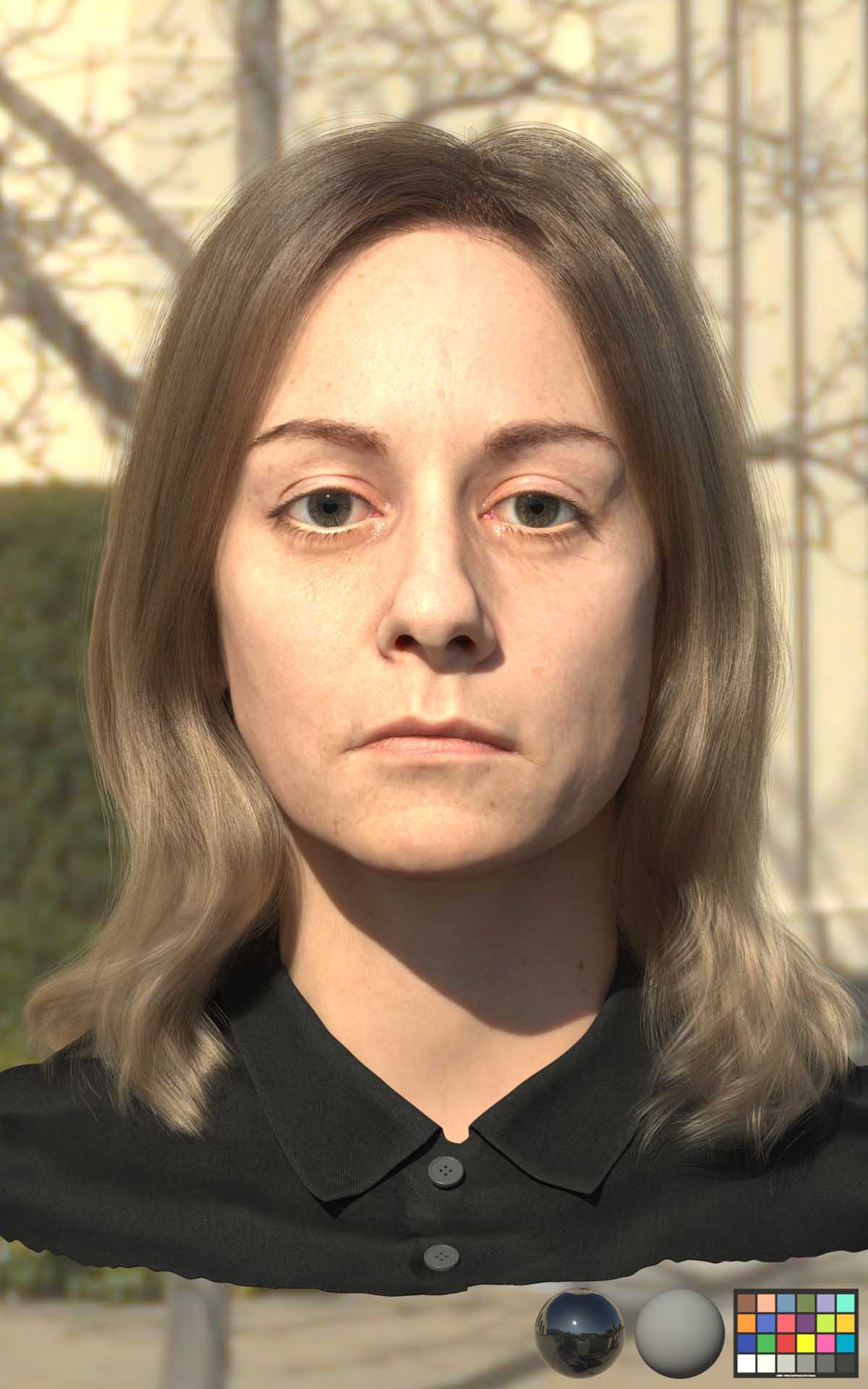}
 &
   \includegraphics[trim= 0 340 0 0, clip, width =0.14\textwidth]{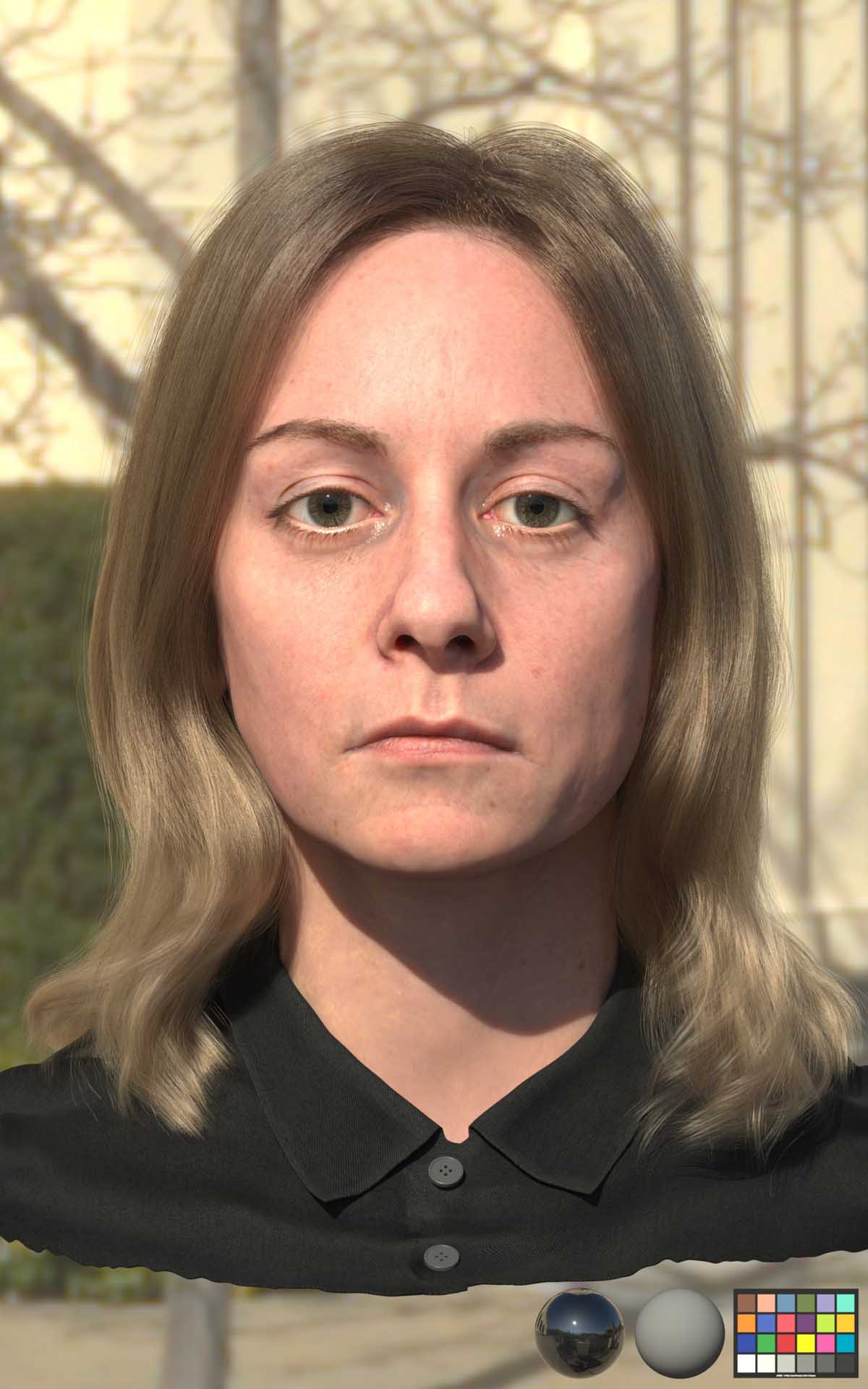}
 &
   \includegraphics[trim= 0 340 0 0, clip, width =0.14\textwidth]{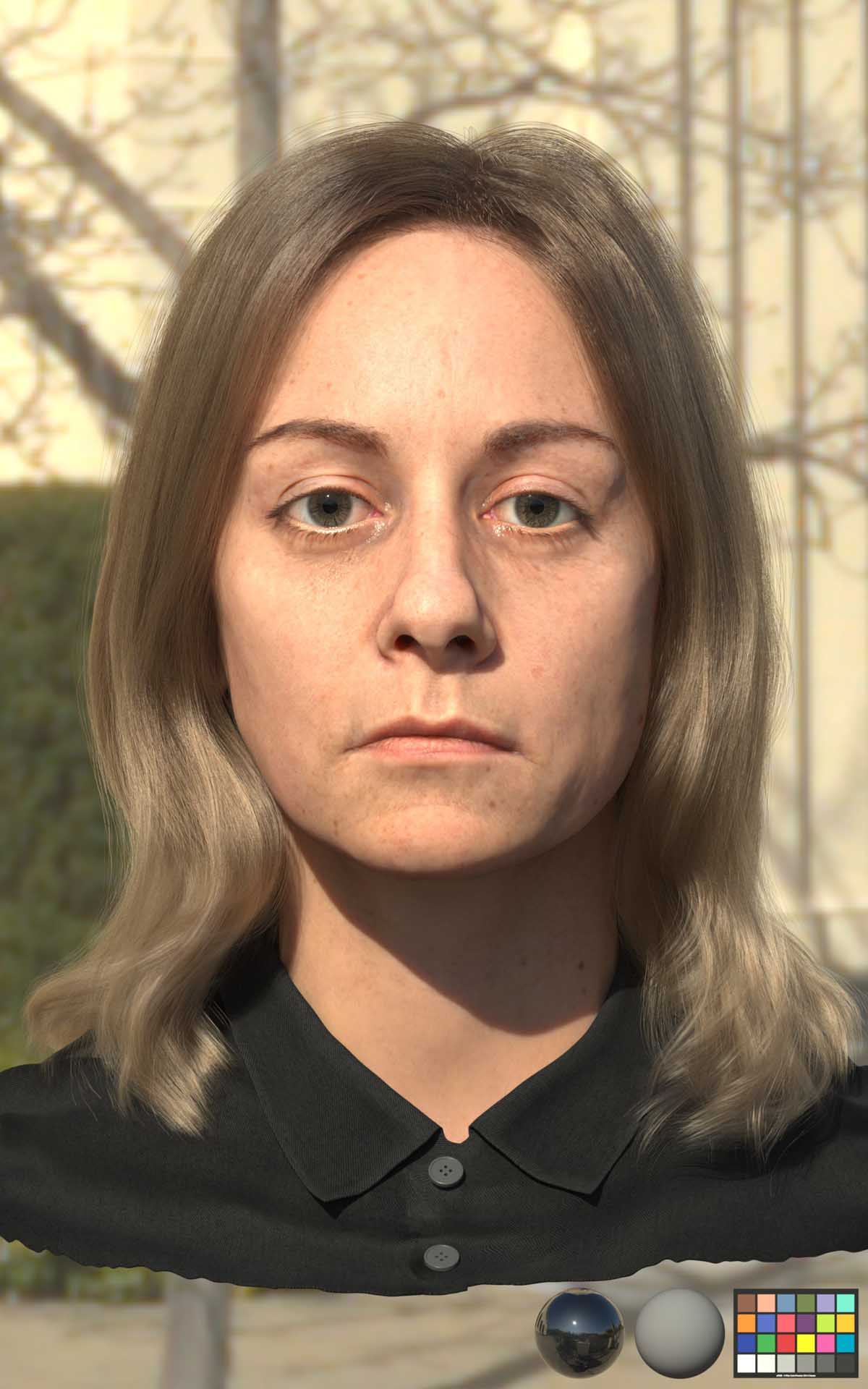}
 &
   \includegraphics[trim= 0 340 0 0, clip, width =0.14\textwidth]{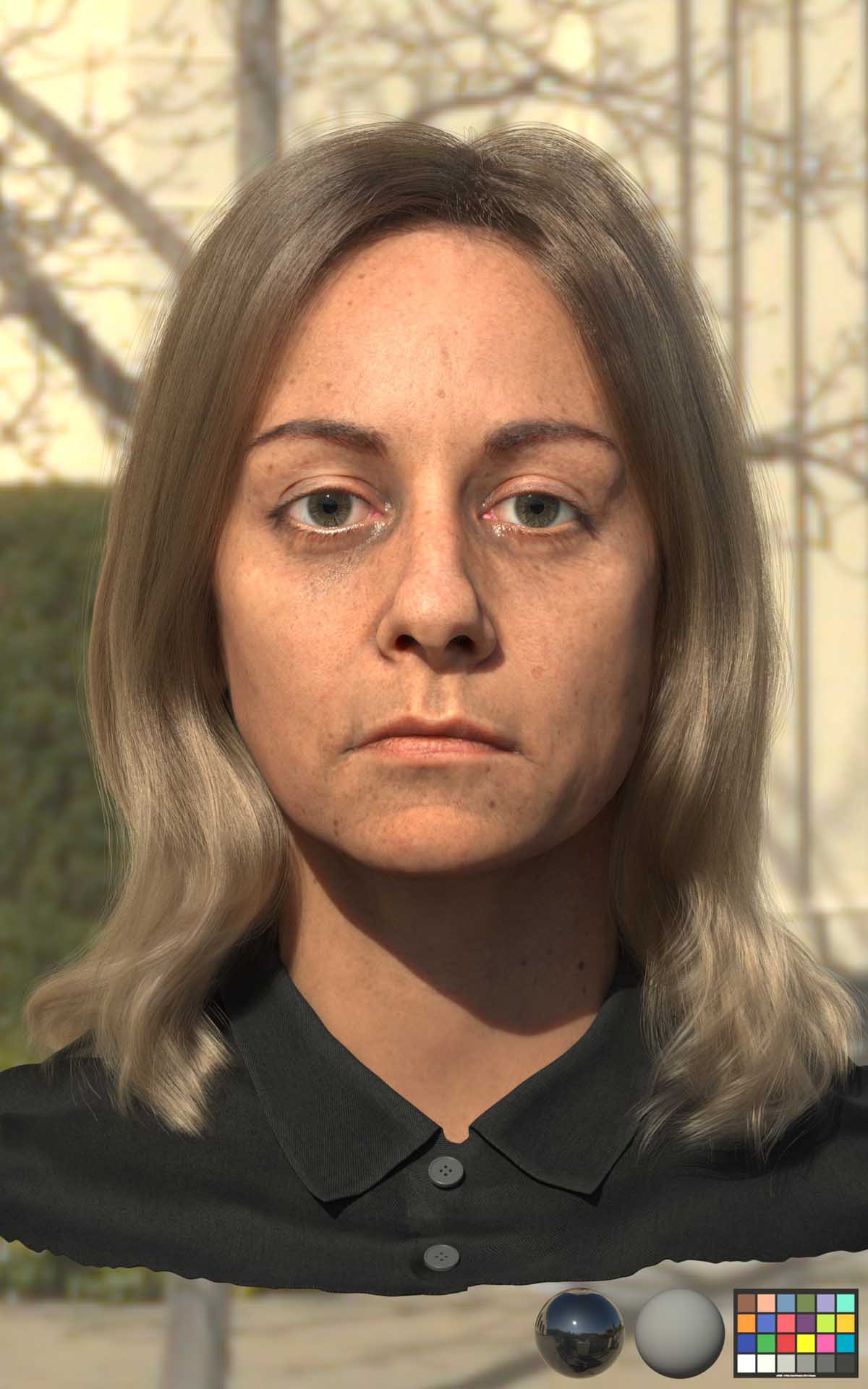}
 &
   \includegraphics[trim= 0 340 0 0, clip, width =0.14\textwidth]{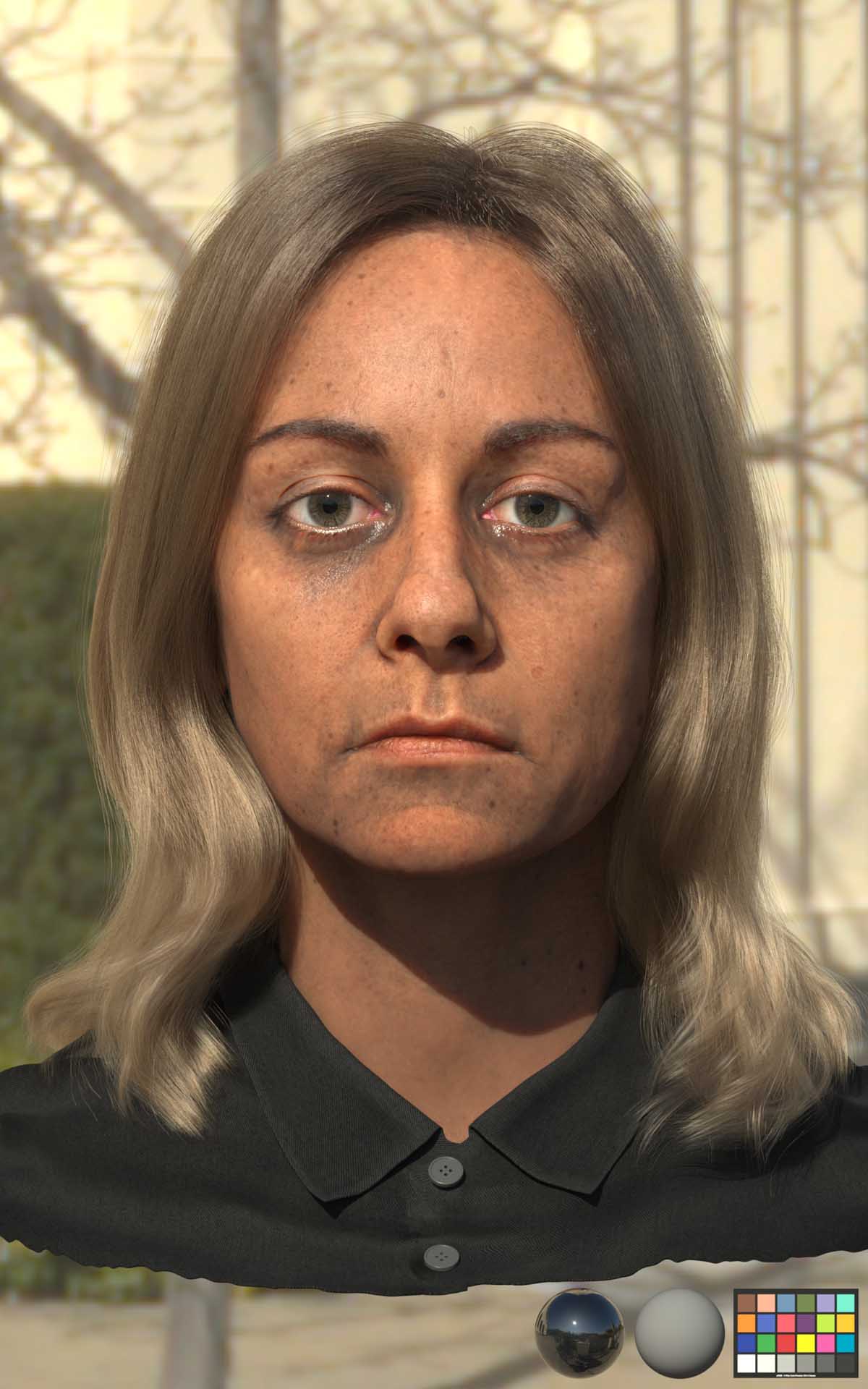}
  
  \\
 & & \textsc{0.125x blood } & \textsc{0.5x blood} & \textsc{3x blood} & \textsc{1.5x melanin} & \textsc{3x melanin} & \textsc{5x melanin} 
  \\
  \begin{sideways}\hspace{0.2cm}\textsc{Type II (Subject C)}\end{sideways}
 &
   \includegraphics[trim= 0 340 0 0, clip, width =0.14\textwidth]{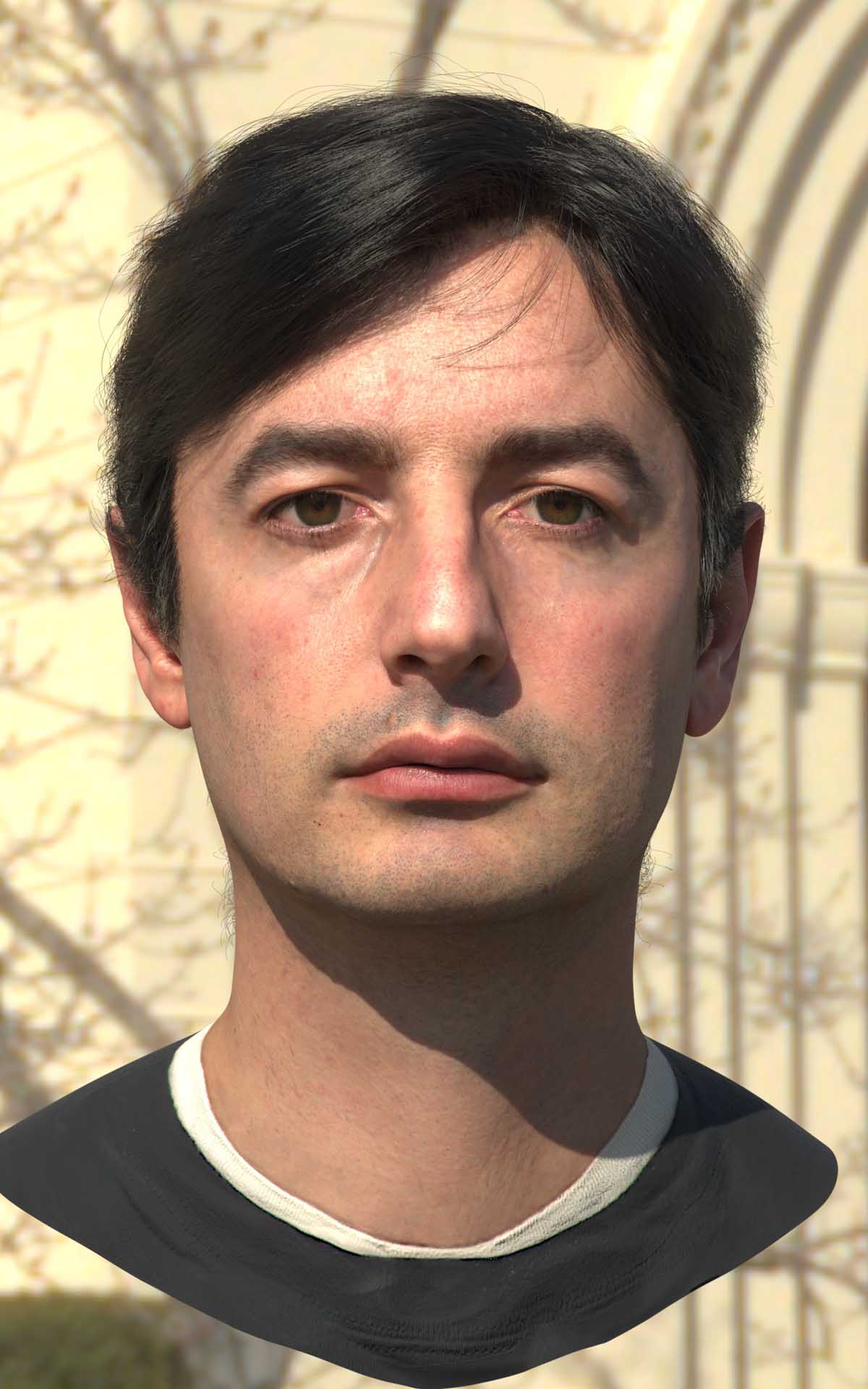}
 &
   \includegraphics[trim= 0 340 0 0, clip, width =0.14\textwidth]{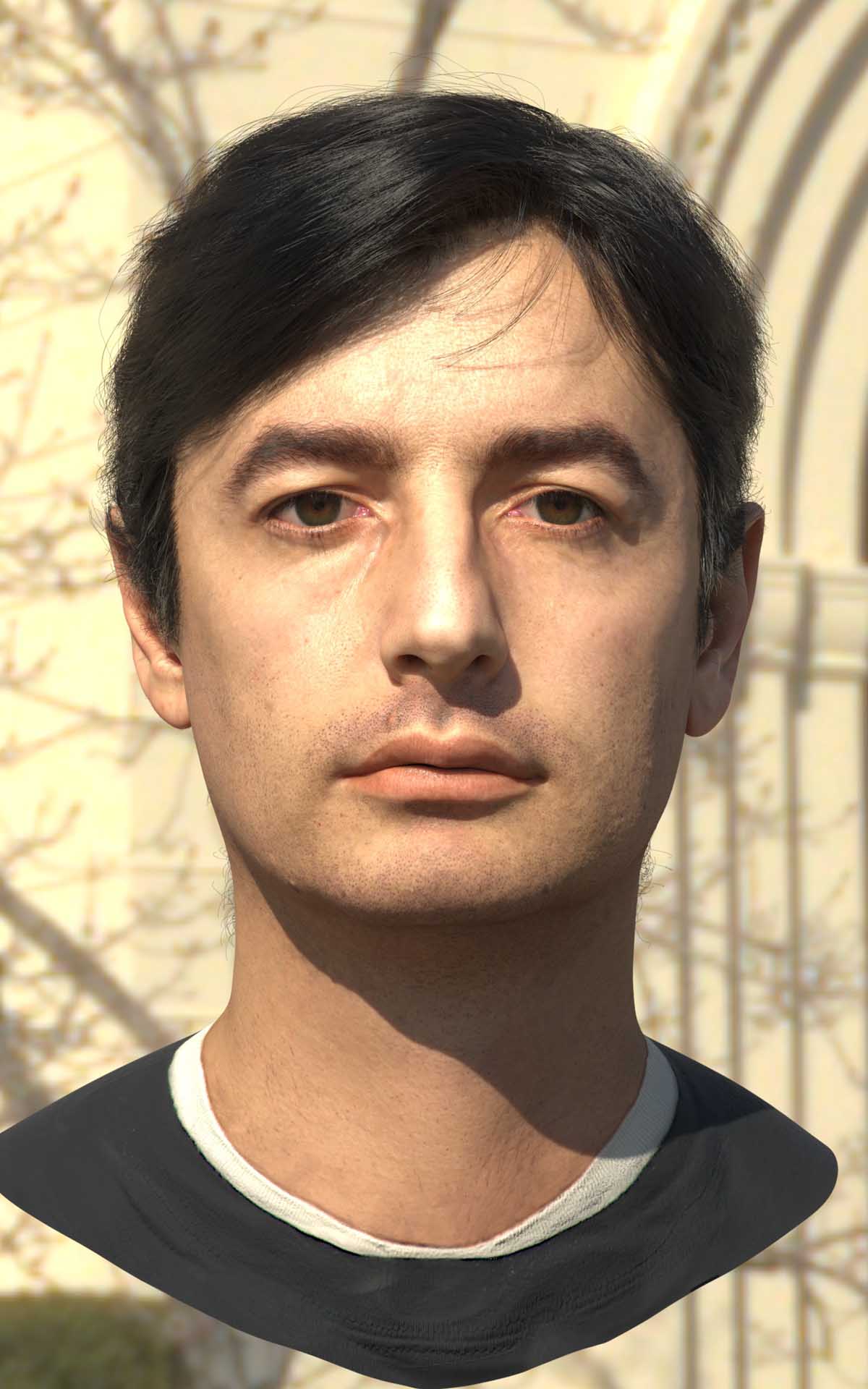}
 &
   \includegraphics[trim= 0 340 0 0, clip, width =0.14\textwidth]{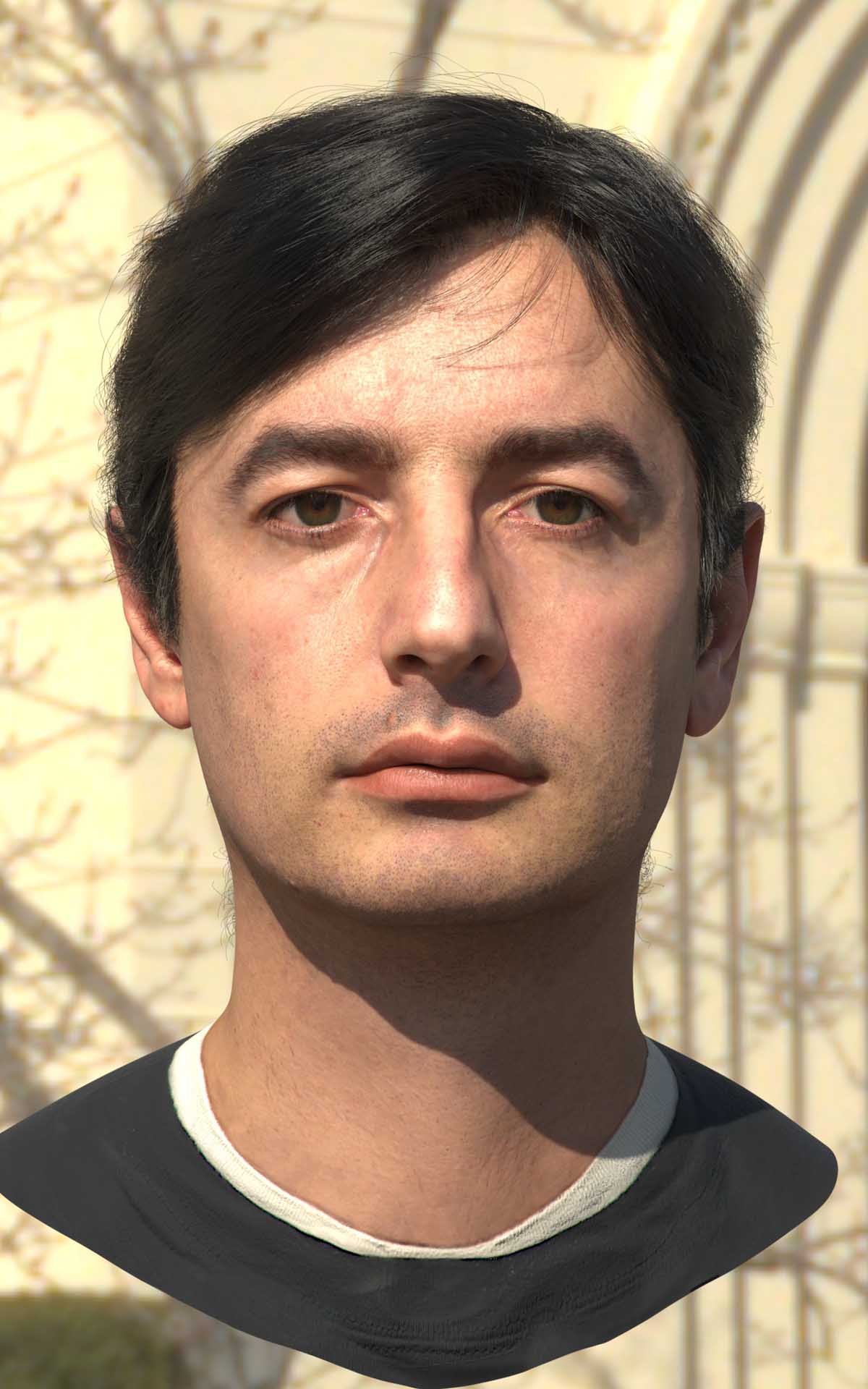}
 &
   \includegraphics[trim= 0 340 0 0, clip, width =0.14\textwidth]{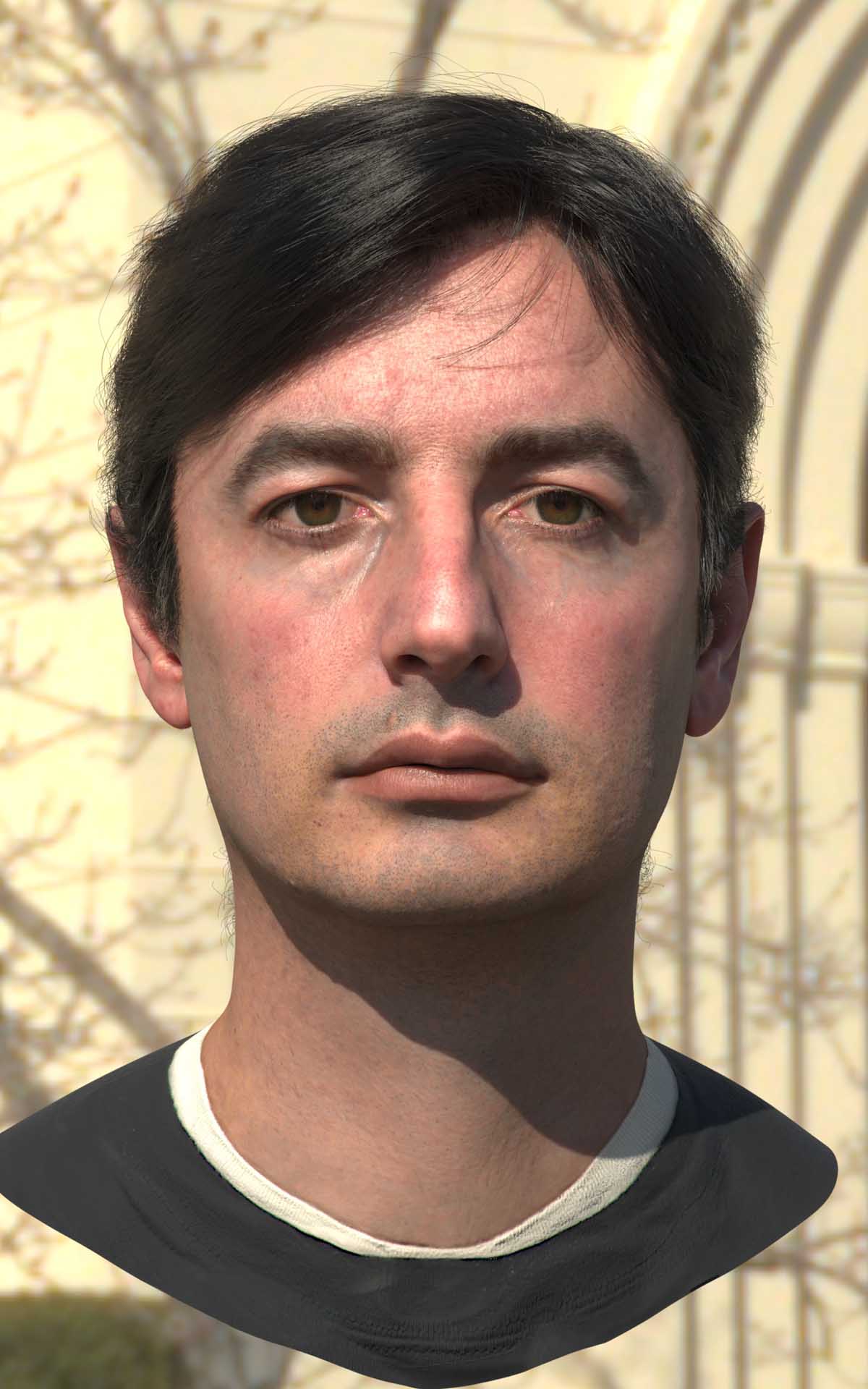}
 &
   \includegraphics[trim= 0 340 0 0, clip, width =0.14\textwidth]{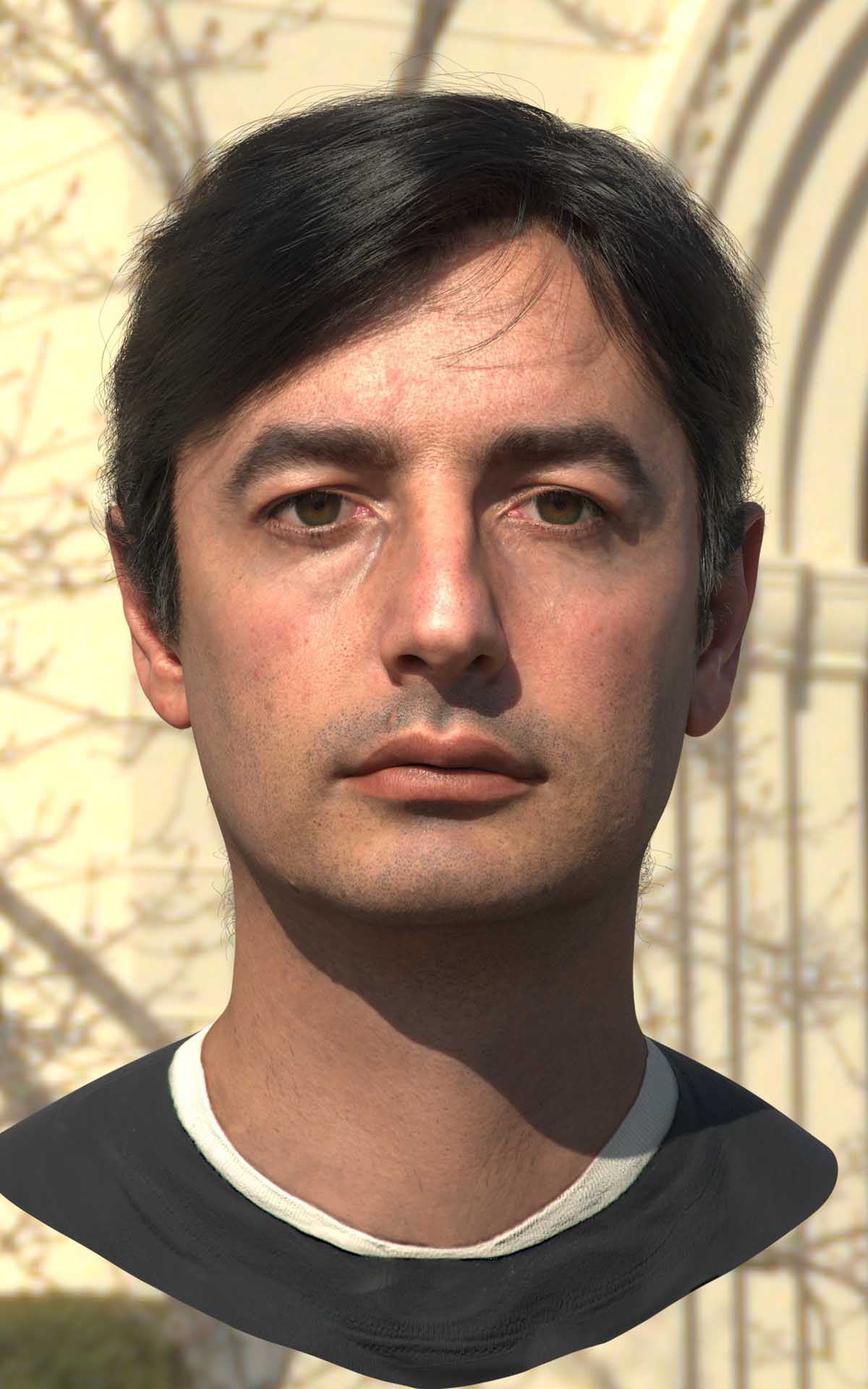}
 &
   \includegraphics[trim= 0 340 0 0, clip, width =0.14\textwidth]{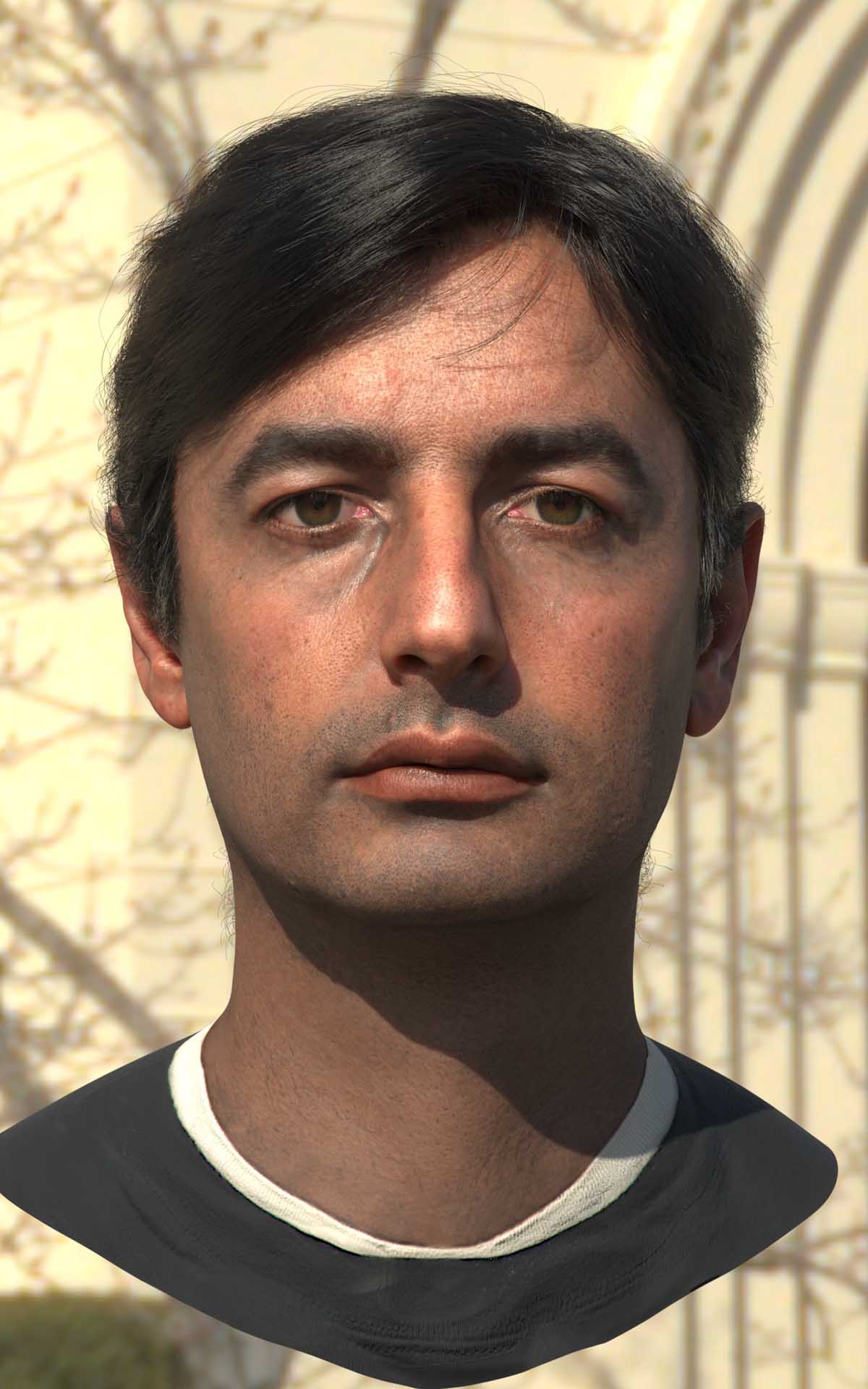}
 &
   \includegraphics[trim= 0 340 0 0, clip, width =0.14\textwidth]{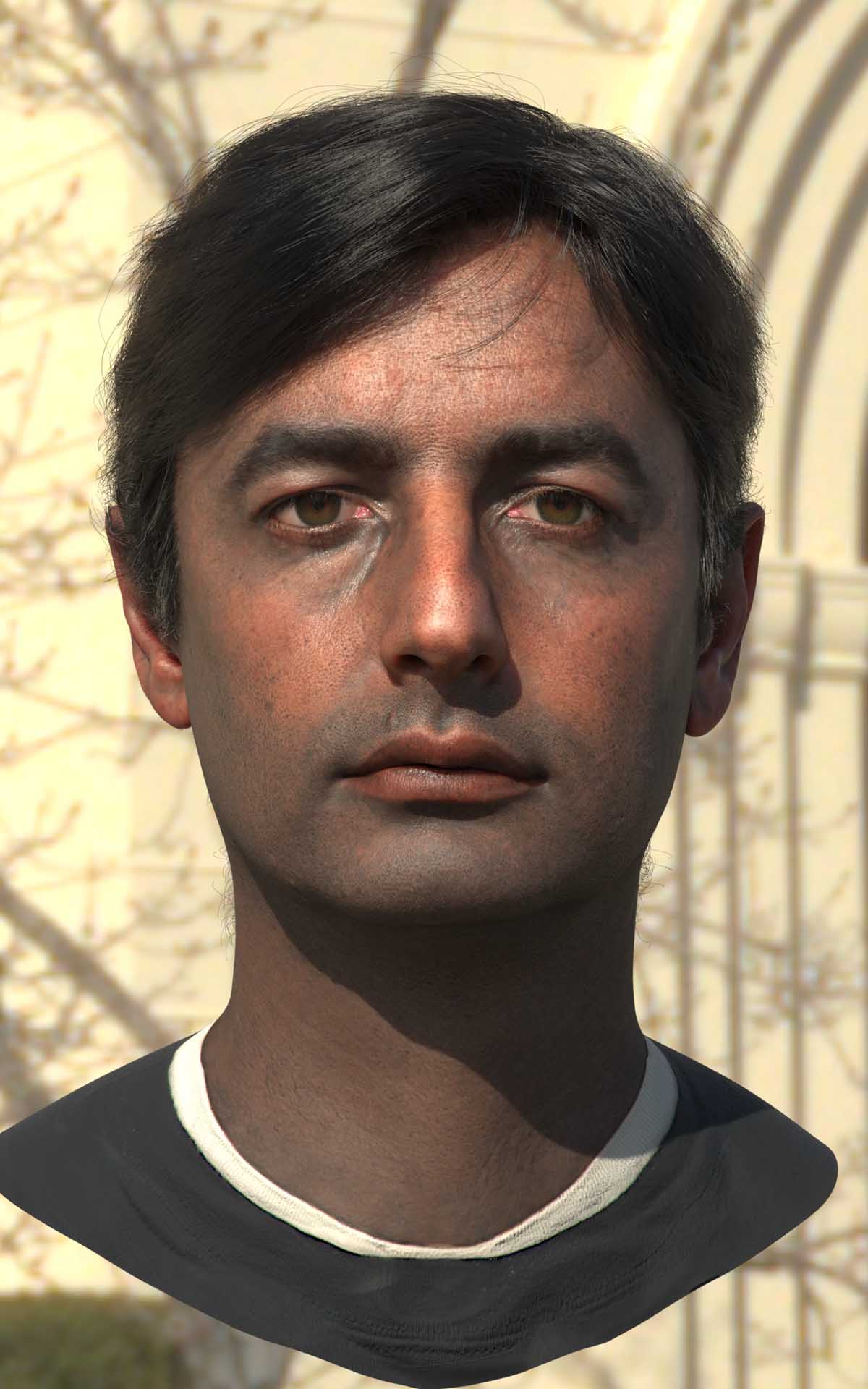}
  
  \\
 & & \textsc{0.125x blood } & \textsc{0.5x blood} & \textsc{3x blood} & \textsc{0.5x melanin} & \textsc{1.5x melanin} & \textsc{3x melanin} 
  \\
  \begin{sideways}\hspace{0.2cm}\textsc{Type III (Subject D)}\end{sideways}
 &
   \includegraphics[trim= 0 340 0 0, clip, width =0.14\textwidth]{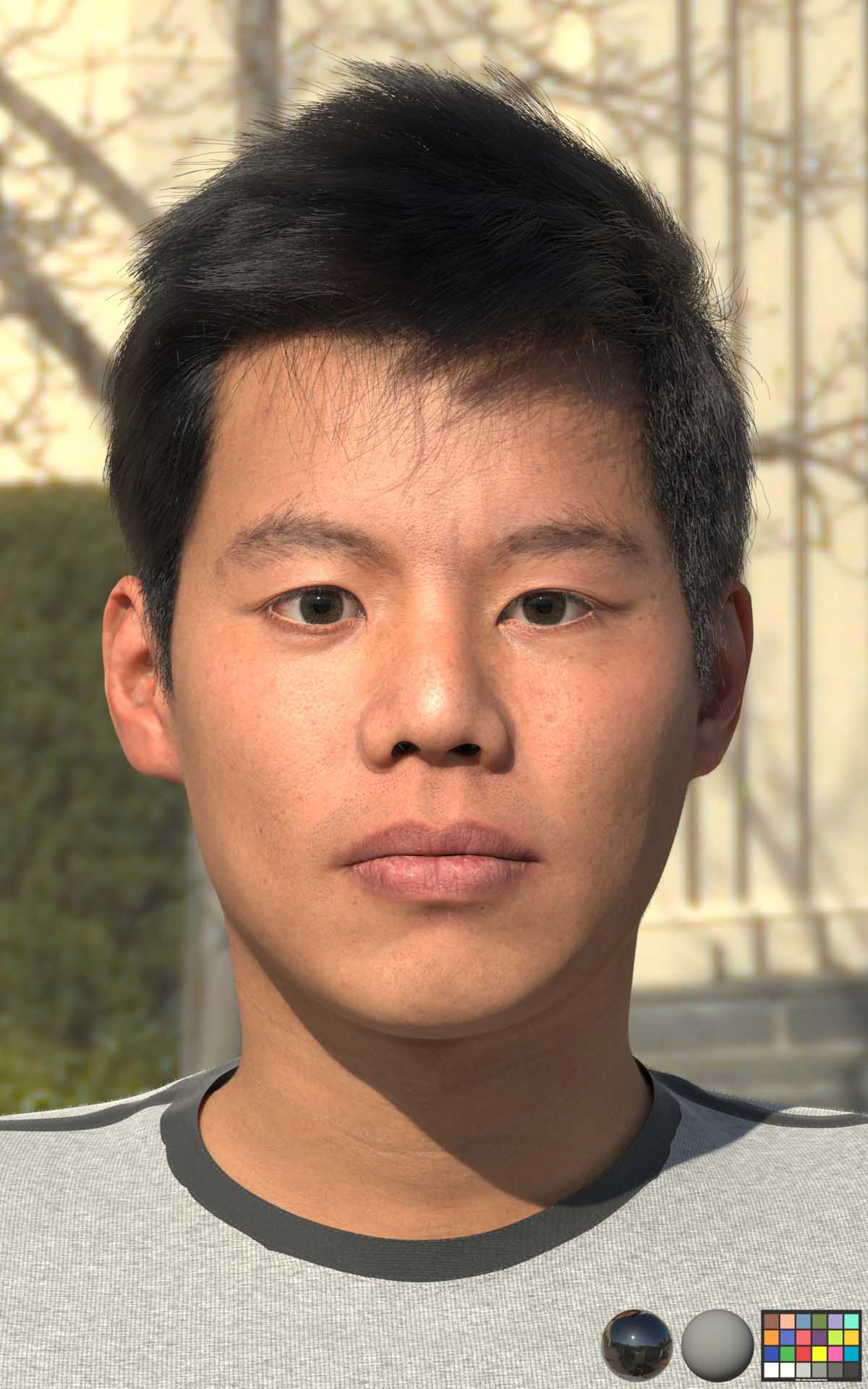}
 &
   \includegraphics[trim= 0 340 0 0, clip, width =0.14\textwidth]{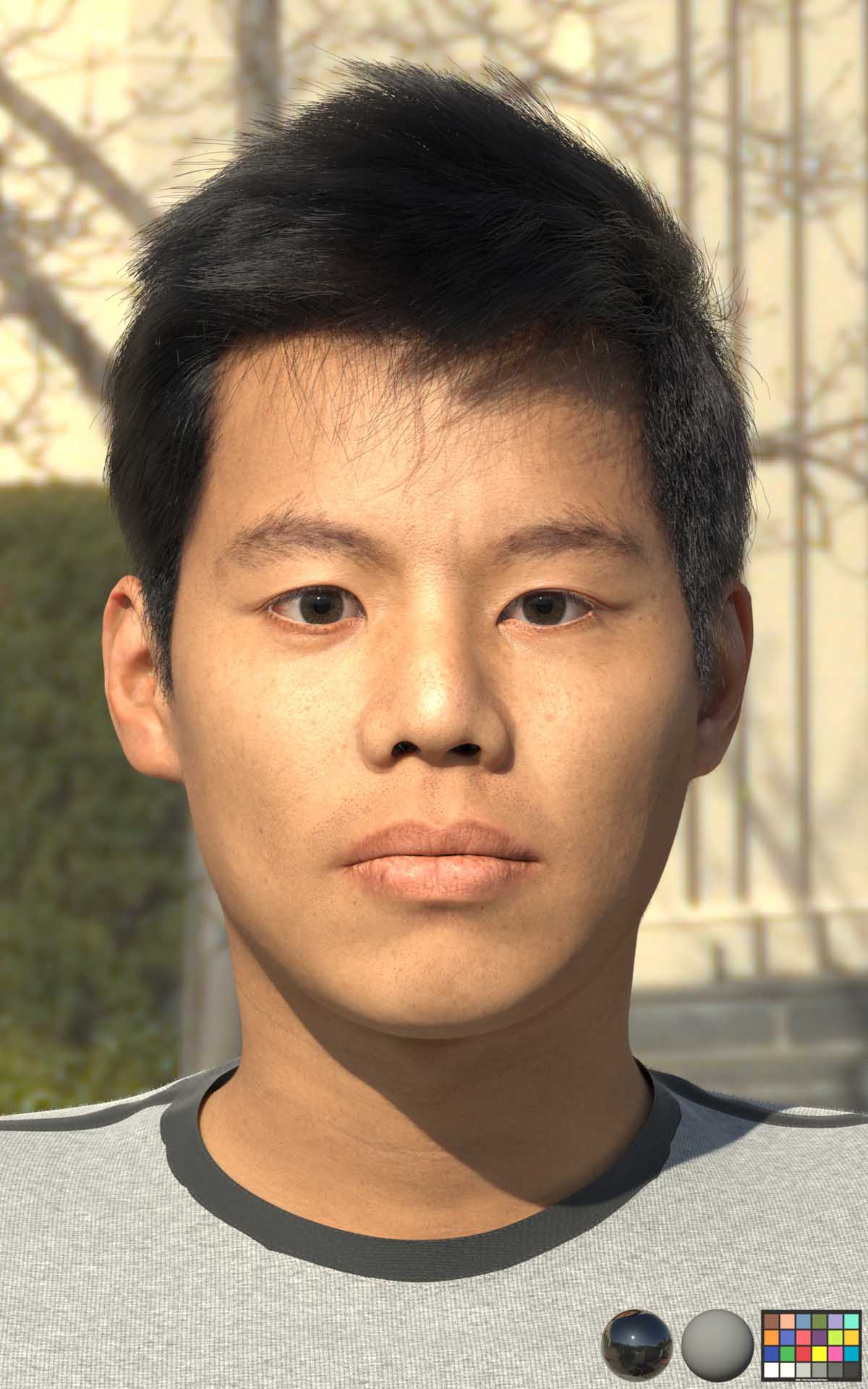}
 &
   \includegraphics[trim= 0 340 0 0, clip, width =0.14\textwidth]{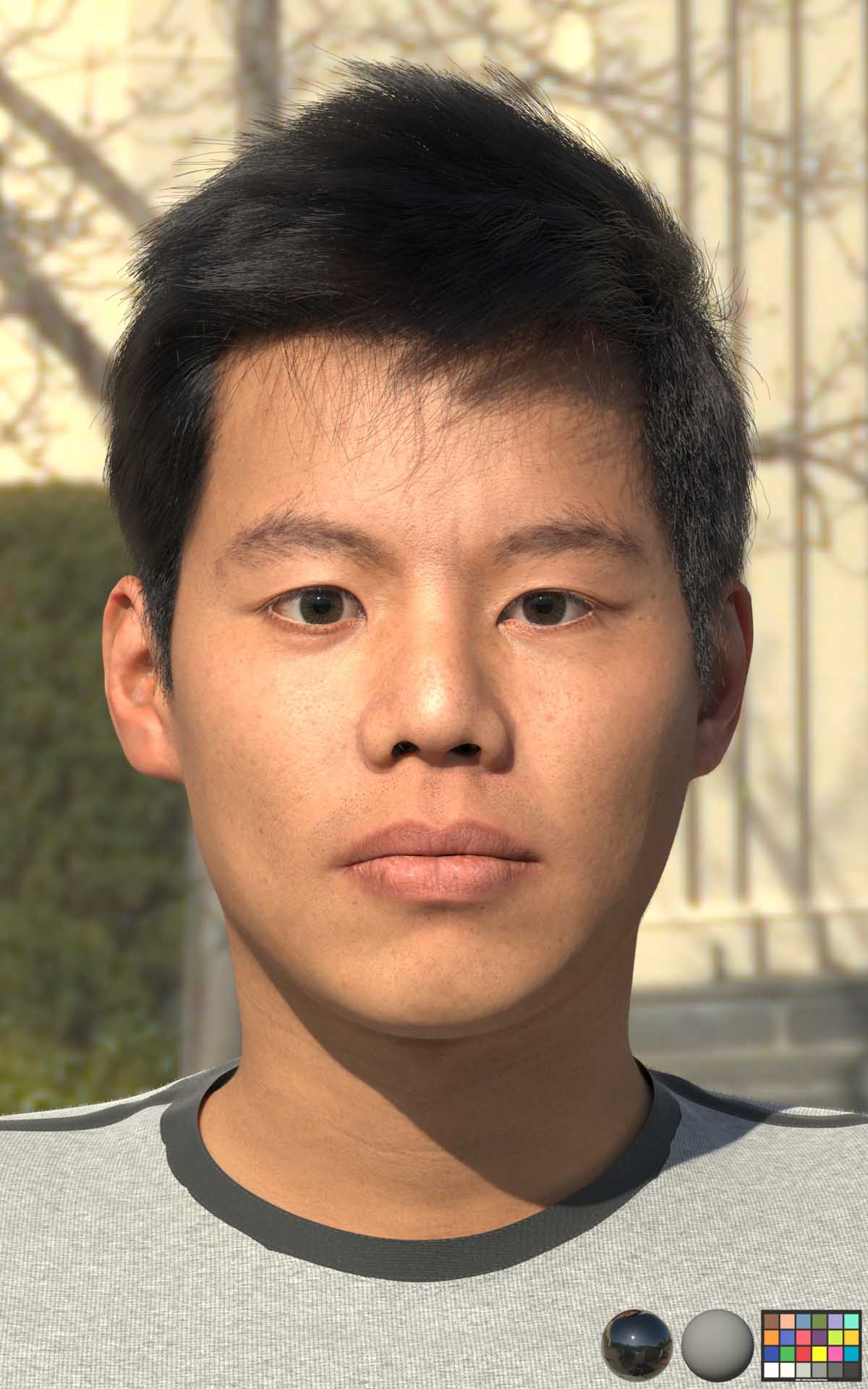}
 &
   \includegraphics[trim= 0 340 0 0, clip, width =0.14\textwidth]{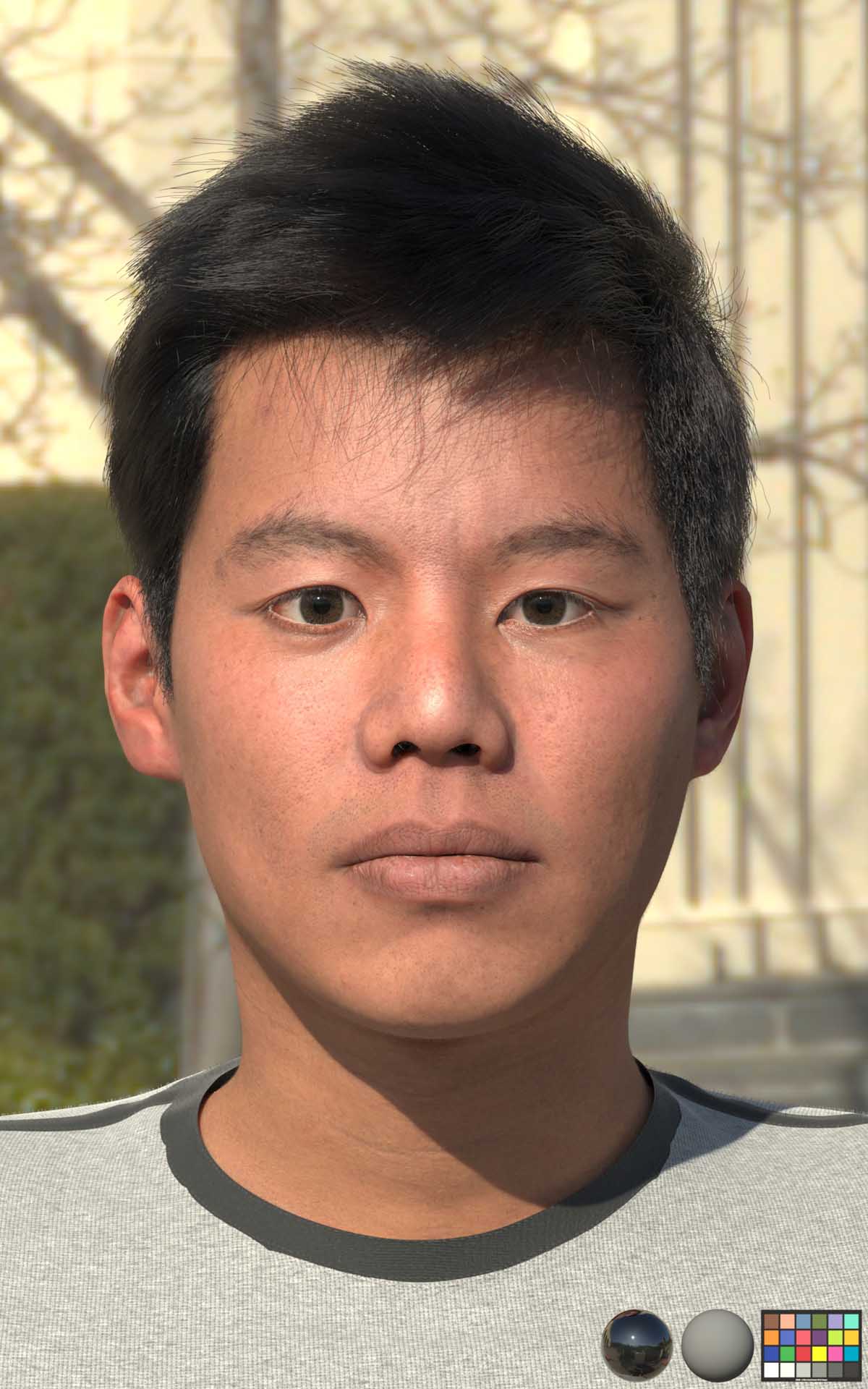}
 &
   \includegraphics[trim= 0 340 0 0, clip, width =0.14\textwidth]{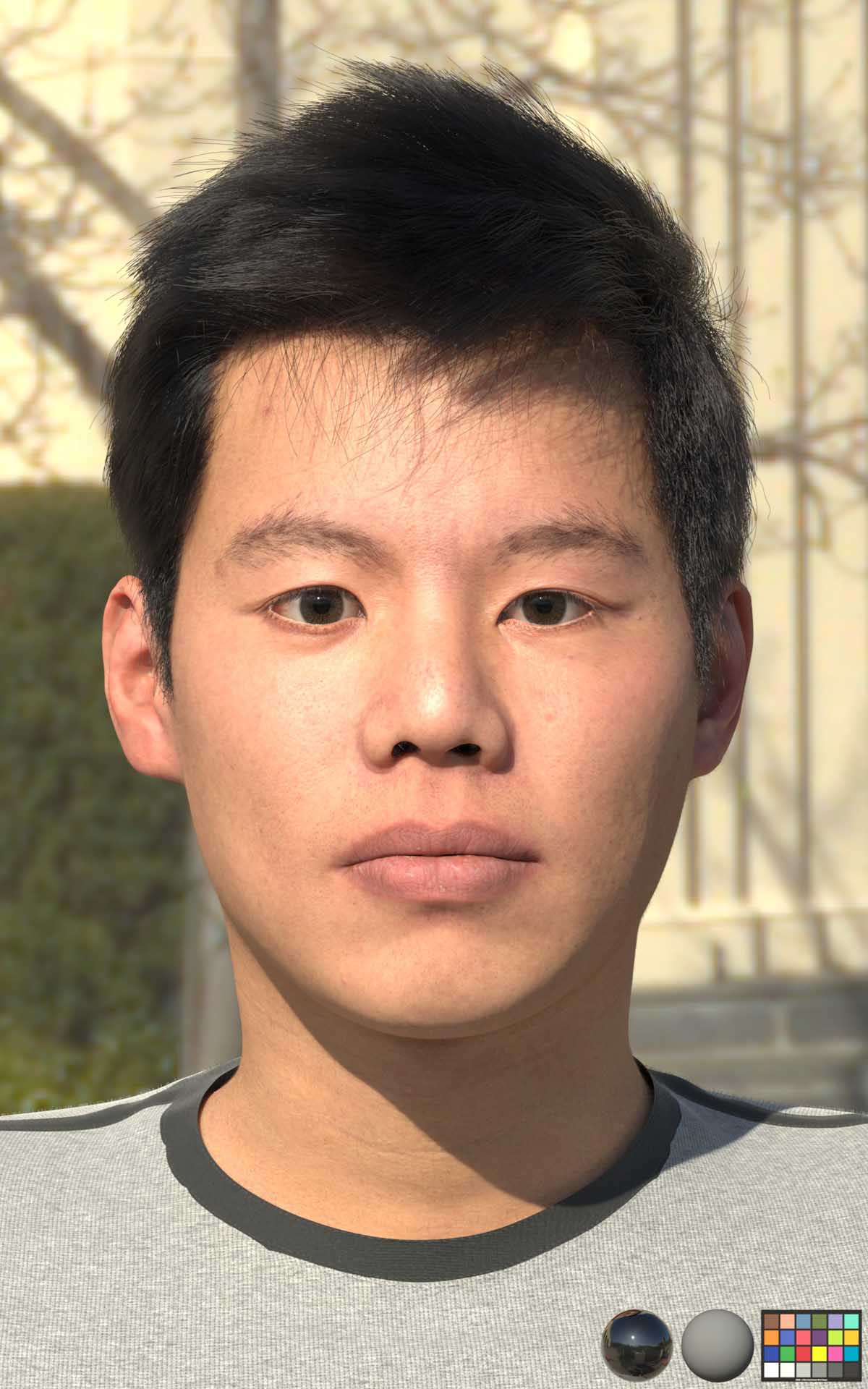}
 &
   \includegraphics[trim= 0 340 0 0, clip, width =0.14\textwidth]{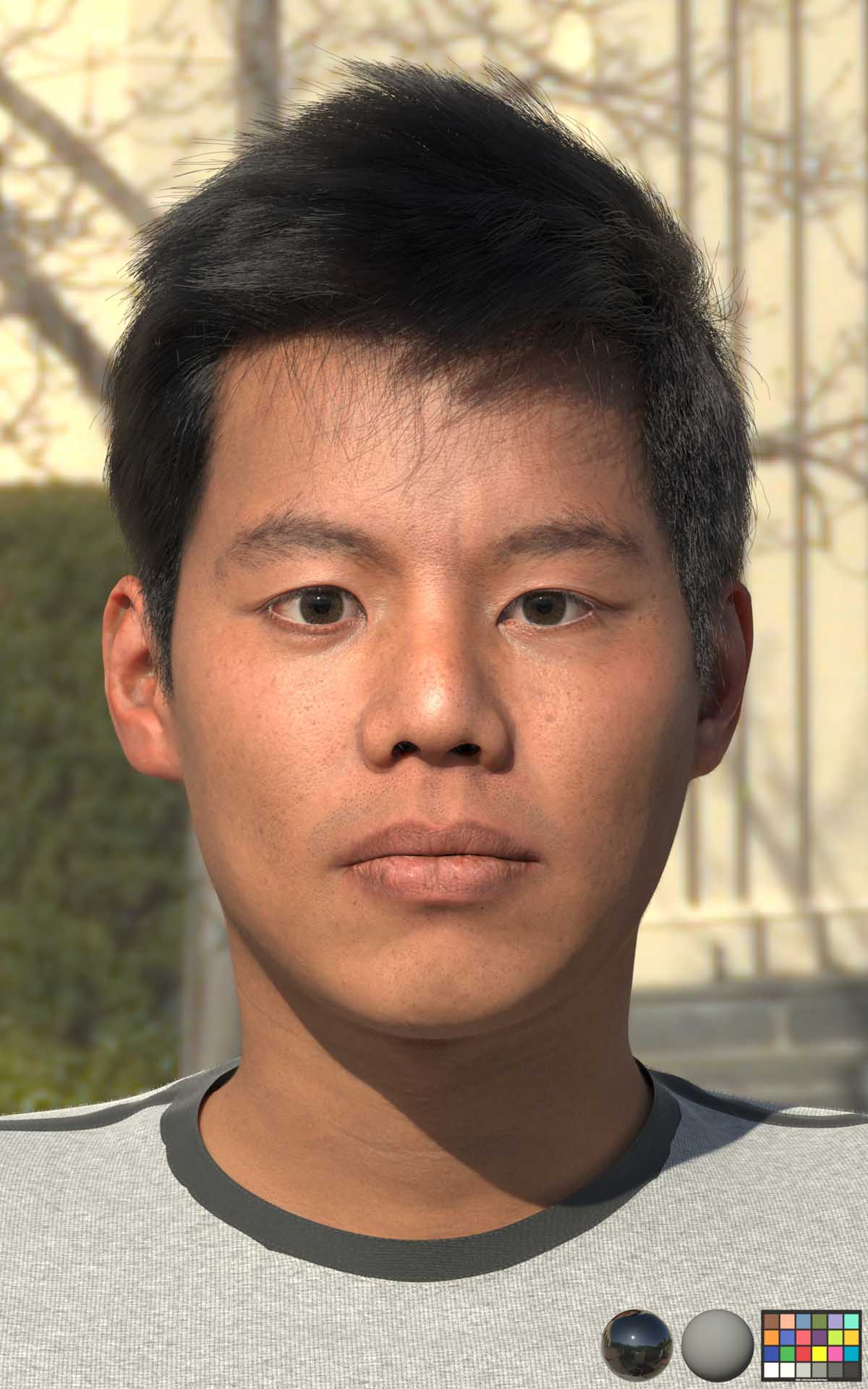}
 &
   \includegraphics[trim= 0 340 0 0, clip, width =0.14\textwidth]{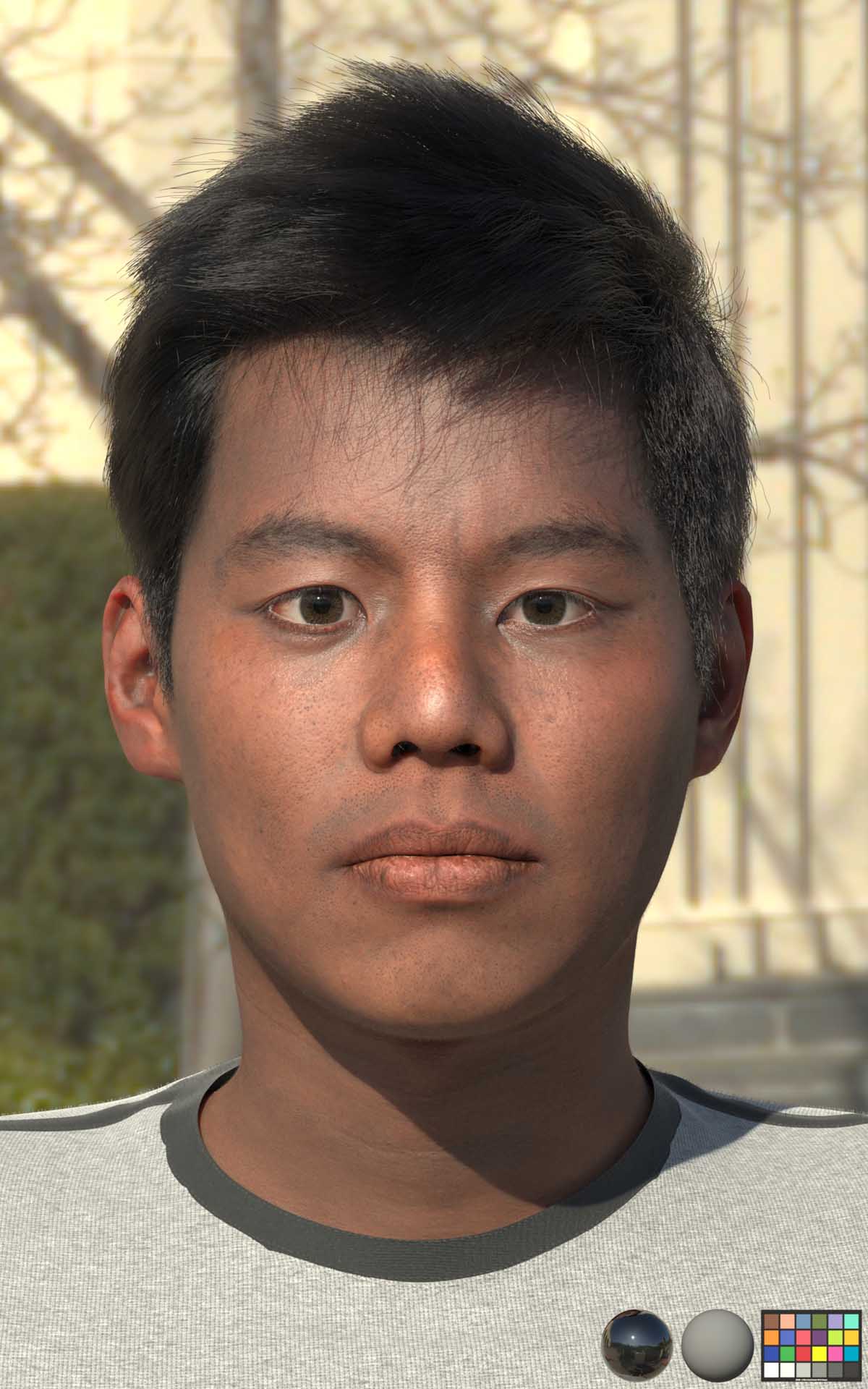}

  \\
 & & \textsc{0.125x blood } & \textsc{0.5x blood} & \textsc{3x blood} & \textsc{0.5x melanin} & \textsc{1.5x melanin} & \textsc{3x melanin} 
  \\
  \begin{sideways}\hspace{0.2cm}\textsc{Type IV (Subject E)}\end{sideways}
 &
   \includegraphics[trim= 0 340 0 0, clip, width =0.14\textwidth]{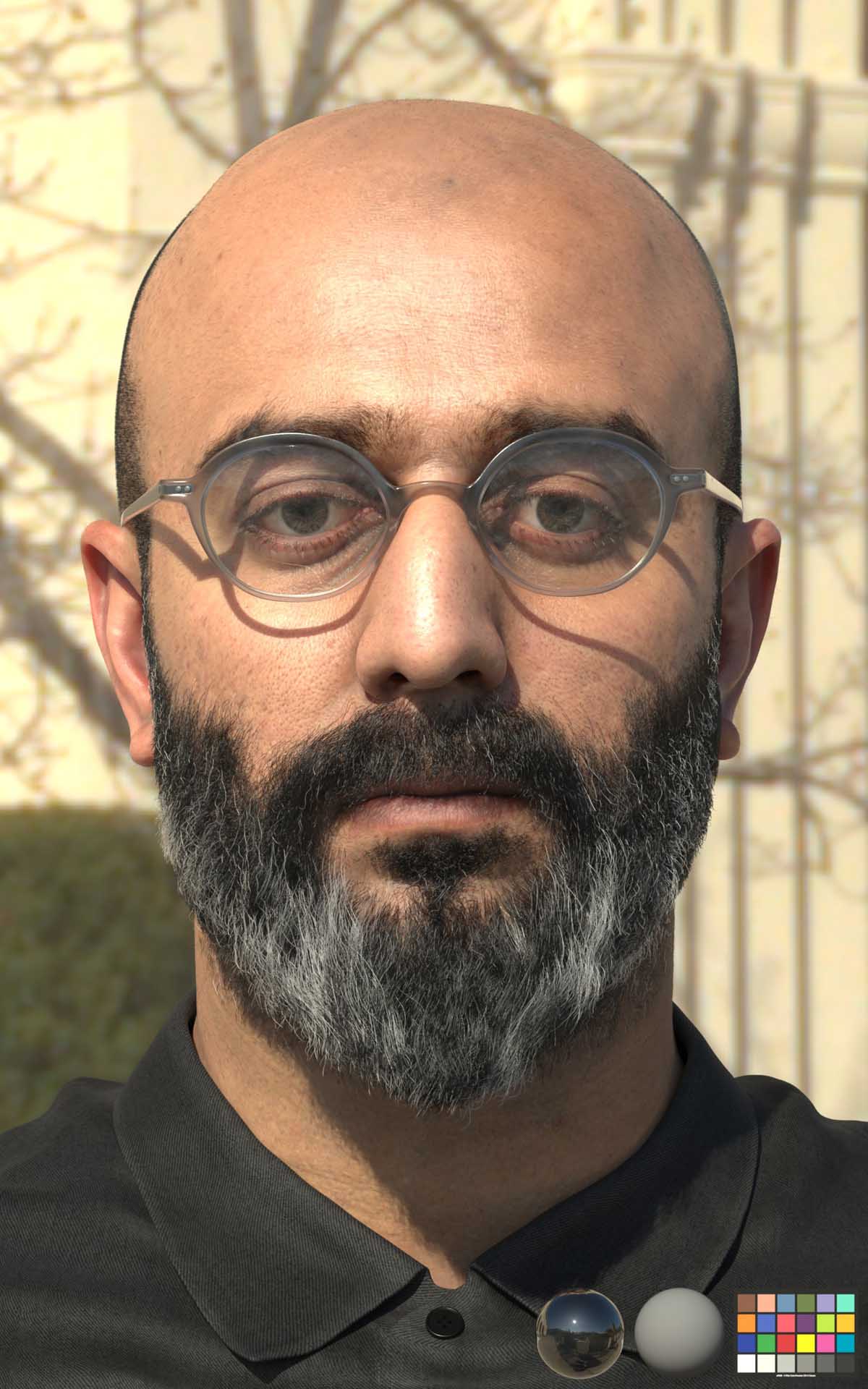}
 &
   \includegraphics[trim= 0 340 0 0, clip, width =0.14\textwidth]{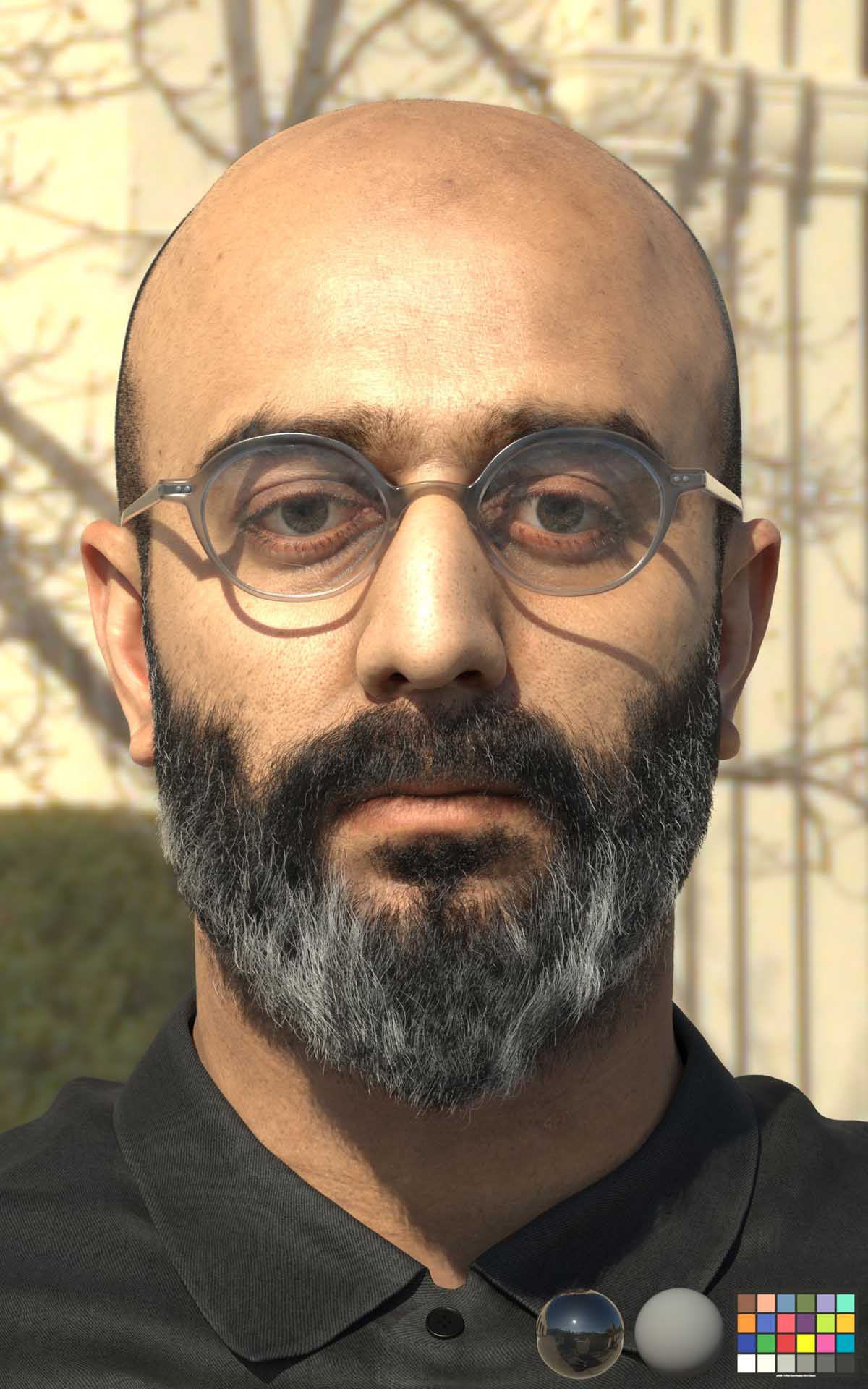}
 &
   \includegraphics[trim= 0 340 0 0, clip, width =0.14\textwidth]{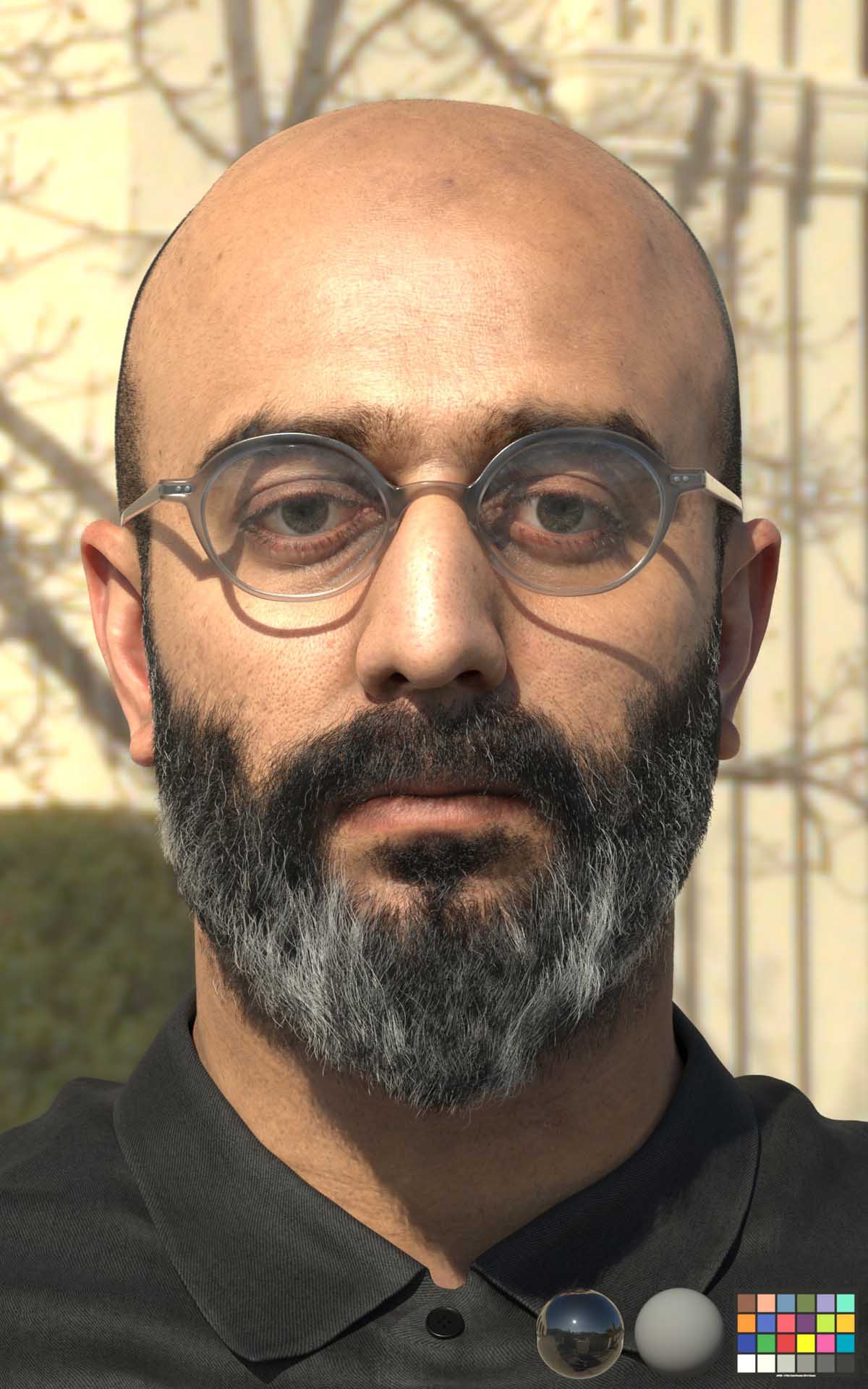}
 &
   \includegraphics[trim= 0 340 0 0, clip, width =0.14\textwidth]{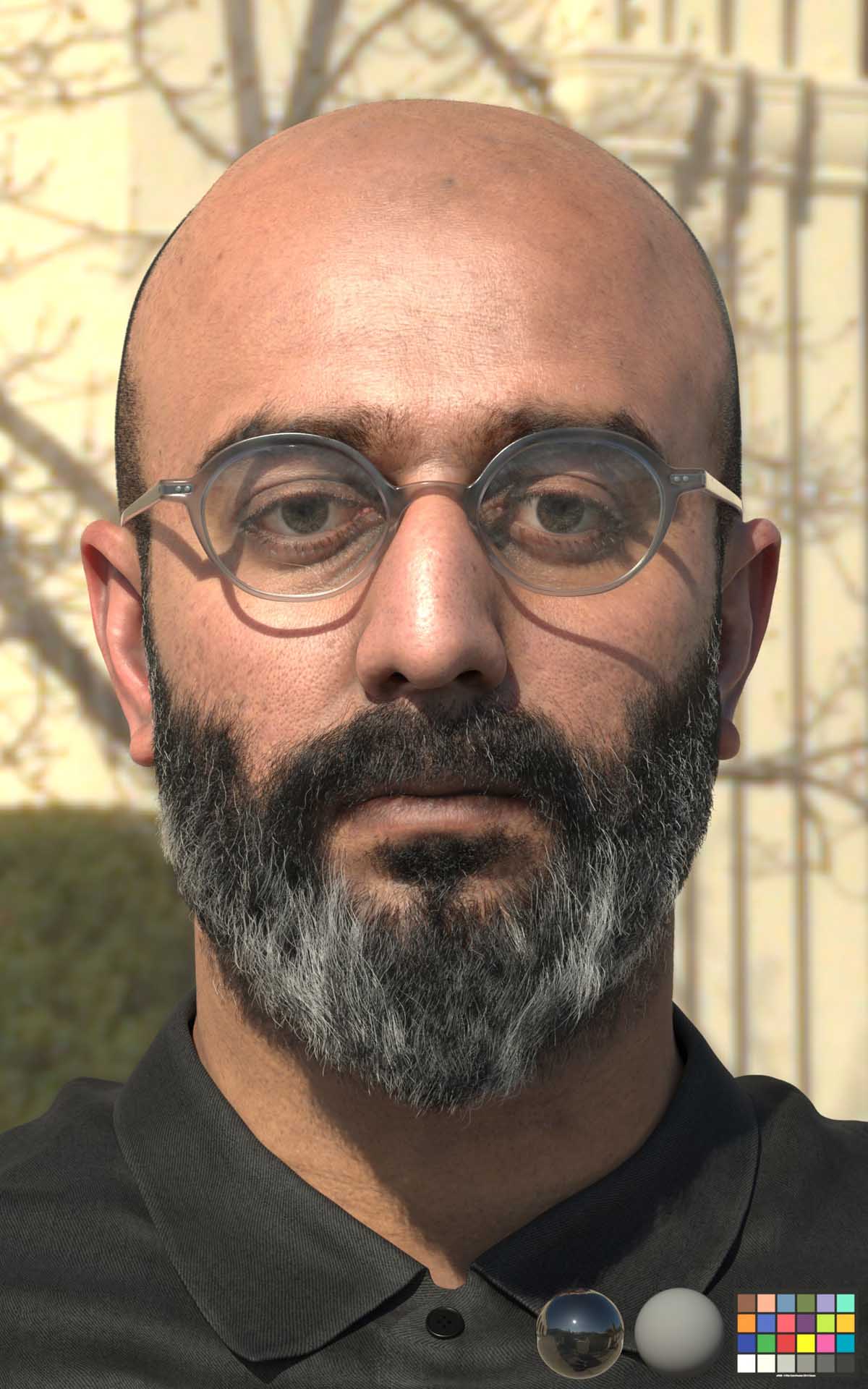}
 &
   \includegraphics[trim= 0 340 0 0, clip, width =0.14\textwidth]{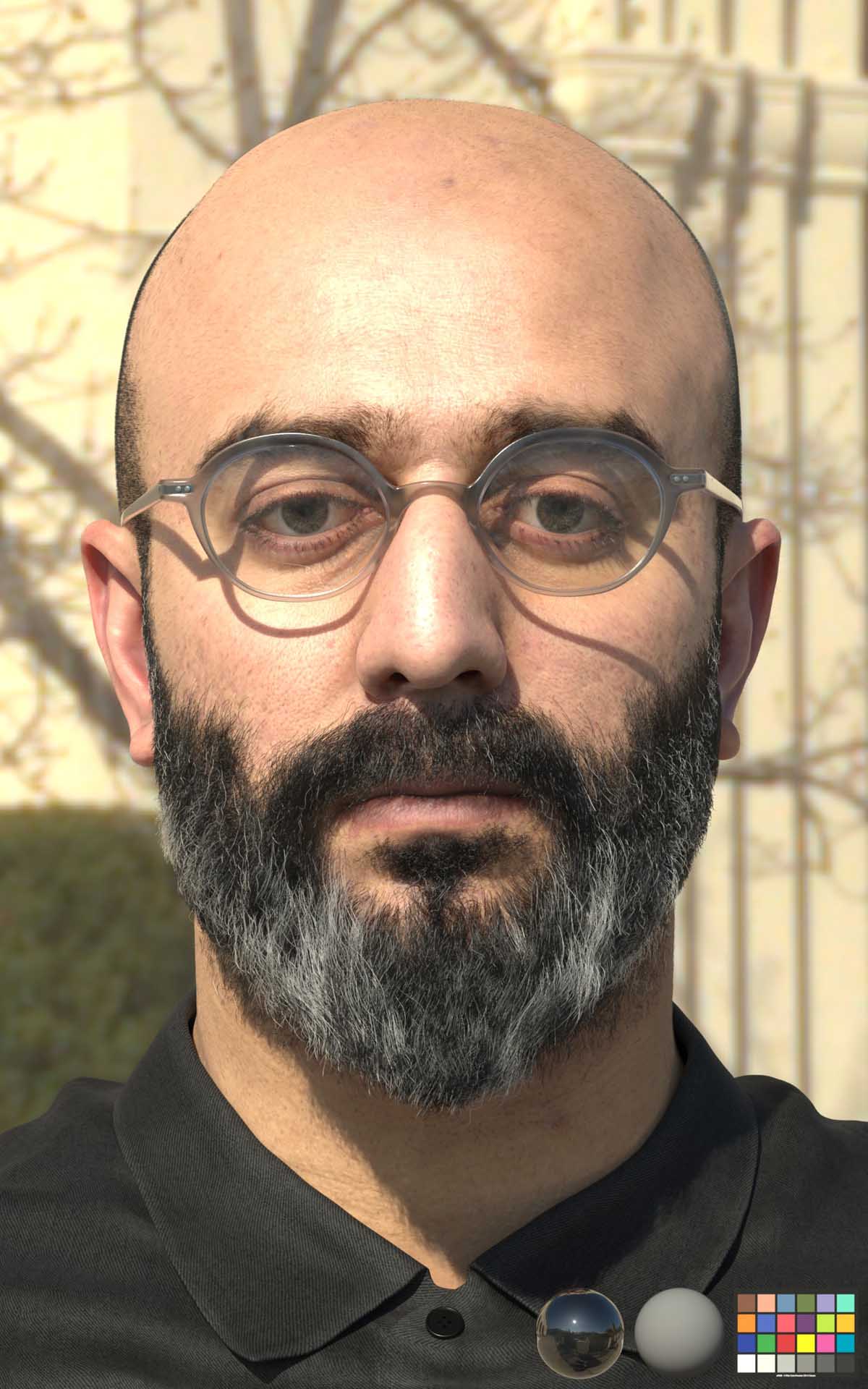}
 &
   \includegraphics[trim= 0 340 0 0, clip, width =0.14\textwidth]{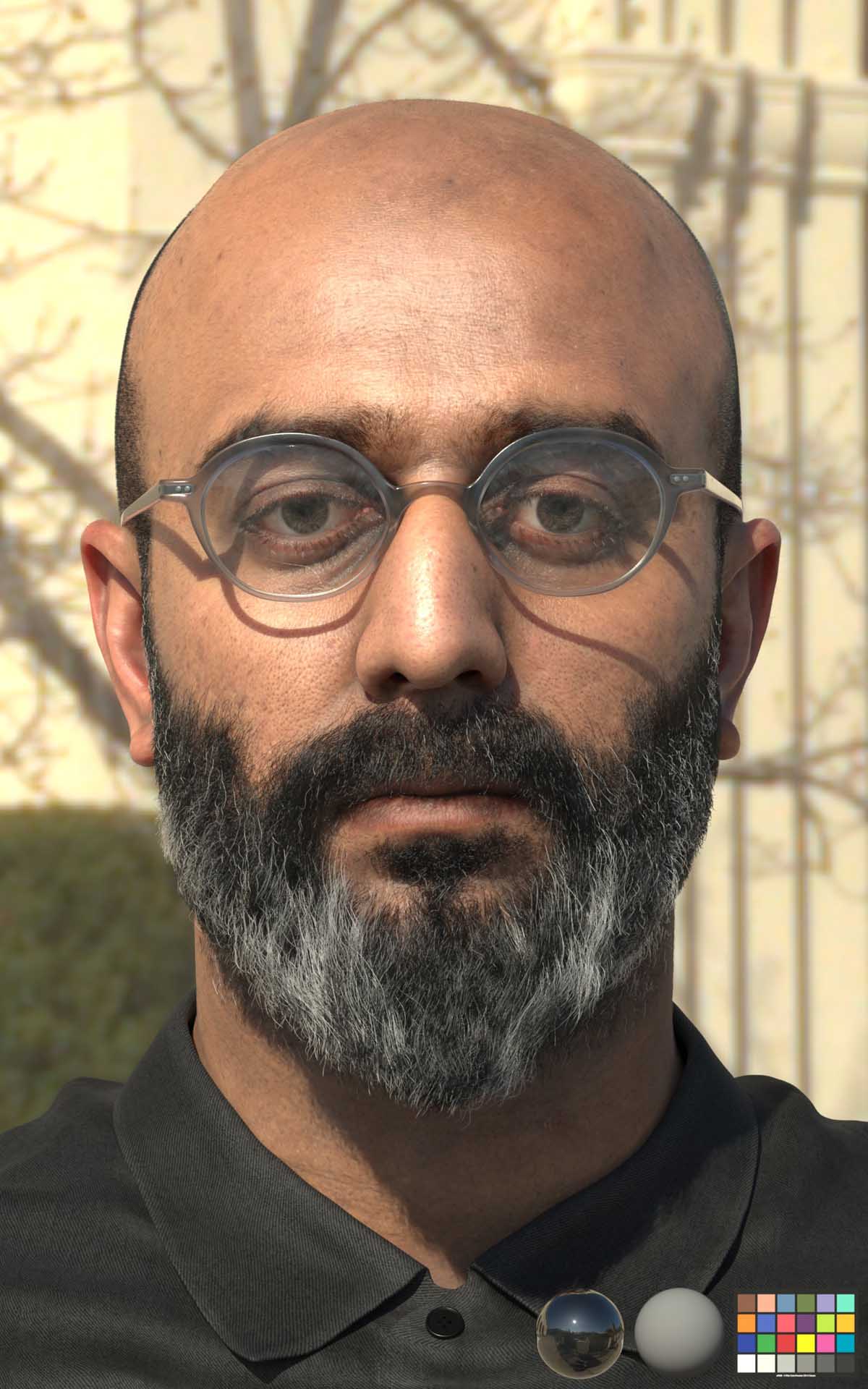}
 &
   \includegraphics[trim= 0 340 0 0, clip, width =0.14\textwidth]{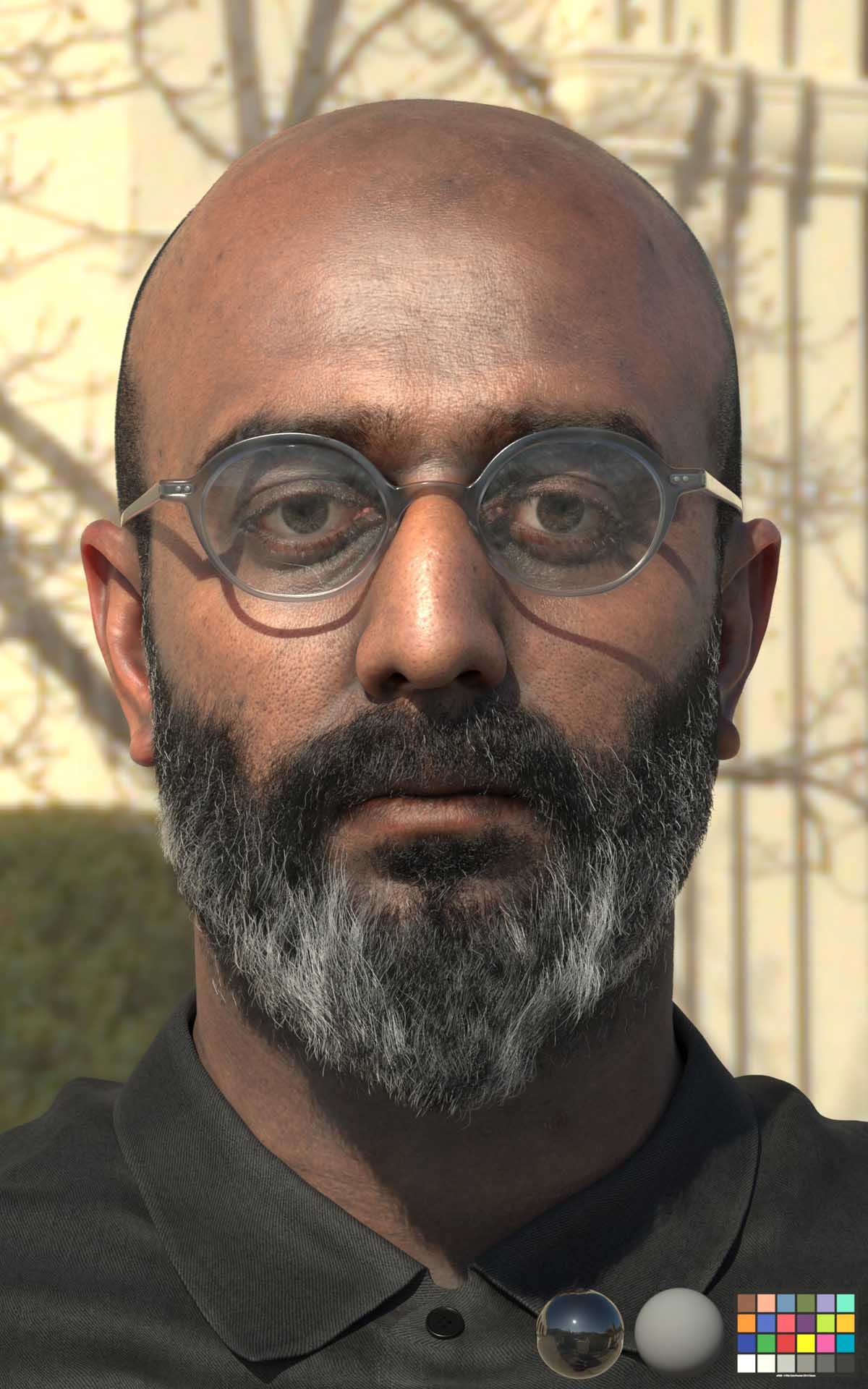}

  \\
 & & \textsc{0.125x blood } & \textsc{0.5x blood} & \textsc{3x blood} & \textsc{0.5x melanin} & \textsc{1.5x melanin} & \textsc{3x melanin} 
  \\
  \begin{sideways}\hspace{0.2cm}\textsc{Type V (Subject F)}\end{sideways}
 &
   \includegraphics[trim= 0 340 0 0, clip, width =0.14\textwidth]{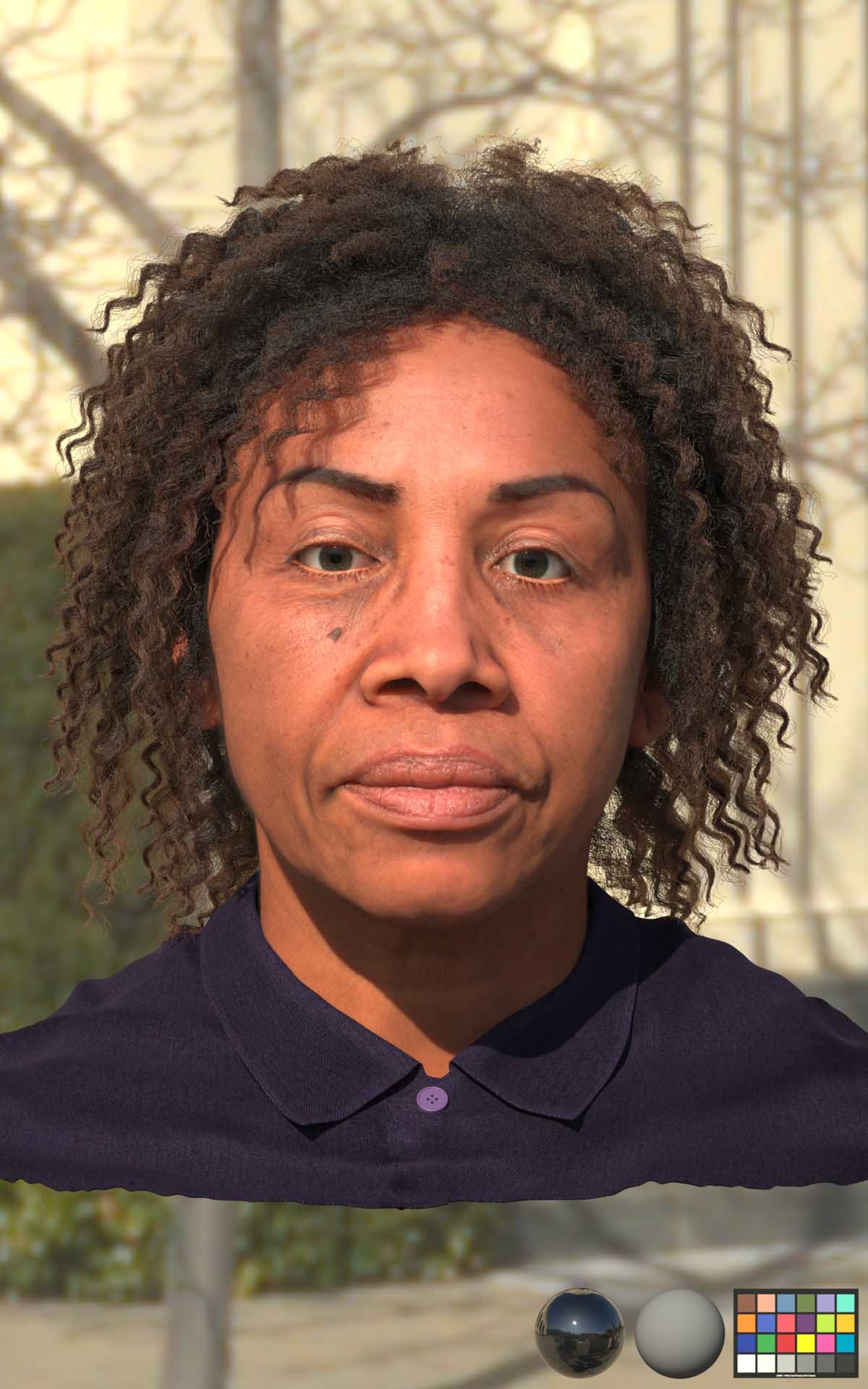}
 &
   \includegraphics[trim= 0 340 0 0, clip, width =0.14\textwidth]{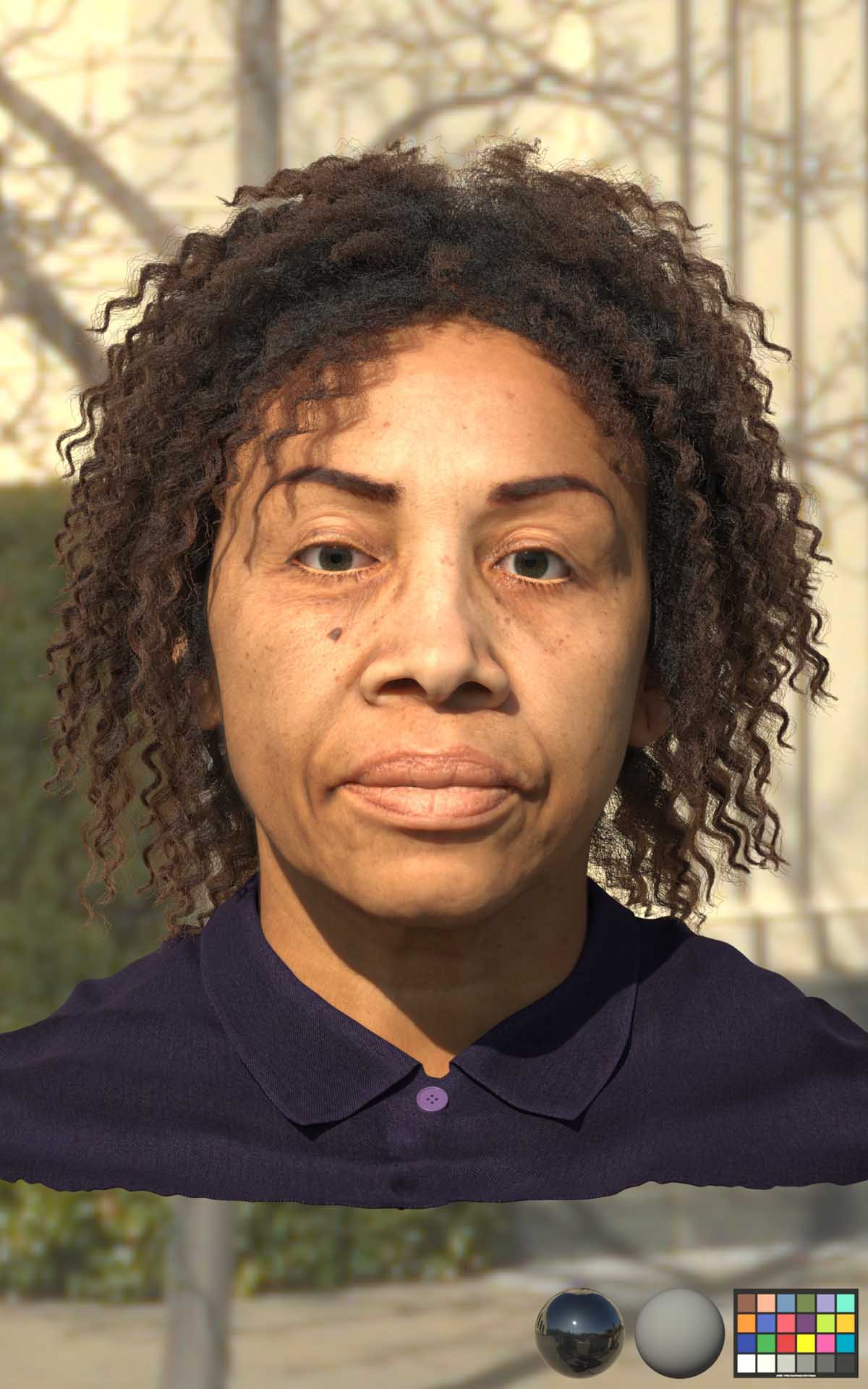}
 &
   \includegraphics[trim= 0 340 0 0, clip, width =0.14\textwidth]{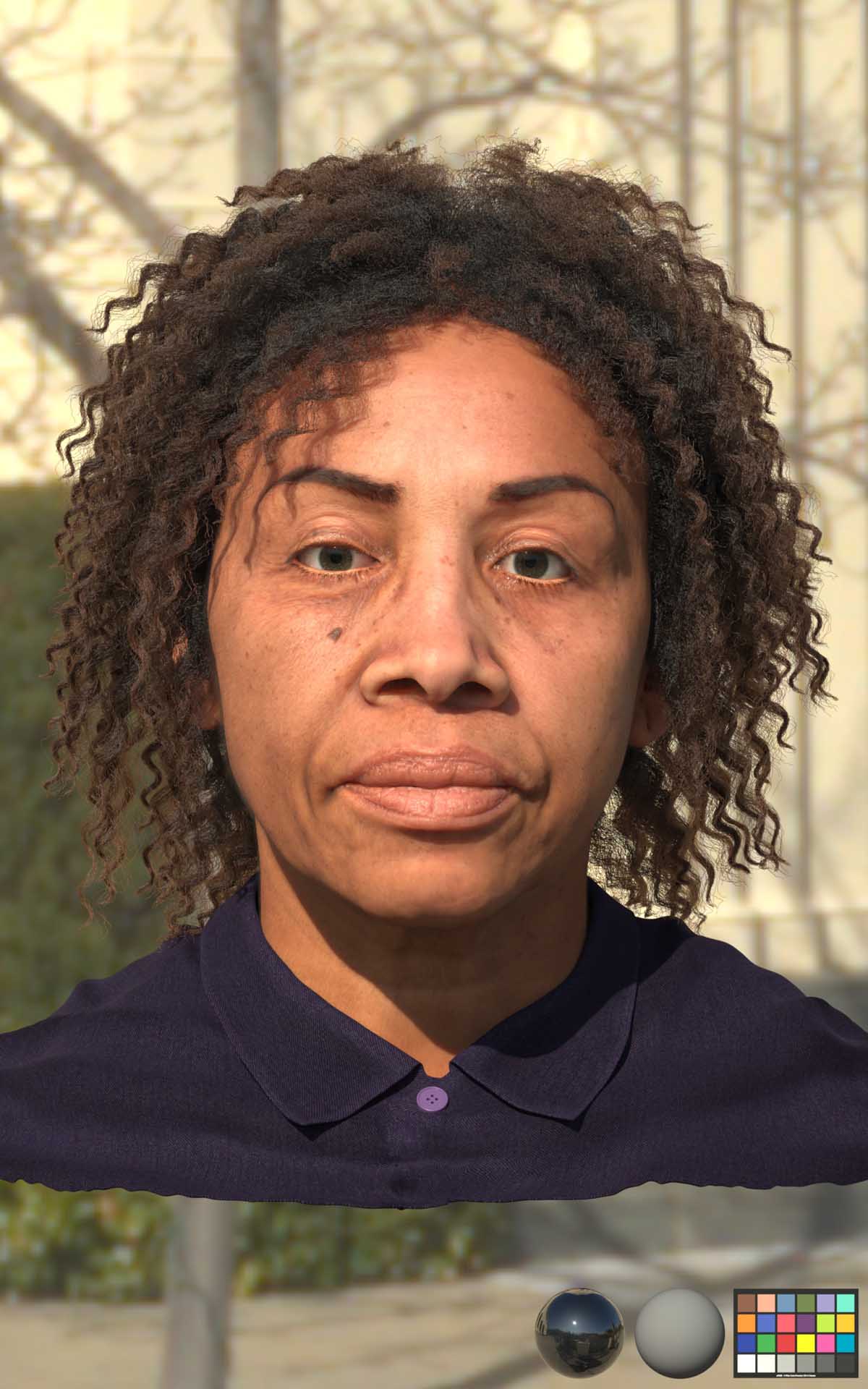}
 &
   \includegraphics[trim= 0 340 0 0, clip, width =0.14\textwidth]{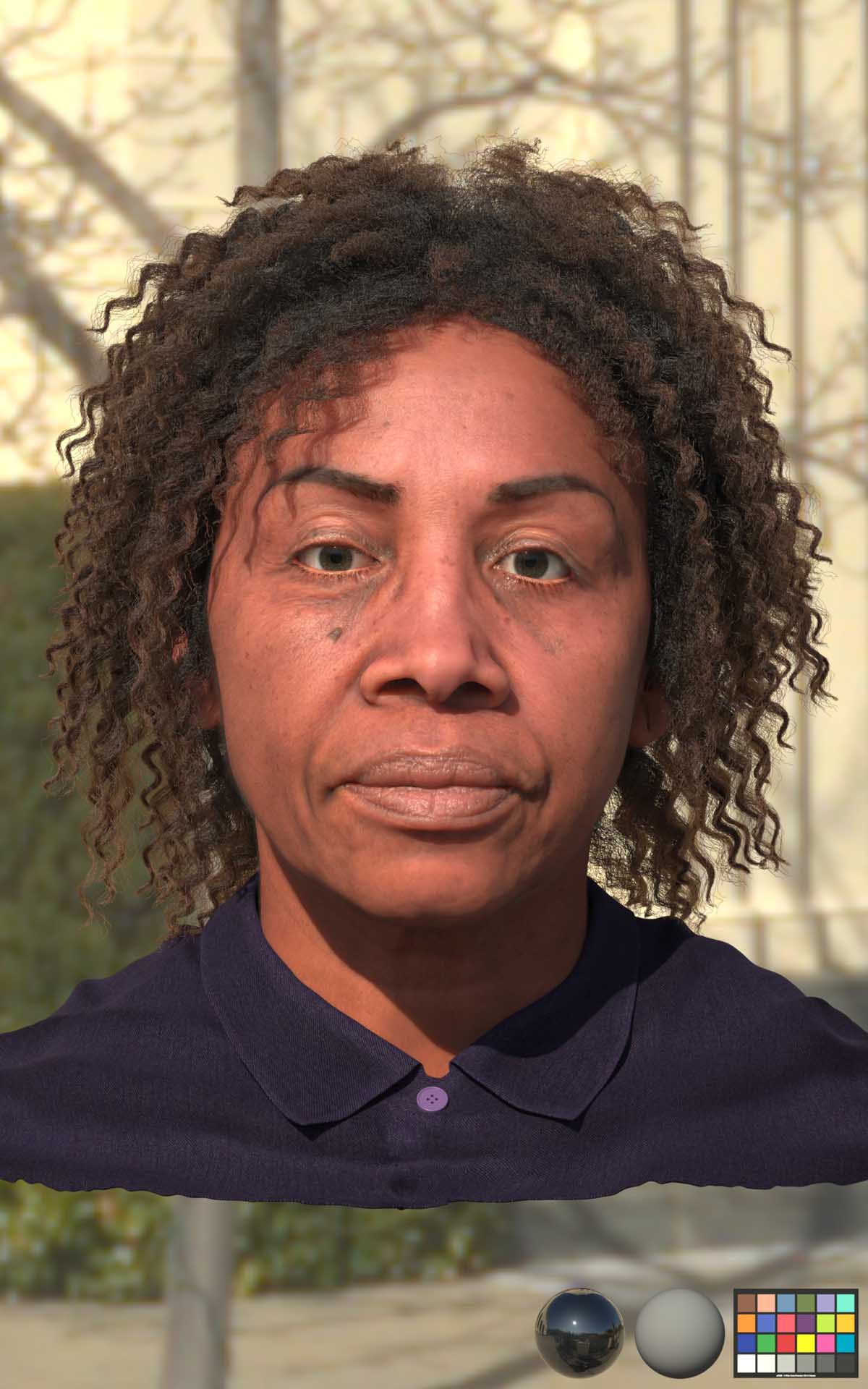}
 &
   \includegraphics[trim= 0 340 0 0, clip, width =0.14\textwidth]{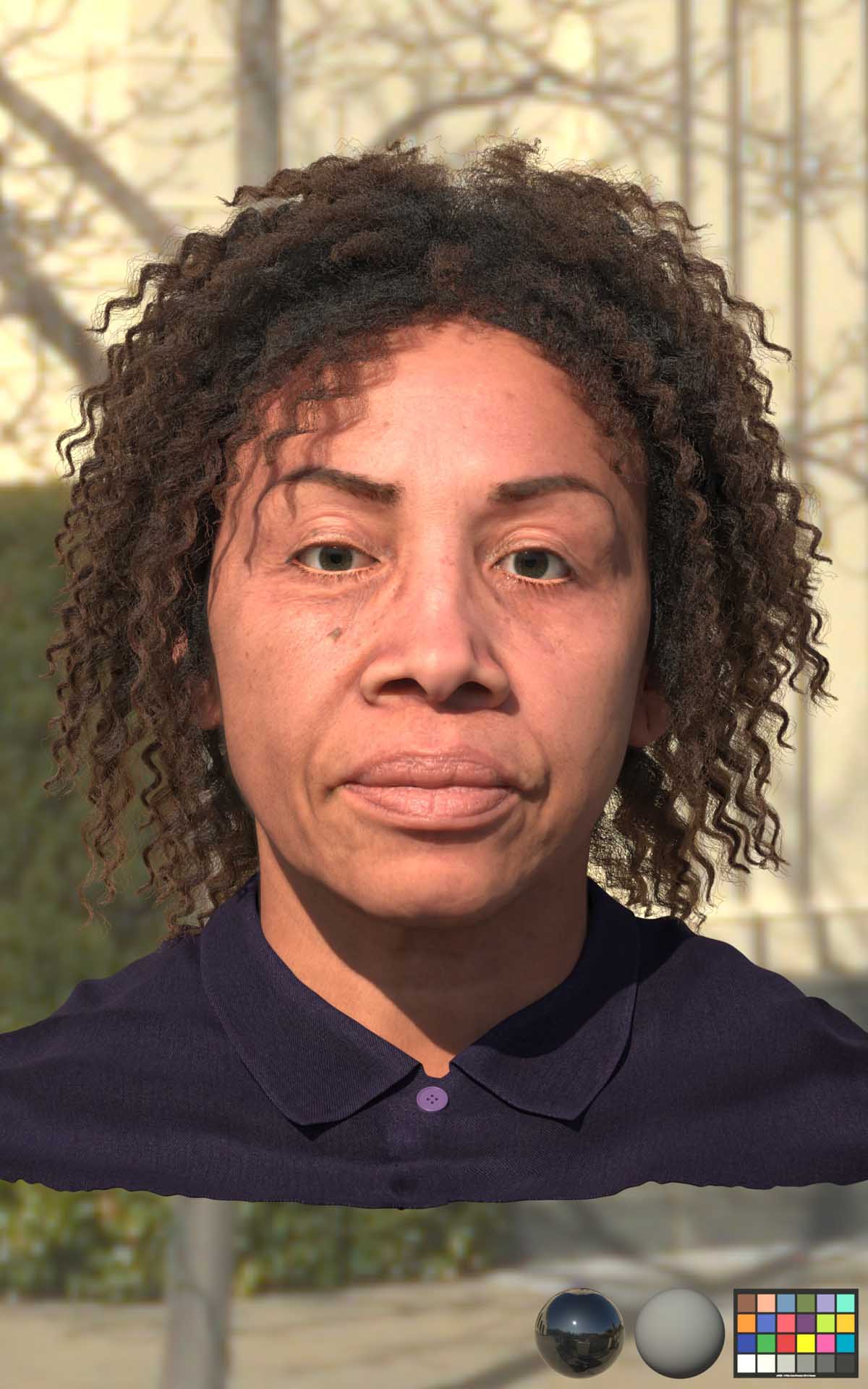}
 &
   \includegraphics[trim= 0 340 0 0, clip, width =0.14\textwidth]{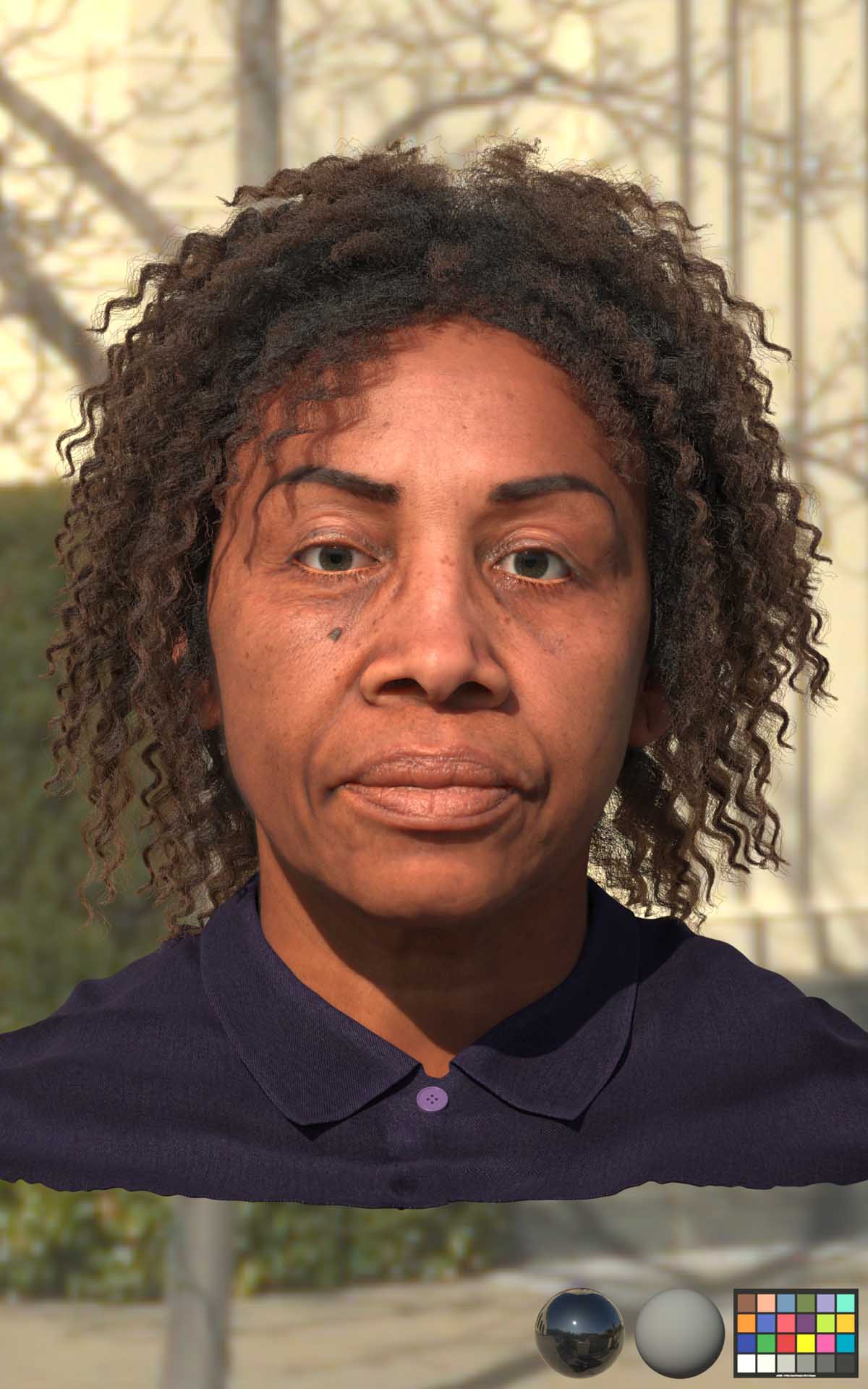}
 &
   \includegraphics[trim= 0 340 0 0, clip, width =0.14\textwidth]{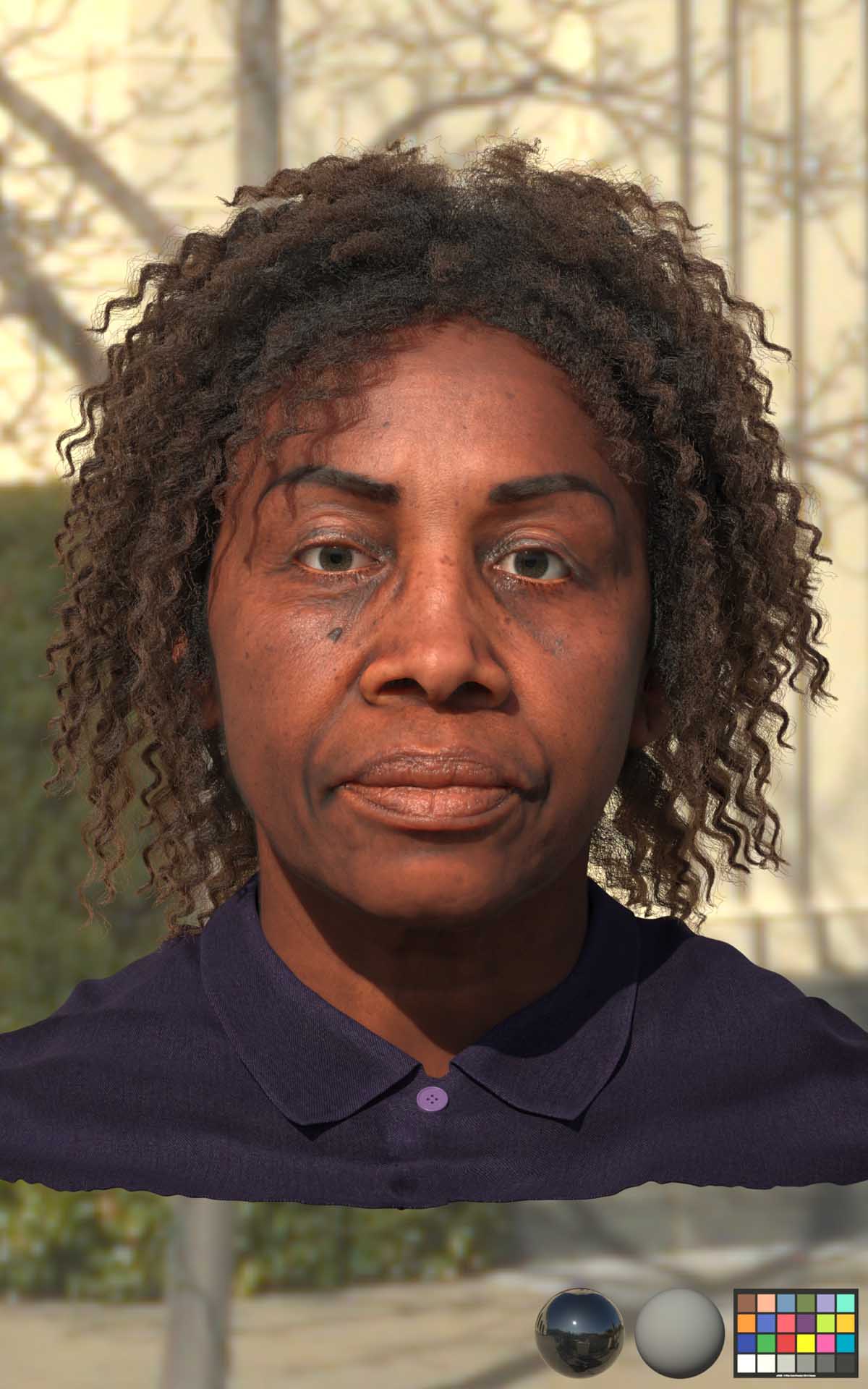}

 \end{tabular}
 \vspace{-3mm}
  \caption{Rendering results under lighting scenario 1 for edited skin parameters inferred by our model over different skin types. From left to right, original, followed by different manipulations per subject, three in blood concentration and three in melanin concentration.}
  \label{fig:editing_1}
  \vspace{-6mm}
\end{figure*}

\begin{figure*}[t!]
 \contourlength{0.1em}%
 \centering
 \hspace*{-4.5mm}%
  \begin{tabular}{l@{\;}c@{\;}c@{\;}c@{\;}c@{\;}c@{\;}c@{\;}c@{\;}c}
  
 & & \textsc{0.125x blood } & \textsc{0.5x blood} & \textsc{3x blood} & \textsc{1.5x melanin} & \textsc{3x melanin} & \textsc{5x melanin} 
  \\
  \begin{sideways}\hspace{0.2cm}\textsc{Type I (Subject A)}\end{sideways}
 &
   \includegraphics[trim= 0 340 0 0, clip, width =0.14\textwidth]{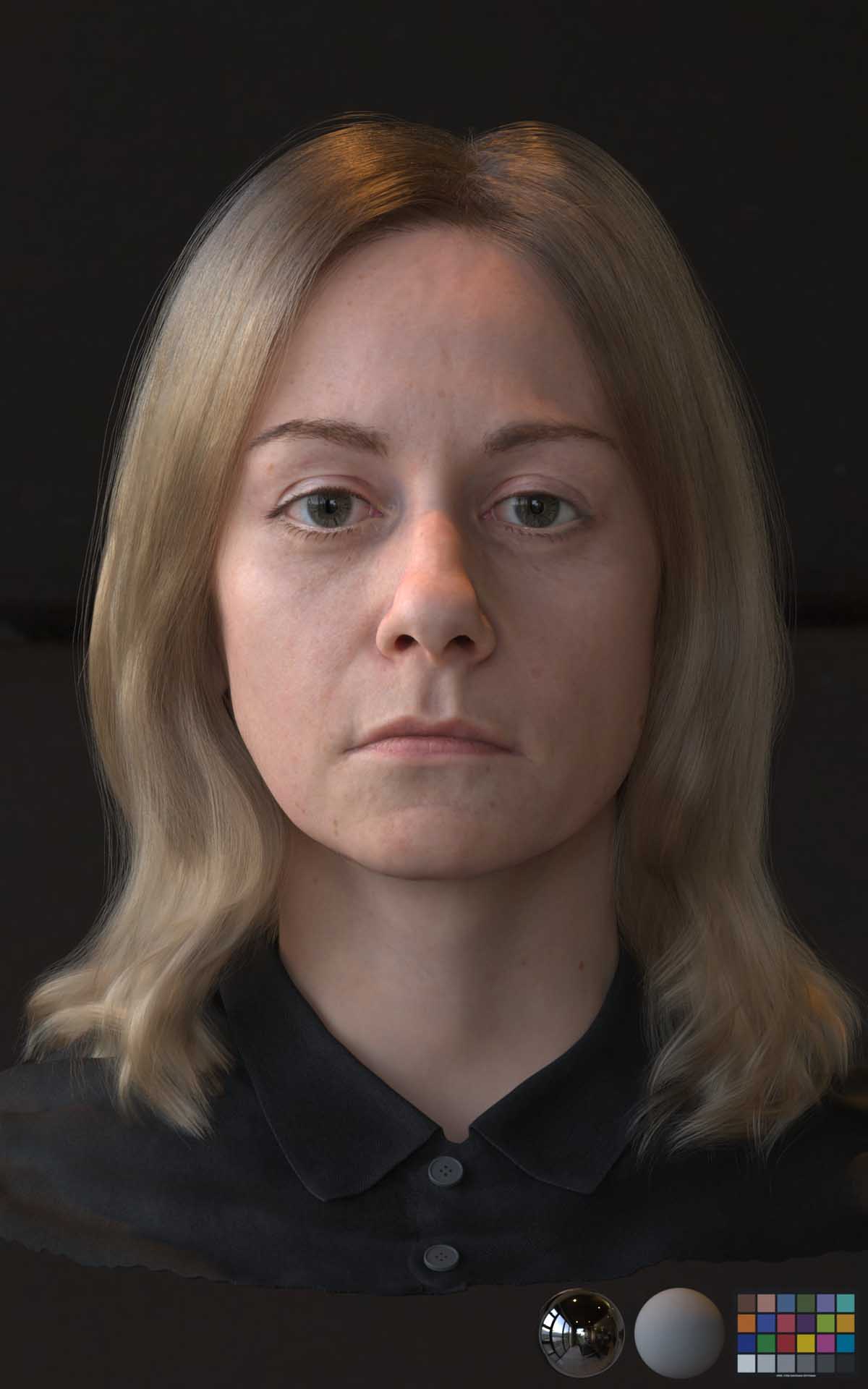}
 &
   \includegraphics[trim= 0 340 0 0, clip, width =0.14\textwidth]{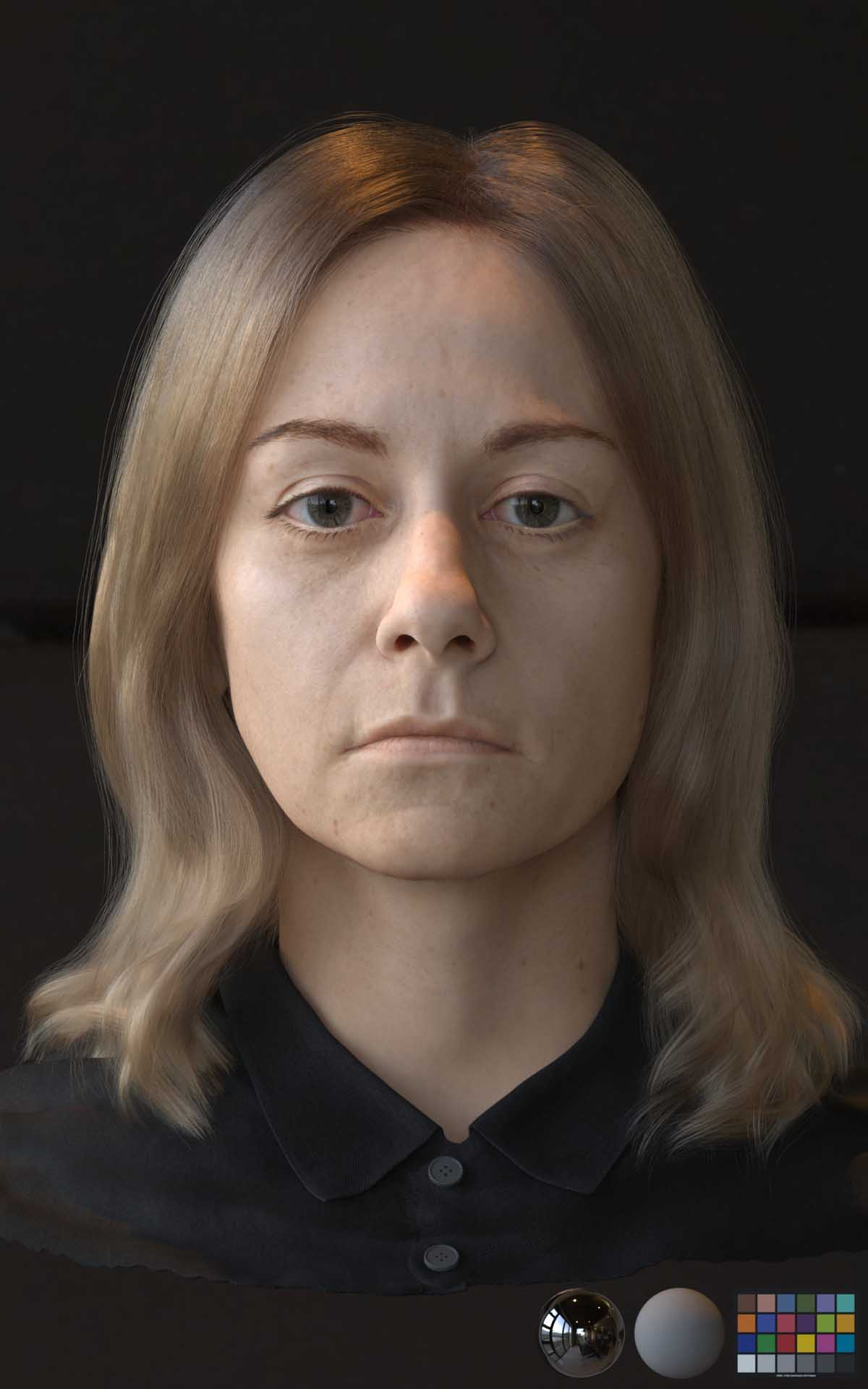}
 &
   \includegraphics[trim= 0 340 0 0, clip, width =0.14\textwidth]{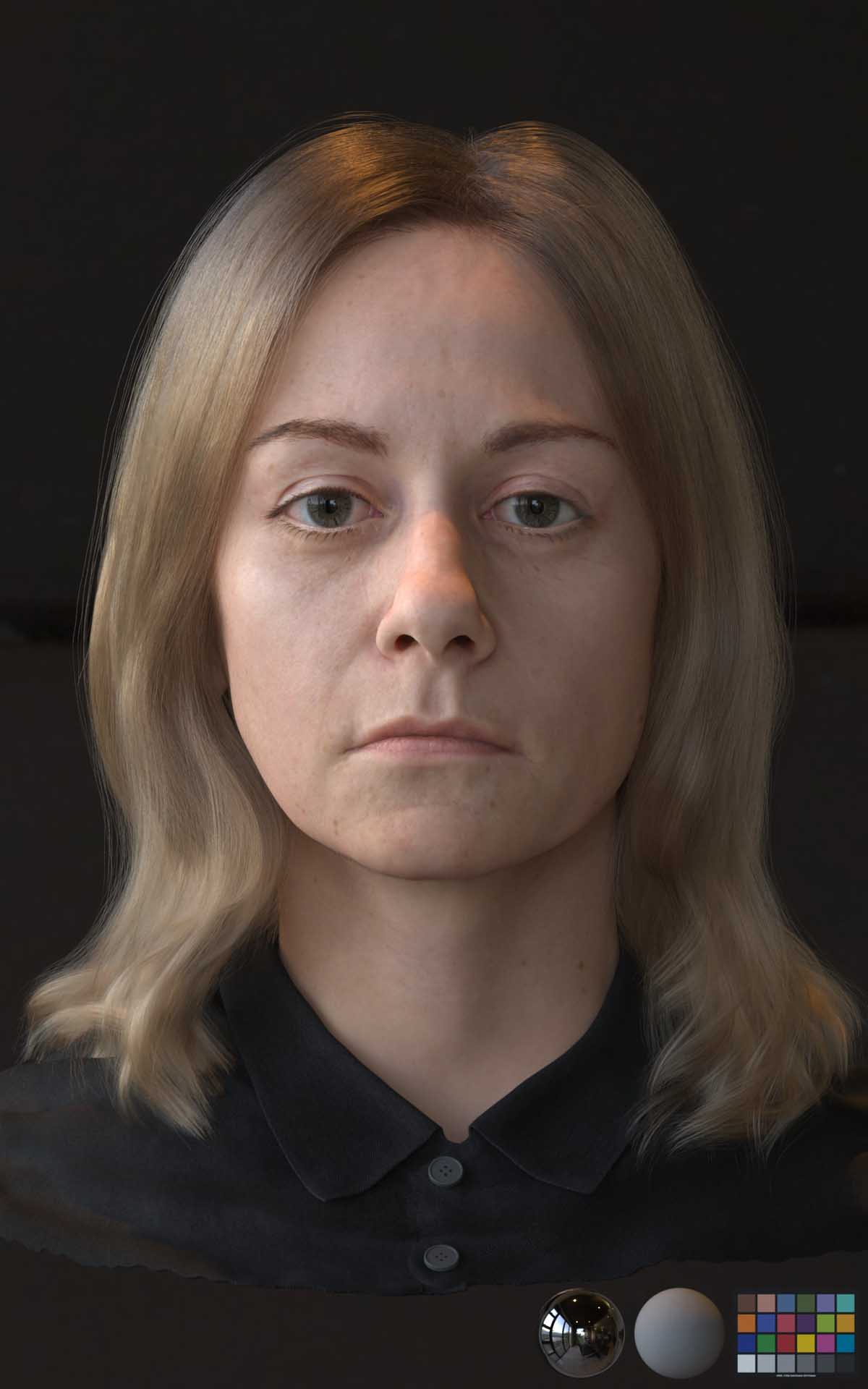}
 &
   \includegraphics[trim= 0 340 0 0, clip, width =0.14\textwidth]{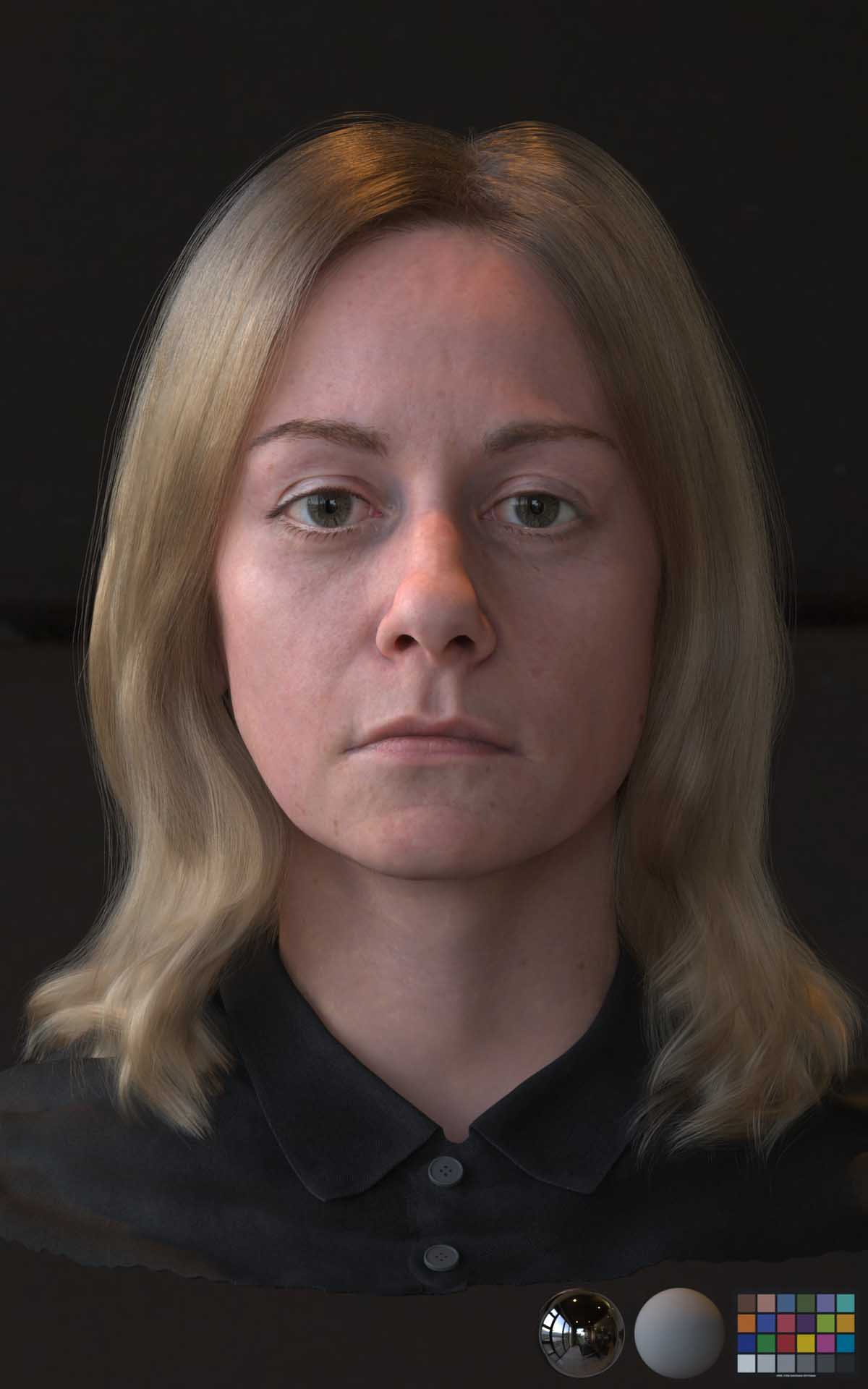}
 &
   \includegraphics[trim= 0 340 0 0, clip, width =0.14\textwidth]{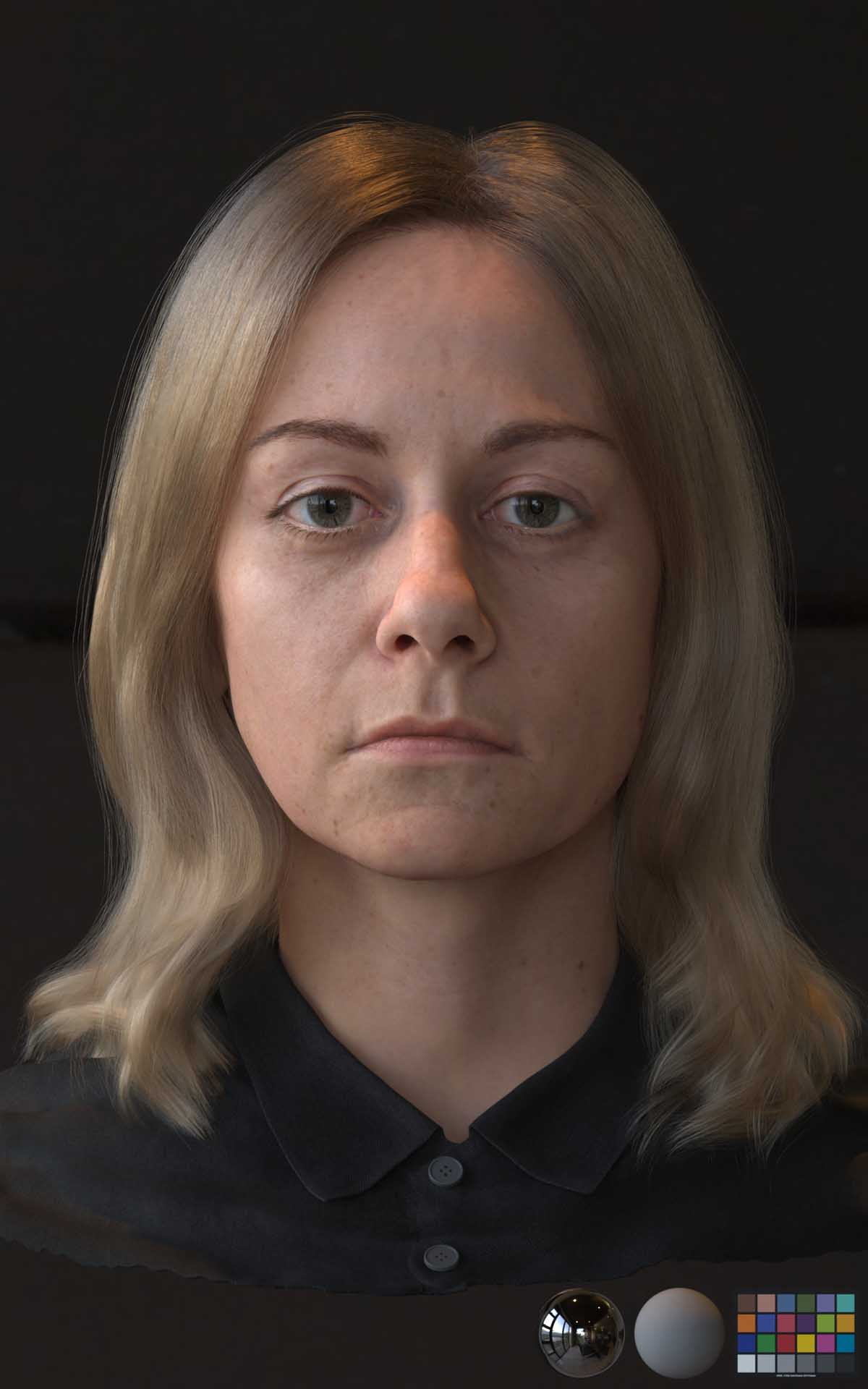}
 &
   \includegraphics[trim= 0 340 0 0, clip, width =0.14\textwidth]{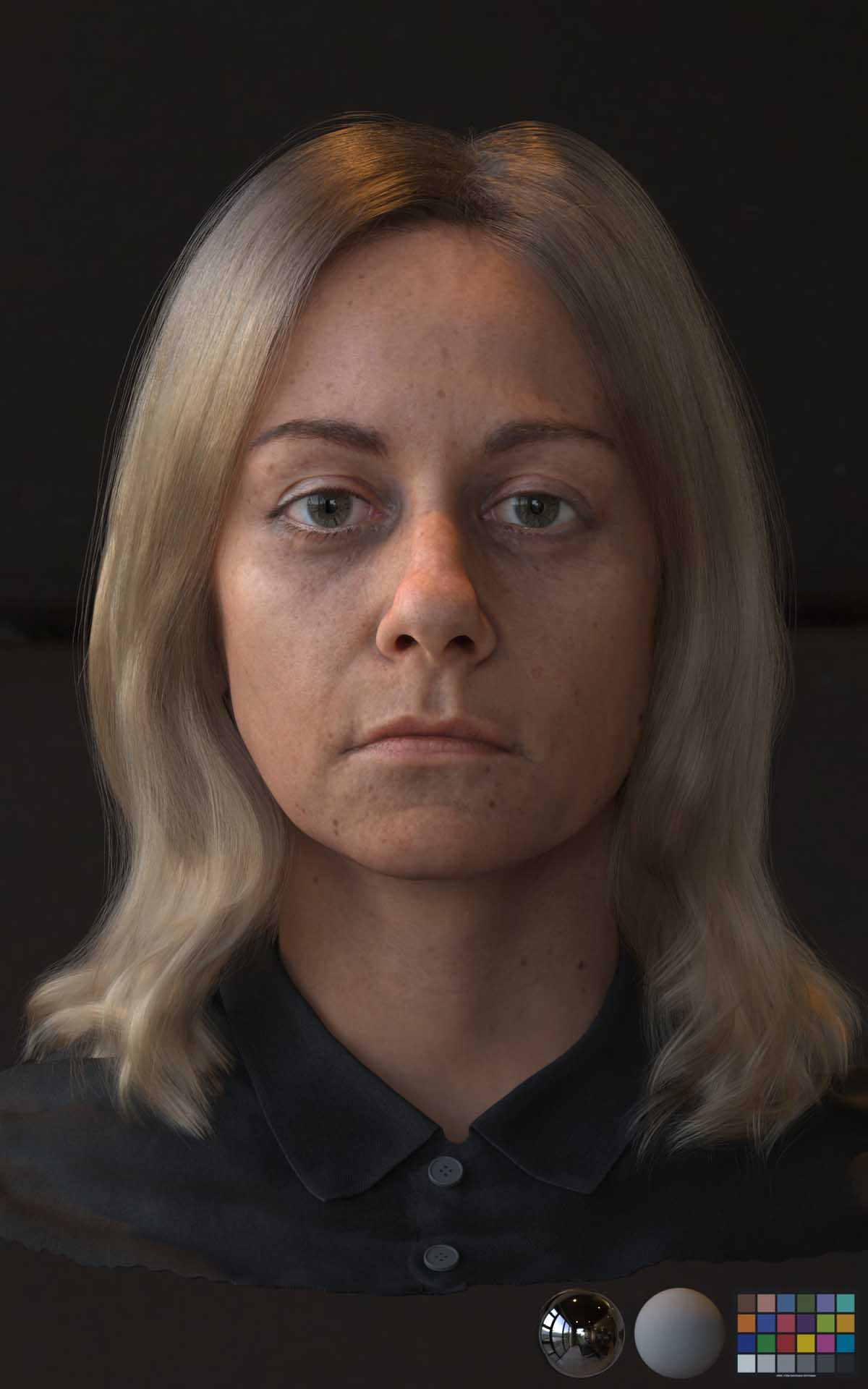}
 &
   \includegraphics[trim= 0 340 0 0, clip, width =0.14\textwidth]{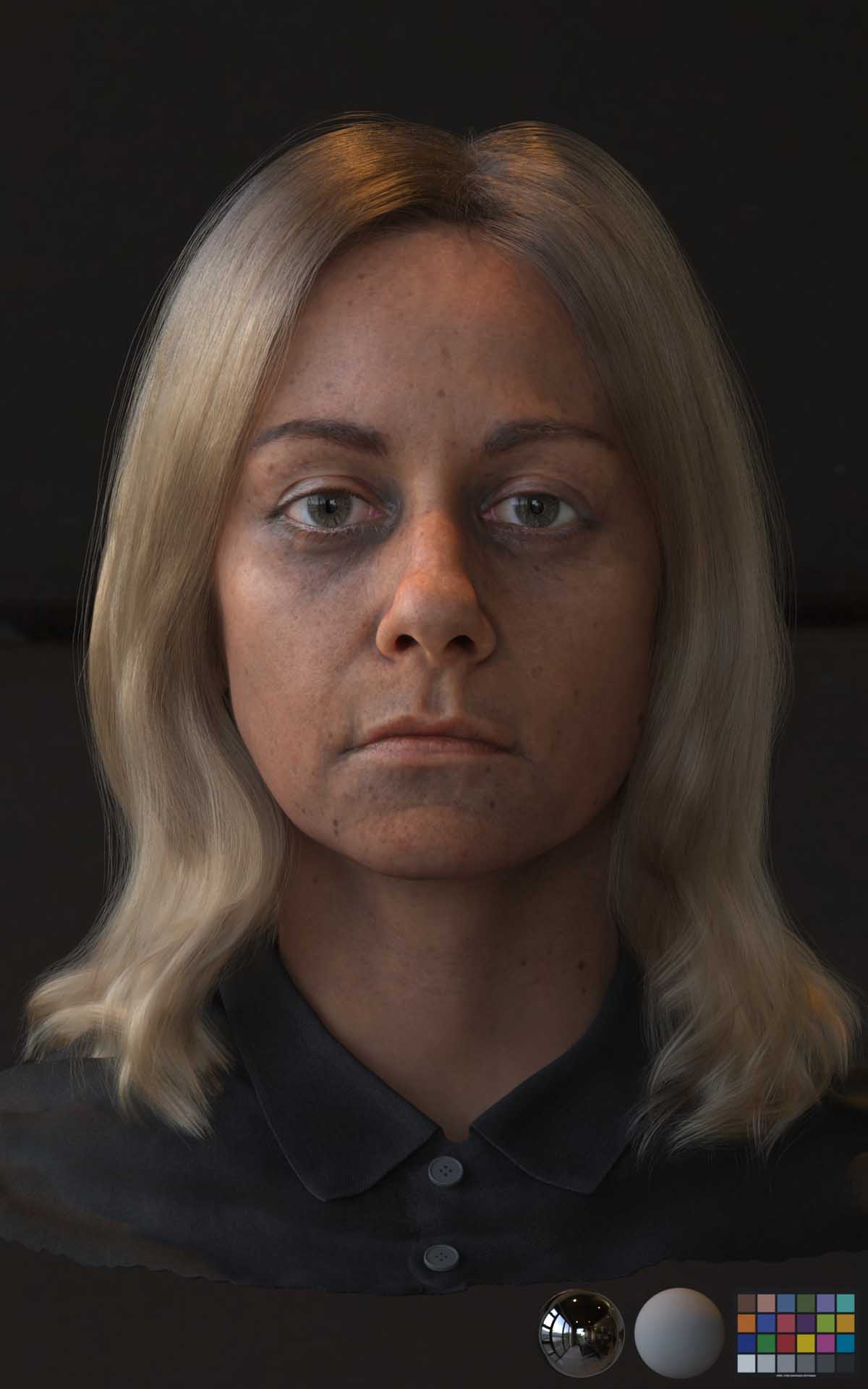}
  
  \\
 & & \textsc{0.125x blood } & \textsc{0.5x blood} & \textsc{3x blood} & \textsc{1.5x melanin} & \textsc{3x melanin} & \textsc{5x melanin} 
  \\
  \begin{sideways}\hspace{0.2cm}\textsc{Type II (Subject C)}\end{sideways}
 &
   \includegraphics[trim= 0 340 0 0, clip, width =0.14\textwidth]{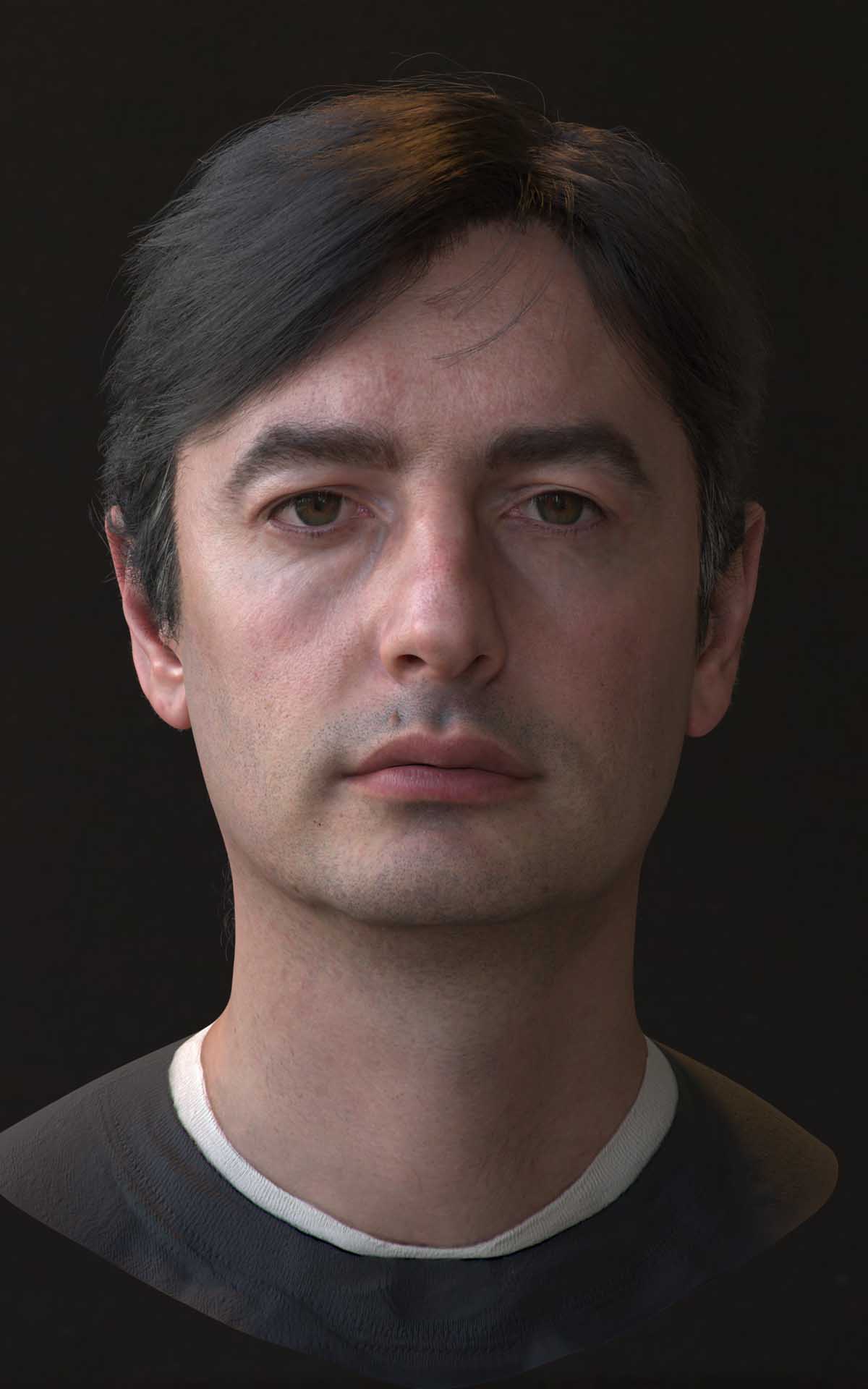}
 &
   \includegraphics[trim= 0 340 0 0, clip, width =0.14\textwidth]{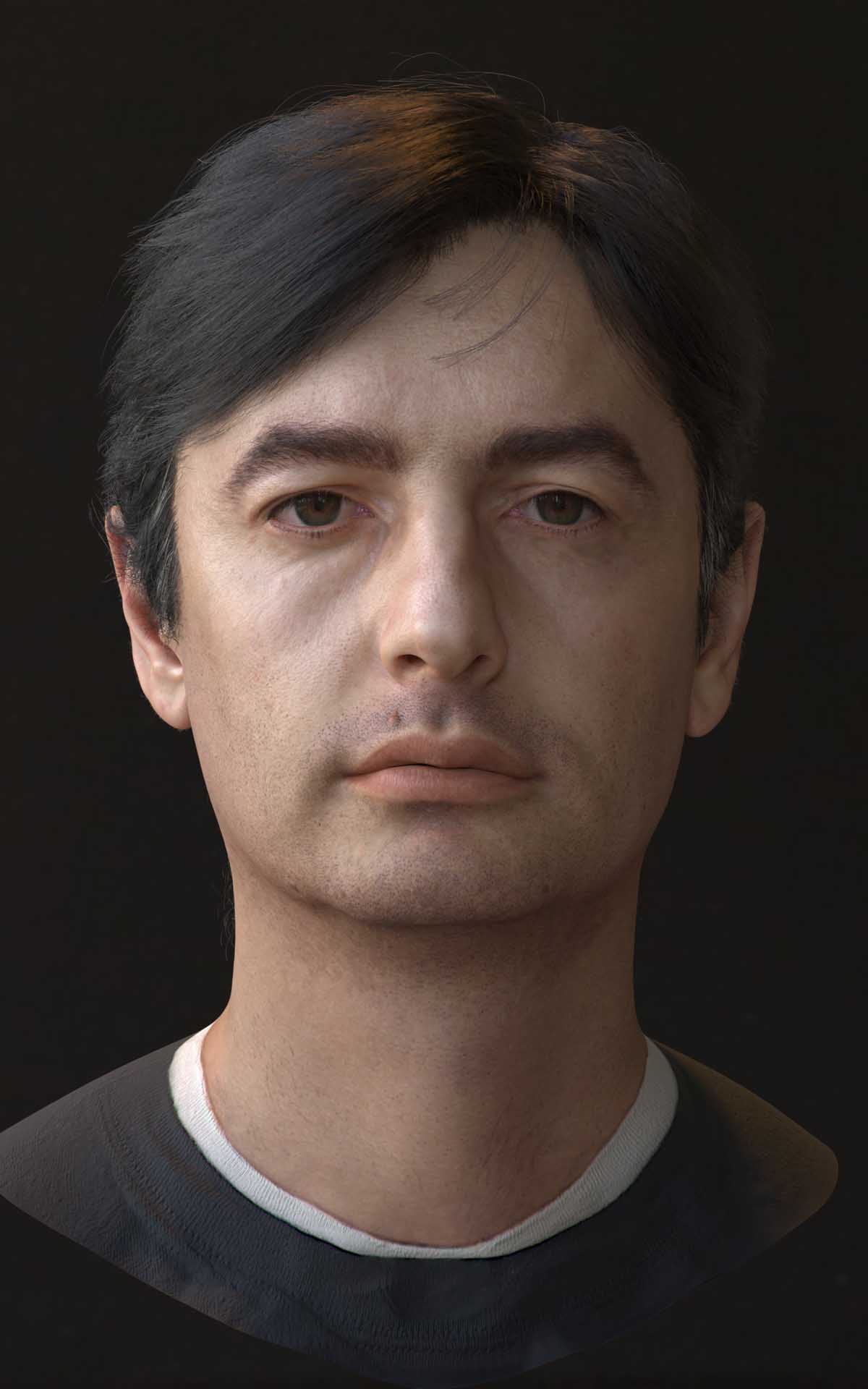}
 &
   \includegraphics[trim= 0 340 0 0, clip, width =0.14\textwidth]{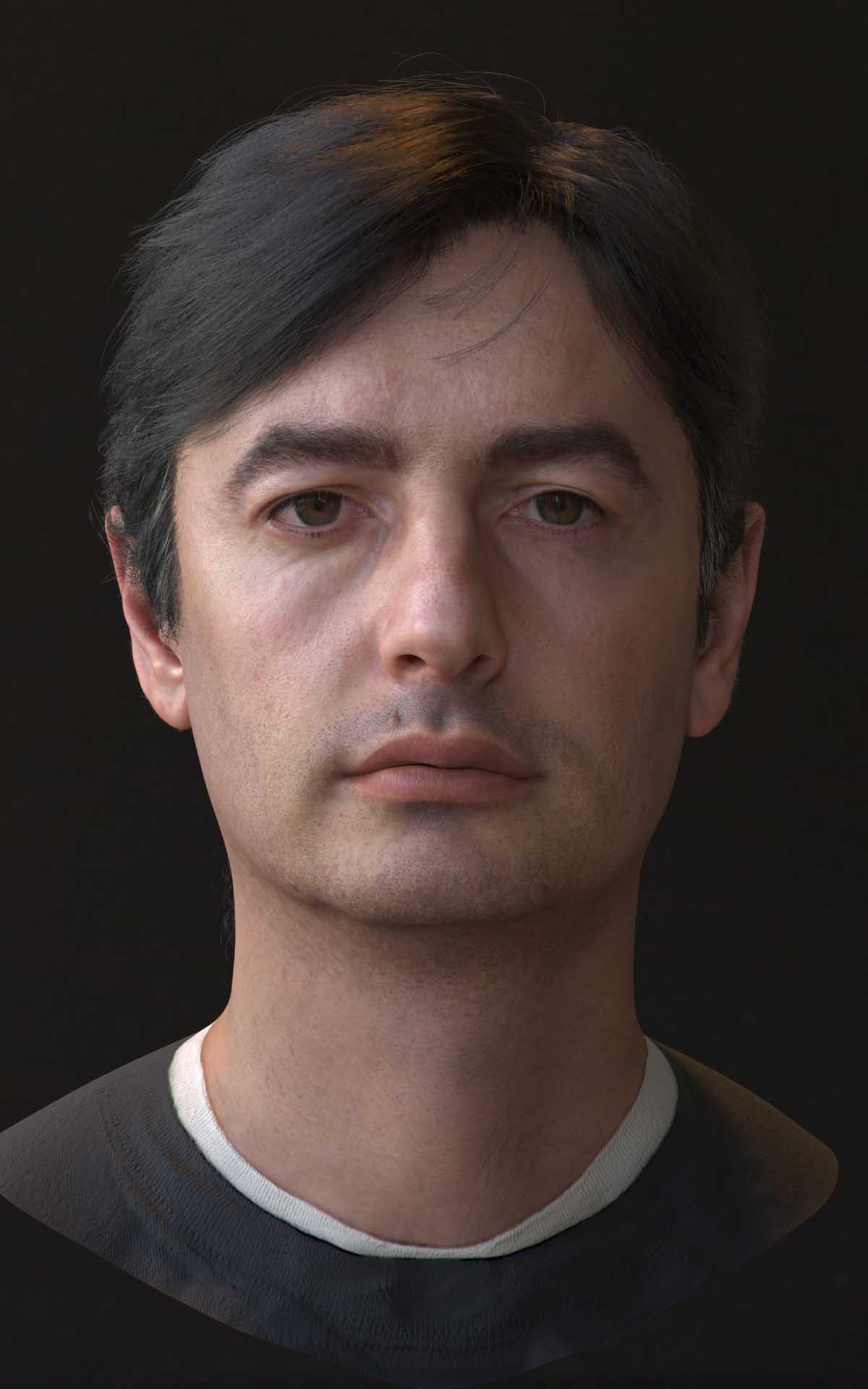}
 &
   \includegraphics[trim= 0 340 0 0, clip, width =0.14\textwidth]{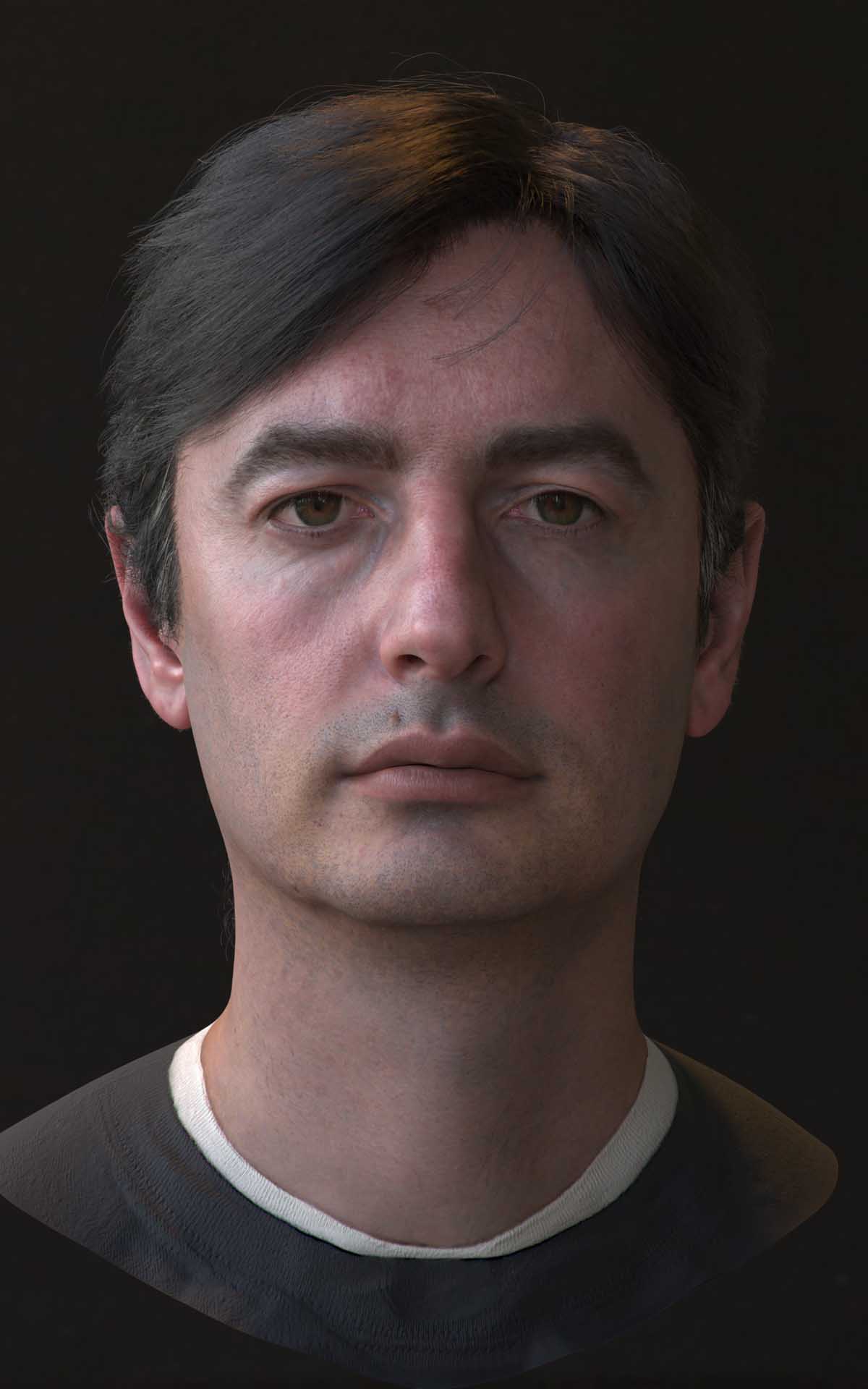}
 &
   \includegraphics[trim= 0 340 0 0, clip, width =0.14\textwidth]{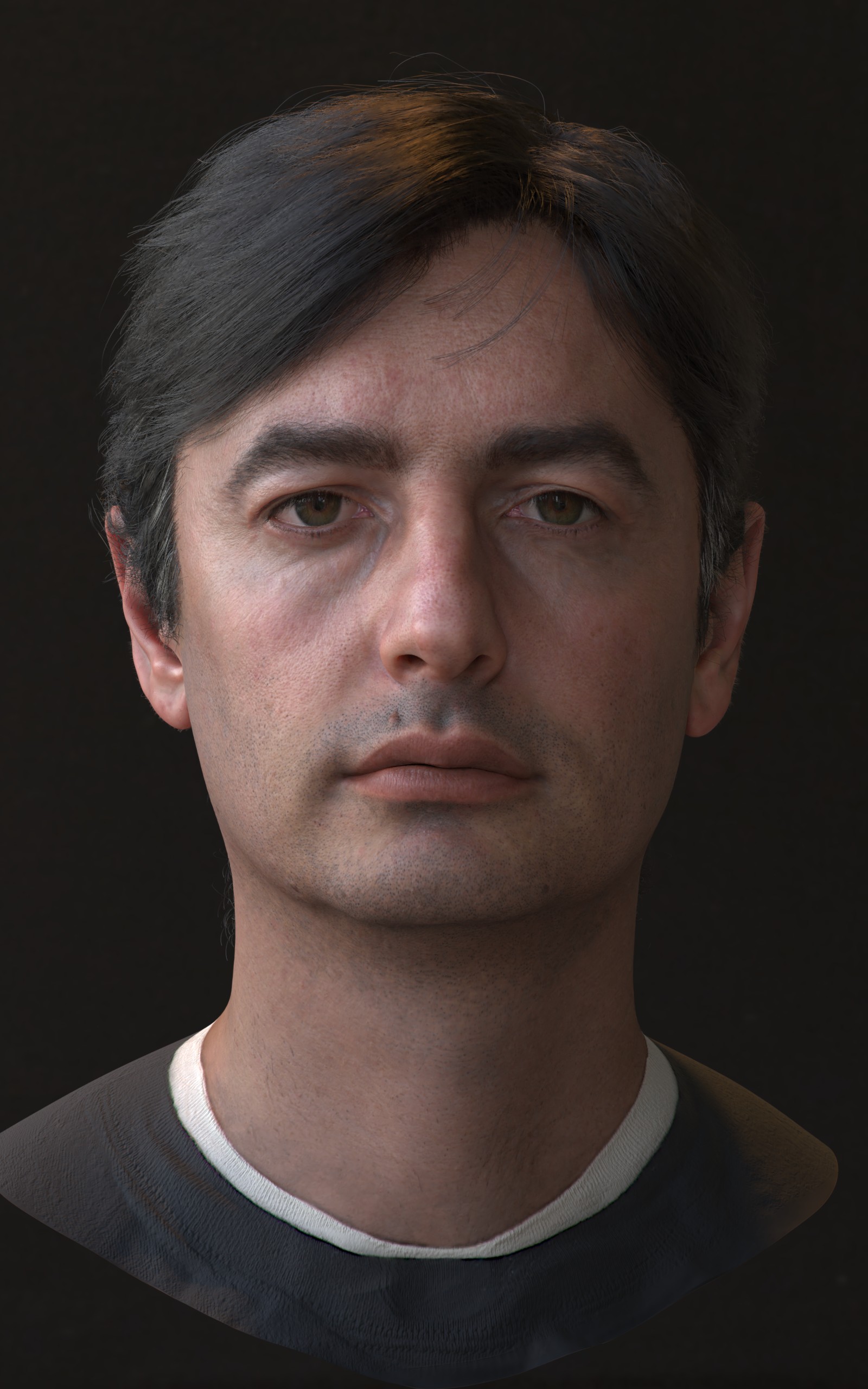}
 &
   \includegraphics[trim= 0 340 0 0, clip, width =0.14\textwidth]{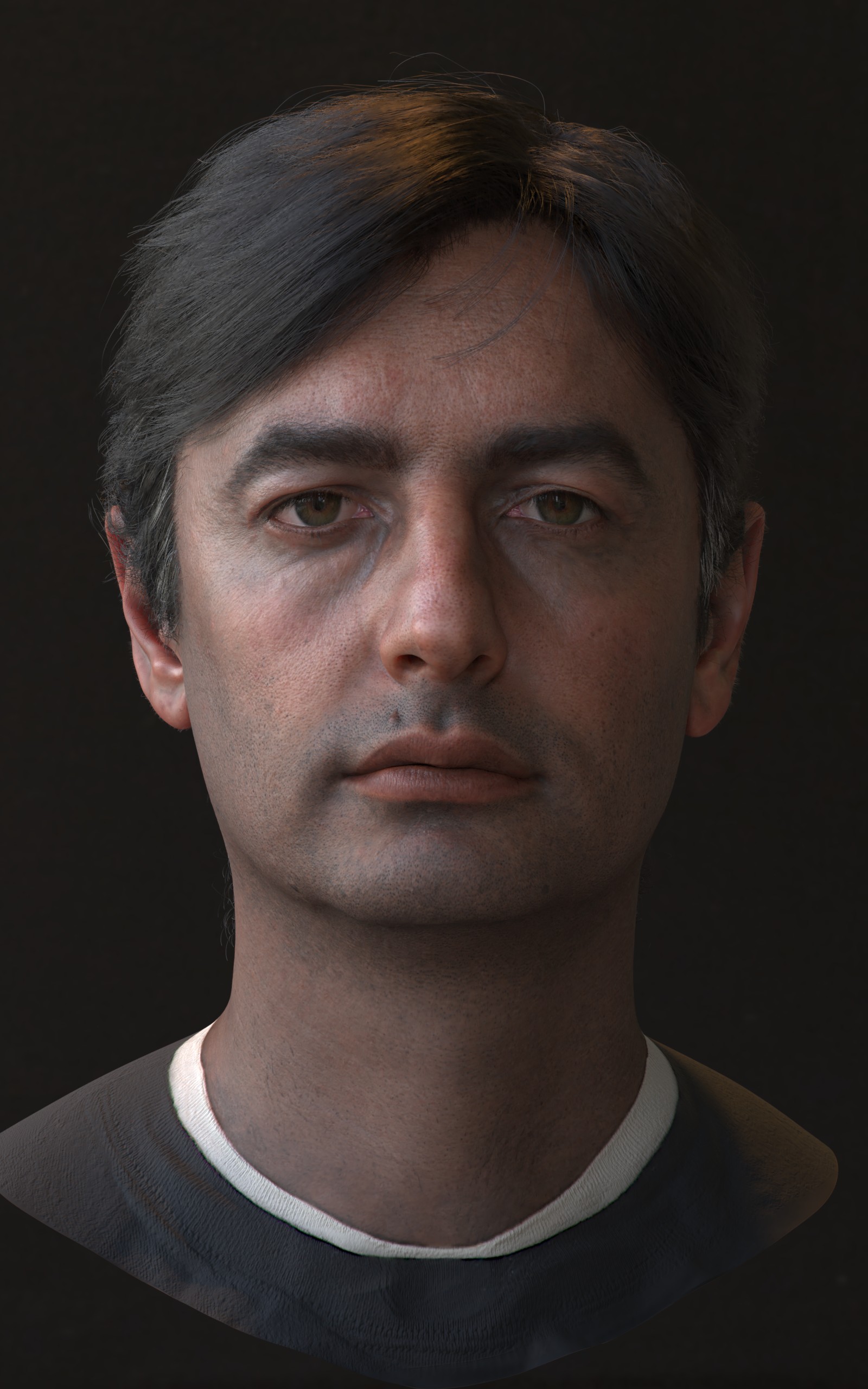}
 &
   \includegraphics[trim= 0 340 0 0, clip, width =0.14\textwidth]{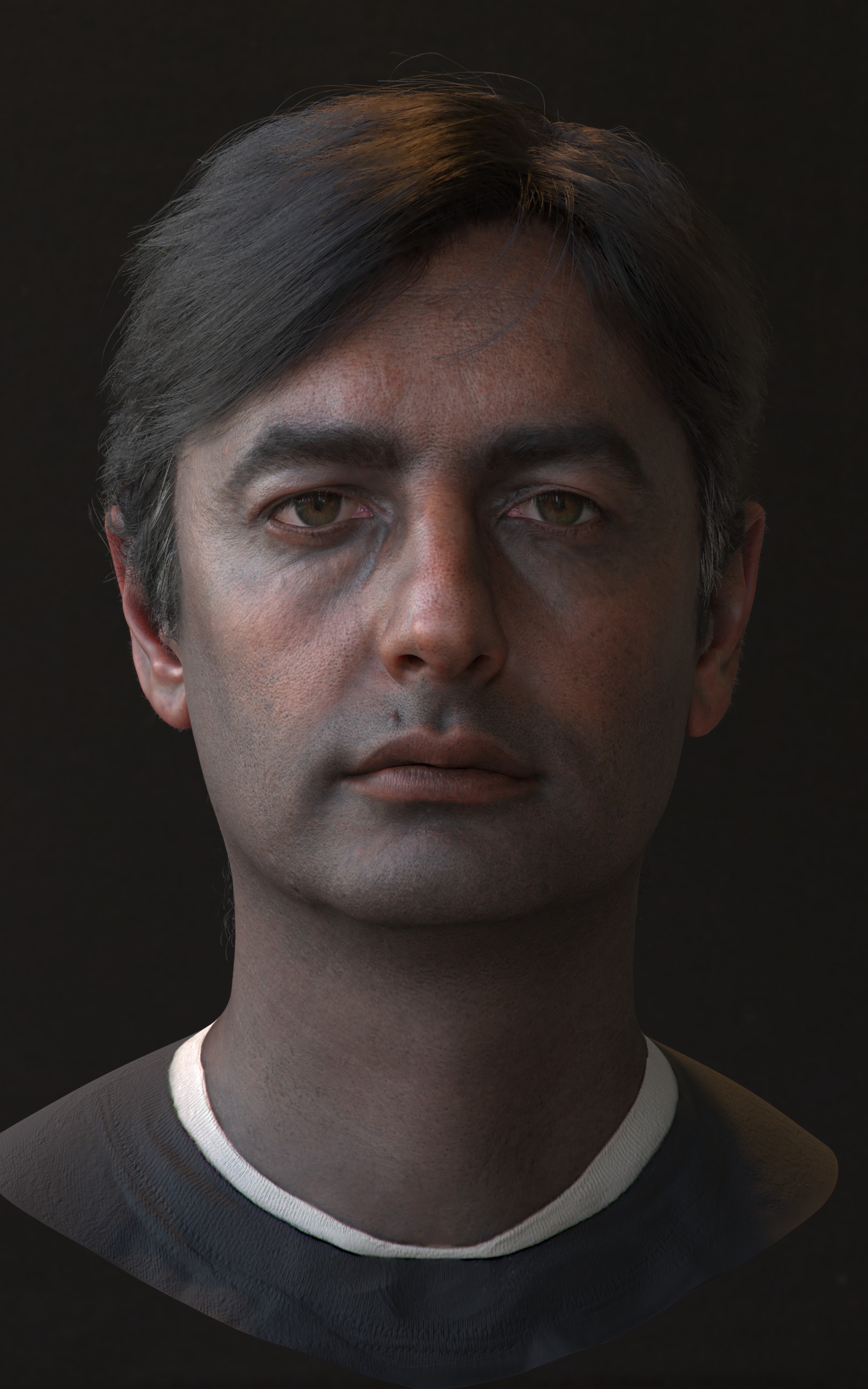}
  
  \\
 & & \textsc{0.125x blood } & \textsc{0.5x blood} & \textsc{3x blood} & \textsc{0.5x melanin} & \textsc{1.5x melanin} & \textsc{3x melanin} 
  \\
  \begin{sideways}\hspace{0.2cm}\textsc{Type III (Subject D)}\end{sideways}
 &
   \includegraphics[trim= 0 340 0 0, clip, width =0.14\textwidth]{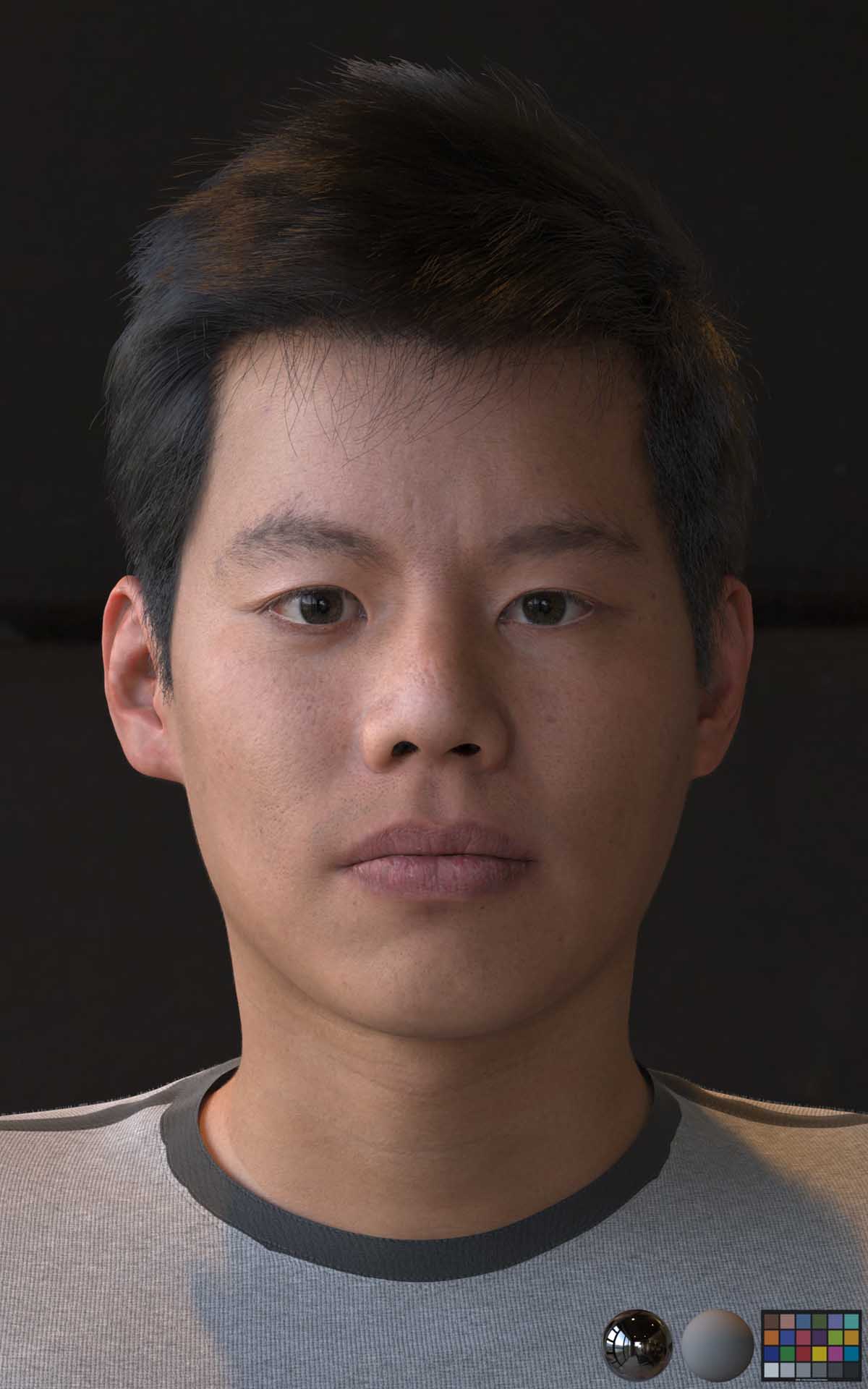}
 &
   \includegraphics[trim= 0 340 0 0, clip, width =0.14\textwidth]{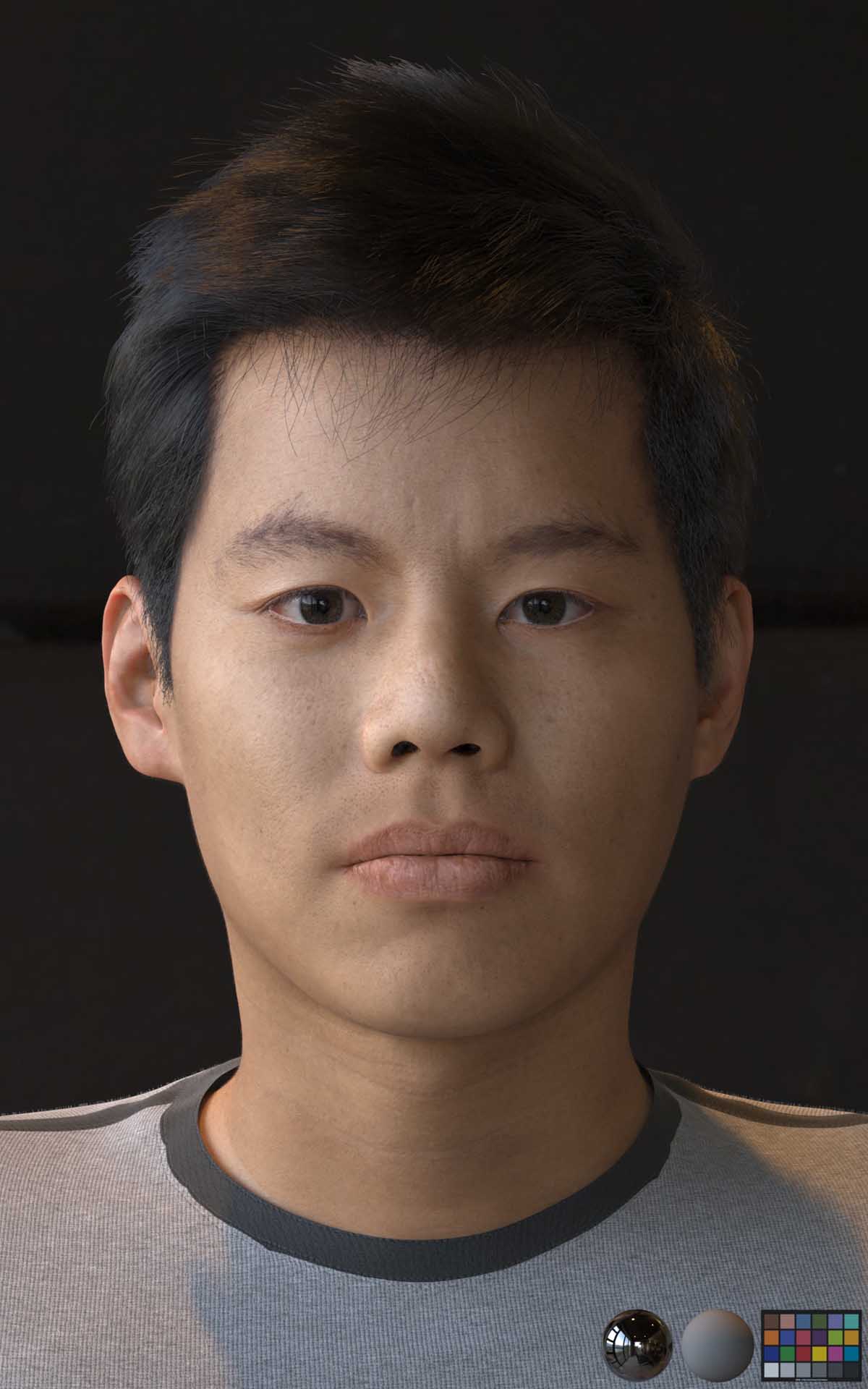}
 &
   \includegraphics[trim= 0 340 0 0, clip, width =0.14\textwidth]{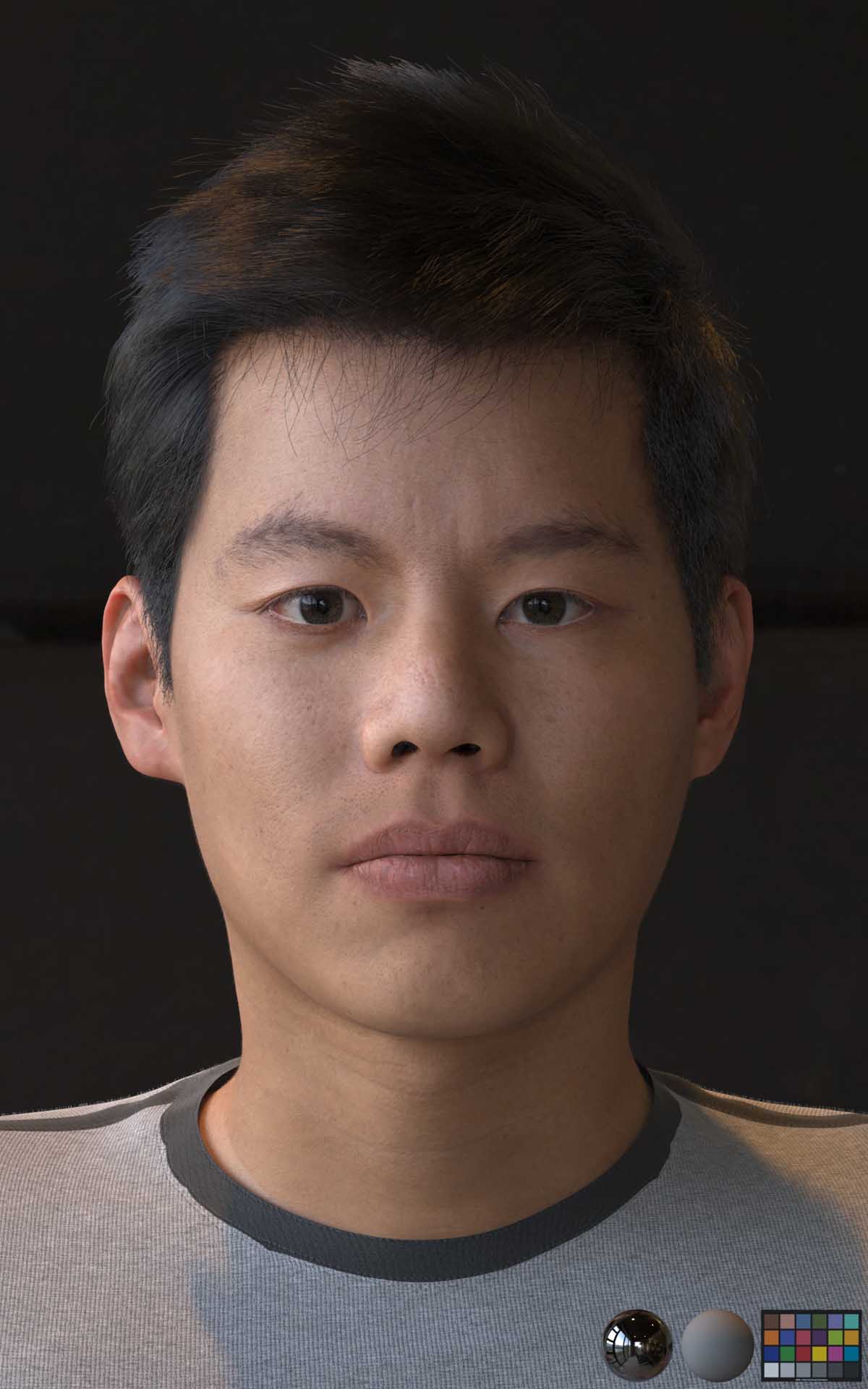}
 &
   \includegraphics[trim= 0 340 0 0, clip, width =0.14\textwidth]{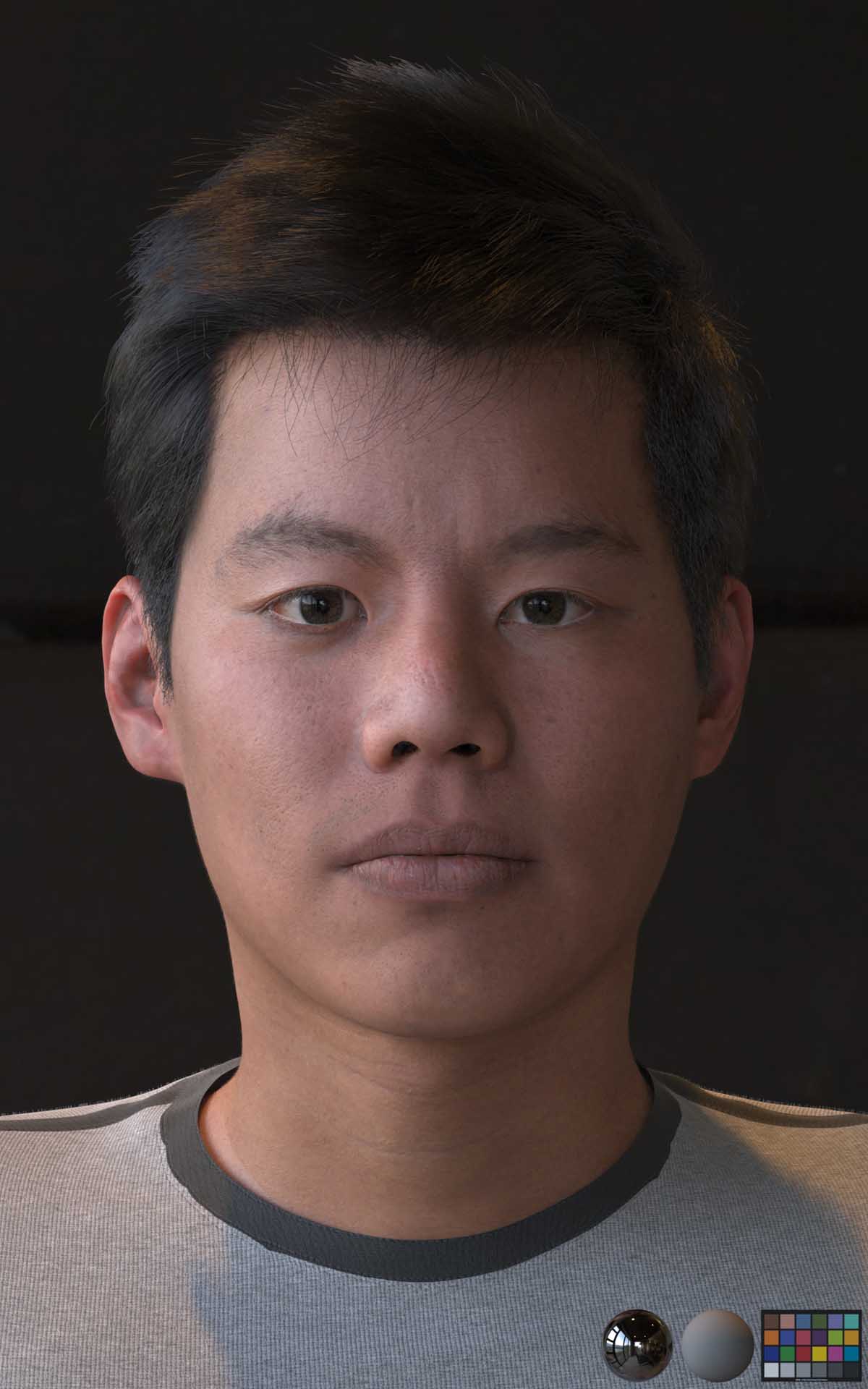}
 &
   \includegraphics[trim= 0 340 0 0, clip, width =0.14\textwidth]{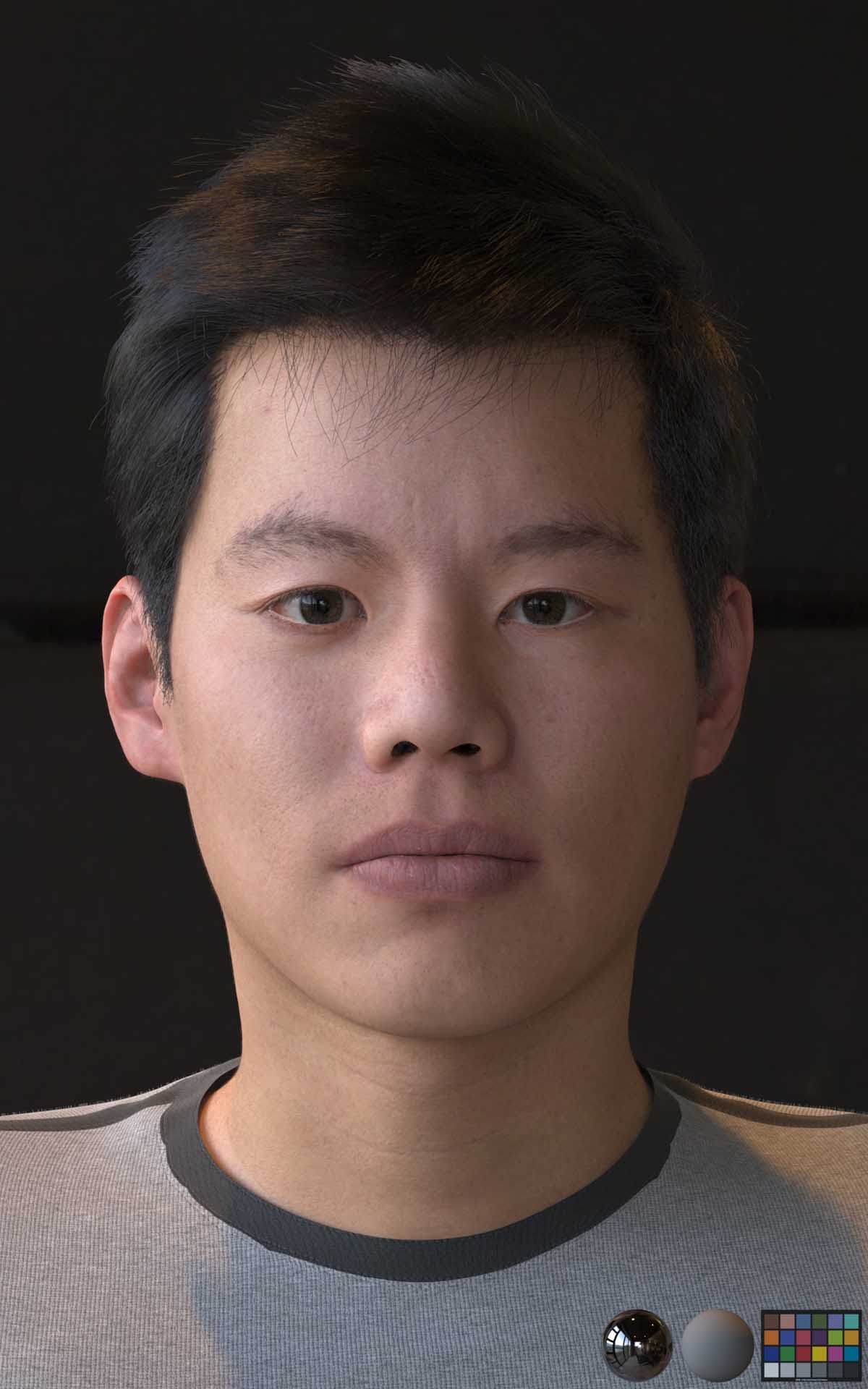}
 &
   \includegraphics[trim= 0 340 0 0, clip, width =0.14\textwidth]{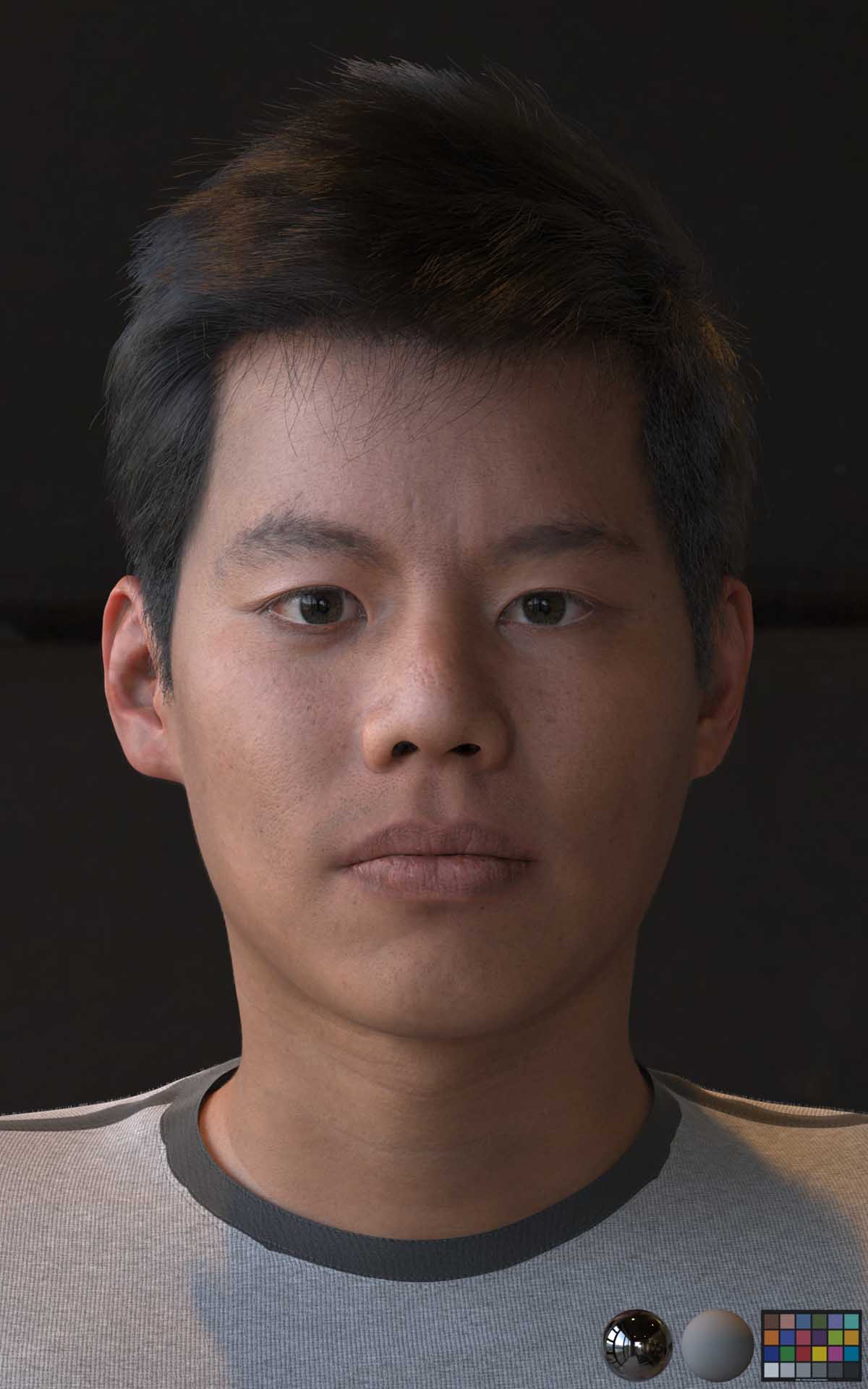}
 &
   \includegraphics[trim= 0 340 0 0, clip, width =0.14\textwidth]{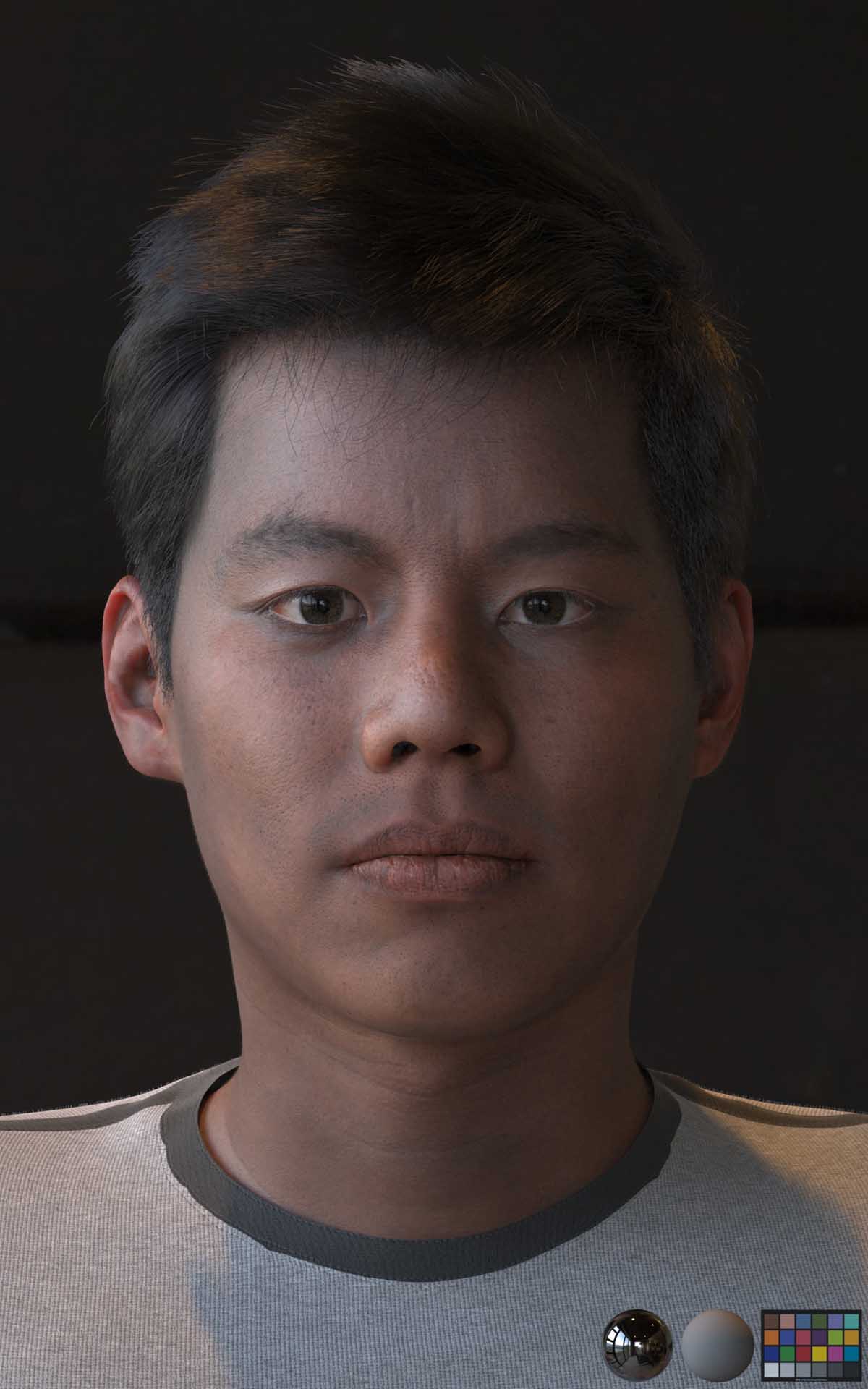}

  \\
 & & \textsc{0.125x blood } & \textsc{0.5x blood} & \textsc{3x blood} & \textsc{0.5x melanin} & \textsc{1.5x melanin} & \textsc{3x melanin} 
  \\
  \begin{sideways}\hspace{0.2cm}\textsc{Type IV (Subject E)}\end{sideways}
 &
   \includegraphics[trim= 0 340 0 0, clip, width =0.14\textwidth]{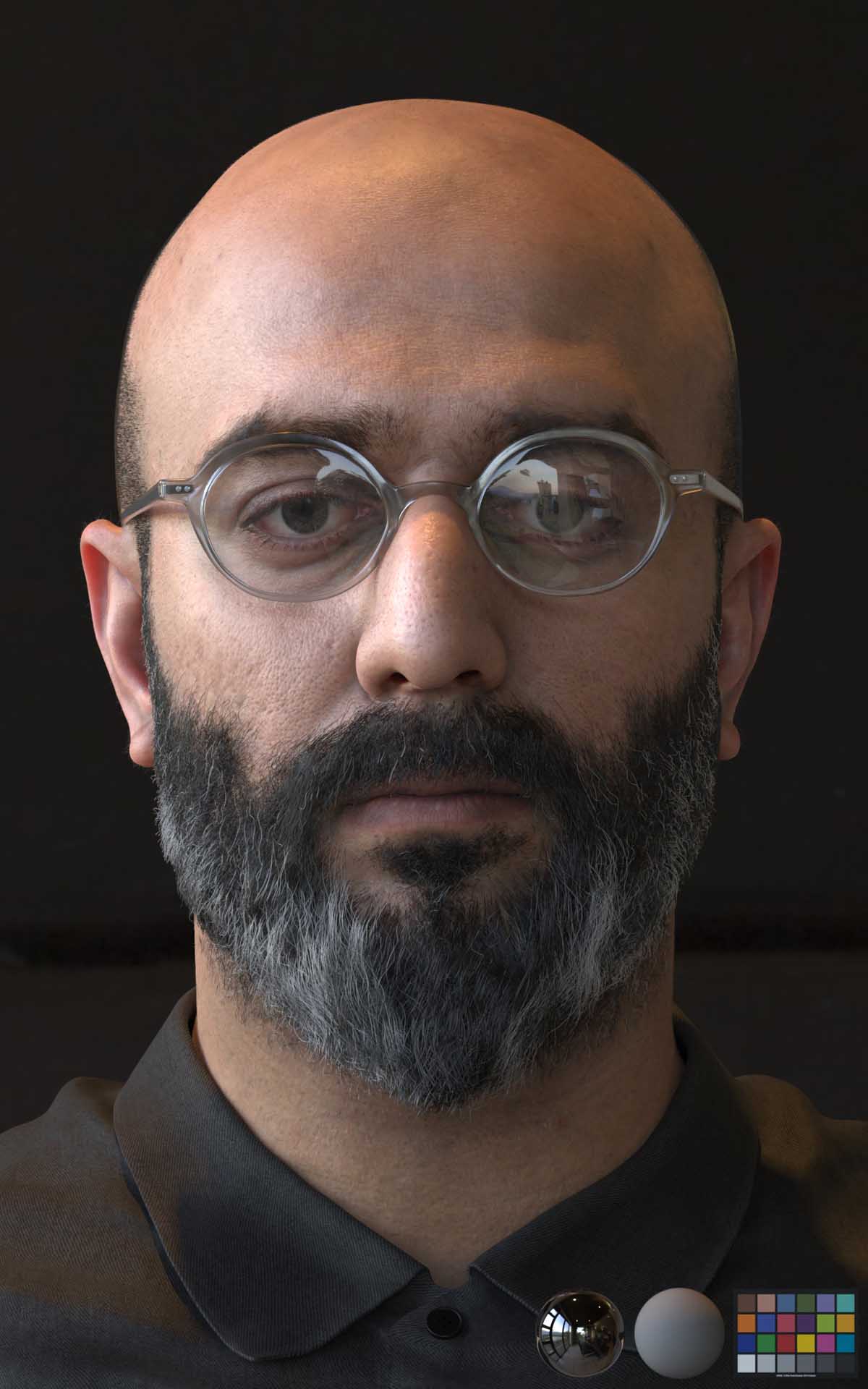}
 &
   \includegraphics[trim= 0 340 0 0, clip, width =0.14\textwidth]{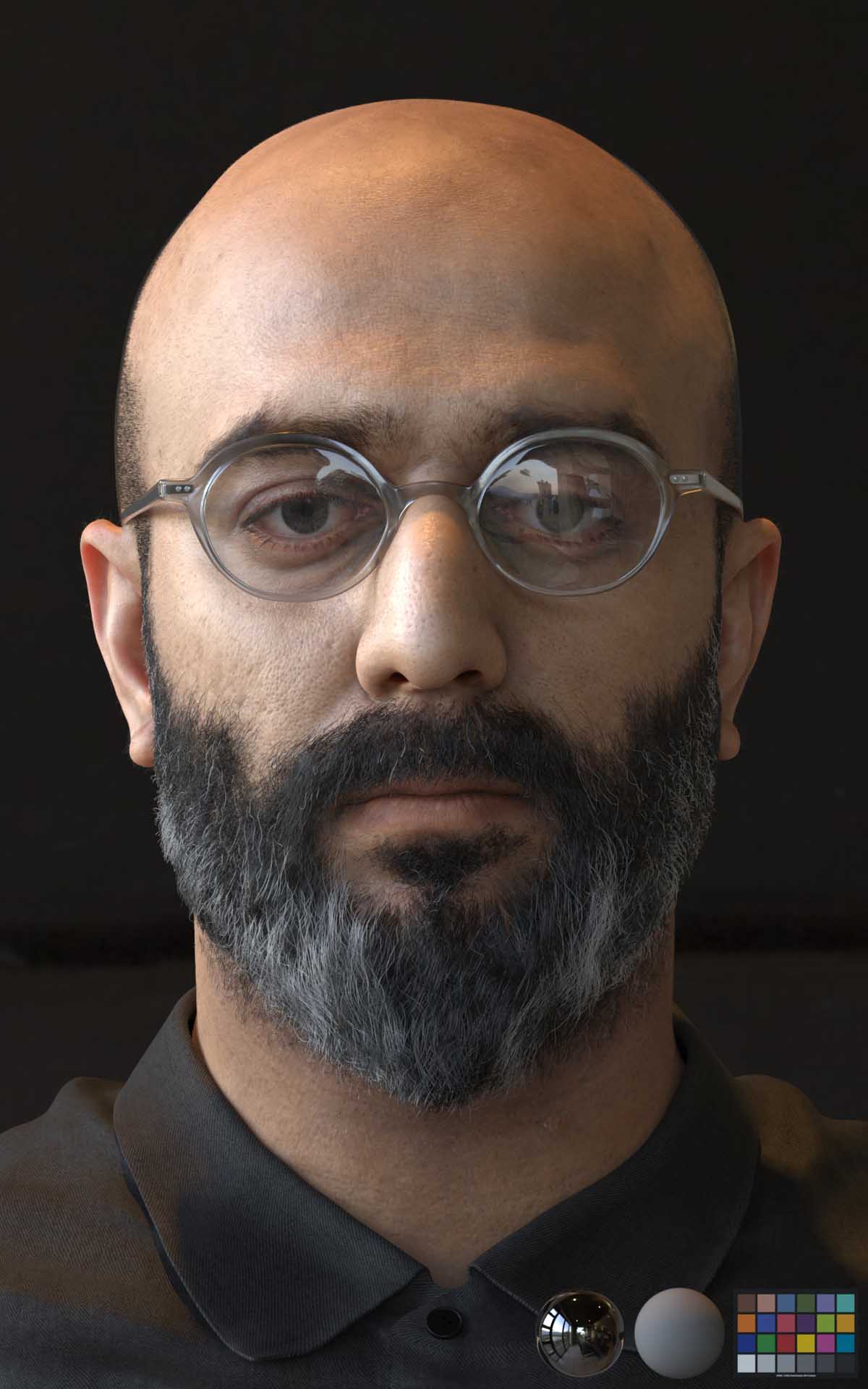}
 &
   \includegraphics[trim= 0 340 0 0, clip, width =0.14\textwidth]{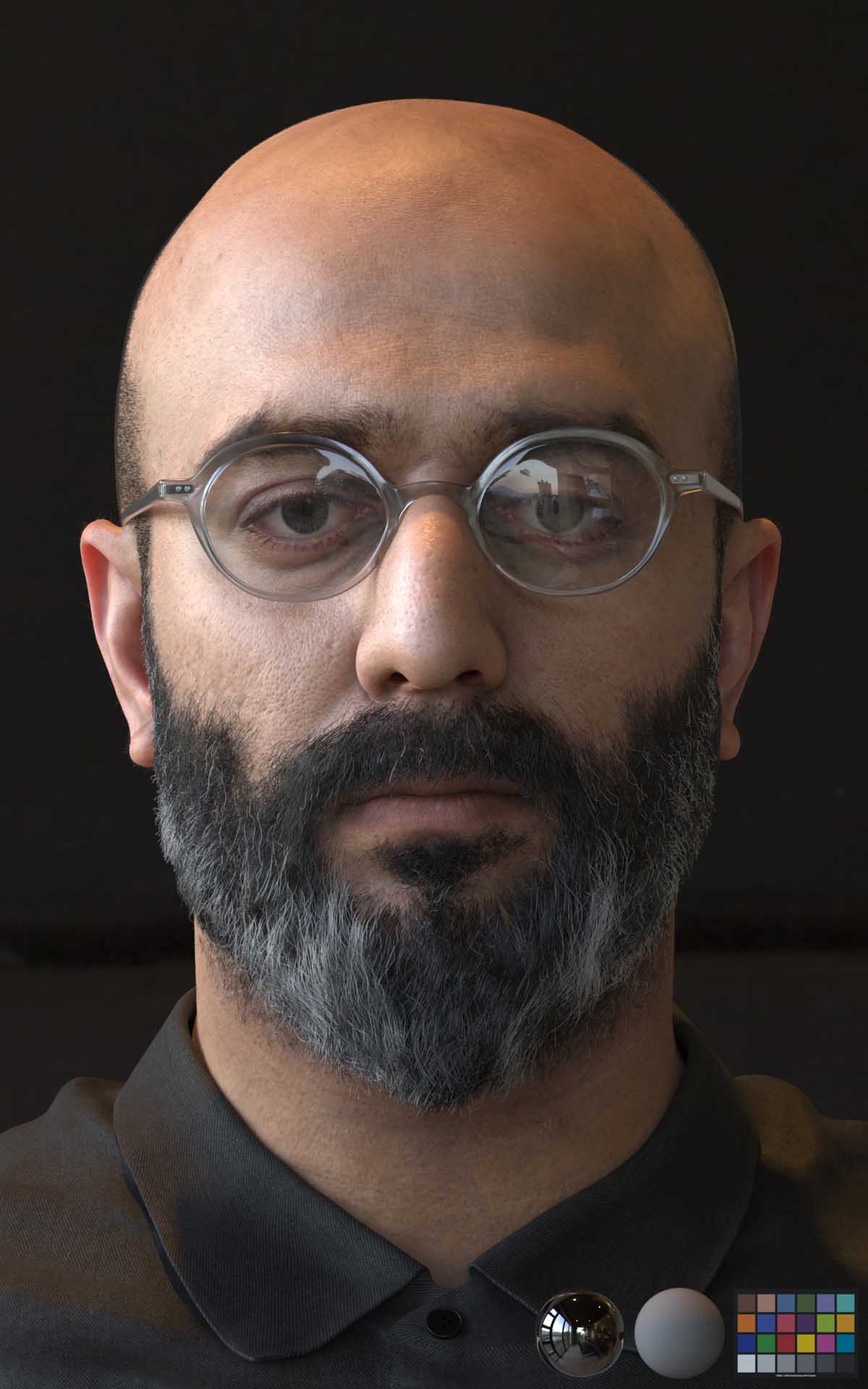}
 &
   \includegraphics[trim= 0 340 0 0, clip, width =0.14\textwidth]{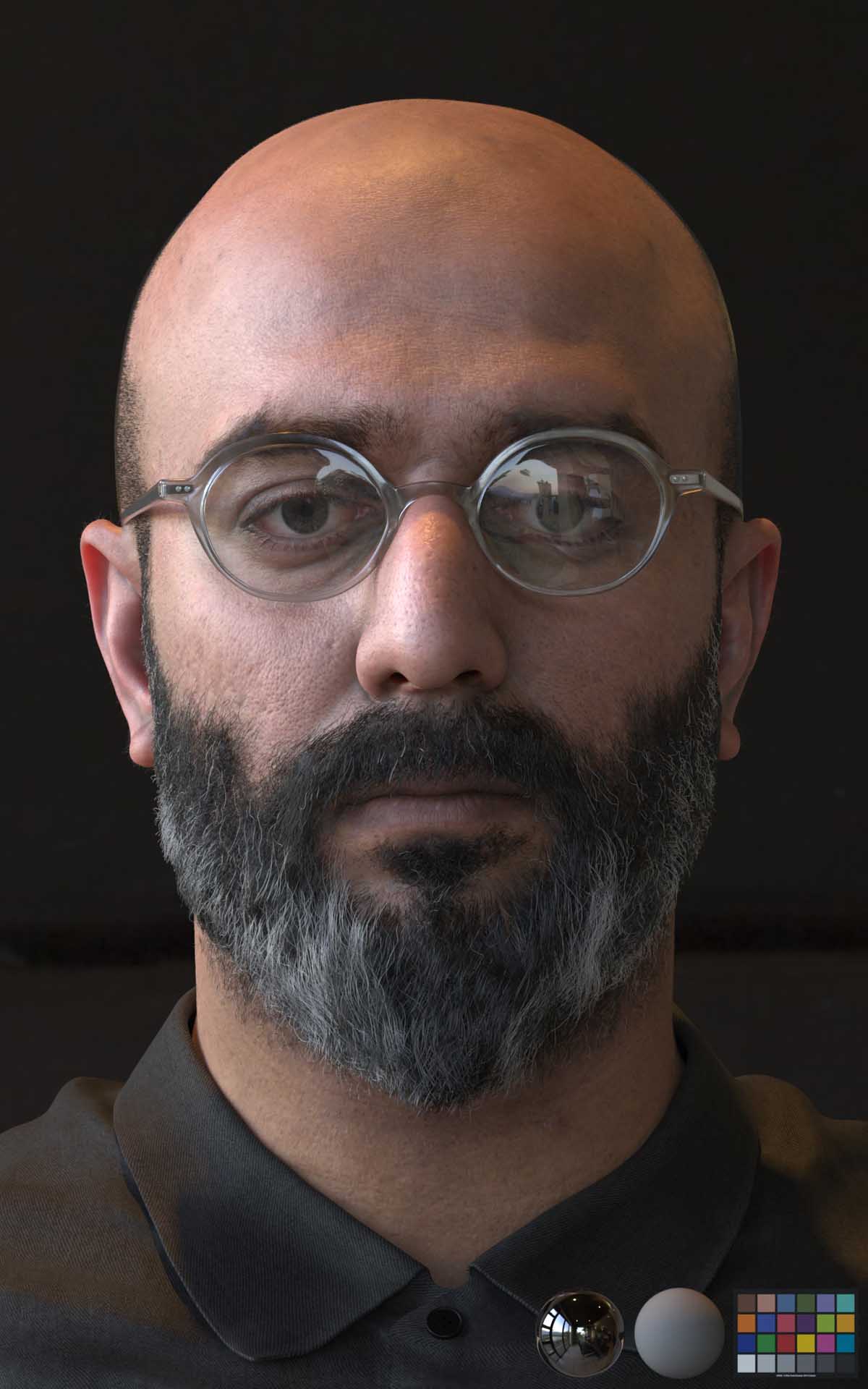}
 &
   \includegraphics[trim= 0 340 0 0, clip, width =0.14\textwidth]{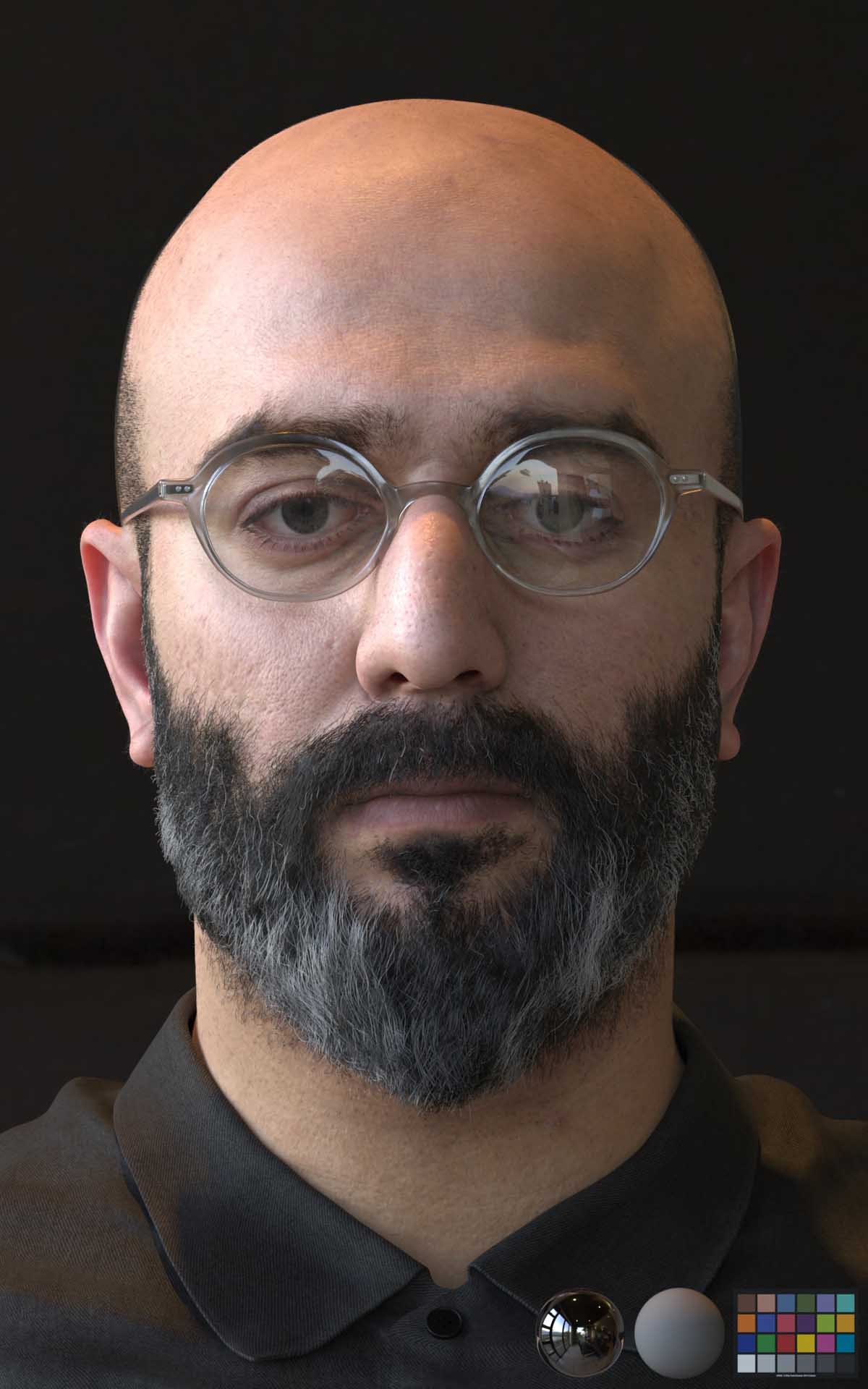}
 &
   \includegraphics[trim= 0 340 0 0, clip, width =0.14\textwidth]{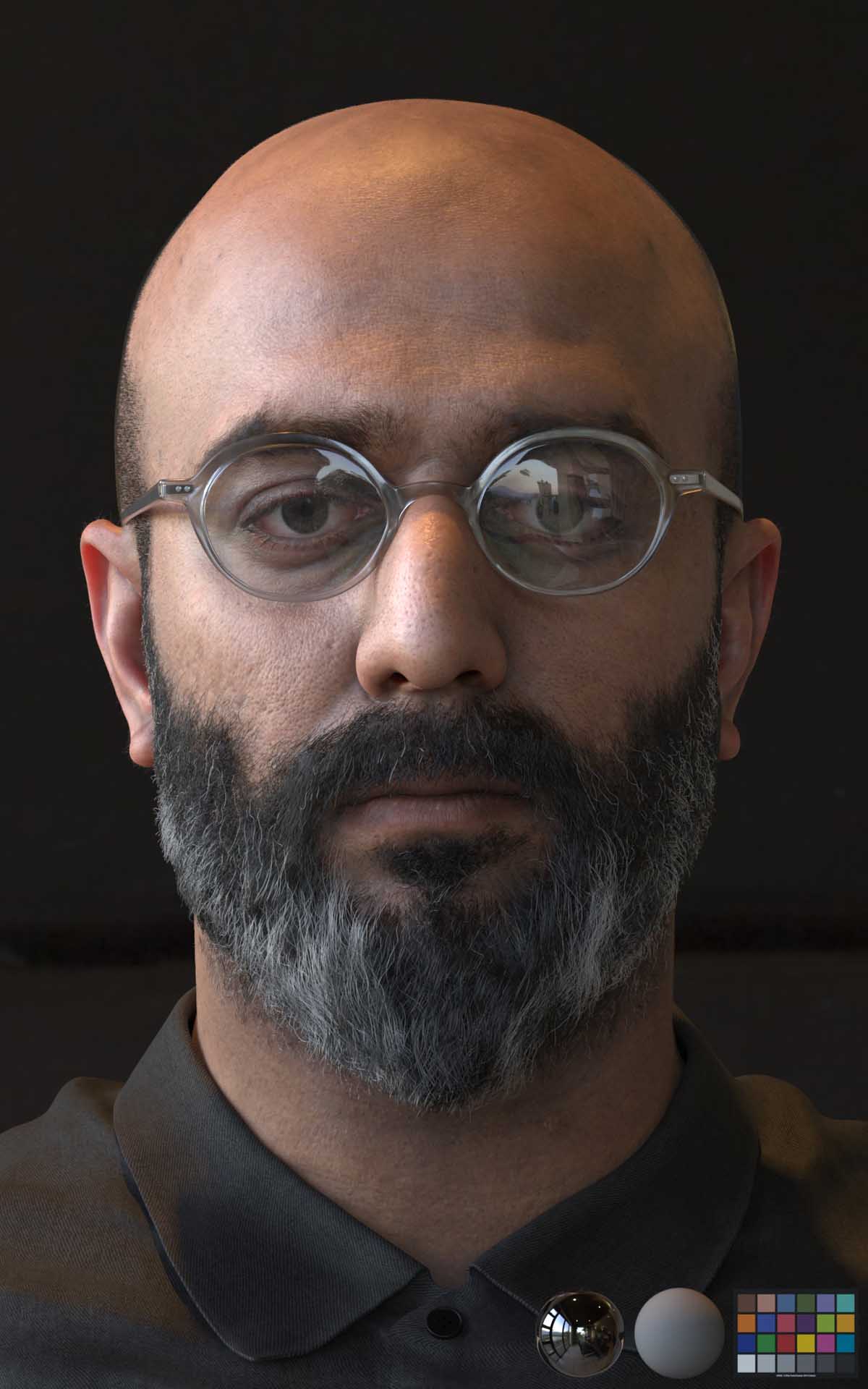}
 &
   \includegraphics[trim= 0 340 0 0, clip, width =0.14\textwidth]{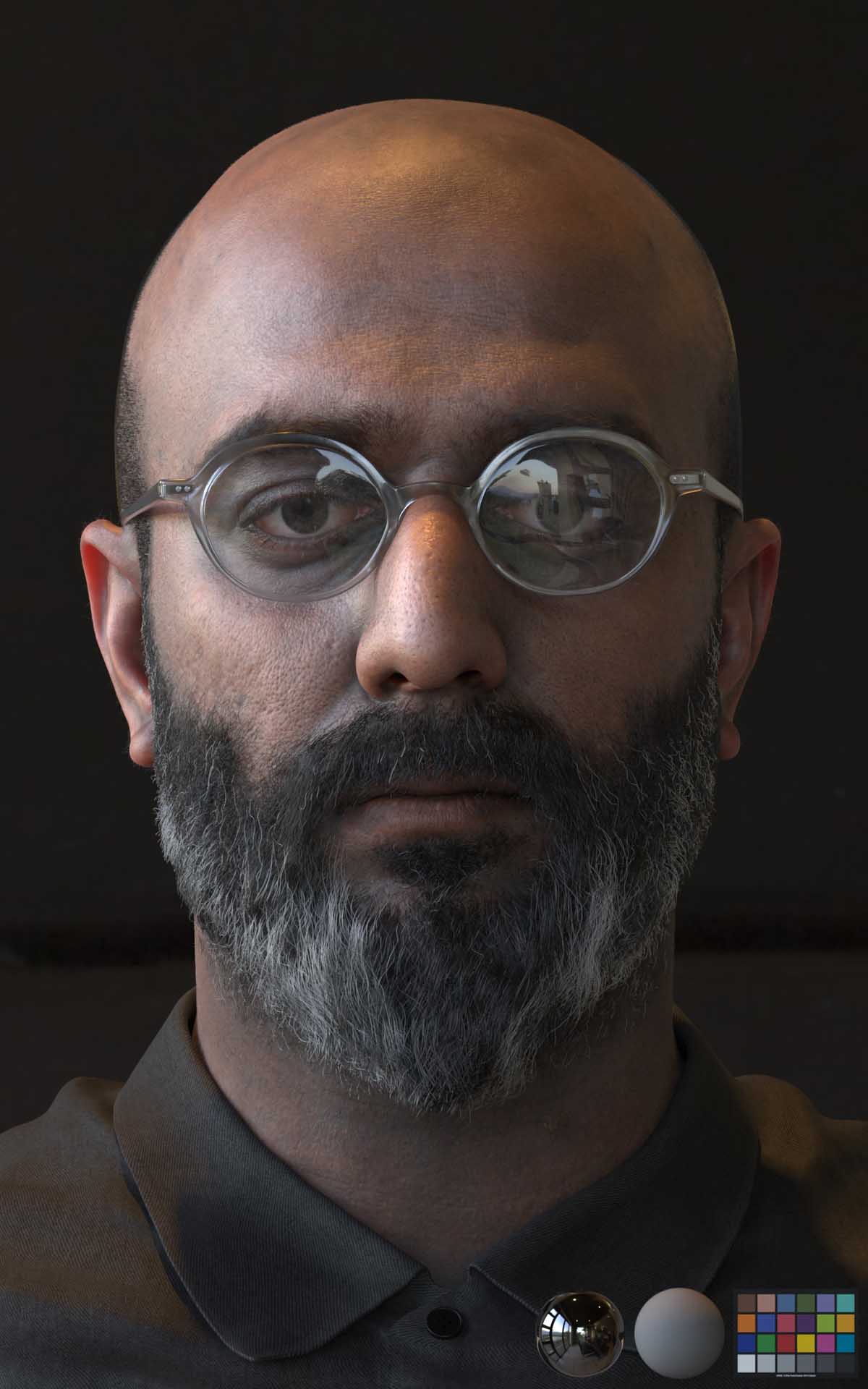}

  \\
 & & \textsc{0.125x blood } & \textsc{0.5x blood} & \textsc{3x blood} & \textsc{0.5x melanin} & \textsc{1.5x melanin} & \textsc{3x melanin} 
  \\
  \begin{sideways}\hspace{0.2cm}\textsc{Type V (Subject F)}\end{sideways}
 &
   \includegraphics[trim= 0 340 0 0, clip, width =0.14\textwidth]{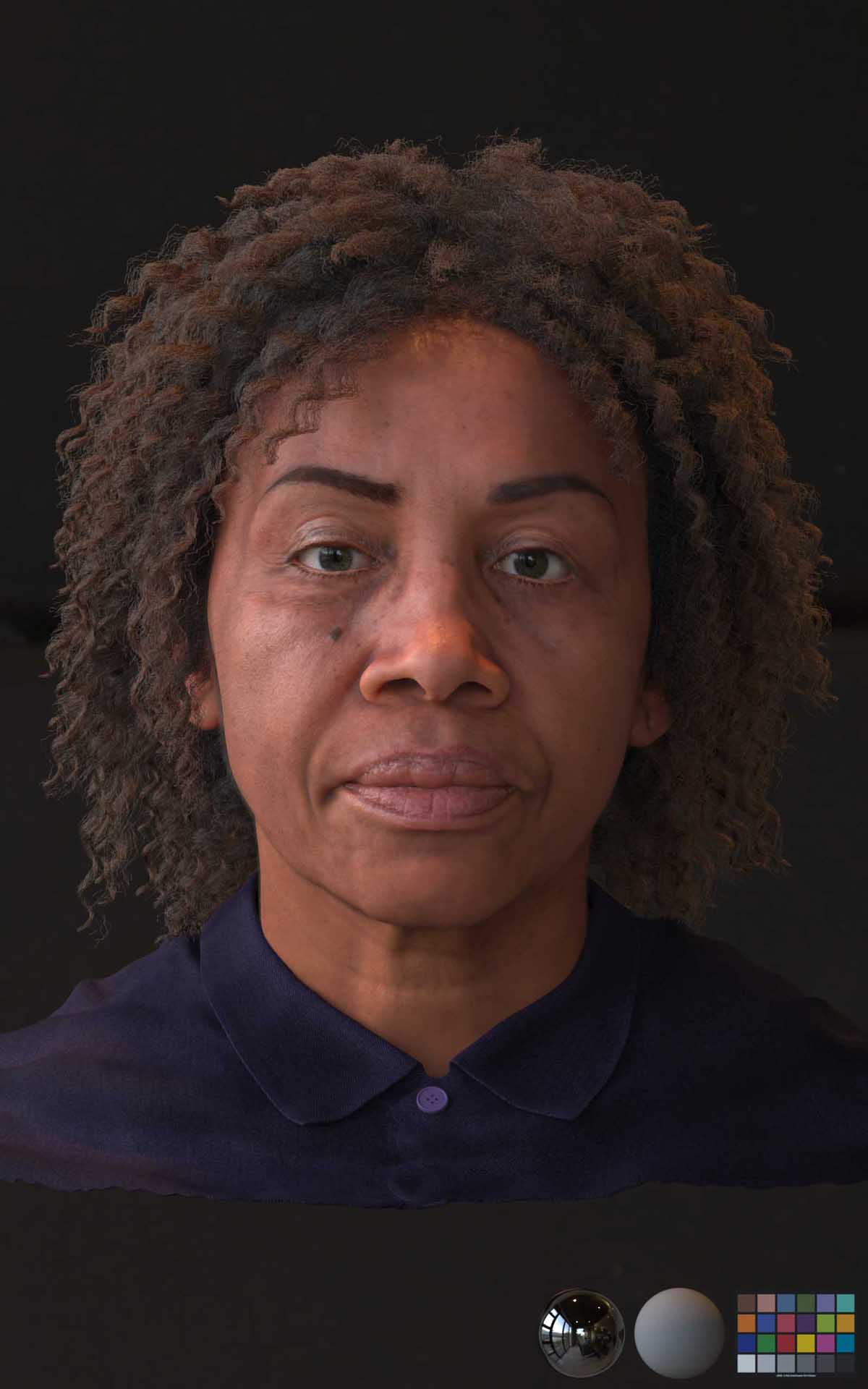}
 &
   \includegraphics[trim= 0 340 0 0, clip, width =0.14\textwidth]{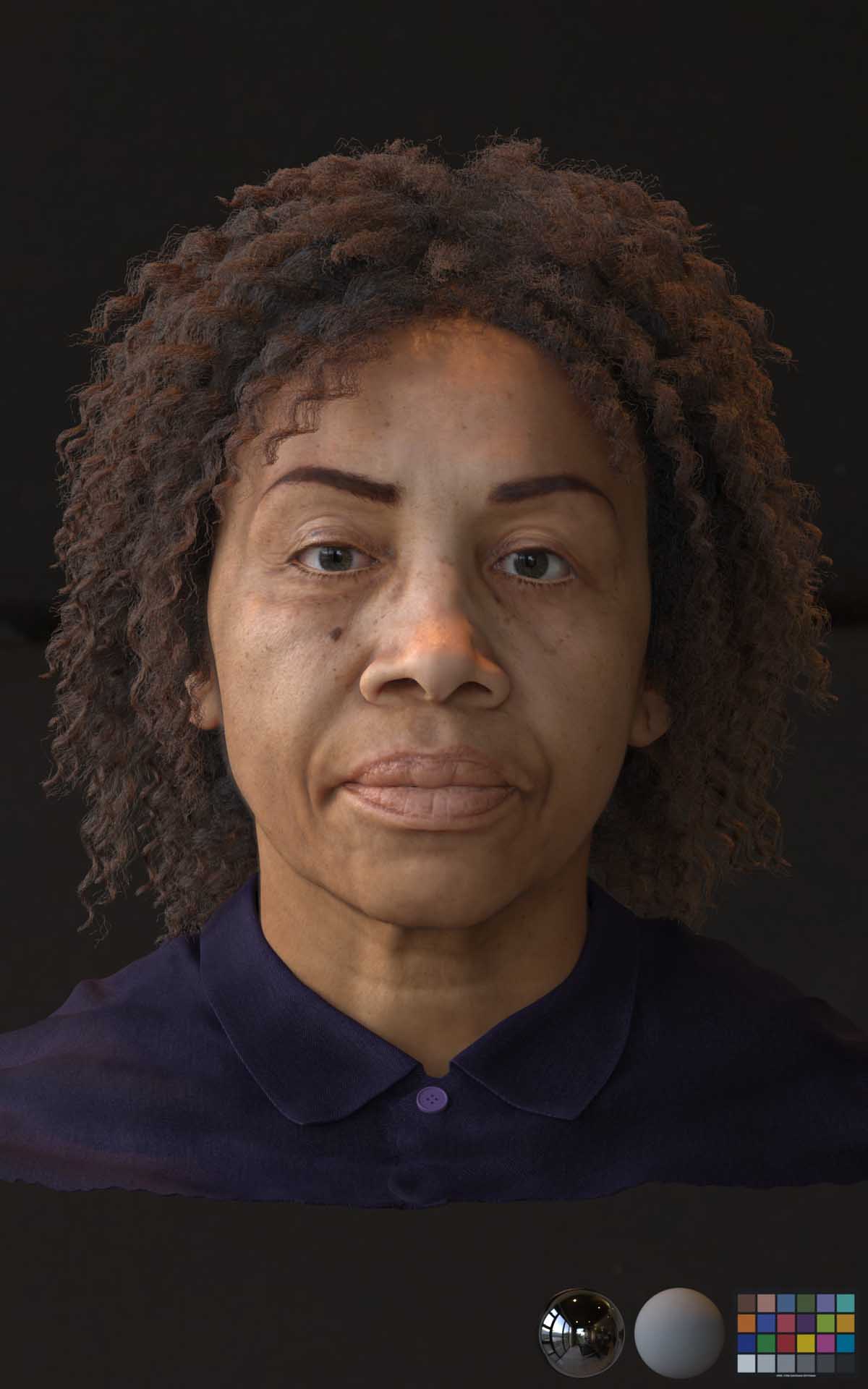}
 &
   \includegraphics[trim= 0 340 0 0, clip, width =0.14\textwidth]{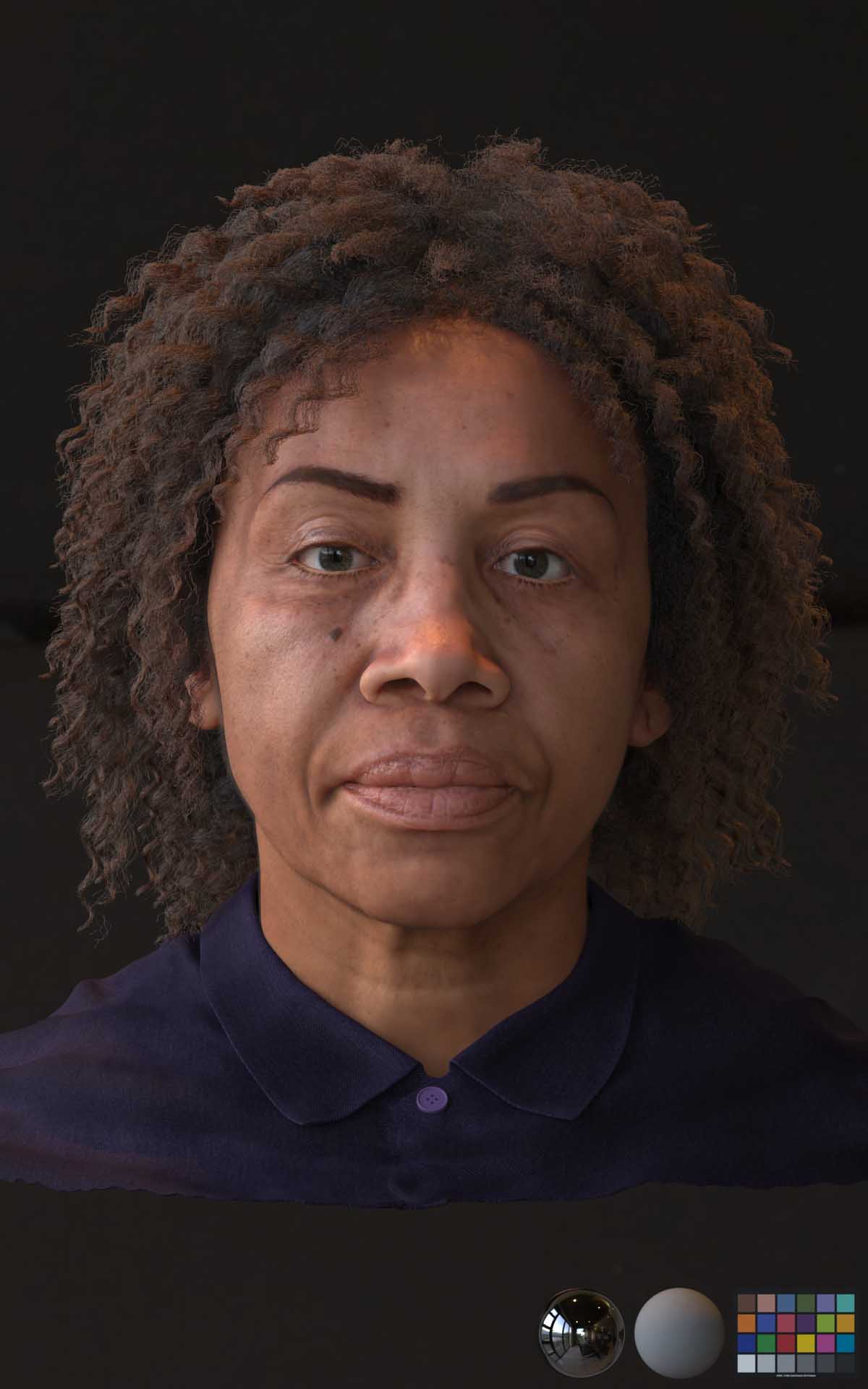}
 &
   \includegraphics[trim= 0 340 0 0, clip, width =0.14\textwidth]{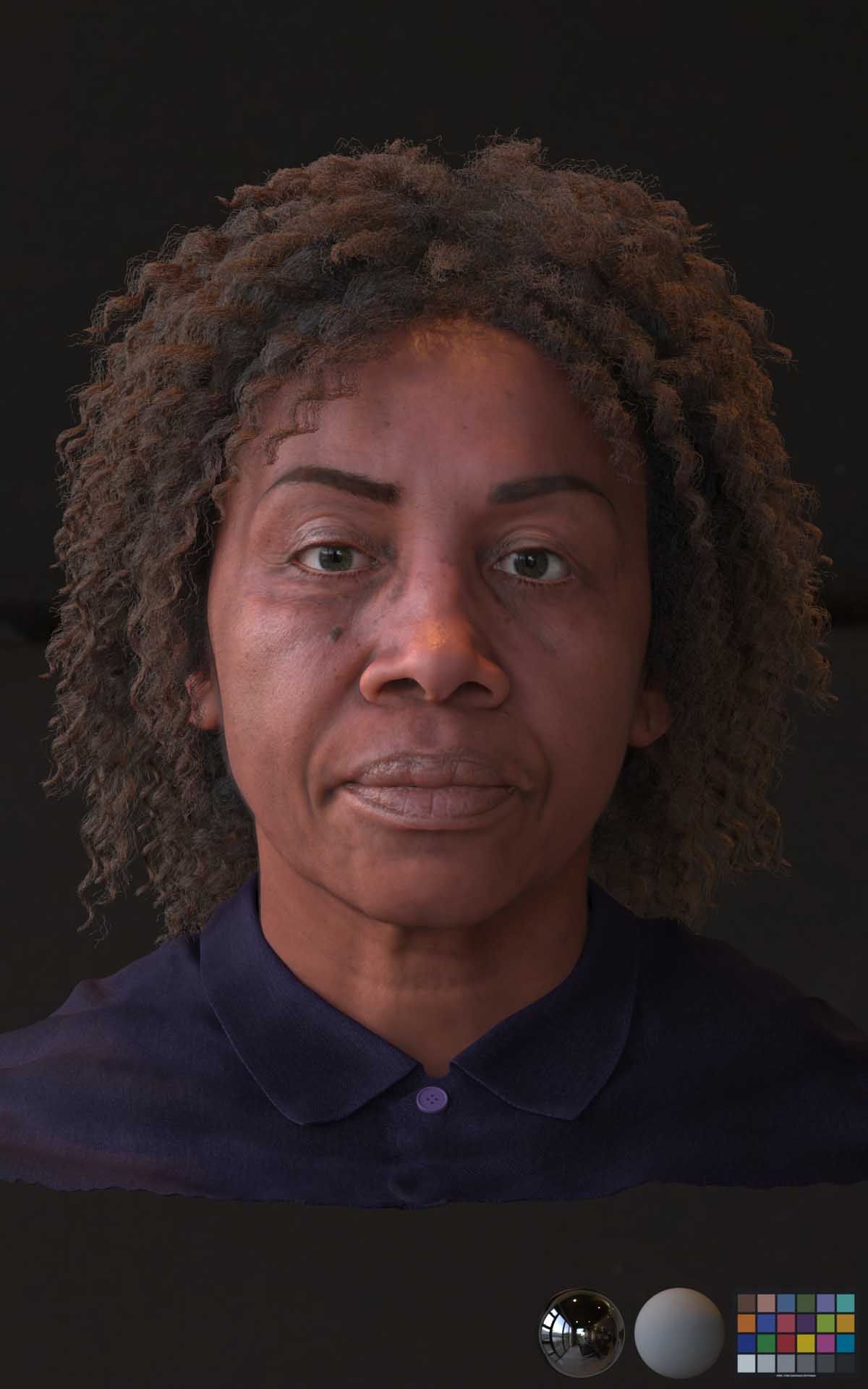}
 &
   \includegraphics[trim= 0 340 0 0, clip, width =0.14\textwidth]{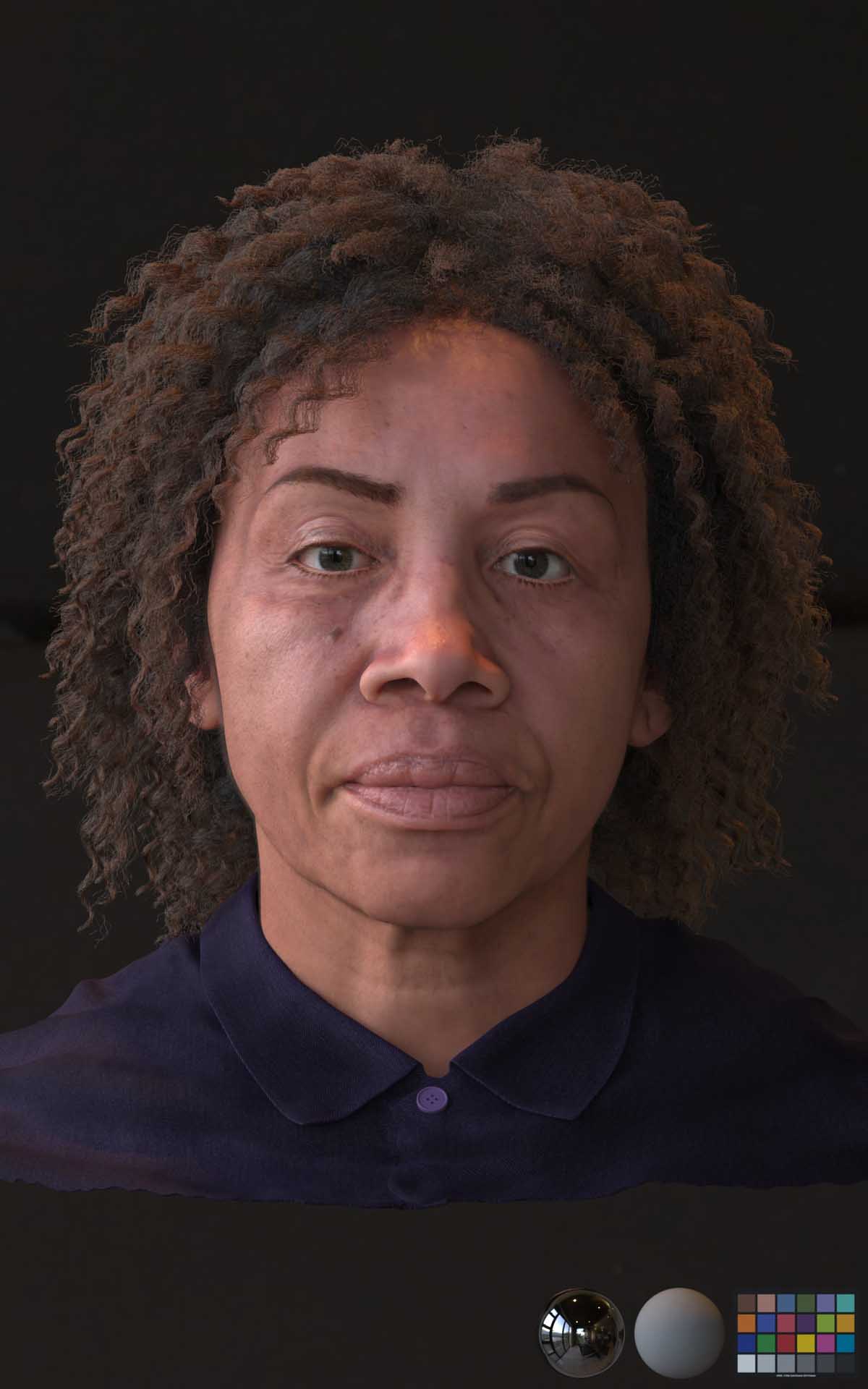}
 &
   \includegraphics[trim= 0 340 0 0, clip, width =0.14\textwidth]{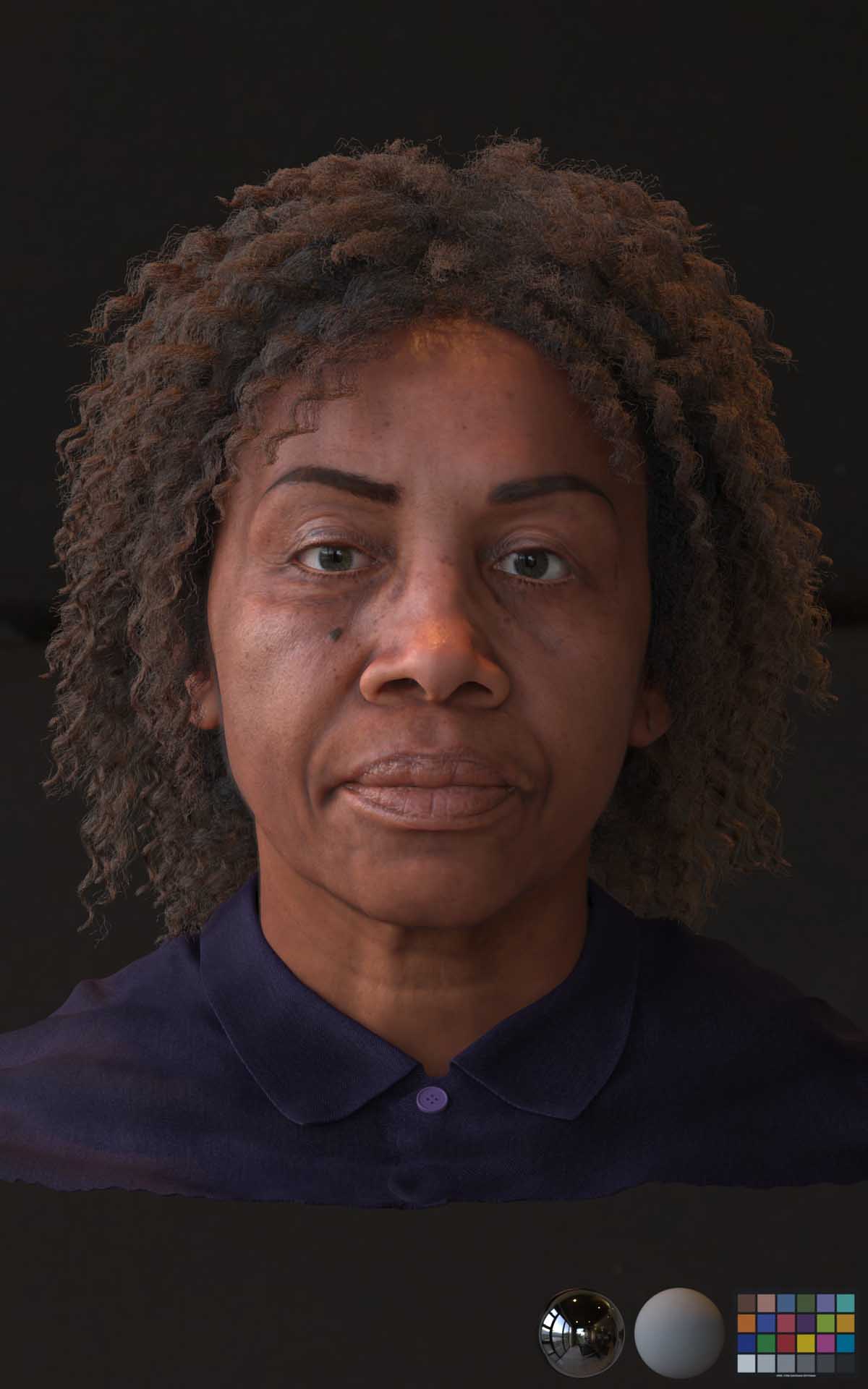}
 &
   \includegraphics[trim= 0 340 0 0, clip, width =0.14\textwidth]{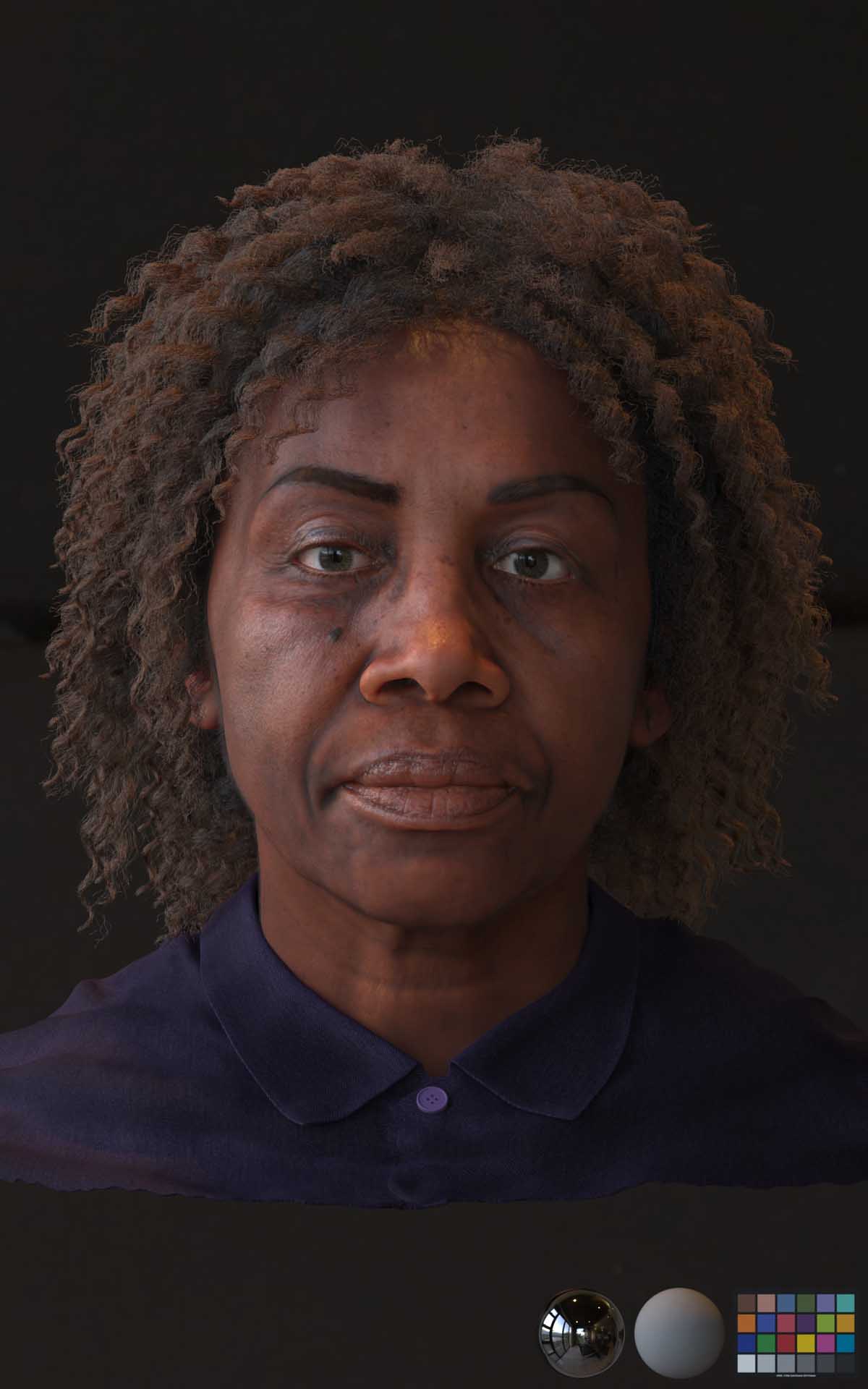}

 \end{tabular}
 \vspace{-3mm}
  \caption{Rendering results under lighting scenario 2 for edited skin parameters inferred by our model over different skin types. From left to right, original, followed by different manipulations per subject, three in blood concentration and three in melanin concentration.}
  \label{fig:editing_2}
  \vspace{-6mm}
\end{figure*}

In Figure~\ref{fig:editing_1} and Figure~\ref{fig:editing_2}, we present two tables of renderings with edited albedos under two different lighting conditions. Consider also Figure 9 in the main paper for a third lighting scenario.


\printbibliography                

